%% file: iclr2023_conference.tex
\PassOptionsToPackage{dvipsnames}{xcolor}
\documentclass{article} %
\usepackage{iclr2023_conference,times}

\usepackage{caption,subcaption}
\usepackage{amsthm,amsmath,amsfonts,amssymb}
\usepackage{graphicx}
\usepackage{enumitem}
\usepackage[framemethod=tikz]{mdframed}

\usepackage[colorlinks=true, citecolor=MidnightBlue]{hyperref}
\usepackage{url}

\input{math_commands.tex}
\input{pkghead.tex}

\title{Clifford Neural Layers for PDE Modeling}

\author{Johannes Brandstetter\\
Microsoft Research AI4Science \\
\texttt{johannesb@microsoft.com} \\
\And
Rianne van den Berg \\
Microsoft Research AI4Science \\
\texttt{rvandenberg@microsoft.com} \\
\And
Max Welling \\
Microsoft Research AI4Science \\
\texttt{maxwelling@microsoft.com} \\
\And
Jayesh K. Gupta \\
Microsoft Autonomous Systems and Robotics Research \\
\texttt{jayesh.gupta@microsoft.com} \\
}

\iclrfinalcopy %

\begin{document}

\maketitle

\begin{abstract}
 Partial differential equations (PDEs) see widespread use in sciences and engineering to describe simulation of physical processes as scalar and vector fields interacting and coevolving over time.
 Due to the computationally expensive nature of their standard solution methods, neural PDE surrogates have become an active research topic to accelerate these simulations.
 However, current methods do not explicitly take into account the relationship between different fields and their internal components, which are often correlated.
 Viewing the time evolution of such correlated fields through the lens of multivector fields allows us to overcome these limitations.
 Multivector fields consist of scalar, vector, as well as higher-order components, such as bivectors and trivectors. Their algebraic properties, such as multiplication, addition and other arithmetic operations can be described by Clifford algebras. 
 To our knowledge, this paper presents the first usage of such multivector representations together with Clifford convolutions and Clifford Fourier transforms in the context of deep learning.
The resulting Clifford neural layers are universally applicable and will find direct use in the areas of fluid dynamics, weather forecasting, and the modeling of physical systems in general.
We empirically evaluate the benefit of Clifford neural layers by replacing convolution and Fourier operations in common neural PDE surrogates by their Clifford counterparts on 2D Navier-Stokes and weather modeling tasks, as well as 3D Maxwell equations. For similar parameter count, 
Clifford neural layers consistently improve generalization capabilities of the tested neural PDE surrogates.
Source code for our PyTorch implementation is available at
\url{https://microsoft.github.io/cliffordlayers/}
\end{abstract}

\input{sections/intro.tex}

\input{sections/background.tex}

\input{sections/approach.tex}

\input{sections/experiments.tex} 

\input{sections/conclusion.tex}

\section*{Reproducibility and ethical statement}
\paragraph{Reproducibility statement.}
We have included error bars, and ablation studies wherever we found it
necessary and appropriate. 
We have described our architectures in Section~\ref{sec:experiments} and provided further implementation details in Appendix Section~\ref{app:experiments}.
We have further include pseudocode for the newly proposed layers in Appendix Section~\ref{app:pseudocode}. 
We open-sourced our PyTorch implementation at
\url{https://microsoft.github.io/cliffordlayers/} for others to use. We aim to develop this codebase further in the future.

\paragraph{Ethical statement.}
Neural PDE surrogates will play an important role in modeling many natural phenomena, and thus developing them further might enable us to achieve shortcuts or alternatives for computationally expensive simulations.
For example, if used as such, PDE surrogates will potentially help to advance different fields of research, especially in the natural sciences. Examples related to this paper are fluid dynamics or weather modeling. Therefore, PDE surrogates might potentially be directly or indirectly related to reducing the
carbon footprint.
On the downside, relying on simulations always requires rigorous cross-checks and
monitoring, especially when we ``learn to simulate''.
\bibliography{literature}
\bibliographystyle{iclr2023_conference}

\clearpage

\appendix
\appendixpage
\begingroup
\hypersetup{linkcolor=black}
\tableofcontents
\endgroup

\strut\newpage

\clearpage

\input{appendix/cliffordalgebra.tex}

\strut\newpage

\clearpage

\input{appendix/cliffordlayers.tex}

\input{appendix/experiments.tex}

\strut\newpage

\clearpage

\input{appendix/relatedwork}

\input{appendix/glossary.tex}

\end{document}

%% file: math_commands.tex
\usepackage{amsmath,amsfonts,bm}

\def\eqref#1{equation~\ref{#1}}

\def\1{\bm{1}}

\def\cin{{c_{\text{in}}}}
\def\cout{{c_{\text{out}}}}

\def\rva{{\mathbf{a}}}

\def\va{{\bm{a}}}
\def\vb{{\bm{b}}}
\def\vc{{\bm{c}}}

\def\ve{{\bm{e}}}
\def\vf{{\bm{f}}}
\def\vg{{\bm{g}}}

\def\vj{{\bm{j}}}
\def\vk{{\bm{k}}}

\def\vs{{\bm{s}}}

\def\vv{{\bm{v}}}
\def\vw{{\bm{w}}}

\def\mF{{\bm{F}}}

\def\mH{{\bm{H}}}

\def\mR{{\bm{R}}}

\def\mW{{\bm{W}}}

\DeclareMathAlphabet{\mathsfit}{\encodingdefault}{\sfdefault}{m}{sl}
\SetMathAlphabet{\mathsfit}{bold}{\encodingdefault}{\sfdefault}{bx}{n}

\def\gF{{\mathcal{F}}}

\def\gL{{\mathcal{L}}}

\def\gX{{\mathcal{X}}}

\def\sZ{{\mathbb{Z}}}

\newcommand{\R}{\mathbb{R}}
\newcommand{\CC}{\mathbb{C}}
\newcommand{\HH}{\mathbb{H}}

\newcommand{\eone}{\eqnmarkbox[OliveGreen]{}{e_1}}
\newcommand{\etwo}{\eqnmarkbox[RoyalPurple]{}{e_2}}
\newcommand{\ethree}{\eqnmarkbox[CadetBlue]{}{e_3}}
\newcommand{\eonetwo}{\eqnmarkbox[Mulberry]{}{e_1e_2}}
\newcommand{\etwoone}{\eqnmarkbox[Mulberry]{}{e_2e_1}}
\newcommand{\eonethree}{\eqnmarkbox[SeaGreen]{}{e_1e_3}}
\newcommand{\ethreeone}{\eqnmarkbox[SeaGreen]{}{e_3e_1}}
\newcommand{\etwothree}{\eqnmarkbox[TealBlue]{}{e_2e_3}}
\newcommand{\ethreetwo}{\eqnmarkbox[TealBlue]{}{e_3e_2}}
\newcommand{\eonetwothree}{\eqnmarkbox[Thistle]{}{e_1e_2e_3}}

\newcommand{\eonethreetwo}{\eqnmarkbox[Thistle]{}{e_1e_3e_2}}
\newcommand{\etwoonethree}{\eqnmarkbox[Thistle]{}{e_1e_3e_2}}
\newcommand{\etwothreeone}{\eqnmarkbox[Thistle]{}{e_1e_3e_2}}

\newcommand{\ethreeonetwo}{\eqnmarkbox[Thistle]{}{e_1e_3e_2}}

%% file: pkghead.tex
\usepackage{tikz}
\usepackage{mathdots}
\usepackage{yhmath}
\usepackage{cancel}
\usepackage{color}
\usepackage[binary-units=true]{siunitx}
\usepackage{array}
\usepackage{multirow}
\usepackage{amssymb}
\usepackage{gensymb}
\usepackage{tabularx}
\usepackage{extarrows}
\usepackage{booktabs}
\usepackage{threeparttable}
\usepackage{annotate-equations}
\usepackage{csquotes}
\usetikzlibrary{fadings}
\usetikzlibrary{patterns}
\usetikzlibrary{shadows.blur}
\usetikzlibrary{shapes}
\usepackage{pgfplots}
\pgfplotsset{compat = newest}
\usepgfplotslibrary{colormaps}
\pgfplotsset{
    colormap={colorblind}{
        rgb255=(232,236,251),
        rgb255=(217,204,227),
        rgb255=(209,187,215),
        rgb255=(202,172,203),
        rgb255=(186,141,180),
        rgb255=(174,118,163),
        rgb255=(170,111,158),
        rgb255=(153,79,136),
        rgb255=(136,46,114),
        rgb255=(25,101,176),
        rgb255=(67,125,191),
        rgb255=(82,137,199),
        rgb255=(97,149,207),
        rgb255=(123,175,222),
        rgb255=(78,178,101),
        rgb255=(144,201,135),
        rgb255=(202,224,171),
        rgb255=(247,240,86),
        rgb255=(247,203,69),
        rgb255=(246,193,65),
        rgb255=(244,167,54),
        rgb255=(241,147,45),
        rgb255=(238,128,38),
        rgb255=(232,96,28),
        rgb255=(230,85,24),
        rgb255=(220,5,12),
        rgb255=(165,23,14),
        rgb255=(114,25,14),
        rgb255=(66,21,10)
    },
}
\usepackage[utf8]{inputenc}
\usepgfplotslibrary{groupplots,dateplot}
\usetikzlibrary{patterns,shapes.arrows}
\pgfplotsset{compat=1.11,
 /pgfplots/ybar legend/.style={
 /pgfplots/legend image code/.code={
 \draw[##1,/tikz/.cd,yshift=-0.25em]
 (0cm,0cm) rectangle (6pt,0.8em);},
 },
}
\usepackage{wrapfig}
\usepackage[plain]{algorithm}
\usepackage[noend]{algpseudocode}
\usepackage{amsthm}
\usepackage{appendix}
\let\appendixpagenameorig\appendixpagename
\renewcommand{\appendixpagename}{\sffamily\appendixpagenameorig}

\definecolor{lightseagreen42157143}{RGB}{42,157,143}
\definecolor{nice-purple}{HTML}{984EA3}
\definecolor{shadecolor}{RGB}{44,62,80}

\newcounter{theo}[section]
\renewcommand{\thetheo}{\arabic{theo}}
\newenvironment{theo}[2][]{%
\refstepcounter{theo}%
\ifstrempty{#1}%
{\mdfsetup{%
frametitle={%
\tikz[baseline=(current bounding box.east),outer sep=0pt]
\node[anchor=east,rectangle,fill=nice-purple!20]
{\strut Theorem~\thetheo};}}
}%
{\mdfsetup{%
frametitle={%
\tikz[baseline=(current bounding box.east),outer sep=0pt]
\node[anchor=east,rectangle,fill=nice-purple!20]
{\strut Theorem~\thetheo:~#1};}}%
}%
\mdfsetup{innertopmargin=10pt,linecolor=nice-purple!20,%
linewidth=2pt,topline=true,%
frametitleaboveskip=\dimexpr-\ht\strutbox\relax
}
\begin{mdframed}[]\relax%
\label{#2}}{\end{mdframed}}

\newcounter{defn}[section]
\renewcommand{\thedefn}{\arabic{defn}}
\newenvironment{defn}[2][]{%
\refstepcounter{defn}%
\ifstrempty{#1}%
{\mdfsetup{%
frametitle={%
\tikz[baseline=(current bounding box.east),outer sep=0pt]
\node[anchor=east,rectangle,fill=lightseagreen42157143!20]
{\strut Definition~\thedefn};}}
}%
{\mdfsetup{%
frametitle={%
\tikz[baseline=(current bounding box.east),outer sep=0pt]
\node[anchor=east,rectangle,fill=lightseagreen42157143!20]
{\strut Definition~\thedefn:~#1};}}%
}%
\mdfsetup{innertopmargin=10pt,linecolor=lightseagreen42157143!20,%
linewidth=2pt,topline=true,%
frametitleaboveskip=\dimexpr-\ht\strutbox\relax
}
\begin{mdframed}[]\relax%
\label{#2}}{\end{mdframed}}

\newtheoremstyle{mystyle}%
  {}%
  {}%
  {\itshape}%
  {}%
  {\bfseries}%
  {.}%
  { }%
  {\thmname{#1}\thmnumber{ #2}\thmnote{ (#3)}}%

\theoremstyle{mystyle}

%% file: sections/intro.tex
\section{Introduction}
\label{sec:intro}
\begin{figure}[!tb]
    \centering
    \begin{subfigure}[b]{0.44\textwidth}
        {\includegraphics[clip, trim=16cm 8cm 1cm 1cm,width=\textwidth]{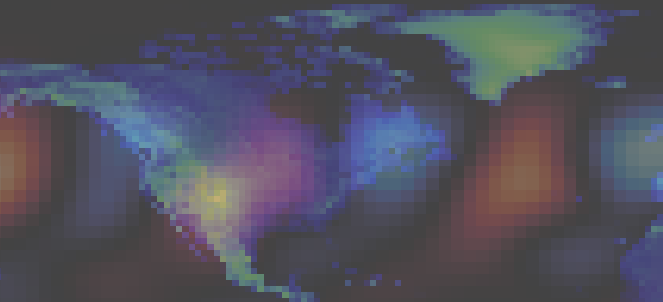}}\caption{Scalar pressure field}
    \end{subfigure}
    \begin{subfigure}[b]{0.44\textwidth}
        \includegraphics[clip, trim=16cm 8cm 1cm 1cm, width=\textwidth]{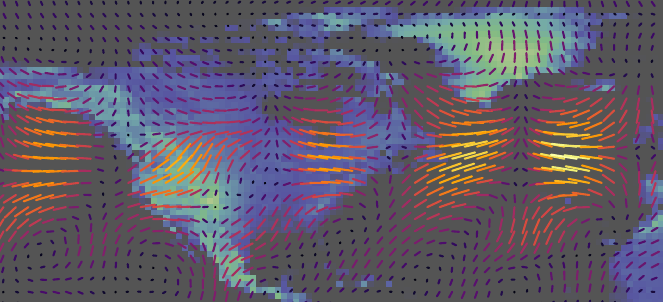}\caption{Vector wind velocity field}
    \end{subfigure}
    \caption{Fields of the Earth's shallow water model. Vector components of the wind velocities (right) are strongly related, i.e. they form a vector field. Additionally, the wind vector field and the scalar pressure field (left) are related since the gradient of the pressure field causes air movement and subsequently influences the wind components. We therefore aim to describe scalar and vector field as one multivector field, which models the dependencies correctly.}
    \label{fig:earth_shallow_water}
\end{figure}

Most scientific phenomena are described by the evolution and interaction of physical quantities over space and time.
The concept of fields is one widely used construct to continuously parameterize these quantities over chosen coordinates~\citep{mcmullin2002origins}.
Prominent examples include (i) fluid mechanics, which has applications in domains ranging from mechanical and civil engineering, to geophysics and meteorology, and (ii) electromagnetism, which provides mathematical models for electric, optical, or radio technologies.
The underlying equations of these examples are famously described in various forms of the Navier-Stokes equations and Maxwell's equations. 
For the majority of these equations, solutions are analytically intractable, and obtaining accurate predictions necessitates falling back on numerical approximation schemes often with prohibitive computation costs. 
Deep learning's success in various fields has led to a surge of interest in scientific applications, especially at augmenting and replacing numerical solving schemes in fluid dynamics with neural networks~\citep{li2020fourier, kochkov2021machine, lu2021learning, rasp2021data, keisler2022forecasting, weyn2020improving, sonderby2020metnet, pathak2022fourcastnet}. \\
Taking weather simulations as our motivating example to ground our discussion, two different kinds of fields emerge: \textbf{scalar} fields such as temperature or humidity, and \textbf{vector} fields such as wind velocity or pressure gradients.
Current deep learning based approaches treat different vector field components the same as scalar fields, and stack all scalar fields along the channel dimension, thereby omitting the geometric relations between different components, both within vector fields as well as between individual vector and scalar fields.
This practice leaves out important inductive bias information present in the input data. 
For example, wind velocities in the $x$- and $y$- directions are strongly related, i.e. they form a vector field.
Additionally, the wind vector field and the scalar pressure field are related since the gradient of the pressure field causes air movement and subsequently influences the wind components. 
In this work, we therefore build neural PDE surrogates which model the relation between different fields (e.g. wind and pressure field) and field components (e.g. $x$- and $y$- component of the wind velocities). 
Figure~\ref{fig:earth_shallow_water} shows an example of a wind vector field as per the Earth's shallow water model in two dimensions, and the related scalar pressure field.

\begin{figure}[!htb]
    \centering
    \resizebox{0.9\textwidth}{!}{%
    \input{tikz/multivector.tex}}
    \vspace*{-10mm}
    \caption{Multivector components of Clifford algebras.}
    \label{fig:multivector}
\end{figure}
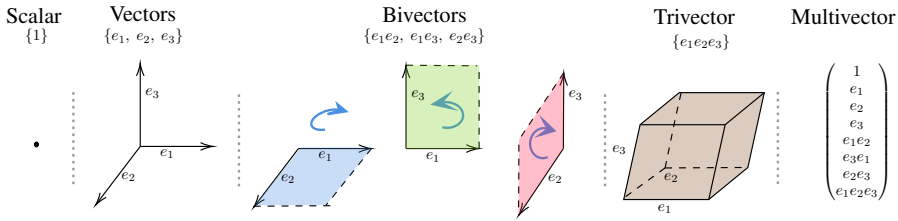

Clifford algebras~\citep{suter2003geometric, hestenes2003oersted, hestenes2012new, dorst2010geometric, renaud2020clifford} are at the core intersection of geometry and algebra, introduced to simplify spatial and geometrical relations between many mathematical concepts. For example, Clifford algebras naturally unify real numbers, vectors, complex numbers, quaternions, exterior algebras, and many more. 
Most notably, in contrast to standard vector analysis where primitives are scalars and vectors, Clifford algebras have additional spatial primitives for representing plane and volume segments. 
An expository example is the cross-product of two vectors in $3$ dimensions, which naturally translates to a plane segment spanned by these two vectors. The cross product is often represented as a vector due to its 3 independent components, but the cross product has a sign flip under reflection that a true vector does not.
In Clifford algebras, different spatial primitives can be summarized into objects called \textbf{multivectors}, as illustrated in Figure~\ref{fig:multivector}.
In this work, we replace operations over feature fields in deep learning architectures by their Clifford algebra counterparts, which operate on multivector feature fields. 
Operations on, and mappings between multivectors are defined by Clifford algebras.
For example, we will endow a convolutional kernel with multivector components, such that it can convolve over multivector feature maps. 

%% file: tikz/multivector.tex
\tikzset{every picture/.style={line width=0.75pt}} %

\begin{tikzpicture}[x=0.75pt,y=0.75pt,yscale=-1,xscale=1]
\draw    (113.55,131.4) -- (113.55,61.54) ;
\draw [shift={(113.55,59.54)}, rotate = 90] [color={rgb, 255:red, 0; green, 0; blue, 0 }  ][line width=0.75]    (10.93,-3.29) .. controls (6.95,-1.4) and (3.31,-0.3) .. (0,0) .. controls (3.31,0.3) and (6.95,1.4) .. (10.93,3.29)   ;
\draw    (113.55,131.4) -- (174.46,131.4) ;
\draw [shift={(176.46,131.4)}, rotate = 180] [color={rgb, 255:red, 0; green, 0; blue, 0 }  ][line width=0.75]    (10.93,-3.29) .. controls (6.95,-1.4) and (3.31,-0.3) .. (0,0) .. controls (3.31,0.3) and (6.95,1.4) .. (10.93,3.29)   ;
\draw    (113.55,131.4) -- (76.17,180.48) ;
\draw [shift={(74.96,182.07)}, rotate = 307.29] [color={rgb, 255:red, 0; green, 0; blue, 0 }  ][line width=0.75]    (10.93,-3.29) .. controls (6.95,-1.4) and (3.31,-0.3) .. (0,0) .. controls (3.31,0.3) and (6.95,1.4) .. (10.93,3.29)   ;
\draw    (249.94,133.24) -- (310.85,133.24) ;
\draw [shift={(312.85,133.24)}, rotate = 180] [color={rgb, 255:red, 0; green, 0; blue, 0 }  ][line width=0.75]    (10.93,-3.29) .. controls (6.95,-1.4) and (3.31,-0.3) .. (0,0) .. controls (3.31,0.3) and (6.95,1.4) .. (10.93,3.29)   ;
\draw    (249.94,133.24) -- (212.56,182.32) ;
\draw [shift={(211.35,183.91)}, rotate = 307.29] [color={rgb, 255:red, 0; green, 0; blue, 0 }  ][line width=0.75]    (10.93,-3.29) .. controls (6.95,-1.4) and (3.31,-0.3) .. (0,0) .. controls (3.31,0.3) and (6.95,1.4) .. (10.93,3.29)   ;
\draw    (342.72,133.24) -- (342.72,63.38) ;
\draw [shift={(342.72,61.38)}, rotate = 90] [color={rgb, 255:red, 0; green, 0; blue, 0 }  ][line width=0.75]    (10.93,-3.29) .. controls (6.95,-1.4) and (3.31,-0.3) .. (0,0) .. controls (3.31,0.3) and (6.95,1.4) .. (10.93,3.29)   ;
\draw    (342.72,133.24) -- (403.63,133.24) ;
\draw [shift={(405.63,133.24)}, rotate = 180] [color={rgb, 255:red, 0; green, 0; blue, 0 }  ][line width=0.75]    (10.93,-3.29) .. controls (6.95,-1.4) and (3.31,-0.3) .. (0,0) .. controls (3.31,0.3) and (6.95,1.4) .. (10.93,3.29)   ;
\draw    (478.32,133.24) -- (478.66,69.92) ;
\draw [shift={(478.67,67.92)}, rotate = 90.3] [color={rgb, 255:red, 0; green, 0; blue, 0 }  ][line width=0.75]    (10.93,-3.29) .. controls (6.95,-1.4) and (3.31,-0.3) .. (0,0) .. controls (3.31,0.3) and (6.95,1.4) .. (10.93,3.29)   ;
\draw    (478.32,133.24) -- (440.81,189.34) ;
\draw [shift={(439.7,191)}, rotate = 303.77] [color={rgb, 255:red, 0; green, 0; blue, 0 }  ][line width=0.75]    (10.93,-3.29) .. controls (6.95,-1.4) and (3.31,-0.3) .. (0,0) .. controls (3.31,0.3) and (6.95,1.4) .. (10.93,3.29)   ;
\draw  [dash pattern={on 4.5pt off 4.5pt}]  (211.35,182.99) -- (277.96,182.53) ;
\draw  [dash pattern={on 4.5pt off 4.5pt}]  (312.85,133.24) -- (273.74,182.07) ;
\draw  [dash pattern={on 4.5pt off 4.5pt}]  (342.72,61.38) -- (404.05,61.84) ;
\draw  [dash pattern={on 4.5pt off 4.5pt}]  (405.63,133.24) -- (405.63,61.84) ;
\draw  [dash pattern={on 4.5pt off 4.5pt}]  (478.67,67.92) -- (440.04,125.68) ;
\draw  [dash pattern={on 4.5pt off 4.5pt}]  (439.73,183.91) -- (440.04,125.68) ;
\draw  [fill={rgb, 255:red, 0; green, 0; blue, 0 }  ,fill opacity=1 ] (22.36,129.71) .. controls (22.36,128.61) and (23.12,127.71) .. (24.07,127.71) .. controls (25.02,127.71) and (25.79,128.61) .. (25.79,129.71) .. controls (25.79,130.81) and (25.02,131.71) .. (24.07,131.71) .. controls (23.12,131.71) and (22.36,130.81) .. (22.36,129.71) -- cycle ;
\draw  [draw opacity=0][line width=1.5]  (272.84,116.87) .. controls (271.74,117.03) and (270.65,117.12) .. (269.58,117.12) .. controls (262.32,117.12) and (259.04,113.2) .. (262.24,108.37) .. controls (265.44,103.54) and (273.92,99.62) .. (281.17,99.62) .. controls (282.24,99.62) and (283.22,99.7) .. (284.11,99.86) -- (275.38,108.37) -- cycle ; \draw  [color={rgb, 255:red, 74; green, 144; blue, 226 }  ,draw opacity=1 ][line width=1.5]  (272.84,116.87) .. controls (271.74,117.03) and (270.65,117.12) .. (269.58,117.12) .. controls (262.32,117.12) and (259.04,113.2) .. (262.24,108.37) .. controls (265.44,103.54) and (273.92,99.62) .. (281.17,99.62) .. controls (282.24,99.62) and (283.22,99.7) .. (284.11,99.86) ;  
\draw  [color={rgb, 255:red, 74; green, 144; blue, 226 }  ,draw opacity=1 ][fill={rgb, 255:red, 74; green, 144; blue, 226 }  ,fill opacity=1 ][line width=1.5]  (283.57,95.01) -- (290.99,101.12) -- (275.14,103.56) -- (285.17,100.2) -- cycle ;

\draw  [draw opacity=0][line width=1.5]  (464.86,140.53) .. controls (463.51,141.18) and (462.03,141.53) .. (460.49,141.53) .. controls (454.28,141.53) and (449.44,135.74) .. (449.67,128.59) .. controls (449.9,121.44) and (455.11,115.65) .. (461.32,115.65) .. controls (462.86,115.65) and (464.31,116.01) .. (465.63,116.65) -- (460.9,128.59) -- cycle ; \draw  [color={rgb, 255:red, 74; green, 144; blue, 226 }  ,draw opacity=1 ][line width=1.5]  (464.86,140.53) .. controls (463.51,141.18) and (462.03,141.53) .. (460.49,141.53) .. controls (454.28,141.53) and (449.44,135.74) .. (449.67,128.59) .. controls (449.9,121.44) and (455.11,115.65) .. (461.32,115.65) .. controls (462.86,115.65) and (464.31,116.01) .. (465.63,116.65) ;  
\draw  [color={rgb, 255:red, 74; green, 144; blue, 226 }  ,draw opacity=1 ][fill={rgb, 255:red, 74; green, 144; blue, 226 }  ,fill opacity=1 ][line width=1.5]  (460.97,108.83) -- (470.49,117.87) -- (458.2,121.48) -- (465.04,116.51) -- cycle ;

\draw  [draw opacity=0][line width=1.5]  (374.02,114.84) .. controls (375.25,115.14) and (376.54,115.31) .. (377.89,115.33) .. controls (385.67,115.46) and (391.94,110.62) .. (391.89,104.53) .. controls (391.84,98.43) and (385.49,93.37) .. (377.7,93.24) .. controls (376.36,93.22) and (375.07,93.34) .. (373.84,93.6) -- (377.79,104.28) -- cycle ; \draw  [color={rgb, 255:red, 74; green, 144; blue, 226 }  ,draw opacity=1 ][line width=1.5]  (374.02,114.84) .. controls (375.25,115.14) and (376.54,115.31) .. (377.89,115.33) .. controls (385.67,115.46) and (391.94,110.62) .. (391.89,104.53) .. controls (391.84,98.43) and (385.49,93.37) .. (377.7,93.24) .. controls (376.36,93.22) and (375.07,93.34) .. (373.84,93.6) ;  
\draw  [color={rgb, 255:red, 74; green, 144; blue, 226 }  ,draw opacity=1 ][fill={rgb, 255:red, 74; green, 144; blue, 226 }  ,fill opacity=1 ][line width=1.5]  (378.37,87.43) -- (366.13,94.94) -- (381.42,98.28) -- (373.01,93.9) -- cycle ;

\draw [color={rgb, 255:red, 155; green, 155; blue, 155 }  ,draw opacity=1 ][line width=1.5]  [dash pattern={on 1.69pt off 2.76pt}]  (55.92,84.56) -- (55.92,165.64) ;
\draw [color={rgb, 255:red, 155; green, 155; blue, 155 }  ,draw opacity=1 ][line width=1.5]  [dash pattern={on 1.69pt off 2.76pt}]  (198.66,86.41) -- (198.66,167.48) ;
\draw [color={rgb, 255:red, 155; green, 155; blue, 155 }  ,draw opacity=1 ][line width=1.5]  [dash pattern={on 1.69pt off 2.76pt}]  (514.27,86.41) -- (514.27,167.48) ;
\draw  [draw opacity=0][fill={rgb, 255:red, 74; green, 144; blue, 226 }  ,fill opacity=0.3 ] (252.12,132.32) -- (313.45,132.32) -- (273.74,182.07) -- (212.41,182.07) -- cycle ;
\draw  [draw opacity=0][fill={rgb, 255:red, 126; green, 211; blue, 33 }  ,fill opacity=0.3 ] (342.72,61.38) -- (405.63,61.38) -- (405.63,133.24) -- (342.72,133.24) -- cycle ;
\draw  [draw opacity=0][fill={rgb, 255:red, 255; green, 39; blue, 77 }  ,fill opacity=0.3 ] (439.7,191) -- (440.04,125.68) -- (478.67,67.92) -- (478.33,133.24) -- cycle ;

\draw [color={rgb, 255:red, 155; green, 155; blue, 155 }  ,draw opacity=1 ][line width=1.5]  [dash pattern={on 1.69pt off 2.76pt}]  (663.27,84.41) -- (663.27,165.48) ;
\draw  [fill={rgb, 255:red, 139; green, 87; blue, 42 }  ,fill opacity=0.3 ] (544.49,112.67) -- (578.22,85) -- (651.53,85) -- (637.4,149.56) -- (603.67,177.23) -- (530.36,177.23) -- cycle ; \draw   (651.53,85) -- (617.81,112.67) -- (544.49,112.67) ; \draw   (617.81,112.67) -- (603.67,177.23) ;
\draw  [dash pattern={on 4.5pt off 4.5pt}]  (566.03,149.96) -- (530.38,175.91) ;
\draw  [dash pattern={on 4.5pt off 4.5pt}]  (566.03,149.96) -- (637.39,149.56) ;
\draw  [dash pattern={on 4.5pt off 4.5pt}]  (566.03,149.96) -- (575.26,105.3) -- (579.89,85) ;
\draw  [color={rgb, 255:red, 255; green, 255; blue, 255 }  ,draw opacity=1 ] (0,10) -- (804,10) -- (804,255) -- (0,255) -- cycle ;

\draw (129.34,132.53) node [anchor=north west][inner sep=0.75pt]   [align=left] {$\displaystyle e_{1}$};
\draw (96.03,151.88) node [anchor=north west][inner sep=0.75pt]   [align=left] {$\displaystyle e_{2}$};
\draw (116.65,80.02) node [anchor=north west][inner sep=0.75pt]   [align=left] {$\displaystyle e_{3}$};
\draw (265.73,134.38) node [anchor=north west][inner sep=0.75pt]   [align=left] {$\displaystyle e_{1}$};
\draw (233.43,153.72) node [anchor=north west][inner sep=0.75pt]   [align=left] {$\displaystyle e_{2}$};
\draw (358.51,135.38) node [anchor=north west][inner sep=0.75pt]   [align=left] {$\displaystyle e_{1}$};
\draw (345.83,81.86) node [anchor=north west][inner sep=0.75pt]   [align=left] {$\displaystyle e_{3}$};
\draw (463.81,153.72) node [anchor=north west][inner sep=0.75pt]   [align=left] {$\displaystyle e_{2}$};
\draw (481.43,81.86) node [anchor=north west][inner sep=0.75pt]   [align=left] {$\displaystyle e_{3}$};
\draw (696,61) node [anchor=north west][inner sep=0.75pt]   [align=left] {$ $};
\draw (690.97,33.08) node [anchor=north west][inner sep=0.75pt]    {$$};
\draw (673.28,10.88) node [anchor=north west][inner sep=0.75pt]   [align=left] {{\Large Multivector}};
\draw (570.76,32.9) node [anchor=north west][inner sep=0.75pt]    {$\{e_{1} e_{2} e_{3}\}$};
\draw (554.86,11) node [anchor=north west][inner sep=0.75pt]   [align=left] {{\Large Trivector}};
\draw (12.84,28.37) node [anchor=north west][inner sep=0.75pt]    {$\{1\}$};
\draw (-3.68,10.08) node [anchor=north west][inner sep=0.75pt]   [align=left] {{\Large Scalar}};
\draw (80.58,28.3) node [anchor=north west][inner sep=0.75pt]    {$\{e_{1} ,\ e_{2} ,\ e_{3}\}$};
\draw (85.48,9.15) node [anchor=north west][inner sep=0.75pt]   [align=left] {{\Large Vectors}};
\draw (304.62,28.3) node [anchor=north west][inner sep=0.75pt]    {$\{e_{1} e_{2} ,\ e_{1} e_{3} ,\ e_{2} e_{3}\}$};
\draw (320.82,9.15) node [anchor=north west][inner sep=0.75pt]   [align=left] {{\Large Bivectors}};
\draw (558.41,179.47) node [anchor=north west][inner sep=0.75pt]   [align=left] {$\displaystyle e_{1}$};
\draw (563.77,152.89) node [anchor=north west][inner sep=0.75pt]   [align=left] {$\displaystyle e_{2}$};
\draw (519.09,128.77) node [anchor=north west][inner sep=0.75pt]   [align=left] {$\displaystyle e_{3}$};
\draw (700,58.4) node [anchor=north west][inner sep=0.75pt]    {$\begin{pmatrix}
1\\
e_{1}\\
e_{2}\\
e_{3}\\
e_{1} e_{2}\\
e_{3} e_{1}\\
e_{2} e_{3}\\
e_{1} e_{2} e_{3}
\end{pmatrix}$};

\end{tikzpicture}

%% file: sections/background.tex
\section{Background: Clifford algebras}\label{sec:background}
We introduce important mathematical concepts and discuss three Clifford algebras, $Cl_{2,0}(\R)$, $Cl_{0,2}(\R), Cl_{3,0}(\R)$, which we later use for the layers introduced in Section~\ref{sec:clifford_layers}. A more detailed introduction as well as connections to complex numbers and quaternions is given in Appendix~\ref{app:clifford_algebras}.

\textbf{Clifford algebras. } Consider the vector space $\R^n$ with standard Euclidean product $\langle.,. \rangle$, where $n=p+q$, and $p$ and $q$ are non-negative integers. A real Clifford algebra $Cl_{p,q}(\R)$ is an associative algebra\footnote{Operations of addition and multiplication are associative.} generated by $p + q$ orthonormal basis elements $e_1,\ldots, e_{p+q}$ of the \emph{generating} vector space $\R^n$, such that the following quadratic relations hold:
\begin{alignat}{2}
    e_i^2 &= +1 \ \mathrm{for} \  1 \leq i \leq p; \ \ \ \ \ e_j^2 &= -1 \  \mathrm{for} \  p < j \leq p+q; \ \ \ \ \ e_i e_j &= - e_j e_i \ \mathrm{for} \ \ i \neq j \ . \label{eq:quadratic_relations}
\end{alignat}

The pair $(p,q)$ is called the \textit{signature} and defines a Clifford algebra $Cl_{p,q}(\R)$, together with the basis elements that span the vector space $G^{p+q}$ of $Cl_{p,q}(\R)$.
Vector spaces of Clifford algebras have scalar elements and vector elements, but can also have elements consisting of multiple basis elements of the generating vector space $\R^n$, which can be interpreted as plane and volume segments.
Exemplary low-dimensional Clifford algebras are:
(i) $Cl_{0,0}(\R)$ which is a one-dimensional algebra that is spanned by the basis element $\{1\}$ and is therefore isomorphic to $\R$, the field of real numbers;
(ii) $Cl_{0,1}(\R)$ which is a two-dimensional algebra with vector space $G^1$ spanned by $\{1,e_1\}$ where the basis vector $e_1$ squares to $-1$, and is therefore isomorphic to $\CC$, the field of complex numbers; (iii) $Cl_{0,2}(\R)$ which is a 4-dimensional algebra with vector space $G^2$ spanned by $\{1,e_1,e_2,e_1e_2\}$, where $e_1,e_2,e_1e_2$ all square to $-1$ and anti-commute. Thus, $Cl_{0,2}(\R)$ is isomorphic to the quaternions $\HH$.

\textbf{Grade, dual, geometric product. }
The \emph{grade} of a Clifford algebra basis element
is the dimension of the subspace it represents.
For example, the basis elements $\{1,e_1,e_2,e_1e_2\}$ of the vector space $G^2$ of the Clifford algebra $Cl_{2,0}(\R)$ have the grades $\{0,1,1,2\}$.
Using the concept of grades, we can divide Clifford algebras into linear subspaces made up of elements of each grade. The grade subspace of
smallest dimension is $M_0$, the subspace of all scalars (elements with 0 basis vectors of the generating vector space).
Elements of $M_1$ are called vectors, elements of $M_2$ are bivectors, and so on. In general, a vector space $G^{p+q}$ of a Clifford algebra $Cl_{p,q}(\R)$ can be written as the direct sum of all of these subspaces:
$
    G^{p+q} = M_0 \oplus M_1 \oplus \ldots \oplus M_{p+q} \ .
$
The elements of a Clifford algebra are called \emph{multivectors}, containing elements of subspaces, i.e. scalars, vectors, bivectors, \ldots, $k$-vectors.
The basis element with the highest grade is called the \emph{pseudoscalar}\footnote{In contrast to scalars, pseudoscalars change sign under reflections.}, which in $\R^2$ corresponds to the bivector $e_1e_2$, and in $\R^3$ to the trivector $e_1e_2e_3$. \\
The \emph{dual} ${\va}^*$ of a multivector $\va$ is defined as
$
    \va^* = \va i_{p+q} \ ,
$
where $i_{p+q}$ represents the respective pseudoscalar of the Clifford algebra. 
This definition allows us to relate different multivectors to each other, which is a useful property when defining Clifford Fourier transforms. For example, for Clifford algebras in $\R^2$ the dual of the scalar is the bivector, and in $\R^3$, the dual of the scalar is the trivector. 
Finally, the \emph{geometric product} is a bilinear operation on multivectors. For arbitrary multivectors $\va$, $\vb$, $\vc \in G^{p+q}$, and scalar $\lambda$, the geometric product has the following properties:
(i) closure, i.e. $\va\vb \in G^{p+q}$, (ii) associativity, i.e. $(\va\vb)\vc = \va(\vb\vc)$; (iii) commutative scalar multiplication, i.e. $\lambda\va = \va\lambda$; (iv) distributive over addition, i.e. $ \va(\vb+\vc) = \va\vb + \va\vc$. 
The geometric product is in general non-commutative, i.e. $\va\vb \neq \vb\va$. Note that Equation~\ref{eq:quadratic_relations} describe the geometric product specifically between basis elements of the generating vector space.

\textbf{Clifford algebras $Cl_{2,0}(\R)$ and $Cl_{0,2}(\R)$. }
The 4-dimensional vector spaces of these Clifford algebras have the basis vectors $\{1,e_1,e_2,e_1e_2\}$ where
$e_1,e_2$ square to $+1$ for $Cl_{2,0}(\R)$ and to $-1$ for $Cl_{0,2}(\R)$. 
For $Cl_{2,0}(\R)$, the geometric product of two multivectors $\va = a_0 + a_1 e_1 + a_2 e_2 + a_{12}e_1e_2$ and $\vb = b_0 + b_1 e_1 + b_2 e_2 + b_{12}e_1e_2$ is given by:
\begin{align}
        \va \vb \ = & \ (a_0 b_0 + a_1 b_1 + a_2 b_2 - a_{12} b_{12})1 + (a_0 b_1 + a_1 b_0 - a_2 b_{12} + a_{12} b_2)e_1 \nonumber \\
    & + (a_0 b_2 + a_1 b_{12} + a_2 b_0 - a_{12} b_1)e_2 + (a_0 b_{12} + a_1 b_2 - a_2 b_1 + a_{12} b_0)e_1e_2 \ ,
\label{eq:geometric_product}
\end{align}
which can be derived by collecting terms that multiply the same basis elements, see Appendix~\ref{app:clifford_algebras}.
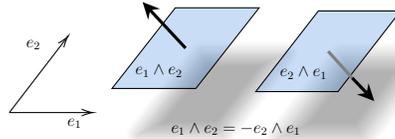
\begin{wrapfigure}[7]{r}{0.45\textwidth}
    \centering
    \resizebox{0.38\textwidth}{!}{%
    \input{tikz/bivector_wedgeprod.tex}}
    \caption{Antisymmetry of bivector exterior (wedge) product.}
    \label{fig:bivector}
\end{wrapfigure}
A vector $x= (x_1,x_2) \in \R^2$ with standard Euclidean product $\langle .,. \rangle$ can be related to $x_1e_1 + x_2e_2 \in \R^2 \subset G^2$. Clifford multiplication of two vectors $x, y \in \R^2 \subset G^2$ yields the geometric product $x y$:
\begin{align}
    x y &= (x_1e_1 + x_2e_2)(y_1e_1 + y_2e_2) \nonumber \\
    &= x_1 y_1 e_1^2 + x_2 y_2 e_2^2 + x_1 y_2 e_1e_2 + x_2 y_1 e_2e_1 \nonumber \\ 
    &=\langle x , y \rangle + x \wedge y \ \label{eq:geometric_product_vectors},
\end{align}
where $\wedge$ is the exterior or wedge product. The asymmetric quantity $x \wedge y = - y \wedge x$ is associated with the \textit{bivector}, which can be interpreted as an oriented plane segment as shown in Figure~\ref{fig:bivector}.
A unit bivector $i_2$, spanned by the (orthonormal) basis vectors $e_1$ and $e_2$ is determined by the product:
\begin{align}
    i_2 = e_1 e_2 = \underbrace{\langle e_1 , e_2 \rangle}_{=0} + \ e_1 \wedge e_2 = - \ e_2 \wedge e_1 = - \ e_2 e_1 \label{eq:bivector}\ ,
\end{align}
which if squared yields $i_2^2 = -1$. Thus, $i_2$ represents a geometric $\sqrt{-1}$. 
From Equation~\ref{eq:bivector}, it follows that 
$
e_2 = e_1 i_2 = -i_2 e_1
$ and
$
e_1 = i_2 e_2 = - e_2 i_2 \ .
$
Using the pseudoscalar $i_2$, the dual of a scalar is a bivector and the dual of a vector is again a vector. The dual pairs of the base vectors are $1 \leftrightarrow e_1 e_2$ and $e_1 \leftrightarrow e_2$. For $Cl_{2,0}(\R)$, these dual pairs allow us to write an arbitrary multivector $\va$ as
\begin{align}
    \va = a_0 + a_1 e_1 + a_2 e_2 + a_{12}e_{1}e_{2}= 1\underbrace{\big(a_0 + a_{12}i_2\big)}_{\text{spinor part}} + e_1\underbrace{\big(a_1 + a_2 i_2\big)}_{\text{vector part}} \ , \label{eq:dual_pairs}
\end{align}
which can be regarded as two complex-valued parts: the spinor\footnote{Spinors are elements of a complex vector space that can be associated with Euclidean space. Unlike vectors, spinors transform to their negative when rotated $360\degree$.} part, which commutes with the base element $1$, i.e. $1 i_2 = i_2 1$, and the vector part, which anti-commutes with the respective base element $\ve_1$, i.e. $e_1 i_2 = e_1e_1e_2 = - e_1 e_2 e_1 = -i_2 e_1$.
For $Cl(0,2)(\R)$, the vector part changes to $e_1\big(a_1 - a_2 i_2\big)$.
This decomposition will be the basis for Clifford Fourier transforms.

The Clifford algebra $Cl_{0,2}(\R)$ is 
isomorphic to the quaternions $\HH$, which are an extension of complex numbers and are commonly written in the literature as $a + b\hat{\imath} + c\hat{\jmath} + d\hat{k}$. Quaternions also form a 4-dimensional algebra spanned by $\{1, \hat{\imath}, \hat{\jmath}, \hat{k}\}$,
where $\hat{\imath}$, $\hat{\jmath}$, $\hat{k}$ all square to $-1$.
The algebra isomorphism to $Cl_{0,2}(\R)$ is easy to verify since $e_1,e_2,e_1e_2$ all square to $-1$ and anti-commute. 
The basis element $1$ is called the scalar and the basis elements $\hat{\imath}$, $\hat{\jmath}$, $\hat{k}$ are called the vector part of a quaternion.
Quaternions have practical uses in applied mathematics, particularly for expressing rotations, which we will use to define the rotational Clifford convolution layer in Section~\ref{sec:clifford_layers}.

\textbf{Clifford algebra $Cl_{3,0}(\R)$. }
The 8-dimensional vector space $G^3$ of the Clifford algebra $Cl_{3,0}(\R)$ has the basis vectors $\{1,e_1,e_2,e_3,e_1e_2, e_3e_1,e_2e_3,e_1e_2e_3\}$, i.e. it consists of one scalar, three vectors $\{e_1,e_2,e_3\}$, three bivectors $\{ e_1e_2,e_3e_1,e_2e_3\}$\footnote{The bivector $e_1e_3$ has negative orientation.}, and one trivector $e_1e_2e_3$. The trivector is the pseudoscalar $i_3$ of the algebra.
The geometric product of two multivectors is defined analogously to the geometric product of $Cl_{2,0}(\R)$, see Appendix~\ref{app:clifford_algebras}.
The dual pairs of $Cl_{3,0}(\R)$ are:
$1 \leftrightarrow e_1e_2e_3 = i_3$,
$e_1 \leftrightarrow e_2e_3$,
$e_2 \leftrightarrow e_3e_1$, and
$e_3 \leftrightarrow e_1e_2$.
An intriguing example of the duality of the multivectors of $Cl_{3,0}(\R)$ emerges when writing the expression of the electromagnetic field $\mF$ in terms of an electric vector field $E$ and a magnetic vector field $B$, such that
$
    \mF = E + B i_3 \ ,
$
where $E = E_x e_1 + E_y e_2 + E_z e_3$ and $B = B_x e_1 + B_y e_2 + B_z e_3$. 
In this way the electromagnetic field $\mF$ decomposes into electric vector and magnetic bivector parts via the pseudoscalar $i_3$~\citep{hestenes2003oersted}. For example, for the base component $B_x e_1$ of $B$ it holds that $B_x e_1 i_3= B_xe_1e_1e_2e_3 = B_xe_2e_3$ which is a bivector and the dual to the base component $E_x e_1$ of $E$.
Consequently, the multivector representing $\mF$ consists of three vectors (the electric field components) and three bivectors (the magnetic field components multiplied by $i_3$). This viewpoint gives Clifford neural layers a natural advantage over their default counterparts as we will see in Section~\ref{sec:experiments}.

%% file: tikz/bivector_wedgeprod.tex
\tikzset{every picture/.style={line width=0.75pt}} %

\begin{tikzpicture}[x=0.75pt,y=0.75pt,yscale=-1,xscale=1]
\draw  [fill={rgb, 255:red, 74; green, 144; blue, 226 }  ,fill opacity=0.3 ][blur shadow={shadow xshift=18.75pt,shadow yshift=-33pt, shadow blur radius=9pt, shadow blur steps=12 ,shadow opacity=30}] (364.54,81.17) -- (444.5,81.17) -- (384.12,158.17) -- (304.17,158.17) -- cycle ;
\draw  [fill={rgb, 255:red, 74; green, 144; blue, 226 }  ,fill opacity=0.3 ][blur shadow={shadow xshift=17.25pt,shadow yshift=-33pt, shadow blur radius=10.5pt, shadow blur steps=14 ,shadow opacity=30}] (500.97,84.17) -- (579.44,84.17) -- (519.7,161.17) -- (441.23,161.17) -- cycle ;
\draw [color={rgb, 255:red, 128; green, 128; blue, 128 }  ,draw opacity=1 ][line width=2.25]    (510.33,122.67) -- (533.33,148.17) ;
\draw [line width=2.25]    (533.33,148.17) -- (550.88,166.55) ;
\draw [shift={(554.33,170.17)}, rotate = 226.33] [fill={rgb, 255:red, 0; green, 0; blue, 0 }  ][line width=0.08]  [draw opacity=0] (16.07,-7.72) -- (0,0) -- (16.07,7.72) -- (10.67,0) -- cycle    ;
\draw [line width=2.25]    (374.33,119.67) -- (336.74,79.32) ;
\draw [shift={(333.33,75.67)}, rotate = 47.02] [fill={rgb, 255:red, 0; green, 0; blue, 0 }  ][line width=0.08]  [draw opacity=0] (16.07,-7.72) -- (0,0) -- (16.07,7.72) -- (10.67,0) -- cycle    ;
\draw    (206.23,181.17) -- (284.18,181.17) ;
\draw [shift={(286.18,181.17)}, rotate = 180] [color={rgb, 255:red, 0; green, 0; blue, 0 }  ][line width=0.75]    (10.93,-3.29) .. controls (6.95,-1.4) and (3.31,-0.3) .. (0,0) .. controls (3.31,0.3) and (6.95,1.4) .. (10.93,3.29)   ;
\draw    (206.23,181.17) -- (261.11,109.75) ;
\draw [shift={(262.33,108.17)}, rotate = 127.55] [color={rgb, 255:red, 0; green, 0; blue, 0 }  ][line width=0.75]    (10.93,-3.29) .. controls (6.95,-1.4) and (3.31,-0.3) .. (0,0) .. controls (3.31,0.3) and (6.95,1.4) .. (10.93,3.29)   ;

\draw (325,134.4) node [anchor=north west][inner sep=0.75pt]  [font=\large]  {$e_{1} \land e_{2}$};
\draw (463,136.4) node [anchor=north west][inner sep=0.75pt]  [font=\large]  {$e_{2} \land e_{1}$};
\draw (221,110.4) node [anchor=north west][inner sep=0.75pt]  [font=\large]  {$e_{2}$};
\draw (359,189.4) node [anchor=north west][inner sep=0.75pt]  [font=\large]  {$e_{1} \land e_{2} =-e_{2} \land e_{1}$};
\draw (260,185) node [anchor=north west][inner sep=0.75pt]  [font=\large] [align=left] {$\displaystyle e_{1}$};

\end{tikzpicture}

%% file: sections/approach.tex
\section{Clifford neural layers}\label{sec:clifford_layers}
Here, we introduce 2D Clifford convolution and 2D Clifford Fourier transform layers.
Appendix~\ref{app:clifford_layers} contains extensions to 3 dimensions. 
In Appendices~\ref{app:clifford_layers},~\ref{app:relatedwork}, related literature is discussed, most notably complex~\citep{bassey2021survey} and quaternion neural networks~\citep{parcollet2020survey}.

\textbf{Clifford CNN layers. }
Regular convolutional neural network (CNN) layers take as input feature maps $f: \sZ^2 \rightarrow \R^\cin$ and convolve\footnote{In deep learning, a convolution operation in the forward pass is implemented as cross-correlation.} them with a set of $\cout$ filters $\{w^i\}_{i=1}^{\cout}$ with $w^i: \sZ^2 \rightarrow \R^\cin$:
\begin{align}
    [f \star w^i ](x) = \sum_{y \in \sZ^2} \big\langle f(y), w^i (y-x)\big\rangle = \sum_{y \in \sZ^2}\sum_{j=1}^\cin f^j(y) w^{i,j} (y-x) \ ,
\end{align}
which can be interpreted as an inner product of input feature maps with the corresponding filters at every point $y \in \sZ^2$. By applying $\cout$ filters, the output feature maps can be interpreted as $\cout$ dimensional feature vectors at every point $y \in \sZ^2$. We now extend CNN layers such that the element-wise product of scalars $f^j(y) w^{i,j} (y-x)$ is replaced by the geometric product of multivector inputs and multivector filters $\vf^j(y)\vw^{i,j} (y-x)$,
where the chosen signature of $Cl$ is reflected in the geometric product.
We replace the feature maps  $f: \sZ^2 \rightarrow \R^\cin$ by multivector feature maps $\vf: \sZ^2 \rightarrow (G^2)^\cin$ and convolve them with a set of $\cout$ multivector filters $\{\vw^i\}_{i=1}^{\cout}: \sZ^2 \rightarrow (G^2)^\cin$:
\begin{align}
    \left[\vf \star \vw^i \right](x) & = \sum_{y \in \sZ^2}\sum_{j=1}^\cin \underbrace{\vf^j(y) \vw^{i,j} (y-x)}_{\vf^j\vw^{i,j} \ : \ G^2 \times G^2 \rightarrow G^2} \ .
    \label{eq:cliff_cnn}
\end{align}
\vspace{-0.25cm}
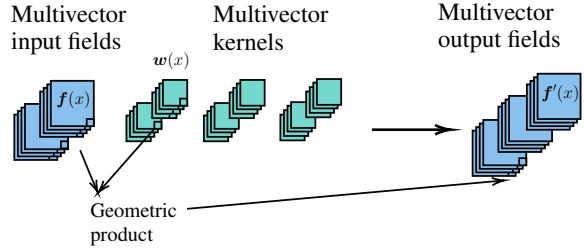
\begin{wrapfigure}[13]{r}{0.55\textwidth}
    \centering
    \resizebox{0.55\textwidth}{!}{%
    \input{tikz/clifford_convolution}}
    \caption{Sketch of Clifford convolution. Multivector input fields are convolved with multivector kernels.} 
    \label{fig:clifford_convolution}
\end{wrapfigure}
Note that each geometric product, indexed by $i\in \{1,...,\cout\}$ and $j\in\{1,...,\cin\}$, now results in a new multivector rather than a scalar.
Hence, the output of a layer is a grid of $\cout$ multivectors. We can e.g. implement a $Cl(2,0)(\R)$ Clifford CNN layer using Equation~\ref{eq:geometric_product} where $\{b_0$, $b_1$, $b_2$, $b_{12}\} \rightarrow \{w^{i,j}_0$, $w^{i,j}_1$, $w^{i,j}_2$, $w^{i,j}_{12}\}$ correspond to 4 different kernels representing one 2D multivector kernel, i.e. 4 different convolution layers, and $\{a_0$, $a_1$, $a_2$, $a_{12}\} \rightarrow \{f^j_0$, $f^j_1$, $f^j_2$, $f^j_{12}\}$ correspond to the scalar, vector and bivector parts of the input multivector field.
The channels of the different layers represent different stacks of scalars, vectors, and bivectors. Analogously, we can implement a $Cl(3,0)(\R)$ CNN layer using Equation~\ref{eq:app_geometric_product_3d} in Appendix~\ref{app:clifford_layers}. A schematic sketch of a Clifford convolution layer is shown in Figure~\ref{fig:clifford_convolution}.

\textbf{Rotational Clifford CNN layers. }
Here we introduce an alternative parameterization to the Clifford CNN layer introduced in Equation~\ref{eq:cliff_cnn} by using the isomorphism of the Clifford algebra $Cl_{0,2}(\R)$ to quaternions. We take advantage of the fact that a quaternion rotation can be realized by a matrix multiplication~\citep{jia2008quaternions, kuipers1999quaternions, schwichtenberg2015physics}. Using the isomorphism, we can represent the feature maps $\vf^j$ and filters $\vw^{i,j}$ as quaternions: $\vf^j$ = $f^j_0 + f^j_1 \hat{\imath} + f^j_2 \hat{\jmath} + f^j_{3} \hat{k}$ and $\vw^{i,j}$ = $w^{i,j}_0 + w^{i,j}_1 \hat{\imath} + w^{i,j}_2 \hat{\jmath} + w^{i,j}_{3} \hat{k}$\footnote{Note that the expansion coefficients for the feature map $\vf^j$ and filters $\vw^{i,j}$ in terms of the basis elements of $G^2$ and in terms of quaternion elements $\hat{\imath}$, $\hat{\jmath}$ and $\hat{k}$ are the same.}. We can now devise an alternative parameterization of the product between the feature map $\vf^j$ and $\vw^{i,j}$. To be more precise, we introduce a composite operation that results in a scalar quantity and a quaternion rotation, where the latter acts on the vector part of the quaternion $\vf^j$ and only produces nonzero expansion coefficients for the vector part of the quaternion output. A quaternion rotation
$\vw^{i,j} \vf^{j} (\vw^{i,j})^{-1}$ acts on the vector part ($\hat{\imath}, \hat{\jmath}, \hat{k}$) of $\vf^j$, and can be algebraically manipulated into a vector-matrix operation $\mR^{i,j} \vf^{j}$, where $\mR^{i,j}:\HH \rightarrow \HH$ is built up from the elements of $\vw^{i,j}$~\citep{kuipers1999quaternions}.
In other words, one can transform the vector part ($\hat{\imath}$, $\hat{\jmath}$, $\hat{k}$) of $\vf^j \in \HH$ via a rotation matrix $\mR^{i,j}$ that is built from
the scalar and vector part ($1, \hat{\imath}$, $\hat{\jmath}$, $\hat{k}$) of $\vw^{i,j} \in \HH$.
Altogether, a rotational multivector filter $\{\vw^{i}_{\text{rot}}\}_{i=1}^\cout:\sZ^2 \rightarrow {(G^2)}^\cin$ acts on the feature map $\vf^j$ through a rotational transformation $\mR^{i,j}(w^{i,j}_{\text{rot},0}, w^{i,j}_{\text{rot},1}, w^{i,j}_{\text{rot},2}, w^{i,j}_{\text{rot},12})$ acting on vector and bivector parts of the multivector feature map $\vf: \sZ^2 \rightarrow (G^2)^\cin$, and an additional scalar response of the multivector filters:
\begin{align}
\left[\vf \star \vw^{i}_{\text{rot}} \right](x)
=\sum_{y \in \sZ^2}\sum_{j=1}^\cin  \underbrace{\big[\vf^j(y) \vw^{i,j}_{\text{rot}}(y-x))\big]_{0}}_{\text{scalar output}} + \mR^{i,j}(y-x) \cdot 
\begin{pmatrix}
f^j_1(y) \\
f^j_2(y) \\
f^j_{12}(y)
\end{pmatrix} \ ,
\label{eq:rot_cliff_cnn}
\end{align}
where $\big[\vf^j(y) \vw^{i,j}_{\text{rot}}(y-x))\big]_0 = f^j_0 w^{i,j}_{\text{rot},0} - f^j_1 w^{i,j}_{\text{rot},1} - f^j_2 w^{i,j}_{\text{rot},2} - f^j_{12}w^{i,j}_{\text{rot},12}$\ , i.e., the scalar output of the geometric product of $Cl_{0,2}(\R)$ as in Equation~\ref{eq:geometric_product_cl02}.
A detailed description of the rotational multivector filters $\mR^{i,j}(y-x)$ is outlined in Appendix~\ref{app:clifford_layers}.
While in principle the Clifford CNN layer in Equation~\ref{eq:cliff_cnn} and the rotational Clifford CNN layer in Equation~\ref{eq:rot_cliff_cnn} are equally flexible, our experiments in Section \ref{sec:experiments} show that rotational Clifford CNN layers lead to better performance. \\
Clifford convolutions satisfy the property of equivariance under translation of the multivector inputs, as shown in theorem~\ref{th:app_translation_cliffordconv} in Appendix~\ref{app:clifford_layers}. Analogous to Theorem~\ref{th:app_translation_cliffordconv}, translation equivariance can be derived for rotational Clifford CNN layers. 

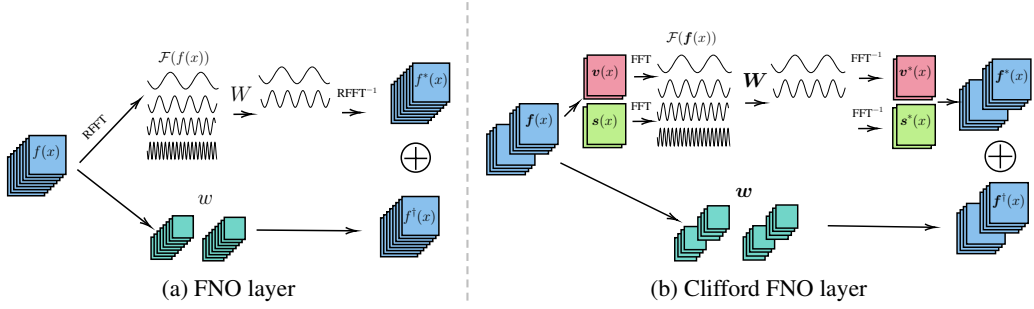
\begin{figure}[tb]
    \centering
    \begin{subfigure}[b]{0.44\textwidth}
    \centering
    \resizebox{0.97\textwidth}{!}{%
    \input{tikz/fourier_operator}} \caption{FNO layer}
    \label{fig:sketch_FNO}
    \end{subfigure}
    \tikz{\draw[-,gray!50, densely dashed, thick](0,2) -- (0,-2);}
    \begin{subfigure}[b]{0.54\textwidth}
    \centering
     \resizebox{0.97\textwidth}{!}{%
    \input{tikz/clifford_fourier}} \caption{Clifford FNO layer}
    \label{fig:sketch_CFNO}
    \end{subfigure}
    \caption{Sketch of Fourier Neural Operator (FNO) and Clifford Fourier Operator (CFNO) layers. The real valued Fast Fourier transform (RFFT) over real valued scalar input fields $f(x)$ is replaced by the complex Fast Fourier transform (FFT) over the complex valued dual parts $\vv(x)$ and $\vs(x)$ of multivector fields $\vf(x)$. Pointwise multiplication in the Fourier space via complex weight tensor $W$ is replaced by the geometric product in the Clifford Fourier space via multivector weight tensor $\mW$. Additionally, the convolution path is replaced by Clifford convolutions with multivector kernels $\vw$.}
\end{figure}

\textbf{Clifford Fourier layers. }
The discrete Fourier transform of an $n$-dimensional complex signal $f(x) = f(x_1,\ldots ,x_n) : \R^n \rightarrow \CC$ at $M_1 \times \ldots \times M_n$ grid points is defined as:
\begin{align}
    \displaystyle \gF\{f\}(\xi_1,\ldots,\xi_n) = \sum_{m_1 = 0}^{M_1} \ldots \sum_{m_n = 0}^{M_n} f(m_1,\ldots, m_n) \cdot e^{-2\pi i \cdot \big(\frac{m_1 \xi_1}{M_1} + \ldots + \frac{m_n \xi_n}{M_n} \big)} \ ,
    \label{eq:FFT}
\end{align}

where $(\xi_1, \ldots, \xi_n) \in \sZ_{M_1} \ldots \times \ldots \sZ_{M_n}$.
In Fourier Neural Operators (FNO)~\citep{li2020fourier},
discrete Fourier transforms on real-valued input fields and respective back-transforms -- implemented as Fast Fourier Transforms on real-valued inputs (RFFTs)\footnote{The FFT of a real-valued signal is Hermitian-symmetric, so the output contains only the positive frequencies below the Nyquist frequency for the last spatial dimension.} -- are interleaved with a weight multiplication by a complex weight matrix of shape $c_{\text{in}} \times c_{\text{out}}$ for each mode, which results in a complex-valued weight tensor of the form 
$W \in \CC^{c_{\text{in}} \times c_{\text{out}} \times (\xi^{\text{max}}_1 \times \ldots \times \xi^{\text{max}}_n)}$, where Fourier modes above cut-off frequencies $(\xi^{\text{max}}_1, \ldots, \xi^{\text{max}}_n)$ are set to zero.
Additionally, a residual connection is usually implemented as convolution layer with kernel size 1. In Figure~\ref{fig:sketch_FNO}, a sketch of an FNO layer is shown.
For $Cl(2,0)(\R)$, the Clifford Fourier transform~\citep{ebling2005, ebling2006visualization, hitzer2012clifford} for multivector valued functions $\vf(x): \R^2 \rightarrow G^2$ and vectors $x, \xi \in \R^2$ is defined as:
\begin{align}
    \hat{\vf}(\xi)= \gF\{\vf\}(\xi) = \frac{1}{2\pi}\int_{\R_{2}} \vf(x) e^{-2\pi i_2 \langle x, \xi\rangle} \ dx \ , \ \forall \xi \in \R^2 \ , \label{eq:clifford_ftransfrom} 
\end{align}
provided that the integral exists. In contrast to standard Fourier transforms, $\vf(x)$ and $\hat{\vf}(\xi)$ represent multivector fields in the spatial and the frequency domain, respectively. Furthermore, $i_2 = e_1e_2$ is used in the exponent.
Inserting the definition of multivector fields, we can rewrite Equation~\ref{eq:clifford_ftransfrom} as:
\begin{align}
     \gF\{\vf\}(\xi) 
     &= \frac{1}{2\pi} \int_{\R_{2}} \Bigg[ 1\bigg(\underbrace{f_0(x) + f_{12}(x)i_2}_{\text{spinor part}}\bigg) + e_1 \bigg(\underbrace{f_1(x) + f_2(x) i_2}_{\text{vector part}}\bigg)\Bigg] e^{-2\pi i_2 \langle x, \xi\rangle} \ dx \nonumber \\
     &= 1\Bigg[\gF\bigg(f_0(x) + f_{12}(x)i_2 \bigg)(\xi)\Bigg] + \ e_1 \Bigg[\gF\bigg(f_1(x) + f_{2}(x)i_2 \bigg)(\xi)\Bigg] \label{eq:clifford_ft_transform_vector_spinor} .
\end{align}
We obtain a Clifford Fourier transform by applying two standard Fourier transforms 
to the dual pairs $\vf_0 = f_0(x) + f_{12}(x)i_2$ and $\vf_1 = f_1(x) + f_{2}(x)i_2$,
which both can be treated as a complex-valued signals $\vf_0, \vf_1: \R^2 \rightarrow \CC$. Consequently, $\vf(x)$ can be understood as an element of $\CC^2$. The 2D Clifford Fourier transform is the linear combination of two classical Fourier transforms.
Discrete versions of 
Equation~\ref{eq:clifford_ft_transform_vector_spinor} are obtained analogously to Equation~\ref{eq:FFT}, see Appendix~\ref{app:clifford_layers}.
Similar to FNO, multivector weight tensors $\mW \in ({G^2})^{\cin \times \cout \times (\xi_1^{\text{max}} \times \xi_2^{\text{max}})}$ are applied,
where again Fourier modes above cut-off frequencies $(\xi^{\text{max}}_1, \xi^{\text{max}}_2)$ are set to zero.
In doing so, we point-wise modify the Clifford Fourier modes
$
    \hat{\vf}(\xi)= \gF\{\vf\}(\xi) = \hat{f}_0(\xi) + \hat{f}_1(\xi)e_1 + \hat{f}_2(\xi)e_2 + \hat{f}_{12}(\xi)e_{12}  
$
via the geometric product. The Clifford Fourier modes follow naturally when combining spinor and vector parts of Equation~\ref{eq:clifford_ft_transform_vector_spinor}.
Finally, the residual connection is replaced by a Clifford convolution with multivector kernel $\vk$. A schematic sketch is shown in Figure~\ref{fig:sketch_CFNO}. 
For $Cl(3,0)(\R)$, Clifford Fourier transforms follow a similar elegant construction, where we apply four separate Fourier transforms to %
\begin{equation}
\begin{aligned}
    &\vf_0(x)=f_0(x) +  f_{123}(x)i_3~
    &&\vf_1(x)=f_1(x) + f_{23}(x)i_3\\
    &\vf_2(x)=f_2(x) + f_{31}(x)i_3~
    &&\vf_3(x)=f_3(x) + f_{12}(x)i_3 \ ,
\end{aligned}
\end{equation}
i.e. scalar/trivector and vector/bivector components are combined into complex fields and then subjected to a Fourier transform. 

%% file: tikz/clifford_convolution.tex
\tikzset{every picture/.style={line width=0.75pt}} %

\begin{tikzpicture}[x=0.75pt,y=0.75pt,yscale=-1,xscale=1]
\draw  [color={rgb, 255:red, 0; green, 0; blue, 0 }  ,draw opacity=1 ][fill={rgb, 255:red, 137; green, 192; blue, 237 }  ,fill opacity=1 ][line width=1.5]  (11,149) -- (59.33,149) -- (59.33,200) -- (11,200) -- cycle ;
\draw  [color={rgb, 255:red, 0; green, 0; blue, 0 }  ,draw opacity=1 ][fill={rgb, 255:red, 137; green, 192; blue, 237 }  ,fill opacity=1 ][line width=1.5]  (15,145) -- (63.33,145) -- (63.33,196) -- (15,196) -- cycle ;
\draw  [color={rgb, 255:red, 0; green, 0; blue, 0 }  ,draw opacity=1 ][fill={rgb, 255:red, 137; green, 192; blue, 237 }  ,fill opacity=1 ][line width=1.5]  (19,141) -- (67.33,141) -- (67.33,192) -- (19,192) -- cycle ;
\draw  [color={rgb, 255:red, 0; green, 0; blue, 0 }  ,draw opacity=1 ][fill={rgb, 255:red, 137; green, 192; blue, 237 }  ,fill opacity=1 ][line width=1.5]  (23,137) -- (71.33,137) -- (71.33,188) -- (23,188) -- cycle ;
\draw  [color={rgb, 255:red, 0; green, 0; blue, 0 }  ,draw opacity=1 ][fill={rgb, 255:red, 137; green, 192; blue, 237 }  ,fill opacity=1 ][line width=1.5]  (40,123) -- (88.33,123) -- (88.33,174) -- (40,174) -- cycle ;
\draw  [color={rgb, 255:red, 0; green, 0; blue, 0 }  ,draw opacity=1 ][fill={rgb, 255:red, 137; green, 192; blue, 237 }  ,fill opacity=1 ][line width=1.5]  (44,119) -- (92.33,119) -- (92.33,170) -- (44,170) -- cycle ;
\draw  [color={rgb, 255:red, 0; green, 0; blue, 0 }  ,draw opacity=1 ][fill={rgb, 255:red, 137; green, 192; blue, 237 }  ,fill opacity=1 ][line width=1.5]  (48,115) -- (96.33,115) -- (96.33,166) -- (48,166) -- cycle ;
\draw  [color={rgb, 255:red, 0; green, 0; blue, 0 }  ,draw opacity=1 ][fill={rgb, 255:red, 137; green, 192; blue, 237 }  ,fill opacity=1 ][line width=1.5]  (52,111) -- (100.33,111) -- (100.33,162) -- (52,162) -- cycle ;
\draw  [color={rgb, 255:red, 0; green, 0; blue, 0 }  ,draw opacity=1 ][fill={rgb, 255:red, 117; green, 216; blue, 204 }  ,fill opacity=1 ][line width=1.5]  (137,150) -- (164.33,150) -- (164.33,179) -- (137,179) -- cycle ;
\draw  [color={rgb, 255:red, 0; green, 0; blue, 0 }  ,draw opacity=1 ][fill={rgb, 255:red, 117; green, 216; blue, 204 }  ,fill opacity=1 ][line width=1.5]  (141,145) -- (168.33,145) -- (168.33,174) -- (141,174) -- cycle ;
\draw  [color={rgb, 255:red, 0; green, 0; blue, 0 }  ,draw opacity=1 ][fill={rgb, 255:red, 117; green, 216; blue, 204 }  ,fill opacity=1 ][line width=1.5]  (145,140) -- (172.33,140) -- (172.33,169) -- (145,169) -- cycle ;
\draw  [color={rgb, 255:red, 0; green, 0; blue, 0 }  ,draw opacity=1 ][fill={rgb, 255:red, 117; green, 216; blue, 204 }  ,fill opacity=1 ][line width=1.5]  (149,136) -- (176.33,136) -- (176.33,165) -- (149,165) -- cycle ;
\draw  [color={rgb, 255:red, 0; green, 0; blue, 0 }  ,draw opacity=1 ][fill={rgb, 255:red, 117; green, 216; blue, 204 }  ,fill opacity=1 ][line width=1.5]  (166,124) -- (193.33,124) -- (193.33,153) -- (166,153) -- cycle ;
\draw  [color={rgb, 255:red, 0; green, 0; blue, 0 }  ,draw opacity=1 ][fill={rgb, 255:red, 117; green, 216; blue, 204 }  ,fill opacity=1 ][line width=1.5]  (170,119) -- (197.33,119) -- (197.33,148) -- (170,148) -- cycle ;
\draw  [color={rgb, 255:red, 0; green, 0; blue, 0 }  ,draw opacity=1 ][fill={rgb, 255:red, 117; green, 216; blue, 204 }  ,fill opacity=1 ][line width=1.5]  (174,114) -- (201.33,114) -- (201.33,143) -- (174,143) -- cycle ;
\draw  [color={rgb, 255:red, 0; green, 0; blue, 0 }  ,draw opacity=1 ][fill={rgb, 255:red, 117; green, 216; blue, 204 }  ,fill opacity=1 ][line width=1.5]  (178,110) -- (205.33,110) -- (205.33,139) -- (178,139) -- cycle ;
\draw  [color={rgb, 255:red, 0; green, 0; blue, 0 }  ,draw opacity=1 ][fill={rgb, 255:red, 117; green, 216; blue, 204 }  ,fill opacity=1 ][line width=1.5]  (224,150) -- (251.33,150) -- (251.33,179) -- (224,179) -- cycle ;
\draw  [color={rgb, 255:red, 0; green, 0; blue, 0 }  ,draw opacity=1 ][fill={rgb, 255:red, 117; green, 216; blue, 204 }  ,fill opacity=1 ][line width=1.5]  (228,145) -- (255.33,145) -- (255.33,174) -- (228,174) -- cycle ;
\draw  [color={rgb, 255:red, 0; green, 0; blue, 0 }  ,draw opacity=1 ][fill={rgb, 255:red, 117; green, 216; blue, 204 }  ,fill opacity=1 ][line width=1.5]  (232,140) -- (259.33,140) -- (259.33,169) -- (232,169) -- cycle ;
\draw  [color={rgb, 255:red, 0; green, 0; blue, 0 }  ,draw opacity=1 ][fill={rgb, 255:red, 117; green, 216; blue, 204 }  ,fill opacity=1 ][line width=1.5]  (236,136) -- (263.33,136) -- (263.33,165) -- (236,165) -- cycle ;
\draw  [color={rgb, 255:red, 0; green, 0; blue, 0 }  ,draw opacity=1 ][fill={rgb, 255:red, 117; green, 216; blue, 204 }  ,fill opacity=1 ][line width=1.5]  (253,124) -- (280.33,124) -- (280.33,153) -- (253,153) -- cycle ;
\draw  [color={rgb, 255:red, 0; green, 0; blue, 0 }  ,draw opacity=1 ][fill={rgb, 255:red, 117; green, 216; blue, 204 }  ,fill opacity=1 ][line width=1.5]  (257,119) -- (284.33,119) -- (284.33,148) -- (257,148) -- cycle ;
\draw  [color={rgb, 255:red, 0; green, 0; blue, 0 }  ,draw opacity=1 ][fill={rgb, 255:red, 117; green, 216; blue, 204 }  ,fill opacity=1 ][line width=1.5]  (261,114) -- (288.33,114) -- (288.33,143) -- (261,143) -- cycle ;
\draw  [color={rgb, 255:red, 0; green, 0; blue, 0 }  ,draw opacity=1 ][fill={rgb, 255:red, 117; green, 216; blue, 204 }  ,fill opacity=1 ][line width=1.5]  (265,110) -- (292.33,110) -- (292.33,139) -- (265,139) -- cycle ;
\draw  [color={rgb, 255:red, 0; green, 0; blue, 0 }  ,draw opacity=1 ][fill={rgb, 255:red, 117; green, 216; blue, 204 }  ,fill opacity=1 ][line width=1.5]  (310,149) -- (337.33,149) -- (337.33,178) -- (310,178) -- cycle ;
\draw  [color={rgb, 255:red, 0; green, 0; blue, 0 }  ,draw opacity=1 ][fill={rgb, 255:red, 117; green, 216; blue, 204 }  ,fill opacity=1 ][line width=1.5]  (314,144) -- (341.33,144) -- (341.33,173) -- (314,173) -- cycle ;
\draw  [color={rgb, 255:red, 0; green, 0; blue, 0 }  ,draw opacity=1 ][fill={rgb, 255:red, 117; green, 216; blue, 204 }  ,fill opacity=1 ][line width=1.5]  (318,139) -- (345.33,139) -- (345.33,168) -- (318,168) -- cycle ;
\draw  [color={rgb, 255:red, 0; green, 0; blue, 0 }  ,draw opacity=1 ][fill={rgb, 255:red, 117; green, 216; blue, 204 }  ,fill opacity=1 ][line width=1.5]  (322,135) -- (349.33,135) -- (349.33,164) -- (322,164) -- cycle ;
\draw  [color={rgb, 255:red, 0; green, 0; blue, 0 }  ,draw opacity=1 ][fill={rgb, 255:red, 117; green, 216; blue, 204 }  ,fill opacity=1 ][line width=1.5]  (339,123) -- (366.33,123) -- (366.33,152) -- (339,152) -- cycle ;
\draw  [color={rgb, 255:red, 0; green, 0; blue, 0 }  ,draw opacity=1 ][fill={rgb, 255:red, 117; green, 216; blue, 204 }  ,fill opacity=1 ][line width=1.5]  (343,118) -- (370.33,118) -- (370.33,147) -- (343,147) -- cycle ;
\draw  [color={rgb, 255:red, 0; green, 0; blue, 0 }  ,draw opacity=1 ][fill={rgb, 255:red, 117; green, 216; blue, 204 }  ,fill opacity=1 ][line width=1.5]  (347,113) -- (374.33,113) -- (374.33,142) -- (347,142) -- cycle ;
\draw  [color={rgb, 255:red, 0; green, 0; blue, 0 }  ,draw opacity=1 ][fill={rgb, 255:red, 117; green, 216; blue, 204 }  ,fill opacity=1 ][line width=1.5]  (351,109) -- (378.33,109) -- (378.33,138) -- (351,138) -- cycle ;
\draw  [color={rgb, 255:red, 0; green, 0; blue, 0 }  ,draw opacity=1 ][fill={rgb, 255:red, 137; green, 192; blue, 237 }  ,fill opacity=1 ][line width=1.5]  (526,167) -- (574.33,167) -- (574.33,218) -- (526,218) -- cycle ;
\draw  [color={rgb, 255:red, 0; green, 0; blue, 0 }  ,draw opacity=1 ][fill={rgb, 255:red, 137; green, 192; blue, 237 }  ,fill opacity=1 ][line width=1.5]  (530,163) -- (578.33,163) -- (578.33,214) -- (530,214) -- cycle ;
\draw  [color={rgb, 255:red, 0; green, 0; blue, 0 }  ,draw opacity=1 ][fill={rgb, 255:red, 137; green, 192; blue, 237 }  ,fill opacity=1 ][line width=1.5]  (534,159) -- (582.33,159) -- (582.33,210) -- (534,210) -- cycle ;
\draw  [color={rgb, 255:red, 0; green, 0; blue, 0 }  ,draw opacity=1 ][fill={rgb, 255:red, 137; green, 192; blue, 237 }  ,fill opacity=1 ][line width=1.5]  (538,155) -- (586.33,155) -- (586.33,206) -- (538,206) -- cycle ;
\draw  [color={rgb, 255:red, 0; green, 0; blue, 0 }  ,draw opacity=1 ][fill={rgb, 255:red, 137; green, 192; blue, 237 }  ,fill opacity=1 ][line width=1.5]  (555,140) -- (603.33,140) -- (603.33,191) -- (555,191) -- cycle ;
\draw  [color={rgb, 255:red, 0; green, 0; blue, 0 }  ,draw opacity=1 ][fill={rgb, 255:red, 137; green, 192; blue, 237 }  ,fill opacity=1 ][line width=1.5]  (559,136) -- (607.33,136) -- (607.33,187) -- (559,187) -- cycle ;
\draw  [color={rgb, 255:red, 0; green, 0; blue, 0 }  ,draw opacity=1 ][fill={rgb, 255:red, 137; green, 192; blue, 237 }  ,fill opacity=1 ][line width=1.5]  (563,132) -- (611.33,132) -- (611.33,183) -- (563,183) -- cycle ;
\draw  [color={rgb, 255:red, 0; green, 0; blue, 0 }  ,draw opacity=1 ][fill={rgb, 255:red, 137; green, 192; blue, 237 }  ,fill opacity=1 ][line width=1.5]  (567,128) -- (615.33,128) -- (615.33,179) -- (567,179) -- cycle ;
\draw  [color={rgb, 255:red, 0; green, 0; blue, 0 }  ,draw opacity=1 ][fill={rgb, 255:red, 137; green, 192; blue, 237 }  ,fill opacity=1 ][line width=1.5]  (587,115) -- (635.33,115) -- (635.33,166) -- (587,166) -- cycle ;
\draw  [color={rgb, 255:red, 0; green, 0; blue, 0 }  ,draw opacity=1 ][fill={rgb, 255:red, 137; green, 192; blue, 237 }  ,fill opacity=1 ][line width=1.5]  (591,111) -- (639.33,111) -- (639.33,162) -- (591,162) -- cycle ;
\draw  [color={rgb, 255:red, 0; green, 0; blue, 0 }  ,draw opacity=1 ][fill={rgb, 255:red, 137; green, 192; blue, 237 }  ,fill opacity=1 ][line width=1.5]  (595,107) -- (643.33,107) -- (643.33,158) -- (595,158) -- cycle ;
\draw  [color={rgb, 255:red, 0; green, 0; blue, 0 }  ,draw opacity=1 ][fill={rgb, 255:red, 137; green, 192; blue, 237 }  ,fill opacity=1 ][line width=1.5]  (599,103) -- (647.33,103) -- (647.33,154) -- (599,154) -- cycle ;
\draw  [fill={rgb, 255:red, 137; green, 192; blue, 237 }  ,fill opacity=1 ][line width=1.5]  (63.33,179) -- (71.33,179) -- (71.33,187) -- (63.33,187) -- cycle ;
\draw  [fill={rgb, 255:red, 117; green, 216; blue, 204 }  ,fill opacity=1 ][line width=1.5]  (168.33,157) -- (176.33,157) -- (176.33,165) -- (168.33,165) -- cycle ;
\draw [fill={rgb, 255:red, 137; green, 192; blue, 237 }  ,fill opacity=1 ][line width=1.5]    (85,189) -- (96,216) -- (103.82,234.24) ;
\draw [shift={(105,237)}, rotate = 246.8] [color={rgb, 255:red, 0; green, 0; blue, 0 }  ][line width=1.5]    (14.21,-4.28) .. controls (9.04,-1.82) and (4.3,-0.39) .. (0,0) .. controls (4.3,0.39) and (9.04,1.82) .. (14.21,4.28)   ;
\draw [fill={rgb, 255:red, 117; green, 216; blue, 204 }  ,fill opacity=1 ][line width=1.5]    (168.33,165) -- (106.98,234.75) ;
\draw [shift={(105,237)}, rotate = 311.34] [color={rgb, 255:red, 0; green, 0; blue, 0 }  ][line width=1.5]    (14.21,-4.28) .. controls (9.04,-1.82) and (4.3,-0.39) .. (0,0) .. controls (4.3,0.39) and (9.04,1.82) .. (14.21,4.28)   ;
\draw [fill={rgb, 255:red, 137; green, 192; blue, 237 }  ,fill opacity=1 ][line width=1.5]    (59.33,187) -- (59.33,191) ;
\draw [fill={rgb, 255:red, 137; green, 192; blue, 237 }  ,fill opacity=1 ][line width=1.5]    (55.33,192) -- (55.33,196) ;
\draw [fill={rgb, 255:red, 137; green, 192; blue, 237 }  ,fill opacity=1 ][line width=1.5]    (50.33,196) -- (50.33,200) ;
\draw [line width=2.25]    (413,166) -- (497.33,166) ;
\draw [shift={(501.33,166)}, rotate = 180] [color={rgb, 255:red, 0; green, 0; blue, 0 }  ][line width=2.25]    (17.49,-5.26) .. controls (11.12,-2.23) and (5.29,-0.48) .. (0,0) .. controls (5.29,0.48) and (11.12,2.23) .. (17.49,5.26)   ;
\draw  [fill={rgb, 255:red, 137; green, 192; blue, 237 }  ,fill opacity=1 ][line width=1.5]  (92.33,154) -- (100.33,154) -- (100.33,162) -- (92.33,162) -- cycle ;
\draw [fill={rgb, 255:red, 137; green, 192; blue, 237 }  ,fill opacity=1 ][line width=1.5]    (81.33,170) -- (81.33,174) ;
\draw [fill={rgb, 255:red, 137; green, 192; blue, 237 }  ,fill opacity=1 ][line width=1.5]    (90.33,162) -- (90.33,166) ;
\draw [fill={rgb, 255:red, 137; green, 192; blue, 237 }  ,fill opacity=1 ][line width=1.5]    (85.33,166) -- (85.33,170) ;
\draw  [fill={rgb, 255:red, 117; green, 216; blue, 204 }  ,fill opacity=1 ][line width=1.5]  (197.33,131) -- (205.33,131) -- (205.33,139) -- (197.33,139) -- cycle ;
\draw [fill={rgb, 255:red, 137; green, 192; blue, 237 }  ,fill opacity=1 ][line width=1.5]    (185.33,149) -- (185.33,153) ;
\draw [fill={rgb, 255:red, 137; green, 192; blue, 237 }  ,fill opacity=1 ][line width=1.5]    (189.33,144) -- (189.33,148) ;
\draw [fill={rgb, 255:red, 137; green, 192; blue, 237 }  ,fill opacity=1 ][line width=1.5]    (195.33,139) -- (195.33,143) ;
\draw [line width=1.5]    (205,255) -- (559.01,223.27) ;
\draw [shift={(562,223)}, rotate = 174.88] [color={rgb, 255:red, 0; green, 0; blue, 0 }  ][line width=1.5]    (14.21,-4.28) .. controls (9.04,-1.82) and (4.3,-0.39) .. (0,0) .. controls (4.3,0.39) and (9.04,1.82) .. (14.21,4.28)   ;
\draw  [fill={rgb, 255:red, 137; green, 192; blue, 237 }  ,fill opacity=1 ][line width=1.5]  (578.33,198) -- (586.33,198) -- (586.33,206) -- (578.33,206) -- cycle ;
\draw [fill={rgb, 255:red, 137; green, 192; blue, 237 }  ,fill opacity=1 ][line width=1.5]    (566.33,214) -- (566.33,218) ;
\draw [fill={rgb, 255:red, 137; green, 192; blue, 237 }  ,fill opacity=1 ][line width=1.5]    (571.33,211) -- (571.33,215) ;
\draw [fill={rgb, 255:red, 137; green, 192; blue, 237 }  ,fill opacity=1 ][line width=1.5]    (576.33,206) -- (576.33,210) ;

\draw (95,247) node [anchor=north west][inner sep=0.75pt]  [font=\LARGE] [align=left] {Geometric\\product};
\draw (6,24) node [anchor=north west][inner sep=0.75pt]  [font=\huge] [align=left] {Multivector \\input fields};
\draw (233,25) node [anchor=north west][inner sep=0.75pt]  [font=\huge] [align=left] {Multivector \\kernels};
\draw (486,22) node [anchor=north west][inner sep=0.75pt]  [font=\huge] [align=left] {Multivector \\output fields};
\draw (348,256) node [anchor=north west][inner sep=0.75pt]   [align=left] {$ $};
\draw (57,121) node [anchor=north west][inner sep=0.75pt]  [font=\Large] [align=left] {$\displaystyle \boldsymbol{f}( x)$};
\draw (165,77) node [anchor=north west][inner sep=0.75pt]  [font=\Large] [align=left] {$\displaystyle \boldsymbol{w}( x)$};
\draw (602,114) node [anchor=north west][inner sep=0.75pt]  [font=\Large] [align=left] {$\displaystyle \boldsymbol{f} '( x)$};

\end{tikzpicture}

%% file: tikz/fourier_operator.tex
\tikzset{every picture/.style={line width=0.75pt}} %

\begin{tikzpicture}[x=0.75pt,y=0.75pt,yscale=-1,xscale=1]
\draw  [color={rgb, 255:red, 0; green, 0; blue, 0 }  ,draw opacity=1 ][fill={rgb, 255:red, 137; green, 192; blue, 237 }  ,fill opacity=1 ][line width=1.5]  (16,172) -- (64.33,172) -- (64.33,223) -- (16,223) -- cycle ;
\draw  [color={rgb, 255:red, 0; green, 0; blue, 0 }  ,draw opacity=1 ][fill={rgb, 255:red, 137; green, 192; blue, 237 }  ,fill opacity=1 ][line width=1.5]  (20,168) -- (68.33,168) -- (68.33,219) -- (20,219) -- cycle ;
\draw  [color={rgb, 255:red, 0; green, 0; blue, 0 }  ,draw opacity=1 ][fill={rgb, 255:red, 137; green, 192; blue, 237 }  ,fill opacity=1 ][line width=1.5]  (24,164) -- (72.33,164) -- (72.33,215) -- (24,215) -- cycle ;
\draw  [color={rgb, 255:red, 0; green, 0; blue, 0 }  ,draw opacity=1 ][fill={rgb, 255:red, 137; green, 192; blue, 237 }  ,fill opacity=1 ][line width=1.5]  (28,160) -- (76.33,160) -- (76.33,211) -- (28,211) -- cycle ;
\draw  [color={rgb, 255:red, 0; green, 0; blue, 0 }  ,draw opacity=1 ][fill={rgb, 255:red, 137; green, 192; blue, 237 }  ,fill opacity=1 ][line width=1.5]  (32,156) -- (80.33,156) -- (80.33,207) -- (32,207) -- cycle ;
\draw  [color={rgb, 255:red, 0; green, 0; blue, 0 }  ,draw opacity=1 ][fill={rgb, 255:red, 137; green, 192; blue, 237 }  ,fill opacity=1 ][line width=1.5]  (36,152) -- (84.33,152) -- (84.33,203) -- (36,203) -- cycle ;
\draw  [color={rgb, 255:red, 0; green, 0; blue, 0 }  ,draw opacity=1 ][fill={rgb, 255:red, 137; green, 192; blue, 237 }  ,fill opacity=1 ][line width=1.5]  (40,148) -- (88.33,148) -- (88.33,199) -- (40,199) -- cycle ;
\draw  [color={rgb, 255:red, 0; green, 0; blue, 0 }  ,draw opacity=1 ][fill={rgb, 255:red, 137; green, 192; blue, 237 }  ,fill opacity=1 ][line width=1.5]  (44,144) -- (92.33,144) -- (92.33,195) -- (44,195) -- cycle ;
\draw [fill={rgb, 255:red, 113; green, 213; blue, 200 }  ,fill opacity=1 ][line width=1.5]    (107.33,188.67) -- (200.97,261.82) ;
\draw [shift={(203.33,263.67)}, rotate = 218] [color={rgb, 255:red, 0; green, 0; blue, 0 }  ][line width=1.5]    (14.21,-4.28) .. controls (9.04,-1.82) and (4.3,-0.39) .. (0,0) .. controls (4.3,0.39) and (9.04,1.82) .. (14.21,4.28)   ;
\draw [fill={rgb, 255:red, 74; green, 144; blue, 226 }  ,fill opacity=0.6 ][line width=1.5]    (110.33,179.67) -- (187.93,98.17) ;
\draw [shift={(190,96)}, rotate = 133.6] [color={rgb, 255:red, 0; green, 0; blue, 0 }  ][line width=1.5]    (14.21,-4.28) .. controls (9.04,-1.82) and (4.3,-0.39) .. (0,0) .. controls (4.3,0.39) and (9.04,1.82) .. (14.21,4.28)   ;
\draw [fill={rgb, 255:red, 144; green, 19; blue, 254 }  ,fill opacity=1 ][line width=1.5]    (312,116.17) -- (336.33,116) ;
\draw [shift={(336.33,116)}, rotate = 180] [color={rgb, 255:red, 0; green, 0; blue, 0 }  ][line width=1.5]    (14.21,-4.28) .. controls (9.04,-1.82) and (4.3,-0.39) .. (0,0) .. controls (4.3,0.39) and (9.04,1.82) .. (14.21,4.28)   ;
\draw   (199.33,72) .. controls (202.59,77.64) and (205.71,83) .. (209.33,83) .. controls (212.95,83) and (216.07,77.64) .. (219.33,72) .. controls (222.59,66.36) and (225.71,61) .. (229.33,61) .. controls (232.95,61) and (236.07,66.36) .. (239.33,72) .. controls (242.59,77.64) and (245.71,83) .. (249.33,83) .. controls (252.95,83) and (256.07,77.64) .. (259.33,72) .. controls (262.59,66.36) and (265.71,61) .. (269.33,61) .. controls (272.95,61) and (276.07,66.36) .. (279.33,72) .. controls (282.59,77.64) and (285.71,83) .. (289.33,83) .. controls (290.37,83) and (291.36,82.56) .. (292.33,81.8) ;
\draw   (200,103) .. controls (201.63,108.64) and (203.19,114) .. (205,114) .. controls (206.81,114) and (208.37,108.64) .. (210,103) .. controls (211.63,97.36) and (213.19,92) .. (215,92) .. controls (216.81,92) and (218.37,97.36) .. (220,103) .. controls (221.63,108.64) and (223.19,114) .. (225,114) .. controls (226.81,114) and (228.37,108.64) .. (230,103) .. controls (231.63,97.36) and (233.19,92) .. (235,92) .. controls (236.81,92) and (238.37,97.36) .. (240,103) .. controls (241.63,108.64) and (243.19,114) .. (245,114) .. controls (246.81,114) and (248.37,108.64) .. (250,103) .. controls (251.63,97.36) and (253.19,92) .. (255,92) .. controls (256.81,92) and (258.37,97.36) .. (260,103) .. controls (261.63,108.64) and (263.19,114) .. (265,114) .. controls (266.81,114) and (268.37,108.64) .. (270,103) .. controls (271.63,97.36) and (273.19,92) .. (275,92) .. controls (276.81,92) and (278.37,97.36) .. (280,103) .. controls (281.63,108.64) and (283.19,114) .. (285,114) .. controls (286.81,114) and (288.37,108.64) .. (290,103) .. controls (290.11,102.62) and (290.22,102.23) .. (290.33,101.85) ;
\draw   (199,132) .. controls (199.82,137.64) and (200.6,143) .. (201.5,143) .. controls (202.4,143) and (203.18,137.64) .. (204,132) .. controls (204.82,126.36) and (205.6,121) .. (206.5,121) .. controls (207.4,121) and (208.18,126.36) .. (209,132) .. controls (209.82,137.64) and (210.6,143) .. (211.5,143) .. controls (212.4,143) and (213.18,137.64) .. (214,132) .. controls (214.82,126.36) and (215.6,121) .. (216.5,121) .. controls (217.4,121) and (218.18,126.36) .. (219,132) .. controls (219.82,137.64) and (220.6,143) .. (221.5,143) .. controls (222.4,143) and (223.18,137.64) .. (224,132) .. controls (224.82,126.36) and (225.6,121) .. (226.5,121) .. controls (227.4,121) and (228.18,126.36) .. (229,132) .. controls (229.82,137.64) and (230.6,143) .. (231.5,143) .. controls (232.4,143) and (233.18,137.64) .. (234,132) .. controls (234.82,126.36) and (235.6,121) .. (236.5,121) .. controls (237.4,121) and (238.18,126.36) .. (239,132) .. controls (239.82,137.64) and (240.6,143) .. (241.5,143) .. controls (242.4,143) and (243.18,137.64) .. (244,132) .. controls (244.82,126.36) and (245.6,121) .. (246.5,121) .. controls (247.4,121) and (248.18,126.36) .. (249,132) .. controls (249.82,137.64) and (250.6,143) .. (251.5,143) .. controls (252.4,143) and (253.18,137.64) .. (254,132) .. controls (254.82,126.36) and (255.6,121) .. (256.5,121) .. controls (257.4,121) and (258.18,126.36) .. (259,132) .. controls (259.82,137.64) and (260.6,143) .. (261.5,143) .. controls (262.4,143) and (263.18,137.64) .. (264,132) .. controls (264.82,126.36) and (265.6,121) .. (266.5,121) .. controls (267.4,121) and (268.18,126.36) .. (269,132) .. controls (269.82,137.64) and (270.6,143) .. (271.5,143) .. controls (272.4,143) and (273.18,137.64) .. (274,132) .. controls (274.82,126.36) and (275.6,121) .. (276.5,121) .. controls (277.4,121) and (278.18,126.36) .. (279,132) .. controls (279.82,137.64) and (280.6,143) .. (281.5,143) .. controls (282.4,143) and (283.18,137.64) .. (284,132) .. controls (284.82,126.36) and (285.6,121) .. (286.5,121) .. controls (287.16,121) and (287.75,123.82) .. (288.33,127.53) ;
\draw   (199,164) .. controls (199.41,169.64) and (199.8,175) .. (200.25,175) .. controls (200.7,175) and (201.09,169.64) .. (201.5,164) .. controls (201.91,158.36) and (202.3,153) .. (202.75,153) .. controls (203.2,153) and (203.59,158.36) .. (204,164) .. controls (204.41,169.64) and (204.8,175) .. (205.25,175) .. controls (205.7,175) and (206.09,169.64) .. (206.5,164) .. controls (206.91,158.36) and (207.3,153) .. (207.75,153) .. controls (208.2,153) and (208.59,158.36) .. (209,164) .. controls (209.41,169.64) and (209.8,175) .. (210.25,175) .. controls (210.7,175) and (211.09,169.64) .. (211.5,164) .. controls (211.91,158.36) and (212.3,153) .. (212.75,153) .. controls (213.2,153) and (213.59,158.36) .. (214,164) .. controls (214.41,169.64) and (214.8,175) .. (215.25,175) .. controls (215.7,175) and (216.09,169.64) .. (216.5,164) .. controls (216.91,158.36) and (217.3,153) .. (217.75,153) .. controls (218.2,153) and (218.59,158.36) .. (219,164) .. controls (219.41,169.64) and (219.8,175) .. (220.25,175) .. controls (220.7,175) and (221.09,169.64) .. (221.5,164) .. controls (221.91,158.36) and (222.3,153) .. (222.75,153) .. controls (223.2,153) and (223.59,158.36) .. (224,164) .. controls (224.41,169.64) and (224.8,175) .. (225.25,175) .. controls (225.7,175) and (226.09,169.64) .. (226.5,164) .. controls (226.91,158.36) and (227.3,153) .. (227.75,153) .. controls (228.2,153) and (228.59,158.36) .. (229,164) .. controls (229.41,169.64) and (229.8,175) .. (230.25,175) .. controls (230.7,175) and (231.09,169.64) .. (231.5,164) .. controls (231.91,158.36) and (232.3,153) .. (232.75,153) .. controls (233.2,153) and (233.59,158.36) .. (234,164) .. controls (234.41,169.64) and (234.8,175) .. (235.25,175) .. controls (235.7,175) and (236.09,169.64) .. (236.5,164) .. controls (236.91,158.36) and (237.3,153) .. (237.75,153) .. controls (238.2,153) and (238.59,158.36) .. (239,164) .. controls (239.41,169.64) and (239.8,175) .. (240.25,175) .. controls (240.7,175) and (241.09,169.64) .. (241.5,164) .. controls (241.91,158.36) and (242.3,153) .. (242.75,153) .. controls (243.2,153) and (243.59,158.36) .. (244,164) .. controls (244.41,169.64) and (244.8,175) .. (245.25,175) .. controls (245.7,175) and (246.09,169.64) .. (246.5,164) .. controls (246.91,158.36) and (247.3,153) .. (247.75,153) .. controls (248.2,153) and (248.59,158.36) .. (249,164) .. controls (249.41,169.64) and (249.8,175) .. (250.25,175) .. controls (250.7,175) and (251.09,169.64) .. (251.5,164) .. controls (251.91,158.36) and (252.3,153) .. (252.75,153) .. controls (253.2,153) and (253.59,158.36) .. (254,164) .. controls (254.41,169.64) and (254.8,175) .. (255.25,175) .. controls (255.7,175) and (256.09,169.64) .. (256.5,164) .. controls (256.91,158.36) and (257.3,153) .. (257.75,153) .. controls (258.2,153) and (258.59,158.36) .. (259,164) .. controls (259.41,169.64) and (259.8,175) .. (260.25,175) .. controls (260.7,175) and (261.09,169.64) .. (261.5,164) .. controls (261.91,158.36) and (262.3,153) .. (262.75,153) .. controls (263.2,153) and (263.59,158.36) .. (264,164) .. controls (264.41,169.64) and (264.8,175) .. (265.25,175) .. controls (265.7,175) and (266.09,169.64) .. (266.5,164) .. controls (266.91,158.36) and (267.3,153) .. (267.75,153) .. controls (268.2,153) and (268.59,158.36) .. (269,164) .. controls (269.41,169.64) and (269.8,175) .. (270.25,175) .. controls (270.7,175) and (271.09,169.64) .. (271.5,164) .. controls (271.91,158.36) and (272.3,153) .. (272.75,153) .. controls (273.2,153) and (273.59,158.36) .. (274,164) .. controls (274.41,169.64) and (274.8,175) .. (275.25,175) .. controls (275.7,175) and (276.09,169.64) .. (276.5,164) .. controls (276.91,158.36) and (277.3,153) .. (277.75,153) .. controls (278.2,153) and (278.59,158.36) .. (279,164) .. controls (279.41,169.64) and (279.8,175) .. (280.25,175) .. controls (280.7,175) and (281.09,169.64) .. (281.5,164) .. controls (281.91,158.36) and (282.3,153) .. (282.75,153) .. controls (283.2,153) and (283.59,158.36) .. (284,164) .. controls (284.41,169.64) and (284.8,175) .. (285.25,175) .. controls (285.7,175) and (286.09,169.64) .. (286.5,164) .. controls (286.91,158.36) and (287.3,153) .. (287.75,153) .. controls (288.2,153) and (288.59,158.36) .. (289,164) .. controls (289.41,169.64) and (289.8,175) .. (290.25,175) .. controls (290.28,175) and (290.31,174.98) .. (290.33,174.94) ;
\draw   (345.33,65) .. controls (348.59,70.64) and (351.71,76) .. (355.33,76) .. controls (358.95,76) and (362.07,70.64) .. (365.33,65) .. controls (368.59,59.36) and (371.71,54) .. (375.33,54) .. controls (378.95,54) and (382.07,59.36) .. (385.33,65) .. controls (388.59,70.64) and (391.71,76) .. (395.33,76) .. controls (398.95,76) and (402.07,70.64) .. (405.33,65) .. controls (408.59,59.36) and (411.71,54) .. (415.33,54) .. controls (418.95,54) and (422.07,59.36) .. (425.33,65) .. controls (428.59,70.64) and (431.71,76) .. (435.33,76) .. controls (436.37,76) and (437.36,75.56) .. (438.33,74.8) ;
\draw   (348,96) .. controls (349.63,101.64) and (351.19,107) .. (353,107) .. controls (354.81,107) and (356.37,101.64) .. (358,96) .. controls (359.63,90.36) and (361.19,85) .. (363,85) .. controls (364.81,85) and (366.37,90.36) .. (368,96) .. controls (369.63,101.64) and (371.19,107) .. (373,107) .. controls (374.81,107) and (376.37,101.64) .. (378,96) .. controls (379.63,90.36) and (381.19,85) .. (383,85) .. controls (384.81,85) and (386.37,90.36) .. (388,96) .. controls (389.63,101.64) and (391.19,107) .. (393,107) .. controls (394.81,107) and (396.37,101.64) .. (398,96) .. controls (399.63,90.36) and (401.19,85) .. (403,85) .. controls (404.81,85) and (406.37,90.36) .. (408,96) .. controls (409.63,101.64) and (411.19,107) .. (413,107) .. controls (414.81,107) and (416.37,101.64) .. (418,96) .. controls (419.63,90.36) and (421.19,85) .. (423,85) .. controls (424.81,85) and (426.37,90.36) .. (428,96) .. controls (429.63,101.64) and (431.19,107) .. (433,107) .. controls (434.81,107) and (436.37,101.64) .. (438,96) .. controls (438.11,95.62) and (438.22,95.23) .. (438.33,94.85) ;
\draw [fill={rgb, 255:red, 74; green, 144; blue, 226 }  ,fill opacity=0.8 ][line width=1.5]    (454.33,114) -- (478.67,113.83) ;
\draw [shift={(478.67,113.83)}, rotate = 180] [color={rgb, 255:red, 0; green, 0; blue, 0 }  ][line width=1.5]    (14.21,-4.28) .. controls (9.04,-1.82) and (4.3,-0.39) .. (0,0) .. controls (4.3,0.39) and (9.04,1.82) .. (14.21,4.28)   ;
\draw  [color={rgb, 255:red, 0; green, 0; blue, 0 }  ,draw opacity=1 ][fill={rgb, 255:red, 137; green, 192; blue, 237 }  ,fill opacity=1 ][line width=1.5]  (519,77) -- (567.33,77) -- (567.33,128) -- (519,128) -- cycle ;
\draw  [color={rgb, 255:red, 0; green, 0; blue, 0 }  ,draw opacity=1 ][fill={rgb, 255:red, 137; green, 192; blue, 237 }  ,fill opacity=1 ][line width=1.5]  (523,73) -- (571.33,73) -- (571.33,124) -- (523,124) -- cycle ;
\draw  [color={rgb, 255:red, 0; green, 0; blue, 0 }  ,draw opacity=1 ][fill={rgb, 255:red, 137; green, 192; blue, 237 }  ,fill opacity=1 ][line width=1.5]  (527,69) -- (575.33,69) -- (575.33,120) -- (527,120) -- cycle ;
\draw  [color={rgb, 255:red, 0; green, 0; blue, 0 }  ,draw opacity=1 ][fill={rgb, 255:red, 137; green, 192; blue, 237 }  ,fill opacity=1 ][line width=1.5]  (531,65) -- (579.33,65) -- (579.33,116) -- (531,116) -- cycle ;
\draw  [color={rgb, 255:red, 0; green, 0; blue, 0 }  ,draw opacity=1 ][fill={rgb, 255:red, 137; green, 192; blue, 237 }  ,fill opacity=1 ][line width=1.5]  (535,61) -- (583.33,61) -- (583.33,112) -- (535,112) -- cycle ;
\draw  [color={rgb, 255:red, 0; green, 0; blue, 0 }  ,draw opacity=1 ][fill={rgb, 255:red, 137; green, 192; blue, 237 }  ,fill opacity=1 ][line width=1.5]  (539,57) -- (587.33,57) -- (587.33,108) -- (539,108) -- cycle ;
\draw  [color={rgb, 255:red, 0; green, 0; blue, 0 }  ,draw opacity=1 ][fill={rgb, 255:red, 137; green, 192; blue, 237 }  ,fill opacity=1 ][line width=1.5]  (543,53) -- (591.33,53) -- (591.33,104) -- (543,104) -- cycle ;
\draw  [color={rgb, 255:red, 0; green, 0; blue, 0 }  ,draw opacity=1 ][fill={rgb, 255:red, 137; green, 192; blue, 237 }  ,fill opacity=1 ][line width=1.5]  (547,49) -- (595.33,49) -- (595.33,100) -- (547,100) -- cycle ;
\draw  [line width=1.5]  (532,173) .. controls (532,162.51) and (540.51,154) .. (551,154) .. controls (561.49,154) and (570,162.51) .. (570,173) .. controls (570,183.49) and (561.49,192) .. (551,192) .. controls (540.51,192) and (532,183.49) .. (532,173) -- cycle ;
\draw  [line width=1.5]  (536,174) -- (566,174)(551,159) -- (551,189) ;
\draw  [color={rgb, 255:red, 0; green, 0; blue, 0 }  ,draw opacity=1 ][fill={rgb, 255:red, 113; green, 213; blue, 200 }  ,fill opacity=1 ][line width=1.5]  (204,277.33) -- (231.33,277.33) -- (231.33,306.33) -- (204,306.33) -- cycle ;
\draw  [color={rgb, 255:red, 0; green, 0; blue, 0 }  ,draw opacity=1 ][fill={rgb, 255:red, 113; green, 213; blue, 200 }  ,fill opacity=1 ][line width=1.5]  (208,273.33) -- (235.33,273.33) -- (235.33,302.33) -- (208,302.33) -- cycle ;
\draw  [color={rgb, 255:red, 0; green, 0; blue, 0 }  ,draw opacity=1 ][fill={rgb, 255:red, 113; green, 213; blue, 200 }  ,fill opacity=1 ][line width=1.5]  (212,269.33) -- (239.33,269.33) -- (239.33,298.33) -- (212,298.33) -- cycle ;
\draw  [color={rgb, 255:red, 0; green, 0; blue, 0 }  ,draw opacity=1 ][fill={rgb, 255:red, 113; green, 213; blue, 200 }  ,fill opacity=1 ][line width=1.5]  (216,265.33) -- (243.33,265.33) -- (243.33,294.33) -- (216,294.33) -- cycle ;
\draw  [color={rgb, 255:red, 0; green, 0; blue, 0 }  ,draw opacity=1 ][fill={rgb, 255:red, 113; green, 213; blue, 200 }  ,fill opacity=1 ][line width=1.5]  (220,261.33) -- (247.33,261.33) -- (247.33,290.33) -- (220,290.33) -- cycle ;
\draw  [color={rgb, 255:red, 0; green, 0; blue, 0 }  ,draw opacity=1 ][fill={rgb, 255:red, 113; green, 213; blue, 200 }  ,fill opacity=1 ][line width=1.5]  (224,257.33) -- (251.33,257.33) -- (251.33,286.33) -- (224,286.33) -- cycle ;
\draw  [color={rgb, 255:red, 0; green, 0; blue, 0 }  ,draw opacity=1 ][fill={rgb, 255:red, 113; green, 213; blue, 200 }  ,fill opacity=1 ][line width=1.5]  (228,253.33) -- (255.33,253.33) -- (255.33,282.33) -- (228,282.33) -- cycle ;
\draw  [color={rgb, 255:red, 0; green, 0; blue, 0 }  ,draw opacity=1 ][fill={rgb, 255:red, 113; green, 213; blue, 200 }  ,fill opacity=1 ][line width=1.5]  (232,249.33) -- (259.33,249.33) -- (259.33,278.33) -- (232,278.33) -- cycle ;
\draw [fill={rgb, 255:red, 137; green, 192; blue, 237 }  ,fill opacity=1 ][line width=1.5]    (342.33,268) -- (470.33,268.98) ;
\draw [shift={(473.33,269)}, rotate = 180.44] [color={rgb, 255:red, 0; green, 0; blue, 0 }  ][line width=1.5]    (14.21,-4.28) .. controls (9.04,-1.82) and (4.3,-0.39) .. (0,0) .. controls (4.3,0.39) and (9.04,1.82) .. (14.21,4.28)   ;
\draw  [color={rgb, 255:red, 0; green, 0; blue, 0 }  ,draw opacity=1 ][fill={rgb, 255:red, 137; green, 192; blue, 237 }  ,fill opacity=1 ][line width=1.5]  (505,252) -- (553.33,252) -- (553.33,303) -- (505,303) -- cycle ;
\draw  [color={rgb, 255:red, 0; green, 0; blue, 0 }  ,draw opacity=1 ][fill={rgb, 255:red, 137; green, 192; blue, 237 }  ,fill opacity=1 ][line width=1.5]  (509,248) -- (557.33,248) -- (557.33,299) -- (509,299) -- cycle ;
\draw  [color={rgb, 255:red, 0; green, 0; blue, 0 }  ,draw opacity=1 ][fill={rgb, 255:red, 137; green, 192; blue, 237 }  ,fill opacity=1 ][line width=1.5]  (513,244) -- (561.33,244) -- (561.33,295) -- (513,295) -- cycle ;
\draw  [color={rgb, 255:red, 0; green, 0; blue, 0 }  ,draw opacity=1 ][fill={rgb, 255:red, 137; green, 192; blue, 237 }  ,fill opacity=1 ][line width=1.5]  (517,240) -- (565.33,240) -- (565.33,291) -- (517,291) -- cycle ;
\draw  [color={rgb, 255:red, 0; green, 0; blue, 0 }  ,draw opacity=1 ][fill={rgb, 255:red, 137; green, 192; blue, 237 }  ,fill opacity=1 ][line width=1.5]  (521,236) -- (569.33,236) -- (569.33,287) -- (521,287) -- cycle ;
\draw  [color={rgb, 255:red, 0; green, 0; blue, 0 }  ,draw opacity=1 ][fill={rgb, 255:red, 137; green, 192; blue, 237 }  ,fill opacity=1 ][line width=1.5]  (525,232) -- (573.33,232) -- (573.33,283) -- (525,283) -- cycle ;
\draw  [color={rgb, 255:red, 0; green, 0; blue, 0 }  ,draw opacity=1 ][fill={rgb, 255:red, 137; green, 192; blue, 237 }  ,fill opacity=1 ][line width=1.5]  (529,228) -- (577.33,228) -- (577.33,279) -- (529,279) -- cycle ;
\draw  [color={rgb, 255:red, 0; green, 0; blue, 0 }  ,draw opacity=1 ][fill={rgb, 255:red, 137; green, 192; blue, 237 }  ,fill opacity=1 ][line width=1.5]  (533,224) -- (581.33,224) -- (581.33,275) -- (533,275) -- cycle ;
\draw  [color={rgb, 255:red, 0; green, 0; blue, 0 }  ,draw opacity=1 ][fill={rgb, 255:red, 113; green, 213; blue, 200 }  ,fill opacity=1 ][line width=1.5]  (272,279.33) -- (299.33,279.33) -- (299.33,308.33) -- (272,308.33) -- cycle ;
\draw  [color={rgb, 255:red, 0; green, 0; blue, 0 }  ,draw opacity=1 ][fill={rgb, 255:red, 113; green, 213; blue, 200 }  ,fill opacity=1 ][line width=1.5]  (276,275.33) -- (303.33,275.33) -- (303.33,304.33) -- (276,304.33) -- cycle ;
\draw  [color={rgb, 255:red, 0; green, 0; blue, 0 }  ,draw opacity=1 ][fill={rgb, 255:red, 113; green, 213; blue, 200 }  ,fill opacity=1 ][line width=1.5]  (280,271.33) -- (307.33,271.33) -- (307.33,300.33) -- (280,300.33) -- cycle ;
\draw  [color={rgb, 255:red, 0; green, 0; blue, 0 }  ,draw opacity=1 ][fill={rgb, 255:red, 113; green, 213; blue, 200 }  ,fill opacity=1 ][line width=1.5]  (284,267.33) -- (311.33,267.33) -- (311.33,296.33) -- (284,296.33) -- cycle ;
\draw  [color={rgb, 255:red, 0; green, 0; blue, 0 }  ,draw opacity=1 ][fill={rgb, 255:red, 113; green, 213; blue, 200 }  ,fill opacity=1 ][line width=1.5]  (288,263.33) -- (315.33,263.33) -- (315.33,292.33) -- (288,292.33) -- cycle ;
\draw  [color={rgb, 255:red, 0; green, 0; blue, 0 }  ,draw opacity=1 ][fill={rgb, 255:red, 113; green, 213; blue, 200 }  ,fill opacity=1 ][line width=1.5]  (292,259.33) -- (319.33,259.33) -- (319.33,288.33) -- (292,288.33) -- cycle ;
\draw  [color={rgb, 255:red, 0; green, 0; blue, 0 }  ,draw opacity=1 ][fill={rgb, 255:red, 113; green, 213; blue, 200 }  ,fill opacity=1 ][line width=1.5]  (296,255.33) -- (323.33,255.33) -- (323.33,284.33) -- (296,284.33) -- cycle ;
\draw  [color={rgb, 255:red, 0; green, 0; blue, 0 }  ,draw opacity=1 ][fill={rgb, 255:red, 113; green, 213; blue, 200 }  ,fill opacity=1 ][line width=1.5]  (300,251.33) -- (327.33,251.33) -- (327.33,280.33) -- (300,280.33) -- cycle ;

\draw (111,142.58) node [anchor=north west][inner sep=0.75pt]  [font=\large,rotate=-314.56] [align=left] {$\displaystyle \text{RFFT}$};
\draw (212.33,32) node [anchor=north west][inner sep=0.75pt]  [font=\Large] [align=left] {$\displaystyle \mathcal{F}( f( x))$};
\draw (446.4,78.86) node [anchor=north west][inner sep=0.75pt]  [font=\large,rotate=-0.4] [align=left] {$\displaystyle \text{RFFT}^{-1}$};
\draw (47,155) node [anchor=north west][inner sep=0.75pt]  [font=\Large] [align=left] {$\displaystyle f( x)$};
\draw (548,60) node [anchor=north west][inner sep=0.75pt]  [font=\Large] [align=left] {$\displaystyle f^{*}( x)$};
\draw (533,237) node [anchor=north west][inner sep=0.75pt]  [font=\Large] [align=left] {$\displaystyle f^{\dagger }( x)$};
\draw (306,78) node [anchor=north west][inner sep=0.75pt]  [font=\huge] [align=left] {$\displaystyle W$};
\draw (264,220) node [anchor=north west][inner sep=0.75pt]  [font=\huge] [align=left] {$\displaystyle w$};

\end{tikzpicture}

%% file: tikz/clifford_fourier.tex
\tikzset{every picture/.style={line width=0.75pt}} %

\begin{tikzpicture}[x=0.75pt,y=0.75pt,yscale=-1,xscale=1]
\draw  [color={rgb, 255:red, 0; green, 0; blue, 0 }  ,draw opacity=1 ][fill={rgb, 255:red, 137; green, 192; blue, 237 }  ,fill opacity=1 ][line width=1.5]  (0,220.2) -- (52.45,220.2) -- (52.45,275.63) -- (0,275.63) -- cycle ;
\draw  [color={rgb, 255:red, 0; green, 0; blue, 0 }  ,draw opacity=1 ][fill={rgb, 255:red, 137; green, 192; blue, 237 }  ,fill opacity=1 ][line width=1.5]  (4.34,215.85) -- (56.79,215.85) -- (56.79,271.28) -- (4.34,271.28) -- cycle ;
\draw  [color={rgb, 255:red, 0; green, 0; blue, 0 }  ,draw opacity=1 ][fill={rgb, 255:red, 137; green, 192; blue, 237 }  ,fill opacity=1 ][line width=1.5]  (8.68,211.5) -- (61.13,211.5) -- (61.13,266.93) -- (8.68,266.93) -- cycle ;
\draw  [color={rgb, 255:red, 0; green, 0; blue, 0 }  ,draw opacity=1 ][fill={rgb, 255:red, 137; green, 192; blue, 237 }  ,fill opacity=1 ][line width=1.5]  (13.02,207.16) -- (65.47,207.16) -- (65.47,262.58) -- (13.02,262.58) -- cycle ;
\draw  [color={rgb, 255:red, 0; green, 0; blue, 0 }  ,draw opacity=1 ][fill={rgb, 255:red, 137; green, 192; blue, 237 }  ,fill opacity=1 ][line width=1.5]  (31.47,191.94) -- (83.92,191.94) -- (83.92,247.37) -- (31.47,247.37) -- cycle ;
\draw  [color={rgb, 255:red, 0; green, 0; blue, 0 }  ,draw opacity=1 ][fill={rgb, 255:red, 137; green, 192; blue, 237 }  ,fill opacity=1 ][line width=1.5]  (35.81,187.59) -- (88.26,187.59) -- (88.26,243.02) -- (35.81,243.02) -- cycle ;
\draw  [color={rgb, 255:red, 0; green, 0; blue, 0 }  ,draw opacity=1 ][fill={rgb, 255:red, 137; green, 192; blue, 237 }  ,fill opacity=1 ][line width=1.5]  (40.15,183.25) -- (92.6,183.25) -- (92.6,238.67) -- (40.15,238.67) -- cycle ;
\draw  [color={rgb, 255:red, 0; green, 0; blue, 0 }  ,draw opacity=1 ][fill={rgb, 255:red, 137; green, 192; blue, 237 }  ,fill opacity=1 ][line width=1.5]  (44.49,178.9) -- (96.94,178.9) -- (96.94,234.33) -- (44.49,234.33) -- cycle ;
\draw  [color={rgb, 255:red, 0; green, 0; blue, 0 }  ,draw opacity=1 ][fill={rgb, 255:red, 190; green, 240; blue, 141 }  ,fill opacity=1 ][line width=1.5]  (131.31,183.25) -- (183.76,183.25) -- (183.76,238.67) -- (131.31,238.67) -- cycle ;
\draw  [color={rgb, 255:red, 0; green, 0; blue, 0 }  ,draw opacity=1 ][fill={rgb, 255:red, 190; green, 240; blue, 141 }  ,fill opacity=1 ][line width=1.5]  (136.74,178.9) -- (189.19,178.9) -- (189.19,234.33) -- (136.74,234.33) -- cycle ;
\draw  [color={rgb, 255:red, 0; green, 0; blue, 0 }  ,draw opacity=1 ][fill={rgb, 255:red, 240; green, 151; blue, 168 }  ,fill opacity=1 ][line width=1.5]  (132.39,116.95) -- (184.85,116.95) -- (184.85,172.38) -- (132.39,172.38) -- cycle ;
\draw  [color={rgb, 255:red, 0; green, 0; blue, 0 }  ,draw opacity=1 ][fill={rgb, 255:red, 240; green, 151; blue, 168 }  ,fill opacity=1 ][line width=1.5]  (136.74,112.6) -- (189.19,112.6) -- (189.19,168.03) -- (136.74,168.03) -- cycle ;
\draw  [color={rgb, 255:red, 0; green, 0; blue, 0 }  ,draw opacity=1 ][fill={rgb, 255:red, 190; green, 240; blue, 141 }  ,fill opacity=1 ][line width=1.5]  (552.9,184.33) -- (605.35,184.33) -- (605.35,239.76) -- (552.9,239.76) -- cycle ;
\draw  [color={rgb, 255:red, 0; green, 0; blue, 0 }  ,draw opacity=1 ][fill={rgb, 255:red, 190; green, 240; blue, 141 }  ,fill opacity=1 ][line width=1.5]  (558.32,179.99) -- (610.78,179.99) -- (610.78,235.41) -- (558.32,235.41) -- cycle ;
\draw  [color={rgb, 255:red, 0; green, 0; blue, 0 }  ,draw opacity=1 ][fill={rgb, 255:red, 240; green, 151; blue, 168 }  ,fill opacity=1 ][line width=1.5]  (553.98,118.04) -- (606.43,118.04) -- (606.43,173.47) -- (553.98,173.47) -- cycle ;
\draw  [color={rgb, 255:red, 0; green, 0; blue, 0 }  ,draw opacity=1 ][fill={rgb, 255:red, 240; green, 151; blue, 168 }  ,fill opacity=1 ][line width=1.5]  (558.32,113.69) -- (610.78,113.69) -- (610.78,169.12) -- (558.32,169.12) -- cycle ;
\draw  [color={rgb, 255:red, 0; green, 0; blue, 0 }  ,draw opacity=1 ][fill={rgb, 255:red, 137; green, 192; blue, 237 }  ,fill opacity=1 ][line width=1.5]  (644.06,159.34) -- (696.51,159.34) -- (696.51,214.76) -- (644.06,214.76) -- cycle ;
\draw  [color={rgb, 255:red, 0; green, 0; blue, 0 }  ,draw opacity=1 ][fill={rgb, 255:red, 137; green, 192; blue, 237 }  ,fill opacity=1 ][line width=1.5]  (648.4,154.99) -- (700.85,154.99) -- (700.85,210.42) -- (648.4,210.42) -- cycle ;
\draw  [color={rgb, 255:red, 0; green, 0; blue, 0 }  ,draw opacity=1 ][fill={rgb, 255:red, 137; green, 192; blue, 237 }  ,fill opacity=1 ][line width=1.5]  (652.74,150.64) -- (705.19,150.64) -- (705.19,206.07) -- (652.74,206.07) -- cycle ;
\draw  [color={rgb, 255:red, 0; green, 0; blue, 0 }  ,draw opacity=1 ][fill={rgb, 255:red, 137; green, 192; blue, 237 }  ,fill opacity=1 ][line width=1.5]  (657.08,146.3) -- (709.53,146.3) -- (709.53,201.72) -- (657.08,201.72) -- cycle ;
\draw  [color={rgb, 255:red, 0; green, 0; blue, 0 }  ,draw opacity=1 ][fill={rgb, 255:red, 137; green, 192; blue, 237 }  ,fill opacity=1 ][line width=1.5]  (675.53,131.08) -- (727.98,131.08) -- (727.98,186.51) -- (675.53,186.51) -- cycle ;
\draw  [color={rgb, 255:red, 0; green, 0; blue, 0 }  ,draw opacity=1 ][fill={rgb, 255:red, 137; green, 192; blue, 237 }  ,fill opacity=1 ][line width=1.5]  (679.87,126.73) -- (732.32,126.73) -- (732.32,182.16) -- (679.87,182.16) -- cycle ;
\draw  [color={rgb, 255:red, 0; green, 0; blue, 0 }  ,draw opacity=1 ][fill={rgb, 255:red, 137; green, 192; blue, 237 }  ,fill opacity=1 ][line width=1.5]  (684.21,122.39) -- (736.66,122.39) -- (736.66,177.81) -- (684.21,177.81) -- cycle ;
\draw  [color={rgb, 255:red, 0; green, 0; blue, 0 }  ,draw opacity=1 ][fill={rgb, 255:red, 137; green, 192; blue, 237 }  ,fill opacity=1 ][line width=1.5]  (688.55,118.04) -- (741,118.04) -- (741,173.47) -- (688.55,173.47) -- cycle ;
\draw [fill={rgb, 255:red, 144; green, 19; blue, 254 }  ,fill opacity=1 ][line width=1.5]    (200.59,142.13) -- (227,141.95) ;
\draw [shift={(227,141.95)}, rotate = 180] [color={rgb, 255:red, 0; green, 0; blue, 0 }  ][line width=1.5]    (14.21,-4.28) .. controls (9.04,-1.82) and (4.3,-0.39) .. (0,0) .. controls (4.3,0.39) and (9.04,1.82) .. (14.21,4.28)   ;
\draw [fill={rgb, 255:red, 144; green, 19; blue, 254 }  ,fill opacity=1 ][line width=1.5]    (351.84,177.99) -- (378.25,177.81) ;
\draw [shift={(378.25,177.81)}, rotate = 180] [color={rgb, 255:red, 0; green, 0; blue, 0 }  ][line width=1.5]    (14.21,-4.28) .. controls (9.04,-1.82) and (4.3,-0.39) .. (0,0) .. controls (4.3,0.39) and (9.04,1.82) .. (14.21,4.28)   ;
\draw [fill={rgb, 255:red, 144; green, 19; blue, 254 }  ,fill opacity=1 ][line width=1.5]    (506.83,144.3) -- (533.24,144.12) ;
\draw [shift={(533.24,144.12)}, rotate = 180] [color={rgb, 255:red, 0; green, 0; blue, 0 }  ][line width=1.5]    (14.21,-4.28) .. controls (9.04,-1.82) and (4.3,-0.39) .. (0,0) .. controls (4.3,0.39) and (9.04,1.82) .. (14.21,4.28)   ;
\draw [fill={rgb, 255:red, 137; green, 192; blue, 237 }  ,fill opacity=1 ][line width=1.5]    (614.03,176.73) -- (640.44,176.54) ;
\draw [shift={(640.44,176.54)}, rotate = 180] [color={rgb, 255:red, 0; green, 0; blue, 0 }  ][line width=1.5]    (14.21,-4.28) .. controls (9.04,-1.82) and (4.3,-0.39) .. (0,0) .. controls (4.3,0.39) and (9.04,1.82) .. (14.21,4.28)   ;
\draw [fill={rgb, 255:red, 144; green, 19; blue, 254 }  ,fill opacity=1 ][line width=1.5]    (198.42,202.99) -- (224.83,202.81) ;
\draw [shift={(224.83,202.81)}, rotate = 180] [color={rgb, 255:red, 0; green, 0; blue, 0 }  ][line width=1.5]    (14.21,-4.28) .. controls (9.04,-1.82) and (4.3,-0.39) .. (0,0) .. controls (4.3,0.39) and (9.04,1.82) .. (14.21,4.28)   ;
\draw [fill={rgb, 255:red, 144; green, 19; blue, 254 }  ,fill opacity=1 ][line width=1.5]    (505.75,212.77) -- (532.15,212.59) ;
\draw [shift={(532.15,212.59)}, rotate = 180] [color={rgb, 255:red, 0; green, 0; blue, 0 }  ][line width=1.5]    (14.21,-4.28) .. controls (9.04,-1.82) and (4.3,-0.39) .. (0,0) .. controls (4.3,0.39) and (9.04,1.82) .. (14.21,4.28)   ;
\draw   (232.6,124.56) .. controls (236.13,130.68) and (239.52,136.51) .. (243.45,136.51) .. controls (247.37,136.51) and (250.76,130.68) .. (254.3,124.56) .. controls (257.84,118.43) and (261.22,112.6) .. (265.15,112.6) .. controls (269.08,112.6) and (272.46,118.43) .. (276,124.56) .. controls (279.54,130.68) and (282.93,136.51) .. (286.86,136.51) .. controls (290.78,136.51) and (294.17,130.68) .. (297.71,124.56) .. controls (301.25,118.43) and (304.63,112.6) .. (308.56,112.6) .. controls (312.49,112.6) and (315.87,118.43) .. (319.41,124.56) .. controls (322.95,130.68) and (326.34,136.51) .. (330.26,136.51) .. controls (331.39,136.51) and (332.47,136.04) .. (333.52,135.21) ;
\draw   (233.32,158.25) .. controls (235.09,164.37) and (236.78,170.2) .. (238.74,170.2) .. controls (240.71,170.2) and (242.4,164.37) .. (244.17,158.25) .. controls (245.94,152.13) and (247.63,146.3) .. (249.6,146.3) .. controls (251.56,146.3) and (253.25,152.13) .. (255.02,158.25) .. controls (256.79,164.37) and (258.49,170.2) .. (260.45,170.2) .. controls (262.41,170.2) and (264.11,164.37) .. (265.88,158.25) .. controls (267.64,152.13) and (269.34,146.3) .. (271.3,146.3) .. controls (273.26,146.3) and (274.96,152.13) .. (276.73,158.25) .. controls (278.5,164.37) and (280.19,170.2) .. (282.15,170.2) .. controls (284.12,170.2) and (285.81,164.37) .. (287.58,158.25) .. controls (289.35,152.13) and (291.04,146.3) .. (293.01,146.3) .. controls (294.97,146.3) and (296.66,152.13) .. (298.43,158.25) .. controls (300.2,164.37) and (301.89,170.2) .. (303.86,170.2) .. controls (305.82,170.2) and (307.51,164.37) .. (309.28,158.25) .. controls (311.05,152.13) and (312.75,146.3) .. (314.71,146.3) .. controls (316.67,146.3) and (318.37,152.13) .. (320.14,158.25) .. controls (321.9,164.37) and (323.6,170.2) .. (325.56,170.2) .. controls (327.52,170.2) and (329.22,164.37) .. (330.99,158.25) .. controls (331.11,157.83) and (331.23,157.41) .. (331.35,157) ;
\draw   (232.23,189.77) .. controls (233.12,195.89) and (233.96,201.72) .. (234.95,201.72) .. controls (235.93,201.72) and (236.77,195.89) .. (237.66,189.77) .. controls (238.54,183.64) and (239.39,177.81) .. (240.37,177.81) .. controls (241.35,177.81) and (242.2,183.64) .. (243.09,189.77) .. controls (243.97,195.89) and (244.82,201.72) .. (245.8,201.72) .. controls (246.78,201.72) and (247.63,195.89) .. (248.51,189.77) .. controls (249.4,183.64) and (250.24,177.81) .. (251.22,177.81) .. controls (252.21,177.81) and (253.05,183.64) .. (253.94,189.77) .. controls (254.82,195.89) and (255.67,201.72) .. (256.65,201.72) .. controls (257.63,201.72) and (258.48,195.89) .. (259.36,189.77) .. controls (260.25,183.64) and (261.09,177.81) .. (262.08,177.81) .. controls (263.06,177.81) and (263.91,183.64) .. (264.79,189.77) .. controls (265.67,195.89) and (266.52,201.72) .. (267.5,201.72) .. controls (268.48,201.72) and (269.33,195.89) .. (270.22,189.77) .. controls (271.1,183.64) and (271.95,177.81) .. (272.93,177.81) .. controls (273.91,177.81) and (274.76,183.64) .. (275.64,189.77) .. controls (276.53,195.89) and (277.37,201.72) .. (278.35,201.72) .. controls (279.34,201.72) and (280.18,195.89) .. (281.07,189.77) .. controls (281.95,183.64) and (282.8,177.81) .. (283.78,177.81) .. controls (284.76,177.81) and (285.61,183.64) .. (286.49,189.77) .. controls (287.38,195.89) and (288.23,201.72) .. (289.21,201.72) .. controls (290.19,201.72) and (291.04,195.89) .. (291.92,189.77) .. controls (292.8,183.64) and (293.65,177.81) .. (294.63,177.81) .. controls (295.61,177.81) and (296.46,183.64) .. (297.35,189.77) .. controls (298.23,195.89) and (299.08,201.72) .. (300.06,201.72) .. controls (301.04,201.72) and (301.89,195.89) .. (302.77,189.77) .. controls (303.66,183.64) and (304.5,177.81) .. (305.48,177.81) .. controls (306.47,177.81) and (307.31,183.64) .. (308.2,189.77) .. controls (309.08,195.89) and (309.93,201.72) .. (310.91,201.72) .. controls (311.89,201.72) and (312.74,195.89) .. (313.62,189.77) .. controls (314.51,183.64) and (315.36,177.81) .. (316.34,177.81) .. controls (317.32,177.81) and (318.17,183.64) .. (319.05,189.77) .. controls (319.93,195.89) and (320.78,201.72) .. (321.76,201.72) .. controls (322.74,201.72) and (323.59,195.89) .. (324.48,189.77) .. controls (325.36,183.64) and (326.21,177.81) .. (327.19,177.81) .. controls (327.9,177.81) and (328.54,180.88) .. (329.18,184.91) ;
\draw   (232.23,224.55) .. controls (232.68,230.67) and (233.1,236.5) .. (233.59,236.5) .. controls (234.08,236.5) and (234.5,230.67) .. (234.95,224.55) .. controls (235.39,218.42) and (235.81,212.59) .. (236.3,212.59) .. controls (236.79,212.59) and (237.22,218.42) .. (237.66,224.55) .. controls (238.1,230.67) and (238.53,236.5) .. (239.02,236.5) .. controls (239.51,236.5) and (239.93,230.67) .. (240.37,224.55) .. controls (240.82,218.42) and (241.24,212.59) .. (241.73,212.59) .. controls (242.22,212.59) and (242.64,218.42) .. (243.09,224.55) .. controls (243.53,230.67) and (243.95,236.5) .. (244.44,236.5) .. controls (244.93,236.5) and (245.36,230.67) .. (245.8,224.55) .. controls (246.24,218.42) and (246.66,212.59) .. (247.16,212.59) .. controls (247.65,212.59) and (248.07,218.42) .. (248.51,224.55) .. controls (248.95,230.67) and (249.38,236.5) .. (249.87,236.5) .. controls (250.36,236.5) and (250.78,230.67) .. (251.22,224.55) .. controls (251.67,218.42) and (252.09,212.59) .. (252.58,212.59) .. controls (253.07,212.59) and (253.5,218.42) .. (253.94,224.55) .. controls (254.38,230.67) and (254.8,236.5) .. (255.29,236.5) .. controls (255.79,236.5) and (256.21,230.67) .. (256.65,224.55) .. controls (257.09,218.42) and (257.52,212.59) .. (258.01,212.59) .. controls (258.5,212.59) and (258.92,218.42) .. (259.36,224.55) .. controls (259.81,230.67) and (260.23,236.5) .. (260.72,236.5) .. controls (261.21,236.5) and (261.63,230.67) .. (262.08,224.55) .. controls (262.52,218.42) and (262.94,212.59) .. (263.43,212.59) .. controls (263.92,212.59) and (264.35,218.42) .. (264.79,224.55) .. controls (265.23,230.67) and (265.66,236.5) .. (266.15,236.5) .. controls (266.64,236.5) and (267.06,230.67) .. (267.5,224.55) .. controls (267.95,218.42) and (268.37,212.59) .. (268.86,212.59) .. controls (269.35,212.59) and (269.77,218.42) .. (270.22,224.55) .. controls (270.66,230.67) and (271.08,236.5) .. (271.57,236.5) .. controls (272.06,236.5) and (272.49,230.67) .. (272.93,224.55) .. controls (273.37,218.42) and (273.79,212.59) .. (274.29,212.59) .. controls (274.78,212.59) and (275.2,218.42) .. (275.64,224.55) .. controls (276.08,230.67) and (276.51,236.5) .. (277,236.5) .. controls (277.49,236.5) and (277.91,230.67) .. (278.35,224.55) .. controls (278.8,218.42) and (279.22,212.59) .. (279.71,212.59) .. controls (280.2,212.59) and (280.63,218.42) .. (281.07,224.55) .. controls (281.51,230.67) and (281.93,236.5) .. (282.42,236.5) .. controls (282.92,236.5) and (283.34,230.67) .. (283.78,224.55) .. controls (284.22,218.42) and (284.65,212.59) .. (285.14,212.59) .. controls (285.63,212.59) and (286.05,218.42) .. (286.49,224.55) .. controls (286.94,230.67) and (287.36,236.5) .. (287.85,236.5) .. controls (288.34,236.5) and (288.76,230.67) .. (289.21,224.55) .. controls (289.65,218.42) and (290.07,212.59) .. (290.56,212.59) .. controls (291.05,212.59) and (291.48,218.42) .. (291.92,224.55) .. controls (292.36,230.67) and (292.79,236.5) .. (293.28,236.5) .. controls (293.77,236.5) and (294.19,230.67) .. (294.63,224.55) .. controls (295.08,218.42) and (295.5,212.59) .. (295.99,212.59) .. controls (296.48,212.59) and (296.9,218.42) .. (297.35,224.55) .. controls (297.79,230.67) and (298.21,236.5) .. (298.7,236.5) .. controls (299.19,236.5) and (299.62,230.67) .. (300.06,224.55) .. controls (300.5,218.42) and (300.92,212.59) .. (301.42,212.59) .. controls (301.91,212.59) and (302.33,218.42) .. (302.77,224.55) .. controls (303.21,230.67) and (303.64,236.5) .. (304.13,236.5) .. controls (304.62,236.5) and (305.04,230.67) .. (305.48,224.55) .. controls (305.93,218.42) and (306.35,212.59) .. (306.84,212.59) .. controls (307.33,212.59) and (307.76,218.42) .. (308.2,224.55) .. controls (308.64,230.67) and (309.06,236.5) .. (309.55,236.5) .. controls (310.05,236.5) and (310.47,230.67) .. (310.91,224.55) .. controls (311.35,218.42) and (311.78,212.59) .. (312.27,212.59) .. controls (312.76,212.59) and (313.18,218.42) .. (313.62,224.55) .. controls (314.07,230.67) and (314.49,236.5) .. (314.98,236.5) .. controls (315.47,236.5) and (315.89,230.67) .. (316.34,224.55) .. controls (316.78,218.42) and (317.2,212.59) .. (317.69,212.59) .. controls (318.18,212.59) and (318.61,218.42) .. (319.05,224.55) .. controls (319.49,230.67) and (319.92,236.5) .. (320.41,236.5) .. controls (320.9,236.5) and (321.32,230.67) .. (321.76,224.55) .. controls (322.21,218.42) and (322.63,212.59) .. (323.12,212.59) .. controls (323.61,212.59) and (324.03,218.42) .. (324.48,224.55) .. controls (324.92,230.67) and (325.34,236.5) .. (325.83,236.5) .. controls (326.32,236.5) and (326.75,230.67) .. (327.19,224.55) .. controls (327.63,218.42) and (328.05,212.59) .. (328.55,212.59) .. controls (329.04,212.59) and (329.46,218.42) .. (329.9,224.55) .. controls (330.34,230.67) and (330.77,236.5) .. (331.26,236.5) .. controls (331.29,236.5) and (331.32,236.48) .. (331.35,236.43) ;
\draw   (387.25,124.56) .. controls (390.79,130.68) and (394.17,136.51) .. (398.1,136.51) .. controls (402.03,136.51) and (405.41,130.68) .. (408.95,124.56) .. controls (412.49,118.43) and (415.87,112.6) .. (419.8,112.6) .. controls (423.73,112.6) and (427.11,118.43) .. (430.65,124.56) .. controls (434.19,130.68) and (437.58,136.51) .. (441.51,136.51) .. controls (445.43,136.51) and (448.82,130.68) .. (452.36,124.56) .. controls (455.9,118.43) and (459.28,112.6) .. (463.21,112.6) .. controls (467.14,112.6) and (470.52,118.43) .. (474.06,124.56) .. controls (477.6,130.68) and (480.99,136.51) .. (484.91,136.51) .. controls (486.04,136.51) and (487.12,136.04) .. (488.17,135.21) ;
\draw   (390.14,158.25) .. controls (391.91,164.37) and (393.6,170.2) .. (395.57,170.2) .. controls (397.53,170.2) and (399.22,164.37) .. (400.99,158.25) .. controls (402.76,152.13) and (404.45,146.3) .. (406.42,146.3) .. controls (408.38,146.3) and (410.07,152.13) .. (411.84,158.25) .. controls (413.61,164.37) and (415.31,170.2) .. (417.27,170.2) .. controls (419.23,170.2) and (420.93,164.37) .. (422.7,158.25) .. controls (424.47,152.13) and (426.16,146.3) .. (428.12,146.3) .. controls (430.09,146.3) and (431.78,152.13) .. (433.55,158.25) .. controls (435.32,164.37) and (437.01,170.2) .. (438.97,170.2) .. controls (440.94,170.2) and (442.63,164.37) .. (444.4,158.25) .. controls (446.17,152.13) and (447.86,146.3) .. (449.83,146.3) .. controls (451.79,146.3) and (453.48,152.13) .. (455.25,158.25) .. controls (457.02,164.37) and (458.71,170.2) .. (460.68,170.2) .. controls (462.64,170.2) and (464.33,164.37) .. (466.1,158.25) .. controls (467.87,152.13) and (469.57,146.3) .. (471.53,146.3) .. controls (473.49,146.3) and (475.19,152.13) .. (476.96,158.25) .. controls (478.73,164.37) and (480.42,170.2) .. (482.38,170.2) .. controls (484.35,170.2) and (486.04,164.37) .. (487.81,158.25) .. controls (487.93,157.83) and (488.05,157.41) .. (488.17,157) ;
\draw  [color={rgb, 255:red, 0; green, 0; blue, 0 }  ,draw opacity=1 ][fill={rgb, 255:red, 137; green, 192; blue, 237 }  ,fill opacity=1 ][line width=1.5]  (640.44,327.79) -- (692.89,327.79) -- (692.89,383.22) -- (640.44,383.22) -- cycle ;
\draw  [color={rgb, 255:red, 0; green, 0; blue, 0 }  ,draw opacity=1 ][fill={rgb, 255:red, 137; green, 192; blue, 237 }  ,fill opacity=1 ][line width=1.5]  (644.78,323.44) -- (697.23,323.44) -- (697.23,378.87) -- (644.78,378.87) -- cycle ;
\draw  [color={rgb, 255:red, 0; green, 0; blue, 0 }  ,draw opacity=1 ][fill={rgb, 255:red, 137; green, 192; blue, 237 }  ,fill opacity=1 ][line width=1.5]  (649.12,319.1) -- (701.57,319.1) -- (701.57,374.52) -- (649.12,374.52) -- cycle ;
\draw  [color={rgb, 255:red, 0; green, 0; blue, 0 }  ,draw opacity=1 ][fill={rgb, 255:red, 137; green, 192; blue, 237 }  ,fill opacity=1 ][line width=1.5]  (653.46,314.75) -- (705.91,314.75) -- (705.91,370.18) -- (653.46,370.18) -- cycle ;
\draw  [color={rgb, 255:red, 0; green, 0; blue, 0 }  ,draw opacity=1 ][fill={rgb, 255:red, 137; green, 192; blue, 237 }  ,fill opacity=1 ][line width=1.5]  (671.91,299.53) -- (724.36,299.53) -- (724.36,354.96) -- (671.91,354.96) -- cycle ;
\draw  [color={rgb, 255:red, 0; green, 0; blue, 0 }  ,draw opacity=1 ][fill={rgb, 255:red, 137; green, 192; blue, 237 }  ,fill opacity=1 ][line width=1.5]  (676.25,295.19) -- (728.7,295.19) -- (728.7,350.61) -- (676.25,350.61) -- cycle ;
\draw  [color={rgb, 255:red, 0; green, 0; blue, 0 }  ,draw opacity=1 ][fill={rgb, 255:red, 137; green, 192; blue, 237 }  ,fill opacity=1 ][line width=1.5]  (680.59,290.84) -- (733.04,290.84) -- (733.04,346.27) -- (680.59,346.27) -- cycle ;
\draw  [color={rgb, 255:red, 0; green, 0; blue, 0 }  ,draw opacity=1 ][fill={rgb, 255:red, 137; green, 192; blue, 237 }  ,fill opacity=1 ][line width=1.5]  (684.93,286.49) -- (737.38,286.49) -- (737.38,341.92) -- (684.93,341.92) -- cycle ;
\draw [fill={rgb, 255:red, 74; green, 144; blue, 226 }  ,fill opacity=0.6 ][line width=1.5]    (105.63,196.29) -- (123.04,178.85) ;
\draw [shift={(125.16,176.73)}, rotate = 134.96] [color={rgb, 255:red, 0; green, 0; blue, 0 }  ][line width=1.5]    (14.21,-4.28) .. controls (9.04,-1.82) and (4.3,-0.39) .. (0,0) .. controls (4.3,0.39) and (9.04,1.82) .. (14.21,4.28)   ;
\draw [fill={rgb, 255:red, 113; green, 213; blue, 200 }  ,fill opacity=1 ][line width=1.5]    (100.2,259.5) -- (257.03,336.43) ;
\draw [shift={(259.73,337.75)}, rotate = 206.13] [color={rgb, 255:red, 0; green, 0; blue, 0 }  ][line width=1.5]    (14.21,-4.28) .. controls (9.04,-1.82) and (4.3,-0.39) .. (0,0) .. controls (4.3,0.39) and (9.04,1.82) .. (14.21,4.28)   ;
\draw [fill={rgb, 255:red, 74; green, 144; blue, 226 }  ,fill opacity=0.6 ][line width=1.5]    (464.38,345.18) -- (607.88,347.31) ;
\draw [shift={(610.88,347.35)}, rotate = 180.85] [color={rgb, 255:red, 0; green, 0; blue, 0 }  ][line width=1.5]    (14.21,-4.28) .. controls (9.04,-1.82) and (4.3,-0.39) .. (0,0) .. controls (4.3,0.39) and (9.04,1.82) .. (14.21,4.28)   ;
\draw  [line width=1.5]  (678.78,250.63) .. controls (678.78,239.22) and (688.01,229.98) .. (699.4,229.98) .. controls (710.79,229.98) and (720.02,239.22) .. (720.02,250.63) .. controls (720.02,262.03) and (710.79,271.28) .. (699.4,271.28) .. controls (688.01,271.28) and (678.78,262.03) .. (678.78,250.63) -- cycle ;
\draw  [line width=1.5]  (683.12,250.63) -- (715.68,250.63)(699.4,234.33) -- (699.4,266.93) ;
\draw  [color={rgb, 255:red, 0; green, 0; blue, 0 }  ,draw opacity=1 ][fill={rgb, 255:red, 117; green, 216; blue, 204 }  ,fill opacity=1 ][line width=1.5]  (255.94,361.48) -- (285.6,361.48) -- (285.6,393) -- (255.94,393) -- cycle ;
\draw  [color={rgb, 255:red, 0; green, 0; blue, 0 }  ,draw opacity=1 ][fill={rgb, 255:red, 117; green, 216; blue, 204 }  ,fill opacity=1 ][line width=1.5]  (260.28,356.05) -- (289.94,356.05) -- (289.94,387.57) -- (260.28,387.57) -- cycle ;
\draw  [color={rgb, 255:red, 0; green, 0; blue, 0 }  ,draw opacity=1 ][fill={rgb, 255:red, 117; green, 216; blue, 204 }  ,fill opacity=1 ][line width=1.5]  (264.62,350.61) -- (294.28,350.61) -- (294.28,382.13) -- (264.62,382.13) -- cycle ;
\draw  [color={rgb, 255:red, 0; green, 0; blue, 0 }  ,draw opacity=1 ][fill={rgb, 255:red, 117; green, 216; blue, 204 }  ,fill opacity=1 ][line width=1.5]  (268.96,346.27) -- (298.62,346.27) -- (298.62,377.78) -- (268.96,377.78) -- cycle ;
\draw  [color={rgb, 255:red, 0; green, 0; blue, 0 }  ,draw opacity=1 ][fill={rgb, 255:red, 117; green, 216; blue, 204 }  ,fill opacity=1 ][line width=1.5]  (287.41,333.23) -- (317.07,333.23) -- (317.07,364.74) -- (287.41,364.74) -- cycle ;
\draw  [color={rgb, 255:red, 0; green, 0; blue, 0 }  ,draw opacity=1 ][fill={rgb, 255:red, 117; green, 216; blue, 204 }  ,fill opacity=1 ][line width=1.5]  (291.75,327.79) -- (321.41,327.79) -- (321.41,359.31) -- (291.75,359.31) -- cycle ;
\draw  [color={rgb, 255:red, 0; green, 0; blue, 0 }  ,draw opacity=1 ][fill={rgb, 255:red, 117; green, 216; blue, 204 }  ,fill opacity=1 ][line width=1.5]  (296.09,322.36) -- (325.75,322.36) -- (325.75,353.88) -- (296.09,353.88) -- cycle ;
\draw  [color={rgb, 255:red, 0; green, 0; blue, 0 }  ,draw opacity=1 ][fill={rgb, 255:red, 117; green, 216; blue, 204 }  ,fill opacity=1 ][line width=1.5]  (300.43,318.01) -- (330.09,318.01) -- (330.09,349.53) -- (300.43,349.53) -- cycle ;
\draw  [color={rgb, 255:red, 0; green, 0; blue, 0 }  ,draw opacity=1 ][fill={rgb, 255:red, 113; green, 213; blue, 200 }  ,fill opacity=1 ][line width=1.5]  (349.27,360.4) -- (378.93,360.4) -- (378.93,391.91) -- (349.27,391.91) -- cycle ;
\draw  [color={rgb, 255:red, 0; green, 0; blue, 0 }  ,draw opacity=1 ][fill={rgb, 255:red, 113; green, 213; blue, 200 }  ,fill opacity=1 ][line width=1.5]  (353.61,354.96) -- (383.27,354.96) -- (383.27,386.48) -- (353.61,386.48) -- cycle ;
\draw  [color={rgb, 255:red, 0; green, 0; blue, 0 }  ,draw opacity=1 ][fill={rgb, 255:red, 113; green, 213; blue, 200 }  ,fill opacity=1 ][line width=1.5]  (357.95,349.53) -- (387.61,349.53) -- (387.61,381.05) -- (357.95,381.05) -- cycle ;
\draw  [color={rgb, 255:red, 0; green, 0; blue, 0 }  ,draw opacity=1 ][fill={rgb, 255:red, 113; green, 213; blue, 200 }  ,fill opacity=1 ][line width=1.5]  (362.29,345.18) -- (391.95,345.18) -- (391.95,376.7) -- (362.29,376.7) -- cycle ;
\draw  [color={rgb, 255:red, 0; green, 0; blue, 0 }  ,draw opacity=1 ][fill={rgb, 255:red, 113; green, 213; blue, 200 }  ,fill opacity=1 ][line width=1.5]  (380.74,332.14) -- (410.4,332.14) -- (410.4,363.66) -- (380.74,363.66) -- cycle ;
\draw  [color={rgb, 255:red, 0; green, 0; blue, 0 }  ,draw opacity=1 ][fill={rgb, 255:red, 113; green, 213; blue, 200 }  ,fill opacity=1 ][line width=1.5]  (385.08,326.7) -- (414.74,326.7) -- (414.74,358.22) -- (385.08,358.22) -- cycle ;
\draw  [color={rgb, 255:red, 0; green, 0; blue, 0 }  ,draw opacity=1 ][fill={rgb, 255:red, 113; green, 213; blue, 200 }  ,fill opacity=1 ][line width=1.5]  (389.42,321.27) -- (419.08,321.27) -- (419.08,352.79) -- (389.42,352.79) -- cycle ;
\draw  [color={rgb, 255:red, 0; green, 0; blue, 0 }  ,draw opacity=1 ][fill={rgb, 255:red, 113; green, 213; blue, 200 }  ,fill opacity=1 ][line width=1.5]  (393.76,316.92) -- (423.42,316.92) -- (423.42,348.44) -- (393.76,348.44) -- cycle ;

\draw (377.95,272.06) node [anchor=north west][inner sep=0.75pt]   [align=left] {$ $};
\draw (50.54,190.94) node [anchor=north west][inner sep=0.75pt]  [font=\Large] [align=left] {$\displaystyle \boldsymbol{f}( x)$};
\draw (141.65,125.73) node [anchor=north west][inner sep=0.75pt]  [font=\Large] [align=left] {$\displaystyle \boldsymbol{v}( x)$};
\draw (141.61,192.03) node [anchor=north west][inner sep=0.75pt]  [font=\Large] [align=left] {$\displaystyle \boldsymbol{s}( x)$};
\draw (690.68,130.25) node [anchor=north west][inner sep=0.75pt]  [font=\Large] [align=left] {$\displaystyle \boldsymbol{f}^{*}( x)$};
\draw (560.33,125.82) node [anchor=north west][inner sep=0.75pt]  [font=\Large] [align=left] {$\displaystyle \boldsymbol{v}^{*}( x)$};
\draw (561.37,192.11) node [anchor=north west][inner sep=0.75pt]  [font=\Large] [align=left] {$\displaystyle \boldsymbol{s}^{*}( x)$};
\draw (194.93,111.43) node [anchor=north west][inner sep=0.75pt]  [font=\large] [align=left] {$\displaystyle \text{FFT}$};
\draw (243.11,79.39) node [anchor=north west][inner sep=0.75pt]  [font=\Large] [align=left] {$\displaystyle \mathcal{F}(\boldsymbol{f}( x))$};
\draw (194.93,175.72) node [anchor=north west][inner sep=0.75pt]  [font=\large] [align=left] {$\displaystyle \text{FFT}$};
\draw (494.73,107.3) node [anchor=north west][inner sep=0.75pt]  [font=\large] [align=left] {$\displaystyle \text{FFT}^{-1}$};
\draw (496.55,182.29) node [anchor=north west][inner sep=0.75pt]  [font=\large] [align=left] {$\displaystyle \text{FFT}^{-1}$};
\draw (686.93,298.79) node [anchor=north west][inner sep=0.75pt]  [font=\Large] [align=left] {$\displaystyle \boldsymbol{f}^{\dagger }( x)$};
\draw (346.73,135.99) node [anchor=north west][inner sep=0.75pt]  [font=\huge] [align=left] {$\displaystyle \boldsymbol{W}$};
\draw (336.99,292.36) node [anchor=north west][inner sep=0.75pt]  [font=\huge] [align=left] {$\displaystyle \boldsymbol{w}$};

\end{tikzpicture}

%% file: sections/experiments.tex
\section{Experiments}\label{sec:experiments}
We assess Clifford neural layers for different architectures in three experimental settings: the incompressible Navier-Stokes equations, shallow water equations for weather modeling, and 3-dimensional Maxwell's equations. 
We replace carefully designed baseline architectures by their Clifford counterparts.
Baseline ResNet architectures comprise 8 residual blocks, each consisting of two convolution layers with $3 \times 3$ kernels, shortcut connections, group normalization~\citep{wu2018group}, and GeLU activation functions~\citep{hendrycks2016gaussian}. 
Baseline 2-dimensional Fourier Neural Operators (FNOs) consist of 8 (4) FNO blocks, GeLU activations and no normalization scheme,
using 16 (8) Fourier modes for the $2$- and $3$-dimensional equations, respectively.
For Clifford networks, we change convolutions and Fourier transforms to their respective Clifford operation, and substitute normalization techniques and activation functions with Clifford counterparts, keeping the number of parameters similar. We evaluate different training set sizes, and report losses for scalar and vector fields.
All datasets share the common trait of containing multiple input and output fields.
More precisely, one scalar and one $2$-dimensional vector field in case of the Navier-Stokes and the shallow water equations, and a $3$-dimensional (electric) vector field and its dual (magnetic) bivector field in case of the Maxwell's equations.

\begin{wrapfigure}[17]{r}{0.37\textwidth}
    \centering
    \vspace{-0.5cm}
    \makebox[0pt]{\rotatebox[origin=l]{90}{
        \hspace{70pt}\textbf{\scriptsize Input}
    }\hspace*{1em}}%
    \makebox[0pt]{\rotatebox[origin=l]{90}{
        \hspace{8pt}\textbf{\scriptsize Target}
    }\hspace*{1em}}%
    \includegraphics[width=0.33\textwidth]{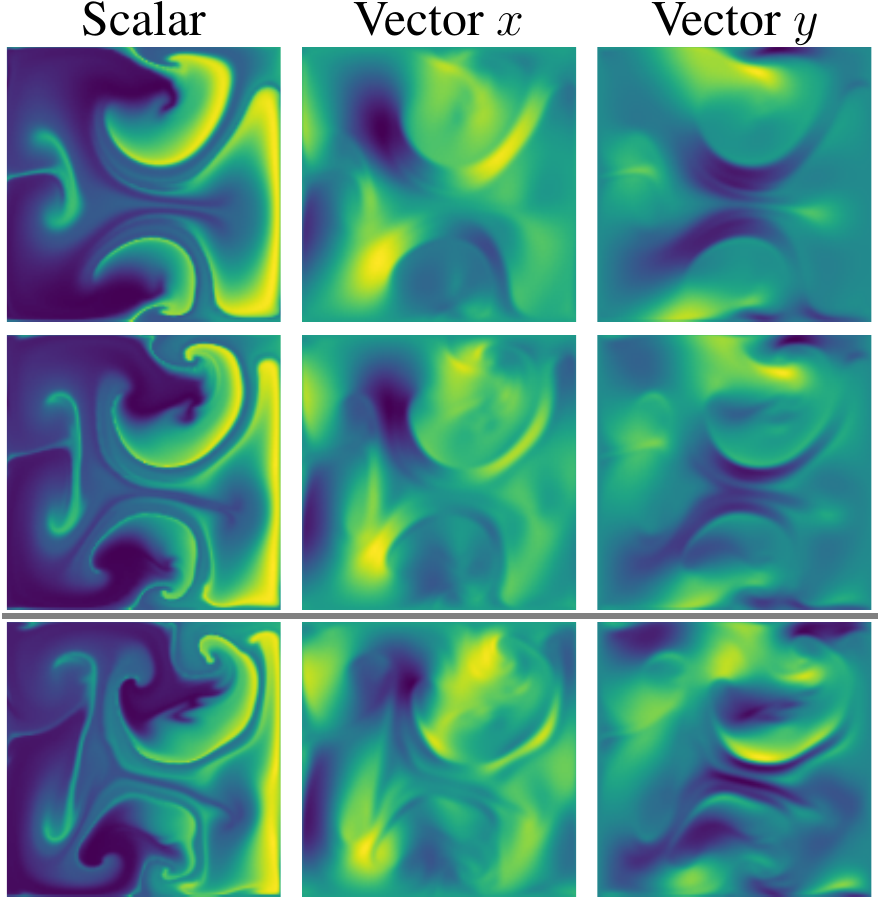}
    \caption{Example input and target fields for the Navier-Stokes experiments. Input fields comprise a $t=2$ timestep history.}
    \label{fig:dataset_example}
\end{wrapfigure}
Example inputs and targets of the neural PDE surrogates are shown in Figure~\ref{fig:dataset_example}. 
The number of input timesteps $t$ vary for different experiments. The \textit{one-step loss} is the mean-squared error at the next timestep summed over fields. The \textit{rollout loss} is the mean-squared error after applying the neural PDE surrogate $5$ times, summing over fields and time dimension. More information on the implementation details of the tested architectures, loss functions, and more detailed results can be found in Appendix~\ref{app:experiments}.

\begin{figure}
    \centering
    \begin{subfigure}[b]{0.47\textwidth}
            \centering
            \vspace{-0.5cm}
            \input{tikz/smoke/mainbar_resnet_2e-4.tex}%
            \caption{Navier-Stokes equations} %
            \label{fig:NS_results_scalar_vector}
    \end{subfigure}\quad
    \begin{subfigure}[b]{0.47\textwidth}
            \centering
            \vspace{-0.5cm}
            \input{tikz/weather/mainbar_resnet_2e-4_hist=2.tex}%
            \caption{Shallow water equations}
            \label{fig:SW_results_scalar_vector}
    \end{subfigure}
    \caption{Results for ResNet based (left) and Fourier based (right) architectures on the $2$-dimensional Navier-Stokes and Shallow water experiments. One-step and rollout loss are shown.}
    \label{fig:my_label}
\end{figure}
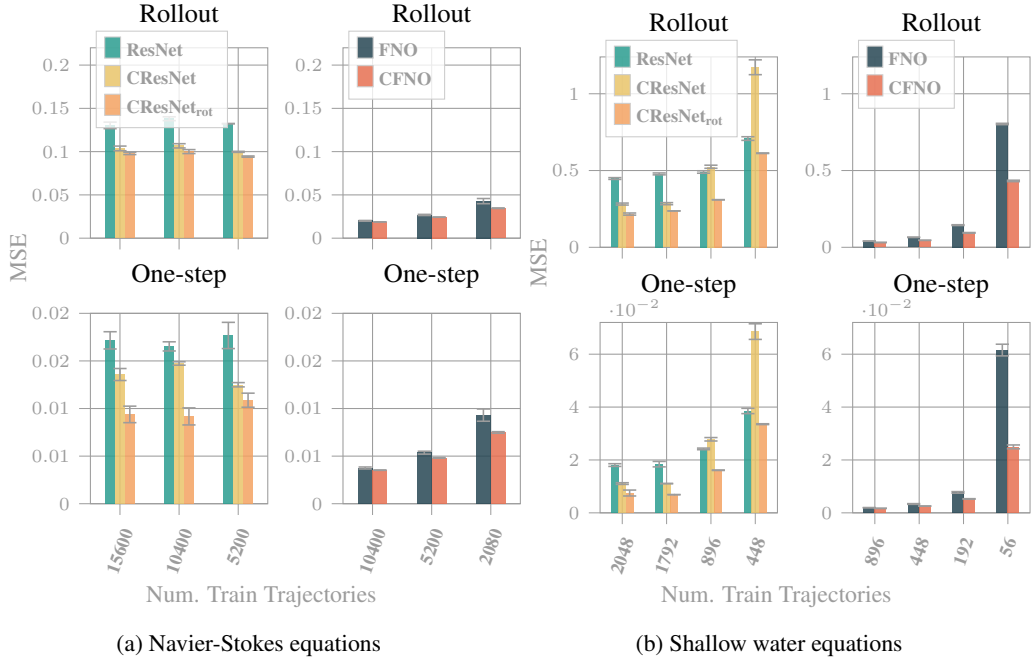

\textbf{Navier-Stokes in 2D. }
The incompressible Navier-Stokes equations~\citep{temam2001navier} conserve the velocity flow fields $v: \gX \rightarrow \R^2$ where $\gX \in \R^2$ via:
\begin{align}
    \frac{\partial v}{\partial t} = -v \cdot \nabla v + \mu \nabla^2 v - \nabla p + f \ , \quad \nabla \cdot v = 0 \ \label{eq:Navier_Stokes},
\end{align}
where $v \cdot \nabla v$ is the convection, i.e. the rate of change of $v$ along $v$, $\mu \nabla^2v$ the viscosity, i.e. the diffusion or net movement of $v$, $\nabla p$ the internal pressure and $f$ an external force, which in our case is a buoyancy force. An additional incompressibility constraint $\nabla \cdot v = 0$ yields mass conservation of the Navier-Stokes equations. In addition to the velocity field, we introduce a scalar field representing a scalar quantity, i.e. smoke, that is being transported via the velocity field. The scalar field is \emph{advected} by the vector field, i.e.\ as the vector field changes, the scalar field is transported along with it,
whereas the scalar field influences the vector field only via an external force term.
We call this \textbf{weak coupling} between vector and scalar fields. We implement the 2D Navier-Stokes equation using \texttt{${\Phi}$Flow}\footnote{\url{https://github.com/tum-pbs/PhiFlow}}\citep{holl2020phiflow}, obtaining data on a grid with spatial resolution of $128 \times 128$ ($\Delta x=0.25$, $\Delta y=0.25$), and temporal resolution of $\Delta t = \SI{1.5}\second$.
Results for one-step loss and rollout loss on the test set are shown in Figure~\ref{fig:NS_results_scalar_vector}. 
For ResNet-like architectures, we observe that both CResNet and CResNet$_{\text{rot}}$ improve upon the ResNet baseline.
Additionally, we observe that rollout losses are also lower for the two Clifford based architectures, which we attribute to better and more stable models that do not overfit to one-step predictions so easily. 
Lastly, while in principle CResNet and CResNet$_{\text{rot}}$ based architectures are equally flexible, CResNet$_{\text{rot}}$ ones in general perform better than CResNet ones. 
For FNO and respective Clifford Fourier based (CFNO) architectures, the loss is in general much lower than for ResNet based architectures. CFNO architectures improve upon FNO architectures for all dataset sizes, and for one-step as well as rollout losses.

\textbf{Shallow water equations. }
This set of coupled equations~\citep{vreugdenhil1994numerical} can be derived from integrating the incompressible Navier–Stokes equations, in cases where the horizontal length scale is much larger than the vertical length scale.
As such, the equations model a thin layer of fluid of constant density in hydrostatic
balance, bounded from below by the bottom topography and from above by a free surface via 3 coupled PDEs, describing
the velocity in $x$- direction, the velocity in the $y$- direction, and the scalar pressure field. 
The shallow water equations can be therefore be used as simplified weather model, as done in this work and exemplified in Figure~\ref{fig:earth_shallow_water}. The relation between vector and scalar components is relatively strong (\textbf{strong coupling} due to the 3-coupled PDEs). We obtain data for the 2D shallow water equations on a grid with spatial resolution of $192 \times 96$ ($\Delta x=1.875\degree$, $\Delta y=3.75\degree$), and temporal resolution of $\Delta t = \SI{6}\hour$.
We observe similar results than for the Navier-Stokes experiments. 
For low number of trajectories, ResNet architectures seem to lack expressiveness, where arguably some data smoothing is learned first. Thus, ResNets need significantly more trajectories compared to (C)FNO architectures to obtain reasonable loss values, which seems to go hand in hand with Clifford layers gaining advantage.   
In general, performance differences between baseline and Clifford architectures are even more pronounced, which we attribute to the stronger coupling of the scalar and the vector fields. 

\textbf{Maxwell's equations in matter in 3D. }
In isotropic media, Maxwell's equations~\citep{griffiths2005introduction} propagate solutions of the displacement field $D$, which is related to the electrical field via $D = \epsilon_0 \epsilon_r E$, where $\epsilon_0$ is the permittivity of free space and $\epsilon_r$ is the permittivity of the medium, and the magnetization field $H$, which is related to the magnetic field $B$ via $H = \mu_0 \mu_r B$, where $\mu_0$ is the permeability of free space and $\mu_r$ is the permeability of the medium.
The electromagnetic field $\mF$ has the intriguing property that the electric field $E$ and the magnetic field $B$ are dual pairs, thus $\mF = E + B i_3$, i.e. \textbf{strong coupling} between the electric field and its dual (bivector) magnetic field. This duality also holds for $D$ and $H$. Concretely, the fields of interest are the vector-valued $D$-field ($D_x, D_y, D_z$) and the vector-valued $H$-field
($H_x, H_y, H_z$).
We obtain data for the 3D Maxwell's equations on a grid with spatial resolution of $32 \times 32 \times 32$ ($\Delta x = \Delta y = \Delta z = 5\cdot 10^{-7}m$), and temporal resolution of $\Delta t = \SI{50}{\second}$. 
We randomly place 18 different light sources outside a cube which emit light with different amplitude and different phase shifts, causing the resulting $D$ and $H$ fields to interfere. The wavelength of the emitted light is $10^{-5}m$.

\begin{wrapfigure}[21]{r}{0.28\textwidth}
    \vspace{-24pt}
    \centering
    \resizebox{0.28\textwidth}{!}{\input{tikz/maxwell/mainbar_fourier_1e-4.tex}}
    \caption{Results for Fourier based architectures on Maxwell equation's.}
    \label{fig:Max_results}
\end{wrapfigure}
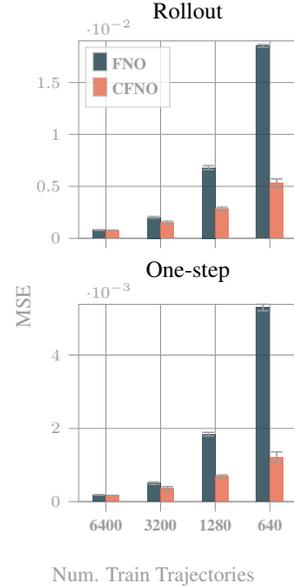

We test FNO based architectures and respective Clifford counterparts (CFNO). 
Due to the vector-bivector character of electric and magnetic field components,
Maxwell's equations are an ideal playground
to stress-test the inductive bias advantages of Clifford base architectures.
Results for one-step loss and rollout loss on the test set are shown in Figure~\ref{fig:Max_results}.
CFNO architectures improve upon FNO architectures,
especially for low numbers of trajectories. Results demonstrate
the much stronger inductive bias of Clifford based 3-dimensional Fourier layers, and their general applicability to 3-dimensional problems, which are structurally even more interesting than 2-dimensional ones. 

%% file: tikz/smoke/mainbar_resnet_2e-4.tex
\begin{tikzpicture}

\definecolor{burlywood233196106}{RGB}{233,196,106}
\definecolor{darkgray176}{RGB}{156,156,156}
\definecolor{lightgray204}{RGB}{204,204,204}
\definecolor{lightseagreen42157143}{RGB}{42,157,143}
\definecolor{sandybrown24416297}{RGB}{244,162,97}
\definecolor{darkslategray387083}{RGB}{38,70,83}
\definecolor{lightgray204}{RGB}{204,204,204}
\definecolor{tomato23111181}{RGB}{231,111,81}

\begin{groupplot}[group style={group size=2 by 2}, height=0.18\textheight, width=0.6\textwidth, yticklabel style={color=darkgray176, font=\bfseries\scriptsize}, xticklabel style={color=darkgray176, font=\bfseries\scriptsize}, yticklabel style={
        /pgf/number format/fixed,
        /pgf/number format/precision=2
},
scaled y ticks=false]
\nextgroupplot[
axis line style={darkgray176},
legend cell align={left},
legend style={
  fill opacity=0.8,
  draw opacity=1,
  text=darkgray176,
  text opacity=1,
  at={(0.03,1.09)},
  anchor=north west,
  draw=lightgray204,
  font=\bfseries\scriptsize
},
tick align=outside,
tick pos=left,
title={Rollout},
x grid style={darkgray176},
xmajorgrids,
xmin=-0.5, xmax=2.5,
xtick style={color=darkgray176},
xticklabels={},
y grid style={darkgray176},
ymajorgrids,
ymin=0, ymax=0.22,
ytick style={color=darkgray176}
]
\draw[draw=none,fill=lightseagreen42157143,fill opacity=0.85] (axis cs:-0.25,0) rectangle (axis cs:-0.0833333333333333,0.130299009382725);
\addlegendimage{ybar,ybar legend,draw=none,fill=lightseagreen42157143,fill opacity=0.85}
\addlegendentry{ResNet}

\draw[draw=none,fill=lightseagreen42157143,fill opacity=0.85] (axis cs:0.75,0) rectangle (axis cs:0.916666666666667,0.138020850718021);
\draw[draw=none,fill=lightseagreen42157143,fill opacity=0.85] (axis cs:1.75,0) rectangle (axis cs:1.91666666666667,0.132339775562286);
\draw[draw=none,fill=burlywood233196106,fill opacity=0.85] (axis cs:-0.0833333333333333,0) rectangle (axis cs:0.0833333333333333,0.103724729269743);
\addlegendimage{ybar,ybar legend,draw=none,fill=burlywood233196106,fill opacity=0.85}
\addlegendentry{CResNet}

\draw[draw=none,fill=burlywood233196106,fill opacity=0.85] (axis cs:0.916666666666667,0) rectangle (axis cs:1.08333333333333,0.106710869818926);
\draw[draw=none,fill=burlywood233196106,fill opacity=0.85] (axis cs:1.91666666666667,0) rectangle (axis cs:2.08333333333333,0.0997475385665894);
\draw[draw=none,fill=sandybrown24416297,fill opacity=0.85] (axis cs:0.0833333333333333,0) rectangle (axis cs:0.25,0.0979935452342033);
\addlegendimage{ybar,ybar legend,draw=none,fill=sandybrown24416297,fill opacity=0.85}
\addlegendentry{CResNet$_\text{rot}$}

\draw[draw=none,fill=sandybrown24416297,fill opacity=0.85] (axis cs:1.08333333333333,0) rectangle (axis cs:1.25,0.100053422152996);
\draw[draw=none,fill=sandybrown24416297,fill opacity=0.85] (axis cs:2.08333333333333,0) rectangle (axis cs:2.25,0.0942723527550697);
\path [draw=darkgray176, semithick]
(axis cs:-0.166666666666667,0.126509562134743)
--(axis cs:-0.166666666666667,0.134088456630707);

\path [draw=darkgray176, semithick]
(axis cs:0.833333333333333,0.135785982012749)
--(axis cs:0.833333333333333,0.140255719423294);

\path [draw=darkgray176, semithick]
(axis cs:1.83333333333333,0.132139414548874)
--(axis cs:1.83333333333333,0.132540136575699);

\addplot [semithick, darkgray176, mark=-, mark size=2.5, mark options={solid}, only marks]
table {%
-0.166666666666667 0.126509562134743
0.833333333333333 0.135785982012749
1.83333333333333 0.132139414548874
};

\addplot [semithick, darkgray176, mark=-, mark size=2.5, mark options={solid}, only marks]
table {%
-0.166666666666667 0.134088456630707
0.833333333333333 0.140255719423294
1.83333333333333 0.132540136575699
};

\path [draw=darkgray176, semithick]
(axis cs:0,0.101037546992302)
--(axis cs:0,0.106411911547184);

\path [draw=darkgray176, semithick]
(axis cs:1,0.104249946773052)
--(axis cs:1,0.109171792864799);

\path [draw=darkgray176, semithick]
(axis cs:2,0.099151685833931)
--(axis cs:2,0.100343391299248);

\addplot [semithick, darkgray176, mark=-, mark size=2.5, mark options={solid}, only marks]
table {%
0 0.101037546992302
1 0.104249946773052
2 0.099151685833931
};

\addplot [semithick, darkgray176, mark=-, mark size=2.5, mark options={solid}, only marks]
table {%
0 0.106411911547184
1 0.109171792864799
2 0.100343391299248
};

\path [draw=darkgray176, semithick]
(axis cs:0.166666666666667,0.0966045185923576)
--(axis cs:0.166666666666667,0.099382571876049);

\path [draw=darkgray176, semithick]
(axis cs:1.16666666666667,0.0977661535143852)
--(axis cs:1.16666666666667,0.102340690791607);

\path [draw=darkgray176, semithick]
(axis cs:2.16666666666667,0.0935615301132202)
--(axis cs:2.16666666666667,0.0949831753969193);

\addplot [semithick, darkgray176, mark=-, mark size=2.5, mark options={solid}, only marks]
table {%
0.166666666666667 0.0966045185923576
1.16666666666667 0.0977661535143852
2.16666666666667 0.0935615301132202
};

\addplot [semithick, darkgray176, mark=-, mark size=2.5, mark options={solid}, only marks]
table {%
0.166666666666667 0.099382571876049
1.16666666666667 0.102340690791607
2.16666666666667 0.0949831753969193
};

\nextgroupplot[
axis line style={darkgray176},
legend cell align={left},
legend style={fill opacity=0.8, 
  at={(0.03,1.09)},
  anchor=north west,
draw opacity=1, text=darkgray176, text opacity=1, draw=lightgray204,  font=\bfseries\scriptsize},
tick align=outside,
tick pos=left,
title={Rollout},
x grid style={darkgray176},
xmajorgrids,
xmin=-0.5, xmax=2.5,
xtick style={color=darkgray176},
xticklabels={},
y grid style={darkgray176},
ymajorgrids,
ymin=0, ymax=0.22,
ytick style={color=darkgray176}
]
\draw[draw=none,fill=darkslategray387083,fill opacity=0.85] (axis cs:-0.25,0) rectangle (axis cs:0,0.0200465209782124);
\addlegendimage{ybar,ybar legend,draw=none,fill=darkslategray387083,fill opacity=0.85}
\addlegendentry{FNO}

\draw[draw=none,fill=darkslategray387083,fill opacity=0.85] (axis cs:0.75,0) rectangle (axis cs:1,0.0268410226951043);
\draw[draw=none,fill=darkslategray387083,fill opacity=0.85] (axis cs:1.75,0) rectangle (axis cs:2,0.0428087487816811);
\draw[draw=none,fill=tomato23111181,fill opacity=0.85] (axis cs:0,0) rectangle (axis cs:0.25,0.018858910848697);
\addlegendimage{ybar,ybar legend,draw=none,fill=tomato23111181,fill opacity=0.85}
\addlegendentry{CFNO}

\draw[draw=none,fill=tomato23111181,fill opacity=0.85] (axis cs:1,0) rectangle (axis cs:1.25,0.0243015394856532);
\draw[draw=none,fill=tomato23111181,fill opacity=0.85] (axis cs:2,0) rectangle (axis cs:2.25,0.0346080536643664);
\path [draw=darkgray176, semithick]
(axis cs:-0.125,0.0196278563494451)
--(axis cs:-0.125,0.0204651856069796);

\path [draw=darkgray176, semithick]
(axis cs:0.875,0.0261708690555498)
--(axis cs:0.875,0.0275111763346587);

\path [draw=darkgray176, semithick]
(axis cs:1.875,0.0398105667012712)
--(axis cs:1.875,0.045806930862091);

\addplot [semithick, darkgray176, mark=-, mark size=2.5, mark options={solid}, only marks]
table {%
-0.125 0.0196278563494451
0.875 0.0261708690555498
1.875 0.0398105667012712
};

\addplot [semithick, darkgray176, mark=-, mark size=2.5, mark options={solid}, only marks]
table {%
-0.125 0.0204651856069796
0.875 0.0275111763346587
1.875 0.045806930862091
};

\path [draw=darkgray176, semithick]
(axis cs:0.125,0.0188005149575856)
--(axis cs:0.125,0.0189173067398085);

\path [draw=darkgray176, semithick]
(axis cs:1.125,0.0241859783887698)
--(axis cs:1.125,0.0244171005825367);

\path [draw=darkgray176, semithick]
(axis cs:2.125,0.0342985319480809)
--(axis cs:2.125,0.0349175753806519);

\addplot [semithick, darkgray176, mark=-, mark size=2.5, mark options={solid}, only marks]
table {%
0.125 0.0188005149575856
1.125 0.0241859783887698
2.125 0.0342985319480809
};

\addplot [semithick, darkgray176, mark=-, mark size=2.5, mark options={solid}, only marks]
table {%
0.125 0.0189173067398085
1.125 0.0244171005825367
2.125 0.0349175753806519
};

\nextgroupplot[
axis line style={darkgray176},
tick align=outside,
tick pos=left,
title={One-step},
x grid style={darkgray176},
xmajorgrids,
xmin=-0.5, xmax=2.5,
xtick style={color=darkgray176},
xtick={0,1,2},
xticklabels={15600,10400,5200},
xticklabel style={rotate=70},
y grid style={darkgray176},
ymajorgrids,
ymin=0, ymax=0.020,
ytick style={color=darkgray176}
]
\draw[draw=none,fill=lightseagreen42157143,fill opacity=0.85] (axis cs:-0.25,0) rectangle (axis cs:-0.0833333333333333,0.0171611784026027);
\draw[draw=none,fill=lightseagreen42157143,fill opacity=0.85] (axis cs:0.75,0) rectangle (axis cs:0.916666666666667,0.0165270520374179);
\draw[draw=none,fill=lightseagreen42157143,fill opacity=0.85] (axis cs:1.75,0) rectangle (axis cs:1.91666666666667,0.0176742104813457);
\draw[draw=none,fill=burlywood233196106,fill opacity=0.85] (axis cs:-0.0833333333333333,0) rectangle (axis cs:0.0833333333333333,0.0135723566636443);
\draw[draw=none,fill=burlywood233196106,fill opacity=0.85] (axis cs:0.916666666666667,0) rectangle (axis cs:1.08333333333333,0.0147318239323795);
\draw[draw=none,fill=burlywood233196106,fill opacity=0.85] (axis cs:1.91666666666667,0) rectangle (axis cs:2.08333333333333,0.0124978306703269);
\draw[draw=none,fill=sandybrown24416297,fill opacity=0.85] (axis cs:0.0833333333333333,0) rectangle (axis cs:0.25,0.00938191823661327);
\draw[draw=none,fill=sandybrown24416297,fill opacity=0.85] (axis cs:1.08333333333333,0) rectangle (axis cs:1.25,0.00917232036590576);
\draw[draw=none,fill=sandybrown24416297,fill opacity=0.85] (axis cs:2.08333333333333,0) rectangle (axis cs:2.25,0.0108740576542914);
\path [draw=darkgray176, semithick]
(axis cs:-0.166666666666667,0.0162548813968897)
--(axis cs:-0.166666666666667,0.0180674754083157);

\path [draw=darkgray176, semithick]
(axis cs:0.833333333333333,0.0160455927252769)
--(axis cs:0.833333333333333,0.0170085113495588);

\path [draw=darkgray176, semithick]
(axis cs:1.83333333333333,0.0162966046482325)
--(axis cs:1.83333333333333,0.0190518163144588);

\addplot [semithick, darkgray176, mark=-, mark size=2.5, mark options={solid}, only marks]
table {%
-0.166666666666667 0.0162548813968897
0.833333333333333 0.0160455927252769
1.83333333333333 0.0162966046482325
};
\addplot [semithick, darkgray176, mark=-, mark size=2.5, mark options={solid}, only marks]
table {%
-0.166666666666667 0.0180674754083157
0.833333333333333 0.0170085113495588
1.83333333333333 0.0190518163144588
};
\path [draw=darkgray176, semithick]
(axis cs:0,0.012942329980433)
--(axis cs:0,0.0142023833468556);

\path [draw=darkgray176, semithick]
(axis cs:1,0.0145516116172075)
--(axis cs:1,0.0149120362475514);

\path [draw=darkgray176, semithick]
(axis cs:2,0.0122731691226363)
--(axis cs:2,0.0127224922180176);

\addplot [semithick, darkgray176, mark=-, mark size=2.5, mark options={solid}, only marks]
table {%
0 0.012942329980433
1 0.0145516116172075
2 0.0122731691226363
};
\addplot [semithick, darkgray176, mark=-, mark size=2.5, mark options={solid}, only marks]
table {%
0 0.0142023833468556
1 0.0149120362475514
2 0.0127224922180176
};
\path [draw=darkgray176, semithick]
(axis cs:0.166666666666667,0.00851487554609776)
--(axis cs:0.166666666666667,0.0102489609271288);

\path [draw=darkgray176, semithick]
(axis cs:1.16666666666667,0.00827652681618929)
--(axis cs:1.16666666666667,0.0100681139156222);

\path [draw=darkgray176, semithick]
(axis cs:2.16666666666667,0.0101350694894791)
--(axis cs:2.16666666666667,0.0116130458191037);

\addplot [semithick, darkgray176, mark=-, mark size=2.5, mark options={solid}, only marks]
table {%
0.166666666666667 0.00851487554609776
1.16666666666667 0.00827652681618929
2.16666666666667 0.0101350694894791
};
\addplot [semithick, darkgray176, mark=-, mark size=2.5, mark options={solid}, only marks]
table {%
0.166666666666667 0.0102489609271288
1.16666666666667 0.0100681139156222
2.16666666666667 0.0116130458191037
};

\nextgroupplot[
axis line style={darkgray176},
tick align=outside,
tick pos=left,
title={One-step},
x grid style={darkgray176},
xmajorgrids,
xmin=-0.5, xmax=2.5,
xtick style={color=darkgray176},
xtick={0,1,2},
xticklabels={10400,5200,2080},
xticklabel style={rotate=70},
y grid style={darkgray176},
ymajorgrids,
ymin=0, ymax=0.020,
ytick style={color=darkgray176}
]
\draw[draw=none,fill=darkslategray387083,fill opacity=0.85] (axis cs:-0.25,0) rectangle (axis cs:0,0.00375463627278805);
\draw[draw=none,fill=darkslategray387083,fill opacity=0.85] (axis cs:0.75,0) rectangle (axis cs:1,0.00536326132714748);
\draw[draw=none,fill=darkslategray387083,fill opacity=0.85] (axis cs:1.75,0) rectangle (axis cs:2,0.00930879730731249);
\draw[draw=none,fill=tomato23111181,fill opacity=0.85] (axis cs:0,0) rectangle (axis cs:0.25,0.00352552028683325);
\draw[draw=none,fill=tomato23111181,fill opacity=0.85] (axis cs:1,0) rectangle (axis cs:1.25,0.00481796295692523);
\draw[draw=none,fill=tomato23111181,fill opacity=0.85] (axis cs:2,0) rectangle (axis cs:2.25,0.00748796761035919);
\path [draw=darkgray176, semithick]
(axis cs:-0.125,0.00365053952712522)
--(axis cs:-0.125,0.00385873301845087);

\path [draw=darkgray176, semithick]
(axis cs:0.875,0.00521129983448025)
--(axis cs:0.875,0.00551522281981471);

\path [draw=darkgray176, semithick]
(axis cs:1.875,0.00866671938656042)
--(axis cs:1.875,0.00995087522806456);

\addplot [semithick, darkgray176, mark=-, mark size=2.5, mark options={solid}, only marks]
table {%
-0.125 0.00365053952712522
0.875 0.00521129983448025
1.875 0.00866671938656042
};
\addplot [semithick, darkgray176, mark=-, mark size=2.5, mark options={solid}, only marks]
table {%
-0.125 0.00385873301845087
0.875 0.00551522281981471
1.875 0.00995087522806456
};
\path [draw=darkgray176, semithick]
(axis cs:0.125,0.00350685575591489)
--(axis cs:0.125,0.0035441848177516);

\path [draw=darkgray176, semithick]
(axis cs:1.125,0.00478344830091218)
--(axis cs:1.125,0.00485247761293829);

\path [draw=darkgray176, semithick]
(axis cs:2.125,0.0074067950706876)
--(axis cs:2.125,0.00756914015003078);

\addplot [semithick, darkgray176, mark=-, mark size=2.5, mark options={solid}, only marks]
table {%
0.125 0.00350685575591489
1.125 0.00478344830091218
2.125 0.0074067950706876
};
\addplot [semithick, darkgray176, mark=-, mark size=2.5, mark options={solid}, only marks]
table {%
0.125 0.0035441848177516
1.125 0.00485247761293829
2.125 0.00756914015003078
};
\end{groupplot};

\draw ({$(current bounding box.south west)!-0.01!(current bounding box.south east)$}|-{$(current bounding box.south west)!0.5!(current bounding box.north west)$}) node[
  scale=0.9,
  anchor=west,
  text=darkgray176,
  rotate=90.0
]{MSE};
\draw ({$(current bounding box.south west)!0.5!(current bounding box.south east)$}|-{$(current bounding box.south west)!-0.05!(current bounding box.north west)$}) node[
  scale=0.9,
  anchor=south,
  text=darkgray176,
  rotate=0.0
]{Num. Train Trajectories};
\end{tikzpicture}

%% file: tikz/weather/mainbar_resnet_2e-4_hist=2.tex
\begin{tikzpicture}

\definecolor{burlywood233196106}{RGB}{233,196,106}
\definecolor{darkgray176}{RGB}{156,156,156}
\definecolor{lightgray204}{RGB}{204,204,204}
\definecolor{lightseagreen42157143}{RGB}{42,157,143}
\definecolor{sandybrown24416297}{RGB}{244,162,97}
\definecolor{darkslategray387083}{RGB}{38,70,83}
\definecolor{lightgray204}{RGB}{204,204,204}
\definecolor{tomato23111181}{RGB}{231,111,81}

\begin{groupplot}[group style={group size=2 by 2}, height=0.18\textheight, width=0.6\textwidth, 
yticklabel style={color=darkgray176, font=\bfseries\scriptsize}, xticklabel style={color=darkgray176, font=\bfseries\scriptsize},]
\nextgroupplot[
axis line style={darkgray176},
legend cell align={left},
legend style={
  fill opacity=0.8,
  draw opacity=1,
  text opacity=1,
  text=darkgray176,
  at={(0.03,1.09)},
  anchor=north west,
  draw=lightgray204,
  font=\bfseries\scriptsize,  
},
tick align=outside,
tick pos=left,
title={Rollout},
x grid style={darkgray176},
xmajorgrids,
xmin=-0.5, xmax=3.5,
xtick style={color=darkgray176},
xticklabels={},
y grid style={darkgray176},
ymajorgrids,
ymin=0, ymax=1.24,
ytick style={color=darkgray176}
]
\draw[draw=none,fill=lightseagreen42157143,fill opacity=0.85] (axis cs:-0.25,0) rectangle (axis cs:-0.0833333333333333,0.44796983897686);
\addlegendimage{ybar,ybar legend,draw=none,fill=lightseagreen42157143,fill opacity=0.85}
\addlegendentry{ResNet}

\draw[draw=none,fill=lightseagreen42157143,fill opacity=0.85] (axis cs:0.75,0) rectangle (axis cs:0.916666666666667,0.478021115064621);
\draw[draw=none,fill=lightseagreen42157143,fill opacity=0.85] (axis cs:1.75,0) rectangle (axis cs:1.91666666666667,0.490442663431168);
\draw[draw=none,fill=lightseagreen42157143,fill opacity=0.85] (axis cs:2.75,0) rectangle (axis cs:2.91666666666667,0.708285570144653);
\draw[draw=none,fill=burlywood233196106,fill opacity=0.85] (axis cs:-0.0833333333333333,0) rectangle (axis cs:0.0833333333333333,0.281607732176781);
\addlegendimage{ybar,ybar legend,draw=none,fill=burlywood233196106,fill opacity=0.85}
\addlegendentry{CResNet}

\draw[draw=none,fill=burlywood233196106,fill opacity=0.85] (axis cs:0.916666666666667,0) rectangle (axis cs:1.08333333333333,0.284194797277451);
\draw[draw=none,fill=burlywood233196106,fill opacity=0.85] (axis cs:1.91666666666667,0) rectangle (axis cs:2.08333333333333,0.524738281965256);
\draw[draw=none,fill=burlywood233196106,fill opacity=0.85] (axis cs:2.91666666666667,0) rectangle (axis cs:3.08333333333333,1.17265355587006);
\draw[draw=none,fill=sandybrown24416297,fill opacity=0.85] (axis cs:0.0833333333333333,0) rectangle (axis cs:0.25,0.216447204351425);
\addlegendimage{ybar,ybar legend,draw=none,fill=sandybrown24416297,fill opacity=0.85}
\addlegendentry{CResNet$_{\text{rot}}$}

\draw[draw=none,fill=sandybrown24416297,fill opacity=0.85] (axis cs:1.08333333333333,0) rectangle (axis cs:1.25,0.236994981765747);
\draw[draw=none,fill=sandybrown24416297,fill opacity=0.85] (axis cs:2.08333333333333,0) rectangle (axis cs:2.25,0.309622228145599);
\draw[draw=none,fill=sandybrown24416297,fill opacity=0.85] (axis cs:3.08333333333333,0) rectangle (axis cs:3.25,0.612715363502502);
\path [draw=darkgray176, semithick]
(axis cs:-0.166666666666667,0.442171990871429)
--(axis cs:-0.166666666666667,0.453767687082291);

\path [draw=darkgray176, semithick]
(axis cs:0.833333333333333,0.471808791160583)
--(axis cs:0.833333333333333,0.484233438968658);

\path [draw=darkgray176, semithick]
(axis cs:1.83333333333333,0.482428193092346)
--(axis cs:1.83333333333333,0.498457133769989);

\path [draw=darkgray176, semithick]
(axis cs:2.83333333333333,0.696405410766602)
--(axis cs:2.83333333333333,0.720165729522705);

\addplot [semithick, darkgray176, mark=-, mark size=2.5, mark options={solid}, only marks]
table {%
-0.166666666666667 0.442171990871429
0.833333333333333 0.471808791160583
1.83333333333333 0.482428193092346
2.83333333333333 0.696405410766602
};

\addplot [semithick, darkgray176, mark=-, mark size=2.5, mark options={solid}, only marks]
table {%
-0.166666666666667 0.453767687082291
0.833333333333333 0.484233438968658
1.83333333333333 0.498457133769989
2.83333333333333 0.720165729522705
};

\path [draw=darkgray176, semithick]
(axis cs:0,0.275118321180344)
--(axis cs:0,0.288097143173218);

\path [draw=darkgray176, semithick]
(axis cs:1,0.277455687522888)
--(axis cs:1,0.290933907032013);

\path [draw=darkgray176, semithick]
(axis cs:2,0.514678657054901)
--(axis cs:2,0.53479790687561);

\path [draw=darkgray176, semithick]
(axis cs:3,1.12437319755554)
--(axis cs:3,1.22093391418457);

\addplot [semithick, darkgray176, mark=-, mark size=2.5, mark options={solid}, only marks]
table {%
0 0.275118321180344
1 0.277455687522888
2 0.514678657054901
3 1.12437319755554
};

\addplot [semithick, darkgray176, mark=-, mark size=2.5, mark options={solid}, only marks]
table {%
0 0.288097143173218
1 0.290933907032013
2 0.53479790687561
3 1.22093391418457
};

\path [draw=darkgray176, semithick]
(axis cs:0.166666666666667,0.209446549415588)
--(axis cs:0.166666666666667,0.223447859287262);

\path [draw=darkgray176, semithick]
(axis cs:1.16666666666667,0.236994981765747)
--(axis cs:1.16666666666667,0.236994981765747);

\path [draw=darkgray176, semithick]
(axis cs:2.16666666666667,0.308647722005844)
--(axis cs:2.16666666666667,0.310596734285355);

\path [draw=darkgray176, semithick]
(axis cs:3.16666666666667,0.610884606838226)
--(axis cs:3.16666666666667,0.614546120166779);

\addplot [semithick, darkgray176, mark=-, mark size=2.5, mark options={solid}, only marks]
table {%
0.166666666666667 0.209446549415588
1.16666666666667 0.236994981765747
2.16666666666667 0.308647722005844
3.16666666666667 0.610884606838226
};

\addplot [semithick, darkgray176, mark=-, mark size=2.5, mark options={solid}, only marks]
table {%
0.166666666666667 0.223447859287262
1.16666666666667 0.236994981765747
2.16666666666667 0.310596734285355
3.16666666666667 0.614546120166779
};

\nextgroupplot[
axis line style={darkgray176},
legend cell align={left},
legend style={
  fill opacity=0.8,
  draw opacity=1,
  text=darkgray176,
  text opacity=1,
  at={(0.03,1.09)},
  anchor=north west,
  draw=lightgray204,
  font=\bfseries\scriptsize
},
tick align=outside,
tick pos=left,
title={Rollout},
x grid style={darkgray176},
xmajorgrids,
xmin=-0.5, xmax=3.5,
xtick style={color=darkgray176},
xticklabels={},
y grid style={darkgray176},
ymajorgrids,
ymin=0, ymax=1.24,
ytick style={color=darkgray176}
]
\draw[draw=none,fill=darkslategray387083,fill opacity=0.85] (axis cs:-0.25,0) rectangle (axis cs:0,0.0404472816735506);
\addlegendimage{ybar,ybar legend,draw=none,fill=darkslategray387083,fill opacity=0.85}
\addlegendentry{FNO}

\draw[draw=none,fill=darkslategray387083,fill opacity=0.85] (axis cs:0.75,0) rectangle (axis cs:1,0.0651295781135559);
\draw[draw=none,fill=darkslategray387083,fill opacity=0.85] (axis cs:1.75,0) rectangle (axis cs:2,0.144377276301384);
\draw[draw=none,fill=darkslategray387083,fill opacity=0.85] (axis cs:2.75,0) rectangle (axis cs:3,0.80318542321523);
\draw[draw=none,fill=tomato23111181,fill opacity=0.85] (axis cs:0,0) rectangle (axis cs:0.25,0.0314969004442294);
\addlegendimage{ybar,ybar legend,draw=none,fill=tomato23111181,fill opacity=0.85}
\addlegendentry{CFNO}

\draw[draw=none,fill=tomato23111181,fill opacity=0.85] (axis cs:1,0) rectangle (axis cs:1.25,0.0454886717100938);
\draw[draw=none,fill=tomato23111181,fill opacity=0.85] (axis cs:2,0) rectangle (axis cs:2.25,0.0941112736860911);
\draw[draw=none,fill=tomato23111181,fill opacity=0.85] (axis cs:3,0) rectangle (axis cs:3.25,0.432298913598061);
\path [draw=darkgray176, semithick]
(axis cs:-0.125,0.0399861969053745)
--(axis cs:-0.125,0.0409083664417267);

\path [draw=darkgray176, semithick]
(axis cs:0.875,0.0637602210044861)
--(axis cs:0.875,0.0664989352226257);

\path [draw=darkgray176, semithick]
(axis cs:1.875,0.141744792461395)
--(axis cs:1.875,0.147009760141373);

\path [draw=darkgray176, semithick]
(axis cs:2.875,0.79883819854832)
--(axis cs:2.875,0.807532647882141);

\addplot [semithick, darkgray176, mark=-, mark size=2.5, mark options={solid}, only marks]
table {%
-0.125 0.0399861969053745
0.875 0.0637602210044861
1.875 0.141744792461395
2.875 0.79883819854832
};

\addplot [semithick, darkgray176, mark=-, mark size=2.5, mark options={solid}, only marks]
table {%
-0.125 0.0409083664417267
0.875 0.0664989352226257
1.875 0.147009760141373
2.875 0.807532647882141
};

\path [draw=darkgray176, semithick]
(axis cs:0.125,0.0310836300189506)
--(axis cs:0.125,0.0319101708695083);

\path [draw=darkgray176, semithick]
(axis cs:1.125,0.0446319874578019)
--(axis cs:1.125,0.0463453559623857);

\path [draw=darkgray176, semithick]
(axis cs:2.125,0.0919993273599173)
--(axis cs:2.125,0.096223220012265);

\path [draw=darkgray176, semithick]
(axis cs:3.125,0.42769256234169)
--(axis cs:3.125,0.436905264854431);

\addplot [semithick, darkgray176, mark=-, mark size=2.5, mark options={solid}, only marks]
table {%
0.125 0.0310836300189506
1.125 0.0446319874578019
2.125 0.0919993273599173
3.125 0.42769256234169
};

\addplot [semithick, darkgray176, mark=-, mark size=2.5, mark options={solid}, only marks]
table {%
0.125 0.0319101708695083
1.125 0.0463453559623857
2.125 0.096223220012265
3.125 0.436905264854431
};

\nextgroupplot[
axis line style={darkgray176},
tick align=outside,
tick pos=left,
title={One-step},
x grid style={darkgray176},
xmajorgrids,
xmin=-0.5, xmax=3.5,
xtick style={color=darkgray176},
xtick={0,1,2,3},
xticklabels={2048,1792,896,448},
xticklabel style={rotate=70},
y grid style={darkgray176},
ymajorgrids,
ymin=0, ymax=0.072,
ytick style={color=darkgray176}
]
\draw[draw=none,fill=lightseagreen42157143,fill opacity=0.85] (axis cs:-0.25,0) rectangle (axis cs:-0.0833333333333333,0.0180717324838042);
\draw[draw=none,fill=lightseagreen42157143,fill opacity=0.85] (axis cs:0.75,0) rectangle (axis cs:0.916666666666667,0.0184197016060352);
\draw[draw=none,fill=lightseagreen42157143,fill opacity=0.85] (axis cs:1.75,0) rectangle (axis cs:1.91666666666667,0.0241788029670715);
\draw[draw=none,fill=lightseagreen42157143,fill opacity=0.85] (axis cs:2.75,0) rectangle (axis cs:2.91666666666667,0.038529496639967);
\draw[draw=none,fill=burlywood233196106,fill opacity=0.85] (axis cs:-0.0833333333333333,0) rectangle (axis cs:0.0833333333333333,0.011072451248765);
\draw[draw=none,fill=burlywood233196106,fill opacity=0.85] (axis cs:0.916666666666667,0) rectangle (axis cs:1.08333333333333,0.01107277860865);
\draw[draw=none,fill=burlywood233196106,fill opacity=0.85] (axis cs:1.91666666666667,0) rectangle (axis cs:2.08333333333333,0.0278224851936102);
\draw[draw=none,fill=burlywood233196106,fill opacity=0.85] (axis cs:2.91666666666667,0) rectangle (axis cs:3.08333333333333,0.0685449503362179);
\draw[draw=none,fill=sandybrown24416297,fill opacity=0.85] (axis cs:0.0833333333333333,0) rectangle (axis cs:0.25,0.00749733252450824);
\draw[draw=none,fill=sandybrown24416297,fill opacity=0.85] (axis cs:1.08333333333333,0) rectangle (axis cs:1.25,0.00688420748338103);
\draw[draw=none,fill=sandybrown24416297,fill opacity=0.85] (axis cs:2.08333333333333,0) rectangle (axis cs:2.25,0.0161148635670543);
\draw[draw=none,fill=sandybrown24416297,fill opacity=0.85] (axis cs:3.08333333333333,0) rectangle (axis cs:3.25,0.0335362777113914);
\path [draw=darkgray176, semithick]
(axis cs:-0.166666666666667,0.017560001462698)
--(axis cs:-0.166666666666667,0.0185834635049105);

\path [draw=darkgray176, semithick]
(axis cs:0.833333333333333,0.0173870921134949)
--(axis cs:0.833333333333333,0.0194523110985756);

\path [draw=darkgray176, semithick]
(axis cs:1.83333333333333,0.0238740108907223)
--(axis cs:1.83333333333333,0.0244835950434208);

\path [draw=darkgray176, semithick]
(axis cs:2.83333333333333,0.0375280454754829)
--(axis cs:2.83333333333333,0.039530947804451);

\addplot [semithick, darkgray176, mark=-, mark size=2.5, mark options={solid}, only marks]
table {%
-0.166666666666667 0.017560001462698
0.833333333333333 0.0173870921134949
1.83333333333333 0.0238740108907223
2.83333333333333 0.0375280454754829
};
\addplot [semithick, darkgray176, mark=-, mark size=2.5, mark options={solid}, only marks]
table {%
-0.166666666666667 0.0185834635049105
0.833333333333333 0.0194523110985756
1.83333333333333 0.0244835950434208
2.83333333333333 0.039530947804451
};
\path [draw=darkgray176, semithick]
(axis cs:0,0.0107529880478978)
--(axis cs:0,0.0113919144496322);

\path [draw=darkgray176, semithick]
(axis cs:1,0.0110111711546779)
--(axis cs:1,0.0111343860626221);

\path [draw=darkgray176, semithick]
(axis cs:2,0.0271749682724476)
--(axis cs:2,0.0284700021147728);

\path [draw=darkgray176, semithick]
(axis cs:3,0.0655909776687622)
--(axis cs:3,0.0714989230036736);

\addplot [semithick, darkgray176, mark=-, mark size=2.5, mark options={solid}, only marks]
table {%
0 0.0107529880478978
1 0.0110111711546779
2 0.0271749682724476
3 0.0655909776687622
};
\addplot [semithick, darkgray176, mark=-, mark size=2.5, mark options={solid}, only marks]
table {%
0 0.0113919144496322
1 0.0111343860626221
2 0.0284700021147728
3 0.0714989230036736
};
\path [draw=darkgray176, semithick]
(axis cs:0.166666666666667,0.00637686531990767)
--(axis cs:0.166666666666667,0.00861779972910881);

\path [draw=darkgray176, semithick]
(axis cs:1.16666666666667,0.00688420748338103)
--(axis cs:1.16666666666667,0.00688420748338103);

\path [draw=darkgray176, semithick]
(axis cs:2.16666666666667,0.0160245858132839)
--(axis cs:2.16666666666667,0.0162051413208246);

\path [draw=darkgray176, semithick]
(axis cs:3.16666666666667,0.0333758853375912)
--(axis cs:3.16666666666667,0.0336966700851917);

\addplot [semithick, darkgray176, mark=-, mark size=2.5, mark options={solid}, only marks]
table {%
0.166666666666667 0.00637686531990767
1.16666666666667 0.00688420748338103
2.16666666666667 0.0160245858132839
3.16666666666667 0.0333758853375912
};
\addplot [semithick, darkgray176, mark=-, mark size=2.5, mark options={solid}, only marks]
table {%
0.166666666666667 0.00861779972910881
1.16666666666667 0.00688420748338103
2.16666666666667 0.0162051413208246
3.16666666666667 0.0336966700851917
};

\nextgroupplot[
axis line style={darkgray176},
tick align=outside,
tick pos=left,
title={One-step},
x grid style={darkgray176},
xmajorgrids,
xmin=-0.5, xmax=3.5,
xtick style={color=darkgray176},
xtick={0,1,2,3},
xticklabels={896,448,192,56},
xticklabel style={rotate=70},
y grid style={darkgray176},
ymajorgrids,
ymin=0, ymax=0.072,
ytick style={color=darkgray176}
]
\draw[draw=none,fill=darkslategray387083,fill opacity=0.85] (axis cs:-0.25,0) rectangle (axis cs:0,0.0019612645264715);
\draw[draw=none,fill=darkslategray387083,fill opacity=0.85] (axis cs:0.75,0) rectangle (axis cs:1,0.00335354392882437);
\draw[draw=none,fill=darkslategray387083,fill opacity=0.85] (axis cs:1.75,0) rectangle (axis cs:2,0.00773894088342786);
\draw[draw=none,fill=darkslategray387083,fill opacity=0.85] (axis cs:2.75,0) rectangle (axis cs:3,0.0615671922763189);
\draw[draw=none,fill=tomato23111181,fill opacity=0.85] (axis cs:0,0) rectangle (axis cs:0.25,0.00169398433839281);
\draw[draw=none,fill=tomato23111181,fill opacity=0.85] (axis cs:1,0) rectangle (axis cs:1.25,0.00256044402097662);
\draw[draw=none,fill=tomato23111181,fill opacity=0.85] (axis cs:2,0) rectangle (axis cs:2.25,0.00526664933810631);
\draw[draw=none,fill=tomato23111181,fill opacity=0.85] (axis cs:3,0) rectangle (axis cs:3.25,0.0249517727643251);
\path [draw=darkgray176, semithick]
(axis cs:-0.125,0.00193743291310966)
--(axis cs:-0.125,0.00198509613983333);

\path [draw=darkgray176, semithick]
(axis cs:0.875,0.00322146224789321)
--(axis cs:0.875,0.00348562560975552);

\path [draw=darkgray176, semithick]
(axis cs:1.875,0.00744888000190258)
--(axis cs:1.875,0.00802900176495314);

\path [draw=darkgray176, semithick]
(axis cs:2.875,0.0593614324910978)
--(axis cs:2.875,0.0637729520615399);

\addplot [semithick, darkgray176, mark=-, mark size=2.5, mark options={solid}, only marks]
table {%
-0.125 0.00193743291310966
0.875 0.00322146224789321
1.875 0.00744888000190258
2.875 0.0593614324910978
};
\addplot [semithick, darkgray176, mark=-, mark size=2.5, mark options={solid}, only marks]
table {%
-0.125 0.00198509613983333
0.875 0.00348562560975552
1.875 0.00802900176495314
2.875 0.0637729520615399
};
\path [draw=darkgray176, semithick]
(axis cs:0.125,0.0016719076421672)
--(axis cs:0.125,0.00171606103461843);

\path [draw=darkgray176, semithick]
(axis cs:1.125,0.00252204231508529)
--(axis cs:1.125,0.00259884572686795);

\path [draw=darkgray176, semithick]
(axis cs:2.125,0.00517335413449837)
--(axis cs:2.125,0.00535994454171426);

\path [draw=darkgray176, semithick]
(axis cs:3.125,0.0242259856313467)
--(axis cs:3.125,0.0256775598973036);

\addplot [semithick, darkgray176, mark=-, mark size=2.5, mark options={solid}, only marks]
table {%
0.125 0.0016719076421672
1.125 0.00252204231508529
2.125 0.00517335413449837
3.125 0.0242259856313467
};
\addplot [semithick, darkgray176, mark=-, mark size=2.5, mark options={solid}, only marks]
table {%
0.125 0.00171606103461843
1.125 0.00259884572686795
2.125 0.00535994454171426
3.125 0.0256775598973036
};
\end{groupplot}

\draw ({$(current bounding box.south west)!-0.01!(current bounding box.south east)$}|-{$(current bounding box.south west)!0.5!(current bounding box.north west)$}) node[
  scale=0.9,
  anchor=west,
  text=darkgray176,
  rotate=90.0
]{MSE};
\draw ({$(current bounding box.south west)!0.5!(current bounding box.south east)$}|-{$(current bounding box.south west)!-0.05!(current bounding box.north west)$}) node[
  scale=0.9,
  anchor=south,
  text=darkgray176,
  rotate=0.0
]{Num. Train Trajectories};
\end{tikzpicture}

%% file: tikz/maxwell/mainbar_fourier_1e-4.tex
\begin{tikzpicture}

\definecolor{darkgray176}{RGB}{156,156,156}
\definecolor{darkslategray387083}{RGB}{38,70,83}
\definecolor{lightgray204}{RGB}{204,204,204}
\definecolor{tomato23111181}{RGB}{231,111,81}

\begin{groupplot}[group style={group size=1 by 2},height=0.2\textheight, width=0.35\textwidth, yticklabel style={color=darkgray176, font=\bfseries\scriptsize}, xticklabel style={color=darkgray176, font=\bfseries\scriptsize},]
\nextgroupplot[
axis line style={darkgray176},
legend cell align={left},
legend style={
  fill opacity=0.8,
  draw opacity=1,
  text=darkgray176,
  text opacity=1,
  at={(0.03,0.97)},
  anchor=north west,
  draw=lightgray204,
  font=\bfseries\scriptsize,
},
tick align=outside,
tick pos=left,
title={Rollout},
x grid style={darkgray176},
xmajorgrids,
xmin=-0.5, xmax=3.5,
xtick style={color=darkgray176},
xtick={0,1,2,3},
xticklabels={},
y grid style={darkgray176},
ymajorgrids,
ymin=0, ymax=0.019,
ytick style={color=darkgray176}
]
\draw[draw=none,fill=darkslategray387083,fill opacity=0.85] (axis cs:-0.25,0) rectangle (axis cs:0,0.000798066088464111);
\addlegendimage{ybar,ybar legend,draw=none,fill=darkslategray387083,fill opacity=0.85}
\addlegendentry{FNO}

\draw[draw=none,fill=darkslategray387083,fill opacity=0.85] (axis cs:0.75,0) rectangle (axis cs:1,0.00202110704655449);
\draw[draw=none,fill=darkslategray387083,fill opacity=0.85] (axis cs:1.75,0) rectangle (axis cs:2,0.00680698361247778);
\draw[draw=none,fill=darkslategray387083,fill opacity=0.85] (axis cs:2.75,0) rectangle (axis cs:3,0.0185260819271207);
\draw[draw=none,fill=tomato23111181,fill opacity=0.85] (axis cs:0,0) rectangle (axis cs:0.25,0.000755452434532344);
\addlegendimage{ybar,ybar legend,draw=none,fill=tomato23111181,fill opacity=0.85}
\addlegendentry{CFNO}

\draw[draw=none,fill=tomato23111181,fill opacity=0.85] (axis cs:1,0) rectangle (axis cs:1.25,0.00150580428695927);
\draw[draw=none,fill=tomato23111181,fill opacity=0.85] (axis cs:2,0) rectangle (axis cs:2.25,0.00282998313196003);
\draw[draw=none,fill=tomato23111181,fill opacity=0.85] (axis cs:3,0) rectangle (axis cs:3.25,0.00529428594745696);
\path [draw=darkgray176, semithick]
(axis cs:-0.125,0.000768150435760617)
--(axis cs:-0.125,0.000827981741167605);

\path [draw=darkgray176, semithick]
(axis cs:0.875,0.00194724577447932)
--(axis cs:0.875,0.00209496831862966);

\path [draw=darkgray176, semithick]
(axis cs:1.875,0.00662089185789227)
--(axis cs:1.875,0.00699307536706328);

\path [draw=darkgray176, semithick]
(axis cs:2.875,0.0184044819325209)
--(axis cs:2.875,0.0186476819217205);

\addplot [semithick, darkgray176, mark=-, mark size=2.5, mark options={solid}, only marks]
table {%
-0.125 0.000768150435760617
0.875 0.00194724577447932
1.875 0.00662089185789227
2.875 0.0184044819325209
};

\addplot [semithick, darkgray176, mark=-, mark size=2.5, mark options={solid}, only marks]
table {%
-0.125 0.000827981741167605
0.875 0.00209496831862966
1.875 0.00699307536706328
2.875 0.0186476819217205
};

\path [draw=darkgray176, semithick]
(axis cs:0.125,0.000755452434532344)
--(axis cs:0.125,0.000755452434532344);

\path [draw=darkgray176, semithick]
(axis cs:1.125,0.00136982230374258)
--(axis cs:1.125,0.00164178627017596);

\path [draw=darkgray176, semithick]
(axis cs:2.125,0.0026701046153903)
--(axis cs:2.125,0.00298986164852977);

\path [draw=darkgray176, semithick]
(axis cs:3.125,0.00486813159659505)
--(axis cs:3.125,0.00572044029831886);

\addplot [semithick, darkgray176, mark=-, mark size=2.5, mark options={solid}, only marks]
table {%
0.125 0.000755452434532344
1.125 0.00136982230374258
2.125 0.0026701046153903
3.125 0.00486813159659505
};

\addplot [semithick, darkgray176, mark=-, mark size=2.5, mark options={solid}, only marks]
table {%
0.125 0.000755452434532344
1.125 0.00164178627017596
2.125 0.00298986164852977
3.125 0.00572044029831886
};

\nextgroupplot[
axis line style={darkgray176},
tick align=outside,
tick pos=left,
title={One-step},
x grid style={darkgray176},
xmajorgrids,
xmin=-0.5, xmax=3.5,
xtick style={color=darkgray176},
xtick={0,1,2,3},
xticklabels={6400,3200,1280,640},
y grid style={darkgray176},
ymajorgrids,
ymin=0, ymax=0.00538419187068939,
ytick style={color=darkgray176}
]
\draw[draw=none,fill=darkslategray387083,fill opacity=0.85] (axis cs:-0.25,0) rectangle (axis cs:0,0.00018444857414579);
\draw[draw=none,fill=darkslategray387083,fill opacity=0.85] (axis cs:0.75,0) rectangle (axis cs:1,0.000504025966317082);
\draw[draw=none,fill=darkslategray387083,fill opacity=0.85] (axis cs:1.75,0) rectangle (axis cs:2,0.00184029177762568);
\draw[draw=none,fill=darkslategray387083,fill opacity=0.85] (axis cs:2.75,0) rectangle (axis cs:3,0.00529624801129103);
\draw[draw=none,fill=tomato23111181,fill opacity=0.85] (axis cs:0,0) rectangle (axis cs:0.25,0.000170920335222036);
\draw[draw=none,fill=tomato23111181,fill opacity=0.85] (axis cs:1,0) rectangle (axis cs:1.25,0.000359014433342963);
\draw[draw=none,fill=tomato23111181,fill opacity=0.85] (axis cs:2,0) rectangle (axis cs:2.25,0.000679968594340608);
\draw[draw=none,fill=tomato23111181,fill opacity=0.85] (axis cs:3,0) rectangle (axis cs:3.25,0.00121880829101428);
\path [draw=darkgray176, semithick]
(axis cs:-0.125,0.000177403169800527)
--(axis cs:-0.125,0.000191493978491053);

\path [draw=darkgray176, semithick]
(axis cs:0.875,0.00047348665450292)
--(axis cs:0.875,0.000534565278131245);

\path [draw=darkgray176, semithick]
(axis cs:1.875,0.00179002247750759)
--(axis cs:1.875,0.00189056107774377);

\path [draw=darkgray176, semithick]
(axis cs:2.875,0.00520830415189266)
--(axis cs:2.875,0.00538419187068939);

\addplot [semithick, darkgray176, mark=-, mark size=2.5, mark options={solid}, only marks]
table {%
-0.125 0.000177403169800527
0.875 0.00047348665450292
1.875 0.00179002247750759
2.875 0.00520830415189266
};
\addplot [semithick, darkgray176, mark=-, mark size=2.5, mark options={solid}, only marks]
table {%
-0.125 0.000191493978491053
0.875 0.000534565278131245
1.875 0.00189056107774377
2.875 0.00538419187068939
};
\path [draw=darkgray176, semithick]
(axis cs:0.125,0.000170920335222036)
--(axis cs:0.125,0.000170920335222036);

\path [draw=darkgray176, semithick]
(axis cs:1.125,0.000309258094822485)
--(axis cs:1.125,0.000408770771863442);

\path [draw=darkgray176, semithick]
(axis cs:2.125,0.000632209703326225)
--(axis cs:2.125,0.00072772748535499);

\path [draw=darkgray176, semithick]
(axis cs:3.125,0.00108130427543074)
--(axis cs:3.125,0.00135631230659783);

\addplot [semithick, darkgray176, mark=-, mark size=2.5, mark options={solid}, only marks]
table {%
0.125 0.000170920335222036
1.125 0.000309258094822485
2.125 0.000632209703326225
3.125 0.00108130427543074
};
\addplot [semithick, darkgray176, mark=-, mark size=2.5, mark options={solid}, only marks]
table {%
0.125 0.000170920335222036
1.125 0.000408770771863442
2.125 0.00072772748535499
3.125 0.00135631230659783
};
\end{groupplot}

\draw ({$(current bounding box.south west)!-0.02!(current bounding box.south east)$}|-{$(current bounding box.south west)!0.375!(current bounding box.north west)$}) node[
  scale=0.9,
  anchor=west,
  text=darkgray176,
  rotate=90.0
]{MSE};
\draw ({$(current bounding box.south west)!0.5!(current bounding box.south east)$}|-{$(current bounding box.south west)!-0.10!(current bounding box.north west)$}) node[
  scale=0.9,
  anchor=south,
  text=darkgray176,
  rotate=0.0
]{Num. Train Trajectories};
\end{tikzpicture}

%% file: sections/conclusion.tex
\section{Conclusion}
We introduced Clifford neural layers that handle the various scalar (e.g. charge density), vector (e.g. electric field), bivector (magnetic field) and higher order fields as proper geometric objects organized as multivectors.
This geometric algebra perspective allowed us to naturally generalize convolution and Fourier transformations to their Clifford counterparts, providing an elegant rule to design new neural network layers.
The multivector viewpoint denotes an inductive bias advantage, leading to a better representation of the relationship between fields and their individual components, which is prominently demonstrated by the fact that our Clifford layers significantly outperformed equivalent standard neural PDE surrogates. 

\textbf{Limitations.} One limitation is the current speed of Fast Fourier Transform (FFT) operations on machine learning accelerators like GPUs. 
While an active area of research, current available versions of \texttt{cuFFT}\footnote{\url{https://developer.nvidia.com/cufft}} kernels wrapped in \texttt{PyTorch}~\citep{pytorch} are not yet as heavily optimized\footnote{For alternative efficient GPU-accelerated multidimensional FFT libraries see e.g. \url{https://github.com/DTolm/VkFFT}}, especially for the gradient pass. 
In contrast to FNO layers, 
which operate on real-valued signals,
Clifford Fourier layers use complex-valued FFT operations where the backward pass is approximately twice as slow. For similar parameter counts, inference times of FNO and CFNO networks are similar. 
Similar to~\citet{grassucci2021lightweight} who
investigated the speed of geometric convolution layers, we found that Clifford convolutions are more parameter efficient since they share parameters among filters, with the downside that the net number of operations is larger, resulting in increased training times by a factor of about $2$.
Finally, from a PDE point of view, the presented approaches to obtain PDE surrogates are limited since the neural networks have to be retrained for different equation parameters or e.g. different $\Delta t$.

\textbf{Future work.} Besides modeling of PDEs, weather, and fluid dynamics, we see potential applications of Clifford layers
for e.g. MRI or radar data, and for
neural implicit representations~\citep{xie2022neural,rella2022neural}. Extensions towards graph networks and attention based models will be useful to explore. 
Furthermore, custom multivector GPU kernels can overcome many of the speed issues as the compute density of Clifford operations is much higher which is better for hardware accelerators~\citep{hoffmann2020algebranets}.
The use of a just-in-time compiled language with better array abstractions like Julia~\citep{bezanson2017julia} could significantly simplify the interface.
Finally, combining the ideas of multivector modeling together with various physics-informed neural network approaches~\citep{raissi2019physics,lutter2018deep,gupta2019general,cranmer2020lagrangian,zubov2021neuralpde} is an attractive next step.

%% file: appendix/cliffordalgebra.tex
\section{Mathematical background}\label{app:clifford_algebras}

This appendix supports Section~\ref{sec:background} of the main paper. We give a more detailed explanation of real Clifford algebras and have a closer look at $Cl_{2, 0}(\R)$, $Cl_{0, 2}(\R)$, and $Cl_{3, 0}(\R)$. For a detailed introduction into Clifford algebras we recommend~\citet{suter2003geometric, hestenes2003oersted, hestenes2012new, dorst2010geometric, renaud2020clifford} 

\subsection{Clifford algebras} 
\paragraph{Vector spaces and algebras over a field.}
A \emph{vector space} over a field $F$ is a set $V$ together with two binary operations that satisfy the axioms for vector addition and scalar multiplication. The axioms of addition ensure that if two elements of $V$ get added together, we end up with another element of $V$. The elements of $F$ are called scalars. Examples of a field $F$ are the real numbers $\R$ and the complex numbers $\CC$. Although it is common practice to refer to the elements of a general vector space $V$ as vectors, to avoid confusion we will reserve the usage of this term to the more specific case of elements of $\R^n$. As we will see below, general vector spaces can consist of more complicated, higher-order objects than scalars, vectors or matrices.

An \emph{algebra over a field} consists of a vector space $V$ over a field $F$ together with an additional \emph{bilinear} law of composition of elements of the vector space, $V \times V \rightarrow V$, that is, if $a$ and $b$ are any two elements of $V$, then $ab: V \times V \rightarrow V$ is an element of $V$, satisfying a pair of distribution laws: $a(\lambda_1 b+\lambda_2 c) = \lambda_1 a b + \lambda_2 a c$ and $(\lambda_1 a + \lambda_2 b) c = \lambda_1 a c + \lambda_2 b c$ for $\lambda_1,\lambda_2 \in F$ and $a, b, c \in V$. Note that general vector spaces don't have bilinear operations defined on their elements.

\paragraph{Clifford algebras over $\R$.} In this manuscript we will focus on Clifford algebras over $\R$. For a more general exposition on Clifford algebras over different fields the reader is referred to~\citet{lounesto1986clifford}.  

A real Clifford algebra is generated by the $n$-dimensional vector space $\R^n$ through a set of relations that hold for the basis elements of the vector space $\R^n$. 
Let us denote the basis elements of $\R^n$ with $e_1, ..., e_n$, and without loss of generality choose these basis elements to be mutually orthonormal. Taking two nonnegative integers $p$ and $q$, such that $p+q=n$, then a real Clifford algebra $Cl_{p,q}(\R)$ with the ``signature'' $(p,q)$, is generated through the following relations that define how the bilinear product of the algebra operates on the basis elements of $\R^n$:
\begin{align}
    e_i^2 &= +1 \ \ &&\mathrm{for} \ \  1 \leq i \leq p \ , \label{eq:app_clifford1}\\
    e_j^2 &= -1 \ \ &&\mathrm{for} \ \ p < j \leq p+q \ , \label{eq:app_clifford2}\\
    e_i e_j &= - e_j e_i \ \ &&\mathrm{for} \ \ i \neq j \label{eq:app_clifford3}\ .
\end{align}
Through these relations we can generate a basis for the vector space of the Clifford algebra, which we will denote with $G$.
Equations~\ref{eq:app_clifford1} and~\ref{eq:app_clifford2} show that the product between two vectors yields a scalar.
According to the aforementioned definition of an algebra over a field, a Clifford algebra with a vector space $G$ is equipped with a bilinear product $G\times G\mapsto G$, that combines two elements from the vector space $G$ and yields another element of the same space $G$. Therefore, both scalars and vectors must be elements of the vector space $G$. 
Equation~\ref{eq:app_clifford3} shows that besides scalar and vector elements, higher order elements consisting of a combination of two basis elements, such as $e_ie_j$ and $e_j e_i$, are also part of the vector space $G$. Finally, by combining Equations~\ref{eq:app_clifford1},~\ref{eq:app_clifford2},~\ref{eq:app_clifford3} we can create even higher order elements such as $e_i e_j e_k$ for $i\neq j\neq k$, or $e_1e_2...e_{p+q}$, which all must be part of the vector space $G$. 

In order to determine what the basis elements are that span the vector space $G$ of $Cl_{p,q}(\R)$, we note that elements $e_{\sigma(1)} e_{\sigma(2)}... e_{\sigma(k)}$ and $e_1 e_2... e_k$ are related through a simple scalar multiplicative factor of plus or minus one, depending on the sign of the permutation $\sigma$. Therefore, it suffices to consider the unordered combinations of basis elements of $\R^n$: the basis of the vector space $G$ is given by $\{1, e_1, e_2, ..., e_{p+q}, e_1e_2, ..., e_{p+q-1}e_{p+q}, ..., e_1e_2...e_{p+q}\}$. 

In summary, we have introduced two different vector spaces. First, the vector space $\R^n$ which \textit{generates} the Clifford algebra, and second the vector space $G$, which is the vector space spanned by the basis elements of the Clifford algebra $Cl_{p, q}(\R)$. Convention is to denote the vector space of a real Clifford algebra with a superscript $n$ of the dimension of the generating vector space, yielding $G^n$ for a generating vector space $\R^n$. Note that the dimension of the vector space $G^n$ is $2^n=2^{p+q}$. 

Exemplary low-dimensional Clifford algebras are:
(i) $Cl_{0,0}(\R)$ which is a one-dimensional algebra that is spanned by the vector $\{1\}$ and is therefore isomorphic to $\R$, the field of real numbers;
(ii) $Cl_{0,1}(\R)$ which is a two-dimensional algebra with vector space $G^1$ spanned by $\{1,e_1\}$ where the basis vector $e_1$ squares to $-1$, and is therefore isomorphic to $\CC$, the field of complex numbers; (iii) $Cl_{0,2}(\R)$ which is a 4-dimensional algebra with vector space $G^2$ spanned by $\{1,e_1,e_2,e_1e_2\}$, where $e_1,e_2$ square to $-1$ and anti-commute. Thus, $Cl_{0,2}(\R)$ is isomorphic to the quaternions $\HH$.

\begin{defn}[Grade of Clifford algebra element]{def:grade}
The grade of a Clifford algebra basis element is the dimension of the subspace it represents.
\end{defn}
For example, the basis elements $\{1,e_1,e_2,e_1e_2\}$ of the Clifford algebras $Cl_{0, 2}(\R)$ and $Cl_{2, 0}(\R)$ have the grades $\{0,1,1,2\}$.
Using the concept of grades, we can divide the vector spaces of Clifford algebras into linear subspaces made up of elements of each grade. The grade subspace of
smallest dimension is $M_0$, the subspace of all scalars (elements with 0 basis vectors).
Elements of $M_1$ are called vectors, elements of $M_2$ are
bivectors, and so on. In general, the vector space $G^{p+q}$ of a Clifford
algebra $Cl_{p, q}$ can be written as the direct sum of all of these subspaces:
\begin{align}
    G^{p+q} = M_0 \oplus M_1 \oplus \ldots \oplus M_{p+q} \ .
\end{align}
The elements of a Clifford algebra are called \textit{multivectors}, containing elements of subspaces, i.e. scalars, vectors, bivectors, trivectors etc.
The basis element with the highest grade is called the \textit{pseudoscalar}\footnote{In contrast to scalars, pseudoscalars change sign under reflection.}, which in $\R^2$ corresponds to the bivector $e_1e_2$, and in $\R^3$ to the trivector $e_1e_2e_3$. The pseudoscalar is often denoted with the symbol $i_{p+q}$. From hereon, only multivectors will be denoted with boldface symbols.

\paragraph{Geometric product.}
Using Equations~\ref{eq:app_clifford1},~\ref{eq:app_clifford2},~\ref{eq:app_clifford3}, we have seen how basis elements of the vector space $G^{p+q}$ of the Clifford algebra are formed using basis elements of the generating vector space $V$. We now, look at how elements of $G^{p+q}$ are combined, i.e. how multivectors are bilinearly operated on.
The \textit{geometric product} is the bilinear operation on multivectors in Clifford algebras. For arbitrary multivectors $\va$, $\vb$, $\vc \in G^{p+q}$, and scalar $\lambda$ the geometric product has the following properties:
\begin{align}
    &\va\vb \in G^{p+q} \ \ && \text{closure} \ , \\
    &(\va\vb)\vc = \va(\vb\vc) \ \ && \text{associativity} \ , \\
    &\lambda\va = \va\lambda \ \ && \text{commutative scalar multiplication} \ , \\
    &\va(\vb+\vc) = \va\vb + \va\vc \ \ && \text{distributive over addition} \ .
\end{align}
The geometric product is in general non-commutative, i.e. $\va\vb \neq \vb\va$. As we describe later, the geometric product is made up of two things: an inner product (that captures similarity) and exterior (wedge) product that captures difference.

\begin{defn}[Dual of a multivector]{def:app_dual}
The dual ${\va}^*$ of a multivector $\va$ is defined as:
\begin{align}
    \va^* = \va i_{p+q} \ ,
\end{align}
where $i_{p+q}$ represents the respective pseudoscalar of the Clifford algebra. 
\end{defn}

This definition allows us to relate different multivectors to each other, which is a useful property when defining Clifford Fourier transforms. For example, for Clifford algebras in $\R^2$ the dual of a scalar is a bivector, and for the Clifford algebra $\R^3$ the dual of a scalar is a trivector.

\subsection{Examples of low-dimensional Clifford algebras}
\subsubsection{Clifford algebra $Cl_{0, 1}(\R)$}
The Clifford algebra $Cl_{0, 1}(\R)$
is a two-dimensional algebra with vector space $G^1$ spanned by $\{1,e_1\}$, and where the basis vector $e_1$ squares to $-1$. $Cl_{0, 1}(\R)$ is thus algebra-isomorphic to $\CC$, the field of complex numbers. This becomes more obvious if we identify the basis element with the highest grade, i.e. $e_1$, as the pseudoscalar $i_1$ which is the imaginary part of the complex numbers.
The geometric product between two multivectors $\va = a_0 + a_1 e_1$ and $\vb = b_0 + b_1 e_1$ is therefore also isomorphic to the product of two complex numbers:
\begin{align}
    \va \vb \ = & \ a_0 b_0 + a_0 b_1 \eone + a_1 b_0 \eone + a_1 b_1 e_1 e_1 \nonumber \\
    = & \ ( a_0 b_0 - a_1 b_1) + (a_0 b_1 + a_1 b_0)\eone \ . 
\end{align}

\subsubsection{Clifford algebra $Cl_{2, 0}(\R)$}
The Clifford algebra $Cl_{2, 0}(\R)$ 
is a 4-dimensional algebra with vector space $G^2$ spanned by the basis vectors $\{1,e_1,e_2,e_1e_2\}$ where $e_1,e_2$ square to $+1$.  
The geometric product of two multivectors $\va = a_0 + a_1 e_1 + a_2 e_2 + a_{12}e_1e_2$ and $\vb = b_0 + b_1 e_1 + b_2 e_2 + b_{12}e_1e_2$ is defined via:
\begin{align}
    \va \vb \  &= \ a_0 b_0 + a_0 b_1 \eone + a_0 b_2 \etwo + a_0 b_{12}\eonetwo \nonumber \\ 
    & + a_1 b_0 \eone + a_1 b_1 e_1 e_1 + a_1 b_2 \eonetwo + a_1 b_{12}e_1 e_1 \etwo \nonumber \\
    & + a_2 b_0 \etwo + a_2 b_1 \etwoone + a_2 b_2 e_2 e_2 + a_2 b_{12}e_2\eone e_2 \nonumber \\
    & + a_{12}b_0 \eonetwo + a_{12}b_1 e_1\etwo e_1 + a_{12}b_2 \eone e_2e_2 + a_{12}b_{12} e_1e_2 e_1e_2 \ .
\end{align}

Using the relations $e_1 e_1 = 1$, $e_2 e_2 = 1$, and $e_i e_j = - e_j e_i$ for $i \neq j \in \{e_1, e_2\}$, from which it follows that $e_1e_2e_1e_2 = -1$, we obtain:
\begin{align}
    \va \vb \ = & \ a_0 b_0 + a_1 b_1 + a_2 b_2 - a_{12} b_{12} \nonumber \\
    & + (a_0 b_1 + a_1 b_0 - a_2 b_{12} + a_{12} b_2) \eone \nonumber \\
    & + (a_0 b_2 + a_1 b_{12} + a_2 b_0 - a_{12} b_1) \etwo \nonumber \\
    & + (a_0 b_{12} + a_1 b_2 - a_2 b_1 + a_{12} b_0) \eonetwo \ . \label{eq:app_geometric_product}
\end{align}

A vector $x \in \R^2 \subset G^2$ is identified with $x_1e_1 + x_2e_2 \in \R^2 \subset G^2$.
Clifford multiplication of two vectors $x, y \in \R^2 \subset G^2$ yields the geometric product $x y$:
\begin{align}
    x y &= (x_1 \eone + x_2 \etwo)(y_1 \eone + y_2 \etwo) \nonumber \\ 
    &= \eqnmarkbox[NavyBlue]{i_0}{x_1 y_1 e_1^2 + x_2 y_2 e_2^2} + \eqnmarkbox[Plum]{w_0}{x_1 y_2 e_1e_2 + x_2 y_1 e_2e_1} \nonumber \\
    &= \eqnmarkbox[NavyBlue]{i}{\langle x , y \rangle} + \eqnmarkbox[Plum]{w}{x \wedge y} \ \label{eq:app_geometric_product_vectors},
\end{align}
\annotate[xshift=-0.25em,yshift=-0.25em]{below,left}{i}{Inner product}
\annotate[yshift=-0.25em]{below}{w}{Outer/Wedge product}

The asymmetric quantity $x \wedge y = - y \wedge x$ is associated with the now often mentioned \textit{bivector}, which can be interpreted as an oriented plane segment.%

Equation~\ref{eq:app_geometric_product_vectors} can be rewritten to express the (symmetric) inner product and the (anti-symmetric) outer product in terms of the geometric product:
\begin{align}
    \eqnmarkbox[Plum]{w}{x \wedge y} &= \frac{1}{2}(xy - yx) \label{eq:app_wedge_prod}\\
    \eqnmarkbox[NavyBlue]{i}{\langle x, y \rangle} & = \frac{1}{2}(xy + yx) \label{eq:app_inner_prod}\ .
\end{align}

From the basis vectors of the vector space $G^2$ of the Clifford algebra $Cl_{2, 0}(\R)$, i.e. $\{1,e_1,e_2,e_1e_2\}$, probably the most interesting is $e_1 e_2$. We therefore have a closer look the unit bivector $i_2 = e_1 e_2$ which is the plane spanned by $e_1$ and $e_2$ and determined by the geometric product:
\begin{align}
    i_2 = e_1 e_2 = \underbrace{\langle e_1 , e_2 \rangle}_{=0} + \  e_1 \wedge e_2 = - \ e_2 \wedge e_1 = - \ e_2 e_1 \label{eq:app_bivector}\ ,
\end{align}
where the inner product $\langle e_1 , e_2 \rangle$ is zero due to the orthogonality of the base vectors. 
The bivector $i_2$ if squared yields $i_2^2 = -1$, and thus $i_2$ represents a true geometric $\sqrt{-1}$. 
From Equation~\ref{eq:app_bivector}, it follows that 
\begin{align}
e_2 &= e_1 i_2 = - \ i_2 e_1 \nonumber \\ 
e_1 &= i_2 e_2 = - \ e_2 i_2 \label{eq:app_anticommutativity}\ .
\end{align}
Using definition~\ref{def:app_dual}, the dual of a multivector $\va \in G^2$ is defined via the bivector as $i_2 \va$. Thus, the dual of a scalar is a bivector and the dual of a vector is again a vector. The dual pairs of the base vectors are $1 \leftrightarrow e_1 e_2$ and $e_1 \leftrightarrow e_2$. These dual pairs allow us to write an arbitrary multivector $\va$ as
\begin{align}
    \va &= a_0 + a_1 e_1 + a_2 e_2 + a_{12}e_{12} \ , \nonumber \\
    \va & = 1\underbrace{\big(a_0 + a_{12}i_2\big)}_{\text{spinor part}} + \  e_1\underbrace{\big(a_1 + a_2 i_2\big)}_{\text{vector part}} \ , \label{eq:app_dual_pairs}
\end{align}
which can be regarded as two complex-valued parts: the spinor part, which commutes with $i_2$ and the vector part, which anti-commutes with $i_2$.

\subsubsection{Clifford algebra $Cl_{0, 2}(\R)$}
The Clifford algebra $Cl_{0, 2}(\R)$ 
is a 4-dimensional algebra with vector space $G^2$ spanned by the basis vectors $\{1,e_1,e_2,e_1e_2\}$ where $e_1,e_2$ square to $-1$. 
The Clifford algebra $Cl_{0, 2}(\R)$ is algebra-isomorphic to the quaternions $\HH$, which are commonly written in literature~\citep{schwichtenberg2015physics} as $a + b\hat{\imath} + c\hat{\jmath} + d\hat{k}$,
where the (imaginary) base elements $\hat{\imath}$, $\hat{\jmath}$, and $\hat{k}$ fulfill the relations:
\begin{align}
    \hat{\imath}^2 &= \hat{\jmath}^2 =  -1 \nonumber \\
    \hat{\imath}\hat{\jmath} &= \hat{k} \nonumber \\
     \hat{\jmath}\hat{\imath} &= -\hat{k} \nonumber \\
     \hat{k}^2 &= \hat{\imath}\hat{\jmath}\hat{\imath}\hat{\jmath} = - \ \hat{\imath}\hat{\jmath}\hat{\jmath}\hat{\imath} = \hat{\imath}\hat{\imath} = -1 \ .
\end{align}
Quaternions also form a 4-dimensional algebra spanned by $\{1, \hat{\imath}, \hat{\jmath}, \hat{k}\}$,
where $\hat{\imath}$, $\hat{\jmath}$, $\hat{k}$ all square to $-1$.
The basis element $1$ is often called the scalar part, and the basis elements $\hat{\imath}$, $\hat{\jmath}$, $\hat{k}$ are called the vector part of a quaternion.

The geometric product of two multivectors $\va = a_0 + a_1 e_1 + a_2 e_2 + a_{12}e_1e_2$ and $\vb = b_0 + b_1 e_1 + b_2 e_2 + b_{12}e_1e_2$ is defined as:
\begin{align}
    \va \vb \  = & \ a_0 b_0 + a_0 b_1 \eone + a_0 b_2 \etwo + a_0 b_{12} \eonetwo \nonumber \\ 
    & + a_1 b_0 \eone + a_1 b_1 e_1 e_1 + a_1 b_2 \eonetwo + a_1 b_{12}e_1e_1\etwo \nonumber \\
    & + a_2 b_0 \etwo + a_2 b_1 \etwoone + a_2 b_2 e_2 e_2 + a_2 b_{12}e_2 \eone e_2 \nonumber \\
    & + a_{12} b_0 \eonetwo + a_{12} b_1 e_1 \etwo e_1 + a_{12}b_2 \eone e_2 e_2 + a_{12} b_{12} e_1e_2 e_1e_2 \ .
\end{align}

Using the relations $e_1 e_1 = - 1$, $e_2 e_2 = - 1$, and $e_i e_j = - \ e_j e_i$ for $i \neq j \in \{e_1, e_2\}$, from which it follows that $e_1e_2e_1e_2 = -1$, we obtain:
\begin{align}
    \va \vb \ = & \ a_0 b_0 - a_1 b_1 - a_2 b_2 - a_{12} b_{12} \nonumber \\
    & + (a_0 b_1 + a_1 b_0 + a_2 b_{12} - a_{12} b_2) \eone \nonumber \\
    & + (a_0 b_2 - a_1 b_{12} + a_2 b_0 + a_{12} b_1) \etwo \nonumber \\
    & + (a_0 b_{12} + a_1 b_2 - a_2 b_1 + a_{12} b_0) \eonetwo \ .
    \label{eq:geometric_product_cl02}
\end{align}

\subsubsection{Clifford algebra $Cl_{3, 0}(\R)$}
The Clifford algebra
is a 8-dimensional algebra with vector space $G^3$ spanned by the basis vectors
$\{1,e_1,e_2,e_3,e_1e_2,e_1e_3,e_2e_3,e_1e_2e_3\}$, i.e. one scalar, three vectors $\{e_1,e_2,e_3\}$, three bivectors $\{ e_1e_2,e_1e_3,e_2e_3\}$, and one trivector $e_1e_2e_3$. The trivector is the pseudoscalar $i_3$ of the algebra.
The geometric product of two multivectors is defined analogously to the geometric product of $Cl_{2, 0}(\R)$,
following the associative and bilinear multiplication of multivectors follows:
\begin{align}
    e_i^2 &= 1 \ \ &&\mathrm{for} \ \  i=1,2,3\\
    e_i e_j &= - e_j e_i \ \ && \mathrm{for} \ \ i, j = 1,2,3, i\neq j \ .
\end{align}
Using Definition~\ref{def:app_dual}, the dual pairs of $Cl_{3, 0}$ are:
\begin{align}
    1 &\leftrightarrow e_1e_2e_3 = i_3 \label{eq:3D_dual_pair_1} \\
    e_1 &\leftrightarrow e_2e_3 \\
    e_2 &\leftrightarrow e_3e_1 \\
    e_3 &\leftrightarrow e_1e_2 \label{eq:3D_dual_pair_4} \ .
\end{align}

The geometric product for $Cl_{3, 0}(\R)$ is defined analogously to the geometric product of $Cl_{2, 0}(\R)$ via:

\begin{align}
    \va \vb \  = & \ a_0 b_0 + a_0 b_1 \eone + a_0 b_2 \etwo + a_0 b_3 \ethree \nonumber \\ 
    & + a_0 b_{12} \eonetwo + a_0 b_{13} \eonethree + a_0 b_{23} \etwothree + a_0 b_{123} \eonetwothree \nonumber \\ 
    & + a_1 b_0\eone + a_1 b_1 e_1 e_1 + a_1 b_2 \eonetwo + a_1 b_3 \eonethree \nonumber \\ & + a_1 b_{12}e_1e_1\etwo + a_1 b_{13}e_1e_1\ethree + a_1 b_{23}\eonetwothree + a_1 b_{123}e_1e_1\etwothree \nonumber \\
    & + a_2 b_0 \etwo + a_2 b_1 \etwoone + a_2 b_2 e_2 e_2 + a_2 b_3 \etwothree \nonumber \\ & + a_2 b_{12}e_2\eone e_2 + a_2 b_{13}\etwoonethree + a_2 b_{23}e_2e_2\ethree - a_2 b_{123}e_2e_2\eonethree \nonumber \\
    & + a_3 b_0\ethree + a_3 b_1 \ethreeone + a_3 b_2 \ethreetwo + a_3 b_3 e_3e_3 \nonumber \\ & + a_3 b_{12}\ethreeonetwo - a_3 b_{13}\eone e_3e_3 - a_3 b_{23}\etwo e_3e_3 + a_3 b_{123}\eonetwo e_3e_3 \nonumber \\
    & + a_{12} b_0 \eonetwo - a_{12} b_1 \etwo e_1e_1 + a_{12}b_2 \eone e_2e_2 + a_{12} b_3 \eonetwothree \nonumber \\ & + a_{12} b_{12}e_1e_2e_1e_2 - a_{12} b_{13} e_1e_1\etwothree + a_{12} b_{23 }e_2e_2 \eonethree \nonumber \\ & + a_{12}b_{123} e_1e_2 e_1e_2\ethree \nonumber \\
    & + a_{13} b_0 \eonethree - a_{13}b_1 \ethree e_1e_1 + a_{13} b_2 \eonethreetwo + a_{13}b_3 \eone e_3e_3 \nonumber \\ & a_{13} b_{12} e_1e_1\ethreetwo + a_{13} b_{13} e_1e_3e_1e_3 - a_{13}b_{23} \eonetwo e_3e_3 \nonumber \\ & + a_{13} b_{123 } e_1e_3e_1\etwo e_3 \nonumber \\ 
    & + a_{23}b_0 \etwothree + a_{23}b_1 \etwothreeone + a_{23}b_2 e_2\ethree e_2 + a_{23}b_3 \etwo e_3e_3 \nonumber \\ & + a_{23}b_{12} e_2\ethreeone e_2 - a_{23}b_{13} \etwoone e_3e_3 + a_{23}b_{23} e_2e_3e_2e_3 \nonumber \\ & + a_{23}b_{123} e_2e_3\eone e_2e_3 \nonumber \\ 
    & + a_{123}b_0 \eonetwothree + a_{123}b_1 e_1\etwothree e_1  - a_{123}b_2 \eonethree e_1e_2 + a_{123}b_3 \eonetwo e_3e_3 \nonumber \\ & + a_{123} b_{12} e_1e_2\etwo e_1e_2 + a_{123}b_{13} e_1\etwo e_3 e_1e_3 + a_{123}b_{23} \eone e_2e_3e_2e_3 \nonumber \\ & + a_{123}b_{123} e_1e_2e_3e_1e_2e_3 \ ,
    \label{eq:app_geometric_product_3d_1}
\end{align}

where minus signs appear to do reordering of basis elements. Equation~\ref{eq:app_geometric_product_3d_1} simplifies to 

\begin{align}
    \va \vb \  = & \ a_0 b_0 + a_1 b_1 + a_2 b_2 + a_3 b_3 - a_{12} b_{12} - a_{13} b_{13} - a_{23} b_{23} - a_{123} b_{123}` \nonumber \\
    & + (a_0 b_1 + a_1 b_0 - a_2 b_{12} - a_3 b_{13} + a_{12} b_2 + a_{13} b_3 - a_{23} b_{123} - a_{123}b_{23}) \eone \nonumber \\
    & + (a_0 b_2 + a_1 b_{12} + a_2 b_{0} - a_3 b_{23} - a_{12} b_1 + a_{13} b_{123} + a_{23} b_3 + a_{123}b_{13}) \etwo \nonumber \\
    & + (a_0 b_3 + a_1 b_{13} + a_2 b_{23} + a_3 b_0 - a_{12} b_{123} - a_{13} b_1 - a_{23} b_2 - a_{123}b_{12}) \ethree \nonumber \\
    & + (a_0 b_{12} + a_1 b_2 - a_2 b_1 + a_3 b_{123} + a_{12} b_0 - a_{13} b_{23} + a_{23} b_{13} + a_{123}b_3) \eonetwo \nonumber \\
    & + (a_0 b_{13} + a_1 b_3 - a_2 b_{123} - a_3 b_1 + a_{12} b_{23} + a_{13} b_0 - a_{23} b_{12} - a_{123}b_2) \eonethree \nonumber \\
    & + (a_0 b_{23} + a_1 b_{123} + a_2 b_{3} - a_3 b_2 - a_{12} b_{13} + a_{13} b_{12} + a_{23} b_0 + a_{123}b_1) \etwothree \nonumber \\
    & + (a_0 b_{123} + a_1 b_{23} - a_2 b_{13} + a_3 b_{12} + a_{12} b_3 - a_{13} b_2 + a_{23} b_1 + a_{123}b_0) \eonetwothree \ . \label{eq:app_geometric_product_3d}
\end{align}

\subsection{The electromagnetic field in 3 dimensions}
Through the lense of $Cl(3,0)(\R)$, an intriguing example of the duality of multivectors is found when writing the expression of the electromagnetic field $\mF$ in terms of an electric vector field $E$ and a magnetic vector field $B$~\citep{hestenes2012clifford, hestenes2003oersted}, such that
\begin{align}
    \mF = E + B i_3 \ . \label{eq:app_emfield}
\end{align}

Both the electric field $E$ and the magnetic field $B$ are described by Maxwell's equations~\citep{griffiths2005introduction}. The two fields are strongly coupled, e.g. temporal changes of electric fields induce magnetic fields and vice versa. Probably the most illustrative co-occurence of electric and magnetic fields is when describing the propagation of light. In standard vector algebra, $E$ is a vector while $B$ is a pseudovector, i.e. the two kinds of fields are distinguished by a difference  in sign under space inversion.  
Equation~\ref{eq:app_emfield} naturally decomposes the electromagnetic field into vector and bivector parts via the pseudoscalar $i_3$. For example, for the base component $B_x e_1$ of $B$ it holds that $B_x e_1 i_3 = B_x e_1e_1e_2e_3 = B_xe_2e_3$, which is a bivector and the dual to the base component $e_1$ of $E$.
Geometric algebra reveals that a pseudovector is nothing else than a bivector represented by its dual, so the magnetic field $B$ in Equation~\ref{eq:app_emfield} is fully represented by the complete bivector $B i_3$, rather than $B$ alone.
Consequently, the multivector representing $\mF$ consists of three vectors (the electric field components) and three bivectors $e_1i_3=e_2 e_3, e_2i_3 = e_3 e_1, e_3i_3 = e_1 e_2$ (the magnetic field components multiplied by $i_3$).

%% file: appendix/cliffordlayers.tex
\section{Clifford neural layers}\label{app:clifford_layers}

This appendix supports Section~\ref{sec:clifford_layers} of the main paper.

Clifford convolutions are related to the work on complex networks by~\citet{Trabelsi2018DeepCN},
and closely related to work on quaternion neural networks
~\citep{Zhu_2018_ECCV, parcollet2018quaternion, gaudet2018deep, parcollet2018quaternionconv, parcollet2018quaternionconv2, parcollet2020survey, nguyen2021quaternion}.
Probably the most related work are (i) by~\citet{zang2022multi} who build geometric algebra convolution networks to process spatial and temporal data, and (ii)~\citet{spellings2021geometric} who build rotation- and permutation-equivariant graph network architectures based on geometric algebra products of node features. Higher order information is built from available node inputs.

\subsection{Clifford convolution layers}
We derive the implementation of translation equivariant Clifford convolution layers for multivectors in $G^2$, i.e. multivectors of Clifford algebras generated by the 2-dimensional vector space $\R^2$. Finally, we make the extension to Clifford algebras generated by the 3-dimensional vector space $\R^3$. 

\paragraph{Regular CNN layers.}
Regular convolutional neural network (CNN) layers take as input feature maps $f: \sZ^2 \rightarrow \R^\cin$ and convolve\footnote{In deep learning, a convolution operation in the forward pass is implemented as cross-correlation.} them with a set of $\cout$ filters $\{w^i\}_{i=1}^\cout: \sZ^2 \rightarrow \R^\cin$:
\begin{align}
    \left[f \star w^i \right](x) & = \sum_{y \in \sZ^2} \big \langle f(y), w^i (y-x)\big \rangle \label{eq:app_cnn_1} \\
    & = \sum_{y \in \sZ^2}\sum_{j=1}^\cin f^j(y) w^{i,j} (y-x) \ \label{eq:app_cnn_2}.
\end{align}
Equation~\ref{eq:app_cnn_1} can be interpreted as inner product of the input feature maps with corresponding filters at every point $y \in \sZ^2$. By applying $\cout$ filters, the output feature maps can be interpreted as $\cout-$ dimensional features vectors at every point $y \in \sZ^2$. 
We now want to extend convolution layers such that the elementwise product of scalars $f^j(y) w^{i,j} (y-x)$ are replaced by the geometric product of multivector inputs and multivector filters $\vf^j(y)\vw^{i,j} (y-x)$.

\paragraph{Clifford CNN layers.}
We replace the feature maps  $f: \sZ^2 \rightarrow \R^\cin$ by multivector feature maps $\vf: \sZ^2 \rightarrow (G^2)^\cin$ and convolve them with a set of $\cout$ multivector filters $\{\vw^i\}_{i=1}^\cout: \sZ^2 \rightarrow (G^2)^\cin$:
\begin{align}
    \left[\vf \star \vw^i \right](x) & = \sum_{y \in \sZ^2}\sum_{j=1}^\cin \underbrace{\vf^j(y) \vw^{i,j} (y-x)}_{\vf^j\vw^{i,j} \ : \ G^2 \times G^2 \rightarrow G^2} \ .
\end{align}

\subsubsection{Translation equivariance of Clifford convolutions}
\begin{theo}[Translation equivariance of Clifford convolutions]{th:app_translation_cliffordconv}
Let $\vf: \sZ^2 \rightarrow (G^2)^\cin$ be a multivector feature map and let $\vw: \sZ^2 \rightarrow (G^2)^\cin$ be a multivector kernel, then for $Cl(2,0)(\R)$ $\left[\left[L_t\vf\right] \star \vw \right](x) = \left[L_t\left[\vf \star \vw\right] \right](x)$.
\end{theo}

\begin{proof}
\begin{footnotesize}
\begin{align}
    &\left[\left[L_t\vf\right] \star \vw \right](x) = \sum_{y 
    \in \sZ^2}\sum_{j=1}^\cin \vf(y-t) \vw (y-x) \nonumber \\
    = & \sum_{y 
    \in \sZ^2}\sum_{j=1}^\cin f_0(y-t) w_0(y-x) + f_1(y-t) w_1(y-x) + f_2(y-t) w_2(y-x) - f_{12}(y-t) w_{12}(y-x) \nonumber \\
    & + \bigg(f_0(y-t) w_1(y-x) + f_1(y-t) w_0(y-x) - f_2(y-t) w_{12}(y-x) + f_{12}(y-t) w_2(y-x)\bigg)e_1 \nonumber \\
    & + \bigg(f_0(y-t) w_2(y-x) + f_1(y-t) w_{12}(y-x) + f_2(y-t) w_0(y-x) - f_{12}(y-t) w_1(y-x)\bigg)e_2 \nonumber \\
    & + \bigg(f_0(y-t) w_{12}(y-x) + f_1(y-t) w_2(y-x) - f_2(f-t) w_1(y-x) + f_{12}(y-t) w_0(y-x)\bigg)e_1e_2 \nonumber \\
    &(\text{using } y \rightarrow y-t)  \nonumber \\
    = & \sum_{y 
    \in \sZ^2}\sum_{j=1}^\cin f_0(y) w_0(y-(x-t)) + f_1(y) w_1(y-(x-t)) + f_2(y) w_2(y-(x-t)) - f_{12}(y) w_{12}(y-(x-t)) \nonumber \\
    & + \bigg(f_0(y) w_1(y-(x-t)) + f_1(y) w_0(y-(x-t)) - f_2(y) w_{12}(y-(x-t)) + f_{12}(y) w_2(y-(x-t))\bigg)e_1 \nonumber \\
    & + \bigg(f_0(y) w_2(y-(x-t)) + f_1(y) w_{12}(y-(x-t)) + f_2(y) w_0(y-(x-t)) - f_{12}(y) w_1(y-(x-t))\bigg)e_2 \nonumber \\
    & + \bigg(f_0(y) w_{12}(y-(x-t)) + f_1(y) w_2(y-(x-t)) - f_2(y) w_1(y-(x-t)) + f_{12}(y) w_0(y-(x-t))\bigg)e_1e_2 \nonumber \\
    = &\left[L_t\left[\vf \star \vw\right] \right](x) \ .
\end{align}
\end{footnotesize}
\end{proof}

\paragraph{Implementation of $Cl_{2,0}(\R)$ and $Cl_{0,2}(\R)$ layers.}
We can implement a $Cl(2,0)(\R)$ Clifford CNN layer using Equation~\ref{eq:app_geometric_product} where $\{b_0$, $b_1$, $b_2$, $b_{12}\} \rightarrow \{w^{i,j}_0$, $w^{i,j}_1$, $w^{i,j}_2$, $w^{i,j}_{12}\}$ correspond to 4 different kernels representing one 2D multivector kernel, i.e. 4 different convolution layers, and $\{a_0$, $a_1$, $a_2$, $a_{12}\} \rightarrow \{f^j_0$, $f^j_1$, $f^j_2$, $f^j_{12}\}$ correspond to the scalar, vector and bivector parts of the input multivector field. The channels of the different layers represent different stacks of scalars, vectors, and bivectors. All kernels have the same number of input and output channels (number of input and output multivectors), and thus the channels mixing occurs for the different terms of Equations~\ref{eq:app_geometric_product},~\ref{eq:app_geometric_product_3d} individually. Lastly, usually not all parts of the multivectors are present in the input vector fields. This can easily be accounted for by just omitting the respective parts of Equations~\ref{eq:app_geometric_product},~\ref{eq:app_geometric_product_3d}. A similar reasoning applies to the output vector fields. For $Cl(0,2)(\R)$, the signs within the geometric product change slightly.

\subsubsection{Rotational Clifford CNN layers}
Here we introduce an alternative parameterization to the Clifford CNN layer introduced in Equation~\ref{eq:cliff_cnn} by using the isomorphism of the Clifford algebra $Cl_{0,2}(\R)$ to quaternions \footnote{We could not find neural rotational quaternion convolutions in existing literature, we however used the codebase of \href{https://github.com/Orkis-Research/Pytorch-Quaternion-Neural-Networks}{https://github.com/Orkis-Research/Pytorch-Quaternion-Neural-Networks} as inspiration.}. We take advantage of the fact that a quaternion rotation can be realized by a matrix multiplication~\citep{jia2008quaternions, kuipers1999quaternions, schwichtenberg2015physics}. Using the isomorphism, we can represent the feature maps $\vf^j$ and filters $\vw^{i,j}$ as quaternions: $\vf^j$ = $f^j_0 + f^j_1 \hat{\imath} + f^j_2 \hat{\jmath} + f^j_{3} \hat{k}$ and $\vw^{i,j}$ = $w^{i,j}_0 + w^{i,j}_1 \hat{\imath} + w^{i,j}_2 \hat{\jmath} + w^{i,j}_{3} \hat{k}$\footnote{Note that the expansion coefficients for the feature map $\vf^j$ and filters $\vw^{i,j}$ in terms of the basis elements of $G^2$ and in terms of quaternion elements $\hat{\imath}$, $\hat{\jmath}$ and $\hat{k}$ are the same.}. Leveraging this quaternion representation, we can devise an alternative parameterization of the product between the feature map $\vf^j$ and $\vw^{i,j}$. To be more precise, we introduce a composite operation that results in a scalar quantity and a quaternion rotation, where the latter acts on the vector part of the quaternion $\vf^j$ and only produces nonzero expansion coefficients for the vector part of the quaternion output. A quaternion rotation
$\vw^{i,j} \vf^{j} (\vw^{i,j})^{-1}$ acts on the vector part ($\hat{\imath}, \hat{\jmath}, \hat{k}$) of $\vf^j$, and can be algebraically manipulated into a vector-matrix operation $\mR^{i,j} \vf^{j}$, where $\mR^{i,j}:\HH \rightarrow \HH$ is built up from the elements of $\vw^{i,j}$~\citep{kuipers1999quaternions}.
In other words, one can transform the vector part ($\hat{\imath}$, $\hat{\jmath}$, $\hat{k}$) of $\vf^j \in \HH$ via a rotation matrix $\mR^{i,j}$ that is built from
the scalar and vector part ($1, \hat{\imath}$, $\hat{\jmath}$, $\hat{k}$) of $\vw^{i,j} \in \HH$.
Altogether, a rotational multivector filter $\{\vw^{i}_{\text{rot}}\}_{i=1}^\cout:\sZ^2 \rightarrow {(G^2)}^\cin$ acts on the feature map $\vf^j$ through a rotational transformation $\mR^{i,j}(w^{i,j}_{\text{rot},0}, w^{i,j}_{\text{rot},1}, w^{i,j}_{\text{rot},2}, w^{i,j}_{\text{rot},12})$ acting on vector and bivector parts of the multivector feature map $\vf: \sZ^2 \rightarrow (G^2)^\cin$, and an additional scalar response of the multivector filters:
\begin{align}
\left[\vf \star \vw^{i}_{\text{rot}} \right](x) &= \sum_{y \in \sZ^2}\sum_{j=1}^\cin \vf^j(y) \vw^{i,j}_{\text{rot}} (y-x) \nonumber \\  
&= \sum_{y \in \sZ^2}\sum_{j=1}^\cin \underbrace{\big[\vf^j(y) \vw^{i,j}_{\text{rot}}(y-x))\big]_{0}}_{\text{scalar output}} + \mR^{i,j}(y-x) \cdot 
\begin{pmatrix}
f^j_1(y) \\
f^j_2(y) \\
f^j_{12}(y)
\end{pmatrix} \ ,
\label{eq:app_rot_cliff_cnn}
\end{align}
where $\big[\vf^j(y) \vw^{i,j}_{\text{rot}}(y-x))\big]_0 = f^j_0 w^{i,j}_{\text{rot},0} - f^j_1 w^{i,j}_{\text{rot},1} - f^j_2 w^{i,j}_{\text{rot},2} - f^j_{12}w^{i,j}_{\text{rot},12}$\ , which is the scalar output of Equation~\ref{eq:geometric_product_cl02}.
The rotational matrix $\mR^{i,j}(y-x)$ in written out form reads:

\begin{footnotesize}
\begin{align}
\mR^{i,j} = 
\begin{pmatrix}
1-2\big[(\hat{w}^{i,j}_{\text{rot},2})^2+(\hat{w}^{i,j}_{\text{rot},12})^2\big] & 2\big[\hat{w}^{i,j}_{\text{rot},1} \hat{w}^{i,j}_{\text{rot},2} - \hat{w}^{i,j}_{\text{rot},0} \hat{w}^{i,j}_{\text{rot},12}\big] & 2\big[\hat{w}^{i,j}_{\text{rot},1} \hat{w}^{i,j}_{\text{rot},12} + \hat{w}^{i,j}_{\text{rot},0} \hat{w}^{i,j}_{\text{rot},2}\big] \\
2\big[\hat{w}^{i,j}_{\text{rot},1} \hat{w}^{i,j}_{\text{rot},2} + \hat{w}^{i,j}_{\text{rot},0} \hat{w}^{i,j}_{\text{rot},12}\big] & 1-2\big[(\hat{w}^{i,j}_{\text{rot},1})^2+(\hat{w}^{i,j}_{\text{rot},12})^2\big] & 2\big[\hat{w}^{i,j}_{\text{rot},2} \hat{w}^{i,j}_{\text{rot},12} - \hat{w}^{i,j}_{\text{rot},0} \hat{w}^{i,j}_{\text{rot},1}\big] \\
2\big[\hat{w}^{i,j}_{\text{rot},1} \hat{w}^{i,j}_{\text{rot},12} - \hat{w}^{i,j}_{\text{rot},0} \hat{w}^{i,j}_{\text{rot},2}\big] & 
2\big[\hat{w}^{i,j}_{\text{rot},2} \hat{w}^{i,j}_{\text{rot},12} + \hat{w}^{i,j}_{\text{rot},0} \hat{w}^{i,j}_{\text{rot},12}\big] & 1-2\big[(\hat{w}^{i,j}_{\text{rot},1})^2+(\hat{w}^{i,j}_{\text{rot},2})^2\big] \\
\end{pmatrix} \ ,
\end{align}
\end{footnotesize}
where $\hat{\vw}^{i,j}_{\text{rot}}(y-x) = \hat{w}^{i,j}_{\text{rot},0}(y-x) + \hat{w}^{i,j}_{\text{rot},1}(y-x) e_1 + \hat{w}^{i,j}_{\text{rot},2} (y-x)e_2 + \hat{w}^{i,j}_{\text{rot},12}(y-x) e_{12}$ is the normalized filter with $\lVert \hat{\vw}^{i,j}_{\text{rot}} \rVert = 1$.
The dependency $(y-x)$ is omitted inside the rotation matrix $\mR^{i,j}$ for clarity.

\subsubsection{3D Clifford convolution layers}

\paragraph{Implementation of $Cl_{3,0}(\R)$ layers.}
Analogously to the 2-dimensional case, we can implement a 3D Clifford CNN layer using Equation~\ref{eq:app_geometric_product_3d},  where $\{b_0$, $b_1$, $b_2$, $b_{12}$, $b_{13}$, $b_{23}$, $b_{123}\}$ correspond to 8 different kernels representing one 3D multivector kernel, i.e. 8 different convolution layers, and $\{a_0$, $a_1$, $a_2$, $a_{12}$, $a_{13}$, $a_{23}$, $a_{123}\}$ correspond to the scalar, vector, bivector, and trivector parts of the input multivector field. Convolution layers for different $3$-dimensional Clifford algebras change the signs in the geometric product.

\subsection{Clifford normalization}
Different normalization schemes have been proposed to stabilize and accelerate training deep neural networks~\citep{ioffe2015batch, ba2016layer, wu2018group, ulyanov2017improved}.
Their standard formulation applies only to real values. 
Simply translating and scaling multivectors such that their mean is $\mathbf 0$ and their variance is $\mathbf 1$ is insufficient because it does not ensure equal variance across all components.

\paragraph{Batch normalization}
\citet{Trabelsi2018DeepCN} extended the batch normalization formulation to apply to complex values.
We build on the same principles to first propose an appropriate batch normalization scheme for multivectors, similar to the work of ~\citet{gaudet2018deep} for quaternions.
For 2D multivectors of the form $\va = a_0 + a_1 e_1 + a_2 e_2 + a_{12} e_1e_2$, we can formulate the problem of batch normalization as that of whitening 4D vectors:
\begin{equation}
    \tilde{\va} = (\mathbf{V})^{-\frac{1}{2}} (\va - \mathbb{E}[\va])
    \label{eq:multivectorbn}
\end{equation}
where the covariance matrix $\mathbf{V}$ is
\begin{equation}
    \mathbf{V} = \begin{pmatrix}
    V_{a_0a_0} & V_{a_0a_1} & V_{a_0a_2} & V_{a_0a_{12}} \\
    V_{a_1a_0} & V_{a_1a_1} & V_{a_1a_2} & V_{a_1a_{12}} \\
    V_{a_2a_0} & V_{a_2a_1} & V_{a_2a_2} & V_{a_2a_{12}} \\
    V_{a_{12}a_0} & V_{a_{12}a_1} & V_{a_{12}a_2} & V_{a_{12}a_{12}} 
    \end{pmatrix} \ .
\end{equation}

The shift parameter $\beta$ is a multivector with $4$ learnable components and the scaling parameter $\mathbf \gamma$ is $4 \times 4$ positive matrix.
The multivector batch normalization is defined as:
\begin{equation}
    BN(\va) = \mathbf{\gamma} \va + \beta
\end{equation}
When the batch sizes are small, it can be more appropriate to use Group Normalization or Layer Normalization.
These can be derived with appropriate application of Eq.~\ref{eq:multivectorbn} along appropriate tensor dimensions.
As such, batch, layer, and group normalization can be easily extended to 3-dimensional Clifford algebras.

\subsection{Clifford initialization}
\citet{parcollet2018quaternion, gaudet2018deep} introduced initialization schemes for quaternions which expands upon deep network initialization schemes proposed by~\citet{glorot2010understanding, he2015delving}.
Similar to Clifford normalization, quaternion initialization schemes can be adapted to Clifford layers in a straight forward way. 
Effectively, tighter bounds are required for the uniform distribution form which Clifford weights are sampled. However, despite intensive studies we did not observe any performance gains over default PyTorch initialization schemes\footnote{The default PyTorch initialization of linear and convolution layers is He Uniform initialization~\citep{he2015delving} for 2-dimensional problems. The gain is calculated using LeakyRelu activation functions with negative part of $5$, which effectively results in Glorot Uniform initialization.} for 2-dimensional experiments. Similar findings are reported in~\citet{hoffmann2020algebranets}. However,  3-dimensional implementations necessitate much smaller initialization values (factor $1/8$).

\subsection{Equivariance under rotations and reflections}\label{app:equivariance_proof}

Clifford convolutions satisfy the property of equivariance under translation of the multivector inputs, as shown in this Appendix~\ref{app:clifford_layers}. However, the current definition of Clifford convolutions is not equivariant under multivector rotations or reflections. Here, we derive a general kernel constraint which allows us to build generalized Clifford convolutions which are equivariant w.r.t rotations or reflections of the multivectors. 
That is, we like to prove equivariance of a Clifford layer under rotations and reflections (i.e. orthogonal transformations) if the multivector kernel multivector filters $\{\vw^i\}_{i=1}^\cout: \sZ^2 \rightarrow (G)^\cin$ satisfies the constraint:
$$
    \vw^{i,j}(Tx)=\mathbf{T}\vw^{i,j}(x)\ ,
$$
for $0 \leq j < \cin$. We first define an orthogonal transformation on a multivector by,
\begin{align}
    \mathbf{T}\vf = \pm \mathbf{u} \vf \mathbf{u}^\dagger,~~~~\mathbf{u}^\dagger \mathbf{u} = 1
\end{align}
where $\mathbf{u}$ and $\vf$ are multivectors which are multiplied using the geometric product. The minus sign is picked by reflections but not by rotations, i.e. it depends on the parity of the transformation. This construction is called a ``versor'' product. The construction can be found in e.g.~\citet{suter2003geometric} for vectors and its extension to arbitrary multivectors. The above construction makes it immediately clear that $\mathbf{T}(\vf\vg)=(\mathbf{T}\vf)(\mathbf{T}\vg)$. When we write $Tx$, we mean an orthogonal transformation of an Euclidean vector (which can in principle also be defined using versors).
To show equivariance, we wish to prove for multivectors $\vf: \sZ^2 \rightarrow (G)^\cin$ and a set of $\cout$ multivector filters $\{\vw^i\}_{i=1}^\cout: \sZ^2 \rightarrow (G)^\cin$ that:
\begin{align}
    \vf'(Tx)=\mathbf{T}\vf(x) \ , \label{eq:app_eqcondition_1}
\end{align}
and 
\begin{align}
    \vw^i(Tx)=\mathbf{T}\vw^i(x) \ , \label{eq:app_eqcondition_2}
\end{align}
Equations~\ref{eq:app_eqcondition_1}, \ref{eq:app_eqcondition_2} yield: 
\begin{align}
    \Rightarrow \left[\vf \star \vw^i \right]'(Tx)=\mathbf{T}\left[\vf \star \vw^i \right](x) \ .
\end{align}
That is: if the input multivector field transforms as a multivector, and the kernel satisfies the stated equivariance constraint, then the output multivector field also transforms properly as a multivector. Note that $\mathbf{T}$ might act differently on the various components (scalars, vectors, pseudoscalars, pseudovectors) under rotations and/or reflections.

Now,
\begin{align}
    &\left[\vf \star \vw^i \right]'(Tx) \nonumber\\
    &=\sum_{y \in \sZ^2}\sum_{j=1}^\cin \vf'^j(y) \vw^{i,j} (y-Tx)) \nonumber\\
    &=\sum_{y \in \sZ^2}\sum_{j=1}^\cin \vf'^j(y) \vw^{i,j} (T(T^{-1}y-x))) \nonumber\\
    &=\sum_{Ty' \in \sZ^2}\sum_{j=1}^\cin \vf'^j(Ty') \vw^{i,j} (T(y'-x))),~~y'=T^{-1}y \nonumber\\
    &=\sum_{y' \in \sZ^2}\sum_{j=1}^\cin \vf'^j(Ty') \vw^{i,j} (T(y'-x))) \nonumber\\
    &=\sum_{y' \in \sZ^2}\sum_{j=1}^\cin \mathbf{T}\vf^j(y') \mathbf{T}\vw^{i,j} (y'-x)) \nonumber\\
    &=\sum_{y' \in \sZ^2}\sum_{j=1}^\cin \mathbf{T}(\vf^j(y')\vw^{i,j} (y'-x))) \nonumber\\
    &=\mathbf{T}\sum_{y' \in \sZ^2}\sum_{j=1}^\cin (\vf^j(y')\vw^{i,j} (y'-x))) \nonumber\\
    &=\mathbf{T}\left[\vf \star \vw_i \right](x)
\end{align}
where in the fourth line we transform variables $y\rightarrow y'$, in the fifth line we use the invariance of the summation ``measure'' under $T$, in the sixth line we use the transformation property of $\vf$ and equivariance for $\vw^i$, in the seventh line we use the property of multivectors, and in the eighth line we use linearity of $\mathbf{T}$.

\subsection{Clifford Fourier layers}
We derive the implementation of Clifford Fourier layers for multivectors in $G^2$ and $G^3$, i.e. multivectors of Clifford algebras generated by the 2-dimensional vector space $\R^2$ and the 3-dimensional vector space $\R^3$. 

\paragraph{Classical Fourier transform.}
In arbitrary dimension $n$, the Fourier transform $\hat{f}(\xi)=\gF\{f\}(\xi)$ for a continuous $n$-dimensional complex-valued signal $f(x) = f(x_1,\ldots,x_n):\R^n \rightarrow \CC$ is defined as:
\begin{align}
    \hat{f}(\xi)= \gF\{f\}(\xi) = \frac{1}{(2\pi)^{n/2}} \int_{\R^{n}} f(x) e^{-2\pi i \langle x, \xi\rangle} \ dx \ , \ \forall \xi \in \R^n \ , \label{eq:app_ndim_ftransfrom}
\end{align}
provided that the integral exists, where $x$ and $\xi$ are $n$-dimensional vectors and $\langle x, \xi\rangle$ is the contraction of $x$ and $\xi$. Usually, $\langle x, \xi\rangle$ is the inner product, and $\xi$ is an element of the dual vector space $\R^{n\star}$.
The inversion theorem states the back-transform from the frequency domain into the spatial domain:
\begin{align}
    f(x) = \gF^{-1}\{\gF\{f\}\}(x) = \frac{1}{(2\pi)^{n/2}} \int_{\R^n} \hat{f}(\xi) e^{2\pi i \langle x, \xi\rangle } \ d\xi \ , \ \forall x \in \R^n \ . \label{eq:app_ndim_ftransfrom_inverse}
\end{align}
We can rewrite the Fourier transform of Equation~\ref{eq:app_ndim_ftransfrom} in coordinates:
\begin{align}
    \hat{f}(\xi_1, \ldots, \xi_n)=\gF\{f\}(\xi_1, \ldots, \xi_n) = \frac{1}{(2\pi)^{n/2}} \int_{\R^{n}} f(x_1, \ldots, x_n) e^{-2\pi i (x_1 \xi_1 + \ldots + x_n \xi_n) }dx_1\ldots dx_n \ . 
\end{align}

\paragraph{Discrete/Fast Fourier transform.}
The discrete counterpart of Equation~\ref{eq:app_ndim_ftransfrom} transforms an n-dimensional complex signal $f(x) = f(x_1,\ldots ,x_n): \R^n \rightarrow \CC$ at $M_1 \times \ldots \times M_n$ grid points into its complex Fourier modes via:
\begin{align}
    \displaystyle \hat{f}(\xi_1,\ldots,\xi_n)= \gF\{f\}(\xi_1,\ldots,\xi_n) = \sum_{m_1 = 0}^{M_1} \ldots \sum_{m_n = 0}^{M_n} f(m_1,\ldots,m_n) \cdot e^{-2\pi i \cdot \bigg(\frac{m_1 \xi_1}{M_1} + \ldots + \frac{m_n \xi_n}{M_n} \bigg)} \ ,
    \label{eq:app_FFT}
\end{align}
where $(\xi_1, \ldots, \xi_n) \in \sZ_{M_1} \ldots \times \ldots \sZ_{M_n}$.
Fast Fourier transforms (FFTs)~\citep{cooley1965algorithm, van1992computational} immensely accelerate the computation of the transformations of Equation~\ref{eq:app_FFT} by factorizing the discrete Fourier transform matrix into a product of sparse (mostly zero) factors.

\subsubsection{2D Clifford Fourier transform}
Analogous to Equation~\ref{eq:app_ndim_ftransfrom}, for $Cl(2,0)(\R)$
the Clifford Fourier transform~\citep{ebling2005, hitzer2012clifford} and the respective inverse transform for multivector valued functions $\vf(x): \R^2 \rightarrow G^2$ and vectors $x, \xi \in \R^2$ are defined as:
\begin{align}
    \hat{\vf}(\xi)= \gF\{\vf\}(\xi) = \frac{1}{2\pi}\int_{\R_{2}} \vf(x) e^{-2\pi i_2 \langle x, \xi\rangle} \ dx \ , \ \forall \xi \in \R^2 \ , \label{eq:app_clifford_ftransfrom} \\
    \vf(x)=\gF^{-1}\{\gF\{\vf\}\}(x) = \frac{1}{2\pi} \int_{\R_{2}} \hat\vf(\xi) e^{2\pi i_2 \langle x, \xi\rangle} \ d\xi \ , \ \forall x \in \R^2 \ , \label{eq:app_clifford_ftransfrom_inverse} 
\end{align}
provided that the integrals exist. The differences to Equations~\ref{eq:app_ndim_ftransfrom} and~\ref{eq:app_ndim_ftransfrom_inverse} are that $\vf(x)$ and $\hat{\vf}(\xi)$ represent multivector fields in the spatial and the frequency domain, respectively, and that the pseudoscalar $i_2 = e_1e_2$ is used in the exponent.
Inserting the definition of multivector fields, we can rewrite Equation~\ref{eq:app_clifford_ftransfrom} as:
\begin{align}
     \gF\{\vf\}(\xi) &= \frac{1}{2\pi}\int_{\R_{2}} \vf(x) e^{-2\pi i_2 \langle x, \xi \rangle} \ dx  \ , \nonumber \\
     &= \frac{1}{2\pi} \int_{\R_{2}} \Bigg[ 1\bigg(\underbrace{f_0(x) + f_{12}(x)i_2}_{\text{spinor part}}\bigg) + \eone\bigg(\underbrace{f_1(x) + f_2(x) i_2}_{\text{vector part}}\bigg)\Bigg] e^{-2\pi i_2 \langle x, \xi\rangle} \ dx \nonumber \\
     &= \frac{1}{2\pi} \int_{\R_{2}} 1\bigg(f_0(x) + f_{12}(x)i_2\bigg) e^{-2\pi i_2 \langle x, \xi\rangle} \ dx \nonumber \\
     & \ \ \ + \ \frac{1}{2\pi} \int_{\R_{2}} \eone\bigg(f_1(x) + f_{2}(x)i_2\bigg) e^{-2\pi i_2 \langle x, \xi\rangle} \ dx  \nonumber \\ 
     &= 1\Bigg[\gF\bigg(f_0(x) + f_{12}(x)i_2 \bigg)(\xi)\Bigg] + \ \eone \Bigg[\gF\bigg(f_1(x) + f_{2}(x)i_2 \bigg)(\xi)\Bigg] \label{eq:app_clifford_ft_transform_vector_spinor} .
\end{align}
We obtain a Clifford Fourier transform by applying two standard Fourier transforms 
for the dual pairs $\vf_0 = f_0(x) + f_{12}(x)i_2$ and $\vf_1 = f_1(x) + f_{2}(x)i_2$,
which both can be treated as a complex-valued signal $\vf_0, \vf_1: \R^2 \rightarrow \CC$. Consequently, $\vf(x)$ can be understood as an element of $\CC^2$. The 2D Clifford Fourier transform is the linear combination of two  classical Fourier transforms.
The discretized versions of the spinor/vector part ($\hat{f}_{s/v}$) reads analogously to Equation~\ref{eq:app_FFT}:
\begin{align}
\displaystyle \hat{f}_{s/v}(\xi_1,\xi_2)= \gF\{f_{s/v}\}(\xi_1, \xi_2) = \sum_{m_1 = 0}^{M_1} \sum_{m_2 = 0}^{M_2} f_{s/v}(m_1,m_2) \cdot e^{-2\pi i_2 \bigg(\frac{m_1 \xi_1}{M_1} + \frac{m_2 \xi_2}{M_2} \bigg)} \ ,
    \label{eq:app_Clifford_FFT}
\end{align}
where again $(\xi_1, \xi_2) \in \sZ_{M_1} \times \sZ_{M_n}$.
Similar to Fourier Neural Operators (FNOs) where weight tensors are applied pointwise in the Fourier space, we apply multivector weight tensors $\mW \in (G^2)^{\cin \times \cout \times (\xi_1^{\text{max}} \times \xi_2^{\text{max}})}$ point-wise. Fourier modes above cut-off frequencies $(\xi^{\text{max}}_1, \xi^{\text{max}}_2)$ are set to zero. In doing so, we modify the Clifford Fourier modes
\begin{align}
    \hat{\vf}(\xi)= \gF\{\vf\}(\xi) = \hat{f}_0(\xi) + \hat{f}_1(\xi)e_1 + \hat{f}_2(\xi)e_2 + \hat{f}_{12}(\xi)e_{12}  
\end{align}
via the geometric product. The Clifford Fourier modes follow naturally when combining spinor and vector parts of Equation~\ref{eq:app_clifford_ft_transform_vector_spinor}.
Analogously to FNOs, higher order modes are cut off.
Finally, the residual connections used in FNO layers is replaced by a multivector weight matrix realized as Clifford convolution, ideally a $Cl_{2,0}(\R)$ convolution layer. 
A schematic sketch of a Clifford Fourier layer is shown in Figure~\ref{fig:sketch_CFNO} in the main paper. For $Cl(0,2)(\R)$, the the vector part changes to $\eone\bigg(f_1(x) - f_2(x) i_2\bigg)$.

\subsubsection{2D Clifford convolution theorem}
In contrast to~\citet{ebling2005}, we proof the 2D Clifford convolution theorem for multivector valued filters applied from the right, such that filter operations are consistent with Clifford convolution layers. 
We first need to show that the Clifford kernel commutes with the spinor and anti-commutes with the vector part of multivectors. We can write the product $\va e^{i_2 s}$ for every scalar $s \in \R$ and multivector $\va \in G^2$ as
\begin{align}
    \rva e^{i_2 s} = \rva\big(\cos(s) + i_2 \sin(s)\big) \ .
\end{align}
For the basis of the spinor part, we obtain $1 i_2 = i_2 1$, and for the basis of the vector part $e_1 i_2 = e_1e_1e_2 = - e_1 e_2 e_1 = -i_2 e_1$.
Thus, the Fourier kernel $e^{-2\pi i_2 \langle x, \xi\rangle}$ commutes with the spinor part,
and anti-commutes with the vector part of $\va$, both for $Cl(2,0)(\R)$ and $Cl(0,2)(\R)$.
We therefore proof the convolution theorem for the commuting spinor and the anti-commuting vector part of $\va$.

\begin{theo}[2D Clifford convolution theorem.]{th:app_clifford_convolution}
Let the field $\vf: \R^2 \rightarrow G^2$ be multivector valued, 
the filter $\vk_\vs: \R^2 \rightarrow G^2$ be spinor valued,
and the filter $\vk_\vv: \R^2 \rightarrow G^2$ be vector valued,
and let $\gF\{\vf\}, \gF\{\vk_\vs\}, \gF\{\vk_\vv\}$ exist, then 
\begin{align}
    \gF\{\vf \star \vk_\vs\}(\xi) &= \gF\{\vf\}(\xi)\cdot\gF\dagger\{\vk_\vs\}(\xi) \ , \nonumber \\
    \gF\{\vf \star \vk_\vv\}(\xi) &= \gF\{\vf\}(\xi)\cdot\gF\{\vk_\vv\}(\xi) \ , \nonumber
\end{align}
where $\gF^{\dagger}\{\vk_\vs\}(\xi) = \gF\{\vk_\vs\}(-\xi)$ and $\gF^{\dagger}\{\vk_\vv\}(\xi) = \gF\{\vk_\vv\}(-\xi)$.
\end{theo}

\begin{proof}
\begin{align}
    \gF\{\vf \star \vk_\vs\}(\xi) &= \frac{1}{(2\pi)^2} \int_{\R^2}\Bigg[ \int_{\R^2} \vf(y) \vk_\vs(y - x) dy \Bigg] e^{-2\pi i_2 \langle x, \xi\rangle} \ dx \nonumber \\
    &= \frac{1}{(2\pi)^2} \int_{\R^2}\vf(y) \Bigg[ \int_{\R^2} \vk_\vs(y - x)  e^{-2\pi i_2 \langle x, \xi\rangle} \ dx \Bigg] dy \nonumber \\
    &= \frac{1}{(2\pi)^2} \int_{\R^2}\vf(y) \Bigg[\underbrace{ \int_{\R^2} \vk_\vs(x)  e^{-2\pi i_2 \langle y-x, \xi\rangle} \ dx }_{\gF^\dagger\{\vk_\vs\}(\xi) e^{-2\pi i_2 \langle y, \xi\rangle} =  e^{-2\pi i_2 \langle y, \xi\rangle} \gF^\dagger\{\vk_\vs\}(\xi)} \Bigg] dy \nonumber \\
    &= \frac{1}{2\pi} \Bigg[\int_{\R^2}\vf(y) e^{-2\pi i_2 \langle y, \xi\rangle} dy \Bigg] \ \gF^\dagger\{\vk_\vs\}(\xi) \nonumber \\
    &= \gF\{\vf\}(\xi) \cdot \gF^\dagger\{\vk_\vs\}(\xi) \ .
\end{align}

\begin{align}
    \gF\{\vf \star \vk_\vv\}(\xi) &= \frac{1}{(2\pi)^2}\int_{\R^2}\Bigg[ \int_{\R^2} \vf(y) \vk_\vv(y-x) dy \Bigg] e^{-2\pi i_2 \langle x, \xi\rangle} \ dx \nonumber \\
    &= \frac{1}{(2\pi)^2} \int_{\R^2}\vf(y) \Bigg[ \int_{\R^2} \vk_\vv(y-x)  e^{-2\pi i_2 \langle x, \xi\rangle} \ dx \Bigg] dy \nonumber \\
    &= \frac{1}{(2\pi)^2} \int_{\R^2}\vf(y) \Bigg[\underbrace{ \int_{\R^2} \vk_\vv(x)  e^{-2\pi i_2 \langle y-x, \xi\rangle} \ dx }_{\gF^\dagger\{\vk_\vv\}(\xi) e^{2\pi i_2 \langle y, \xi\rangle} = e^{-2\pi i_2 \langle y, \xi\rangle} \gF\{\vk_\vv\}(\xi) \text{ , where } -\xi \rightarrow \ \xi} \Bigg] dy \nonumber \\
    &= \frac{1}{2\pi} \Bigg[\int_{\R^2}\vf(y) e^{-2\pi i_2 \langle y, \xi\rangle} dy \Bigg] \ \gF\{\vk_\vv\}(\xi) \nonumber \\
    &= \gF\{\vf\}(\xi) \cdot \gF\{\vk_\vv\}(\xi) \ .
\end{align}
\end{proof}

\subsubsection{3D Clifford Fourier transform}
For $Cl(3,0)(\R)$, analogous to Equation~\ref{eq:app_ndim_ftransfrom}, the Clifford Fourier transform~\citep{ebling2005} and the respective inverse transform for multivector valued functions $\vf: \R^3 \rightarrow G^3$ and vectors $x, \xi \in \R^3$ are defined as:
\begin{align}
    \hat{\vf}(\xi)= \gF\{\vf\}(\xi) = \frac{1}{(2\pi)^{3/2}}\int_{\R_{3}} \vf(x) e^{-2\pi i_3 \langle x, \xi\rangle} \ dx \ , \ \forall \xi \in \R^3 \ , \label{eq:app_clifford_ftransfrom_3d} \\
    \vf(x)=\gF^{-1}\{\gF\{\vf\}\}(x) = \frac{1}{(2\pi)^{3/2}} \int_{\R_{3}} \hat\vf(\xi) e^{2\pi i_3 \langle x, \xi\rangle} \ d\xi \ , \ \forall x \in \R^3 \ , \label{eq:app_clifford_ftransfrom_inverse_3d} 
\end{align}
provided that the integrals exist.
A multivector valued function $\vf: \R^3 \rightarrow G^3$,
\begin{align}
    \vf = f_0 + f_1 e_1 + f_2 e_2 + f_3 e_3 + f_{12} e_{12} + f_{13}e_{13} + f_{23}e_{23} + f_{123}e_{123}
\end{align}
can be expressed via the pseudoscalar $i_3 = e_1 e_2 e_3$ as:
\begin{align}
    \vf &= (f_0 + f_{123}i_3)1 \nonumber \\
    & + (f_1 + f_{23}i_3)e_1 \nonumber \\
    & + (f_2 + f_{31}i_3)e_2 \nonumber \\
    & + (f_3 + f_{12}i_3)e_3 \ ,
    \label{eq:app_multivector_3d}
\end{align}
We obtain a 3-dimensional Clifford Fourier transform by applying four standard Fourier transforms for the four dual pairs $\vf_0 = f_0(x) + f_{123}(x)i_3$, $\vf_1 = f_1(x) + f_{23}(x)i_3$, $\vf_2 = f_2(x) + f_{31}(x)i_3$, and $\vf_3 = f_3(x) + f_{12}(x)i_3$, which all can be treated as a complex-valued signal $\vf_0, \vf_1,
\vf_2, \vf_3: \R^3 \rightarrow \CC$. Consequently, $\vf(x)$ can be understood as an element of $\CC^4$. The 3D Clifford Fourier transform is the linear combination of four  classical Fourier transforms:

\begin{align}
     \gF\{\vf\}(\xi) &= \frac{1}{(2\pi)^{3/2}}\int_{\R_{3}} \vf(x) e^{-2\pi i_3 \langle x, \xi \rangle} \ dx  \ , \nonumber \\
     &= \frac{1}{(2\pi)^{3/2}} \int_{\R_{3}} \Bigg[ 1\bigg( f_0(x) + f_{123}(x)i_3\bigg) + \eone\bigg(f_1(x) + f_{23}(x) i_3\bigg) \nonumber \\
     & \ \ \ \ \ \ \ \ \ \ \ \ \ \ + \etwo \bigg(f_2(x) + f_{31}(x) i_3\bigg)
     + \ethree \bigg(f_3(x) + f_{12}(x) i_3\bigg) \Bigg] e^{-2\pi i_3 \langle x, \xi\rangle} \ dx \nonumber \\
     &= \frac{1}{(2\pi)^{3/2}} \int_{\R_{3}} 1\bigg(f_0(x) + f_{123}(x)i_3\bigg) e^{-2\pi i_3 \langle x, \xi\rangle} \ dx \nonumber \\
     & \ \ \ + \ \frac{1}{(2\pi)^{3/2}} \int_{\R_{3}} \eone\bigg(f_1(x) + f_{23}(x)i_3\bigg) e^{-2\pi i_3 \langle x, \xi\rangle} \ dx  \nonumber \\
     & \ \ \ + \ \frac{1}{(2\pi)^{3/2}} \int_{\R_{3}} \etwo\bigg(f_2(x) + f_{31}(x)i_3\bigg) e^{-2\pi i_3 \langle x, \xi\rangle} \ dx  \nonumber \\
     & \ \ \ + \ \frac{1}{(2\pi)^{3/2}} \int_{\R_{3}} \ethree\bigg(f_3(x) + f_{12}(x)i_3\bigg) e^{-2\pi i_3 \langle x, \xi\rangle} \ dx  \nonumber \\
     &= 1\Bigg[\gF\bigg(f_0(x) + f_{12}(x)i_3 \bigg)(\xi)\Bigg] + \ \eone\Bigg[\gF\bigg(f_1(x) + f_{23}(x)i_3 \bigg)(\xi)\Bigg]  \nonumber \\
     & \ \ \ + \  \etwo\Bigg[\gF\bigg(f_2(x) + f_{31}(x)i_3 \bigg)(\xi)\Bigg] + \ \ethree\Bigg[\gF\bigg(f_3(x) + f_{12}(x)i_3 \bigg)(\xi)\Bigg] \ .
     \label{eq:app_clifford_ft_transform_3d}
\end{align}

Analogous to the 2-dimensional Clifford Fourier transform,
we apply multivector weight tensors $\mW \in (G^3)^{\cin \times \cout \times (\xi_1^{\text{max}} \times \xi_2^{\text{max}} \times \xi_3^{\text{max}})}$ point-wise. Fourier modes above cut-off frequencies $(\xi^{\text{max}}_1, \xi^{\text{max}}_2, \xi_3^{\text{max}})$ are set to zero. In doing so, we modify the Clifford Fourier modes
\begin{equation}
    \begin{aligned}
    \hat{\vf}(\xi) &= \gF\{\vf\}(\xi) \\&= \hat{f}_0(\xi) + \hat{f}_1(\xi)e_1 + \hat{f}_2(\xi)e_2
    + \hat{f}_3(\xi)e_3 
    + \hat{f}_{12}(\xi)e_{12} + \hat{f}_{31}(\xi)e_{31} +
    \hat{f}_{23}(\xi)e_{23} + \hat{f}_{123}(\xi)e_{123} 
    \end{aligned}
\end{equation}
via the geometric product. The Clifford Fourier modes follow naturally when combining the four parts of Equation~\ref{eq:app_clifford_ft_transform_3d} . Finally, the residual connections used in FNO layers is replaced by a multivector weight matrix realized as Clifford convolution, ideally a $Cl_{3,0}(\R)$ convolution layer.
For other $3$-dimensional Clifford algebras, the signs of the dual pairs in Equation~\ref{eq:app_multivector_3d} change accordingly.

\subsubsection{3D Clifford convolution theorem} This theorem adapted from~\citet{ebling2005}).
First, again let's check if the Clifford kernel commutes with the different parts of multivectors. We can write the product $\va e^{i_3 s}$ for every scalar $s \in \R$ and multivector $\va \in G^3$ as
\begin{align}
    \rva e^{i_3 s} = \rva\big(\cos(s) + i_3 \sin(s)\big) \ .
\end{align}
First, we check again if the different basis vectors of the Fourier transforms of Equation~\ref{eq:app_clifford_ft_transform_3d} commute with the pseudoscalar $i_3$:
\begin{align}
    1i_3 & = i_3 1 \ \checkmark \nonumber \\
    e_1 i_3 & = e_1 e_1e_2e_3 = - e_1 e_2e_1e_3 = e_1 e_2e_3e_1 = i_3 e_1 \ \checkmark \nonumber \\
    e_2 i_3 & = e_2 e_1e_2e_3 = - e_1 e_2e_2e_3 = e_1 e_2e_3e_2 = i_3 e_2 \ \checkmark \nonumber \\
    e_3 i_3 & = e_3 e_1e_2e_3 = - e_1 e_3e_2e_3 = e_1 e_2e_3e_3 = i_3 e_3 \ \checkmark 
\end{align}
In contrast to the 2-dimensional Clifford Fourier transform, now all four parts of the multivector of Equation~\ref{eq:app_multivector_3d} commute with $i_3$. This holds for all $3$-dimensional Clifford algebras.

\strut\newpage

\clearpage

\begin{theo}[3D Clifford convolution theorem.]{th:app_clifford_convolution_3d}
Let the field $\vf: \R^3 \rightarrow G^3$ be multivector valued, 
the filter $\vk_\va: \R^3 \rightarrow G^3$ be multivector valued,
and let $\gF\{\vf\}, \gF\{\vk_\va\}$ exist, then 
\begin{align}
    \gF\{\vf \star \vk_\va\}(\xi) &= \gF\{\vf\}(\xi)\cdot\gF^\dagger\{\vk_\va\}(\xi) \ , \nonumber
\end{align}
where $\gF^{\dagger}\{\vk_\va\}(\xi) = \gF\{\vk_\va\}(-\xi)$.
\end{theo}

\begin{proof}
\begin{align}
    \gF\{\vf \star \vk_\va\}(\xi) &= \frac{1}{(2\pi)^{3}} \int_{\R^3}\Bigg[ \int_{\R^3} \vf(y) \vk_\va(y - x) dy \Bigg] e^{-2\pi i_3 \langle x, \xi\rangle} \ dx \nonumber \\
    &= \frac{1}{(2\pi)^{3}} \int_{\R^3}\vf(y) \Bigg[ \int_{\R^3} \vk_\va(y - x)  e^{-2\pi i_3 \langle x, \xi\rangle} \ dx \Bigg] dy \nonumber \\
    &= \frac{1}{(2\pi)^{3}} \int_{\R^3}\vf(y) \Bigg[\underbrace{ \int_{\R^3} \vk_\va(x)  e^{-2\pi i_3 \langle y-x, \xi\rangle} \ dx }_{\gF^\dagger\{\vk_\va\}(\xi) e^{-2\pi i_3 \langle y, \xi\rangle} =  e^{-2\pi i_3 \langle y, \xi\rangle} \gF^\dagger\{\vk_\va\}(\xi)} \Bigg] dy \nonumber \\
    &= \frac{1}{(2\pi)^{3/2}} \Bigg[\int_{\R^3}\vf(y) e^{-2\pi i_3 \langle y, \xi\rangle} dy \Bigg] \ \gF^\dagger\{\vk_\va\}(\xi) \nonumber \\
    &= \gF\{\vf\}(\xi) \cdot \gF^\dagger\{\vk_\va\}(\xi) \ .
\end{align}
\end{proof}

\subsubsection{Implementation of Clifford Fourier layers}
We implement a 2D Clifford Fourier layer by applying two standard Fourier transforms on the dual pairs of
Equation~\ref{eq:clifford_ft_transform_vector_spinor}. These dual pairs can be treated as complex valued inputs.
Similarly, we implement a 3D Clifford Fourier layer by applying four standard Fourier transforms on the dual pairs of 
e.g. $Cl_{3,0}$ (Equation~\ref{eq:3D_dual_pair_1} - Equation~\ref{eq:3D_dual_pair_4}). Since Clifford convolution theorems hold both for the vector and the spinor parts and for the four dual pairs for $Cl_{2,0}$ and $Cl_{3,0}$, respectively, we multiply the modes in the Fourier space using the geometric product. Finally, we apply an inverse Fourier transformation and resemble the multivectors in the spatial domain.

\subsection{Pseudocode}\label{app:pseudocode}

Algorithm~\ref{app:algo_clifford_conv} sketches the implementation of a Clifford convolution, Algorithm~\ref{app:algo_clifford_rotconv} of a rotational Clifford convolution, and Algorithm~\ref{app:algo_clifford_fourier} of a Clifford Fourier layer.
\input{appendix/cliffordlayer_algo.tex}

%% file: appendix/cliffordlayer_algo.tex
\begin{algorithm}
    \caption{Pseudocode for 2D Clifford convolution using Cl$_{2,0}$.}
    \label{app:algo_clifford_conv}
    \begin{mdframed}[backgroundcolor=shadecolor!5,rightline=false,leftline=false]
        \begin{algorithmic}[1]
            \Function{CliffordKernel2d}{$W$}
            \State kernel $\gets \begin{bmatrix}
                W[0] & W[1] & W[2] & -W[3] \\
                W[1] & W[0] & -W[3] & W[2] \\
                W[2] & W[3] & W[0] & -W[1] \\
                W[3] & W[2]& -W[1] & W[0]
            \end{bmatrix}$
            \State \Return{kernel}
            \EndFunction
            \Function{CliffordConv2D}{$W$, $x$}
                \State kernel $\gets$ \Call{CliffordKernel2D}{$W$}
                \State input $\gets$ \Call{view\_as\_realvector}{$x$}
                \State output $\gets$ \Call{Conv2D}{kernel, input}
                \State \Return{\Call{view\_as\_multivector}{output}}
            \EndFunction
        \end{algorithmic}
    \end{mdframed}
\end{algorithm}

\begin{algorithm}
    \caption{Pseudocode for 2D rotational Clifford convolution using Cl$_{0,2}$.}
    \label{app:algo_clifford_rotconv}
    \begin{mdframed}[backgroundcolor=shadecolor!5,rightline=false,leftline=false]
        \begin{algorithmic}[1]
            \Function{CliffordKernel2D$_\text{rot}$}{$W$}
                \State $sq_{12} \gets W[1]^2 + W[2]^2$
                \State $sq_{13} \gets W[1]^2 + W[3]^2$
                \State $sq_{23} \gets W[2]^2 + W[3]^2$
                \State $sumsq \gets W[0]^2+W[1]^2+W[2]^2+W[3]^2 + \epsilon$
                \State $rot_{12} \gets W[0] W[1] / sumsq$
                \State $rot_{13} \gets W[0] W[2] / sumsq$
                \State $rot_{14} \gets W[0] W[3] / sumsq$
                \State $rot_{23} \gets W[1] W[2] / sumsq$
                \State $rot_{24} \gets W[1] W[3] / sumsq$
                \State $rot_{34} \gets W[2] W[3] / sumsq$
                \State kernel $\gets \begin{bmatrix}
                    W[0] & -W[1] & -W[2] & -W[3] \\
                    W[5] & W[4](1.0 - sq_{23}) & W[4](rot_{23} - rot_{14}) & W[4](rot_{24} + rot_{13}) \\
                    W[5] & W[4](rot_{23} + rot_{14}) & W[4](1.0 - sq_{13}) & W[4](rot_{34} - rot_{12}) \\
                    W[5] & W[4](rot_{24} - rot_{13}) & W[4](rot_{34} + rot_{12}) & W[4](1.0 - sq_{12})
                \end{bmatrix}$
                \State \Return{kernel}
            \EndFunction
            \Function{CliffordConv2D$_\text{rot}$}{$W$, $x$}
                \State kernel $\gets$ \Call{CliffordKernel2D$_\text{rot}$}{$W$}
                \State input $\gets$ \Call{view\_as\_realvector}{$x$}
                \State output $\gets$ \Call{Conv2D}{kernel, input}
                \State \Return{\Call{view\_as\_multivector}{output}}
            \EndFunction
        \end{algorithmic}
    \end{mdframed}
\end{algorithm}

\begin{algorithm}
    \caption{Pseudocode for 2D Clifford Fourier layer using Cl$_{2,0}$.}
    \label{app:algo_clifford_fourier}
    \begin{mdframed}[backgroundcolor=shadecolor!5,rightline=false,leftline=false]
        \begin{algorithmic}[1]
            \Function{CliffordSpectralConv2D}{$W$, $x$, $m_1, m_2$}
                \State $x_v, x_v \gets$ \Call{view\_as\_dual\_parts} {$x$}
                \State $f(x_v) \gets \Call{FFT2}{x_v}$ \Comment{Complex 2D FFT of vector part}
                \State $f(x_s) \gets \Call{FFT2}{x_s}$
                \Comment{Complex 2D FFT of scalar part}
                \State $f^{*}(x_v) \gets \begin{bmatrix}
                    f(x_v)[\ldots,:m_1, :m_2] & f(x_v)[\ldots,:m_1, -m_2:] \\
                    f(x_v)[\ldots,-m_1:, :m_2] & f(x_v)[\ldots,-m_1:, -m_2:]
                \end{bmatrix}$ \Comment{Vector modes}
                \State $f^{*}(x_s) \gets \begin{bmatrix}
                    f(x_s)[\ldots,:m_1, :m_2] & f(x_s)[\ldots,:m_1, -m_2:] \\
                    f(x_s)[\ldots,-m_1:, :m_2] & f(x_s)[\ldots,-m_1:, -m_2:]
                \end{bmatrix}$ \Comment{Scalar modes}
                \State $f^*(x) \gets f^{*}(x_s).r + f^{*}(x_v).r + f^{*}(x_v).i + f^{*}(x_s).i $ \Comment{Multivector Fourier modes}
                \State $\hat{f}^*(x) \gets f^*(x) W$ \Comment{Geometric product in the Fourier space}
                \State $\hat{x}_v \gets \Call{IFFT2}{\hat{f}^*(x)[1]+ \hat{f}^*(x)[2]}$
                \Comment{Inverse 2D FFT of vector part}
                \State $\hat{x}_2 \gets \Call{IFFT2}{\hat{f}^*(x)[0]+ \hat{f}^*(x)[3]}$
                \Comment{Inverse 2D FFT of scalar part}
                \State $\hat{x} \gets$ \Call{view\_as\_multivector} {$\hat{x}_v, \hat{x}_s$}
                \State \Return{$\hat{x}$}
            \EndFunction
            \Function{CliffordFourierLayer2D}{$W_{f}, W_c, x$}
                \State $y_1 \gets \Call{CliffordSpectralConv}{W_f, x, m_1, m_2}$
                \State $x_2 \gets \Call{view\_as\_realvector}{x}$
                \State $y_2 \gets \Call{CliffordConv}{W_c, x_2}$
                \State $y_2 \gets \Call{view\_as\_multivector}{y_2}$
                \State out $\gets$ \Call{activation}{$y_1 + y_2$}
                \State \Return{out}
                \State 
            \EndFunction
        \end{algorithmic}
    \end{mdframed}
\end{algorithm}

%% file: appendix/experiments.tex
\section{Experiments}\label{app:experiments}
This appendix supports Section~\ref{sec:experiments} of the main paper.

\subsection{Loss function and metrics}
We report the summed MSE (SMSE) loss defined as:
\begin{align}
    \gL_{\text{SMSE}} = \frac{1}{N_y}\sum_{y \in \sZ^2 (\text or \sZ^3)} \sum_{j=1}^{N_t} \sum_{i=1}^{N_{\text{fields}}} \lVert u_i(y,t_j) - \hat{u}_i(y,t_j)\rVert^2_2 \ ,
    \label{eq:app_loss}
\end{align}
where $u$ is the target, $\hat{u}$ the model output, $N_{\text{fields}}$ comprises scalar fields as well as individual vector field components, and $N_y$ is the total number of spatial points. Equation~\ref{eq:app_loss} is used for training with $N_t=1$, and further allows us to define four metrics:
\begin{itemize}
    \item \textit{One-step} loss where $N_t = 1$ and $N_{\text{fields}}$ comprises all scalar and vector components.
    \item \textit{Vector} loss where $N_t = 1$ and $N_{\text{fields}}$ comprises only vector components.
    \item \textit{Scalar} loss where $N_t = 1$ and $N_{\text{fields}}$ comprises only the scalar field.
    \item \textit{Rollout} loss where $N_t = 5$ and $N_{\text{fields}}$ comprises all scalar and vector components.
\end{itemize}
For Maxwell's equation, \textit{electric} and \textit{magnetic} loss are defined
analogously to the vector and the scalar loss
for Navier-Stokes and shallow water experiments.

\subsection{Models}
We experiment with two architecture families: ResNet models~\citep{he2016deep} and Fourier Neural Operators (FNOs)~\citep{li2020fourier}.
All baseline models are fine-tuned for all individual experiments with respect to number of blocks, number of channels, number of modes (FNO), learning rates, normalization and initialization procedures, and activation functions. The best models are reported, and for reported Clifford results each convolution layer is substituted with a Clifford convolution, each Fourier layer with a Clifford Fourier layer, each normalization with a Clifford normalization and each non-linearity with a Clifford non-linearity. A Clifford non-linearity in this context is a the application of the corresponding default linearity to the different multivector components.

\paragraph{ResNet architectures.}
For Navier-Stokes and shallow water experiments, we use ResNet architectures with 8 residual blocks, each consisting of two convolution layers with $3 \times 3$ kernels, shortcut connections, group normalization~\citep{wu2018group}, and GeLU activation functions~\citep{hendrycks2016gaussian}. We further use two embedding and two output layers, i.e. the overall architectures could be classified as Res-20 networks. In contrast to standard residual networks for image classification, we don't use any down-projection techniques, e.g. convolution layers with strides larger than 1 or via pooling layers. In contrast, the spatial resolution stays constant throughout the network. We therefore also use the same number of hidden channels throughout the network, that is 128 channels per layer. Overall this results in roughly 2.4 million parameters. Increasing the number of residual blocks or the number of channels did not increase the performance significantly.

\paragraph{Clifford ResNet architectures.}
For every ResNet-based experiment, we replaced the fine-tuned ResNet architectures with two Clifford counterparts: each CNN layer is replaced with a (i) Clifford CNN layer, and (ii) with a rotational Clifford CNN layer.
To keep the number of weights similar, instead of 128 channels the resulting architectures have 64 multivector channels, resulting again in roughly 1.6 million floating point parameters.
Additionally for both architectures, GeLU activation functions are replaced with Clifford GeLU activation functions, group normalization is replaced with Clifford group normalization. Using Clifford initialization techniques did not improve results. 

\paragraph{Fourier Neural Operator architectures.}
For Navier-Stokes and shallow water experiments, we used 2-dimensional Fourier Neural Operators (FNOs) consisting of 8 FNO blocks, two embedding and two output layers. 
Each FNO block comprised a convolution path with a $1 \times 1$ kernel and an FFT path. We used 16 Fourier modes (for $x$ and $y$ components) for point-wise weight multiplication, and overall use 128 hidden channels. We used GeLU activation functions~\citep{hendrycks2016gaussian}. Additional shortcut connections or normalization techniques, such as batchnorm or group, norm did not improve performance, neither did larger numbers of hidden channels, nor more FNO blocks. 
Overall this resulted in roughly 140 million parameters for FNO based architectures.

For 3-dimensional Maxwell experiments, we used 3-dimensional Fourier Neural Operators (FNOs) consisting of 4 FNO blocks, two embedding and two output layers. 
Each FNO block comprised a 3D convolution path with a $1 \times 1$ kernel and an FFT path. We used 6 Fourier modes (for $x$, $y$, and $z$ components) for point-wise weight multiplication, and overall used 96 hidden channels. Interestingly, using more layers or more Fourier modes degraded performances. Similar to the 2D experiments, we applied GeLU activation functions, and neither apply shortcut connections nor normalization techniques, such as batchnorm or groupnorms. Overall this resulted in roughly 65 million floating point parameters for FNO based architectures. 

\paragraph{Clifford Fourier Neural Operator architectures.}
For every FNO-based experiment, we replaced the fine-tuned FNO architectures with respective Clifford counterparts: each FNO layer is replaced by its Clifford counterpart.
To keep the number of weights similar, instead of 128 channels the resulting architectures have 48 multivector channels, resulting in roughly the same number of parameters.
Additionally, GeLU activation functions are replaced with Clifford GeLU activation functions. Using Clifford initialization techniques did not improve results.

For 3-dimensional Maxwell experiments, we replaced each 3D Fourier transform layer with a 3D Clifford Fourier layer and each 3D convolution with a respective Clifford convolution. We also use 6 Fourier modes (for $x$, $y$, and $z$ components) for point-wise weight multiplication, and overall used 32 hidden multivector channels, which results in roughly the same number of parameters (55 millions). In contrast to 2-dimensional implementations, Clifford initialization techniques proved important for 3-dimensional architectures. Most notably, too large initial values of the weights of Clifford convolution layers hindered gradient flows through the Clifford Fourier operations.

\subsection{Training and model selection.}
We optimized models using the Adam
optimizer~\citep{kingma2014adam} with learning rates
$[10^{-4}, 2\cdot 10^{-4}, 5\cdot 10^{-4}]$ for 50 epochs and minimized the
summed mean squared error (SMSE) which is outlined
in Equation~\ref{eq:app_loss}. We used cosine annealing as learning rate scheduler~\citep{loshchilov2016sgdr} with a linear warmup.
For baseline ResNet models, we optimized number of layers, number of channels, and normalization procedures. We further tested different activation functions. For baseline FNO models, we optimized number of layers, number of channels, and number of Fourier modes. Larger numbers of layers or channels did not improve the performances for both ResNet and FNO models. For the respective Clifford counterparts, we exchanged convolution and Fourier layers by Clifford convolution and Clifford Fourier layers. We further used Clifford normalization schemes. We decreased the number of layers to obtain similar numbers of parameters. We could have optimized Clifford architectures slightly more by e.g. using different numbers of hidden layers than the baseline models did. However, this would (i) slightly be against the argument of having ``plug- and play'' replace layers, and (ii) would have added quite some computational overhead. Finally, we are quite confident that the used architectures are very close to the optimum for the current tasks. %

\paragraph{Computational resources.}
All FNO and CFNO experiments used $4\times$\SI{16}{\giga\byte} NVIDIA V100 machines for training. All ResNet and Clifford ResNet experiments used $8\times$\SI{32}{\giga\byte} NVIDIA V100 machines. Average training times varied between \SI{3}{\hour} and \SI{48}{\hour}, depending on task and number of trajectories. Clifford runs on average took twice as long to train for equivalent architectures and epochs.

\strut\newpage

\clearpage

\subsection{Navier-Stokes in 2D}
The incompressible Navier-Stokes equations are built upon momentum and mass conservation of fluids. 
Momentum conservation yields for the velocity flow field $v$
\begin{align}
    \frac{\partial v}{\partial t} = -v \cdot \nabla v + \mu \nabla^2 v - \nabla p + f \ \label{eq:Navier_Stokes},
\end{align}
where $v \cdot \nabla v$ is the convection, $\mu \nabla^2 v$ the viscosity, $\nabla p$ the internal pressure and $f$ an external force. Convection is the rate of change of a vector field along a vector field (in this case along itself),
viscosity is the diffusion of a vector field, i.e. the net movement form higher valued regions to lower concentration regions, $\mu$ is the viscosity coefficient.
The incompressibility constrained yields mass conservation via 
\begin{align}
    \nabla \cdot v = 0 \ .
\end{align}
Additional to the velocity field $v(x)$, we introduce a scalar field $s(x)$ 
representing a scalar quantity that is being transported through the velocity field. For example, $v$ might represent velocity of air inside a room, and $s$ might represent concentration of smoke. As the vector field changes, the scalar field is transported along it, i.e. the scalar field is \textit{advected} by the vector field. Similar to convection, advection is the transport of a scalar field along a vector field:
\begin{align}
    \frac{ds}{dt} = -v \cdot \nabla s \ .
\end{align}
We implement the 2D Navier-Stokes equation using \texttt{${\Phi}$Flow}\footnote{\url{https://github.com/tum-pbs/PhiFlow}} \citep{holl2020phiflow}. Solutions are propagated where we solve for the pressure field and subtract its spatial gradients afterwards. Semi-Lagrangian advection (convection) is used for $v$, and MacCormack advection for $s$. Additionally, we express the external buoyancy force $f$ in Equation~\ref{eq:Navier_Stokes} as force acting on the scalar field.
Solutions are obtained using Boussinesq approximation~\citep{kleinstreuer1997engineering}, which ignores density differences except where they appear in terms multiplied by the acceleration due to gravity. The essence of the Boussinesq approximation is that the difference in inertia is negligible but gravity is sufficiently strong to make the specific weight appreciably different between the two fluids.

\begin{figure}[!htb]
    \centering
    \input{tikz/smoke/bar_resnet_2e-4.tex}
    \caption{Results on Navier-Stokes equations obtained by ResNet based architectures. Unrolled loss, one-step loss, scalar loss and vector loss are reported for ResNet, CResNet, and CResNet$_{\text{rot}}$ architectures. Models are trained on training sets with increasing number of trajectories. ResNet based architectures have a much higher loss than FNO based architectures in the low data regime, where possibly smearing and averaging operations are learned first.}
    \label{fig:app_resnet_NS_results}
\end{figure}
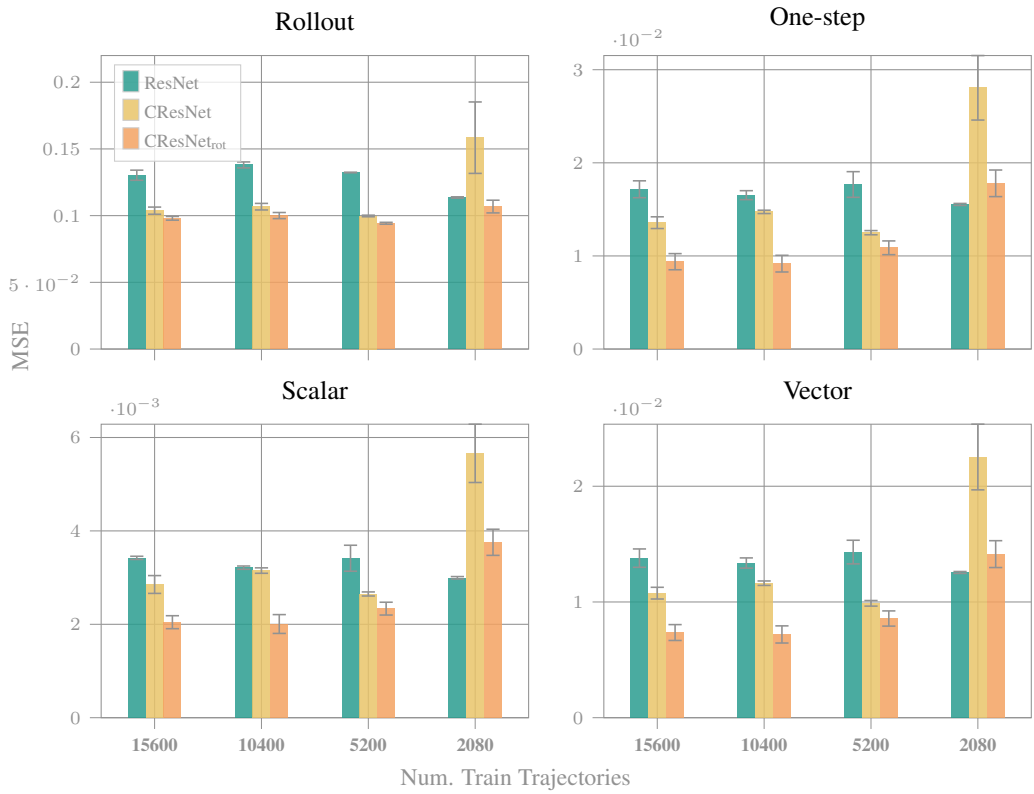

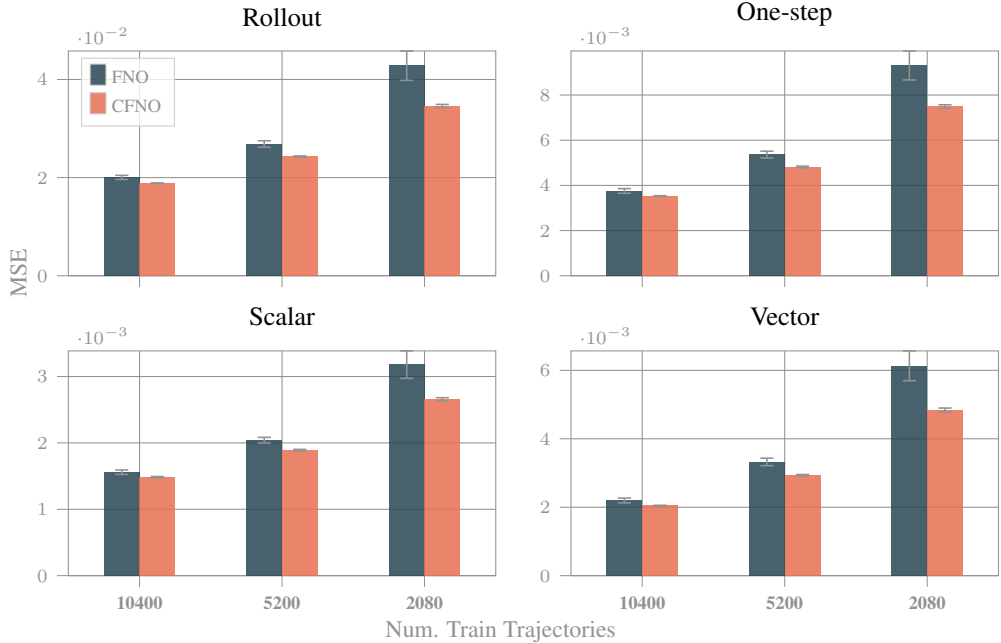
\begin{figure}[!htb]
    \centering
    {\input{tikz/smoke/bar_fourier_2e-4.tex}}
    \caption{Results on Navier-Stokes equations obtained by Fourier based architectures. Rollout loss, one-step loss, scalar loss and vector loss are reported for FNO and CFNO architectures. Models are trained on three training sets with increasing number of trajectories.}
    \label{fig:app_fourier_NS_results}
\end{figure}

\paragraph{Equation details.}
We obtain data for the 2D Navier-Stokes equations on a grid with spatial resolution of $128 \times 128$ ($\Delta x=0.25$, $\Delta y=0.25$), and temporal resolution of $\Delta t = \SI{1.5}\second$. The equation is solved on a closed domain with Dirichlet boundary conditions ($v=0$) for the velocity, and Neumann boundaries $\frac{\partial s}{\partial x} = 0$ for the scalar smoke field. The viscosity parameter is set to $\nu=0.01$, and a buoyancy factor of $(0,0.5)^T$ is used. The scalar field is initialized with random Gaussian noise fluctuations, and the velocity field is initialized to $0$. 
We run the simulation for $\SI{21}\second$ and sample every $\SI{1.5}\second$. Trajectories contain scalar and vector fields at $14$ different time points. 

\begin{figure}[!htb]
    \centering
    \begin{subfigure}[c]{0.8\textwidth}
        \makebox[0pt]{\rotatebox[origin=c]{90}{
            \textbf{\bigskip \small Prediction}
        }\hspace*{1em}}%
        \centering
        \begin{subfigure}[c]{\textwidth}
            \centering
            {\includegraphics[width=\textwidth]{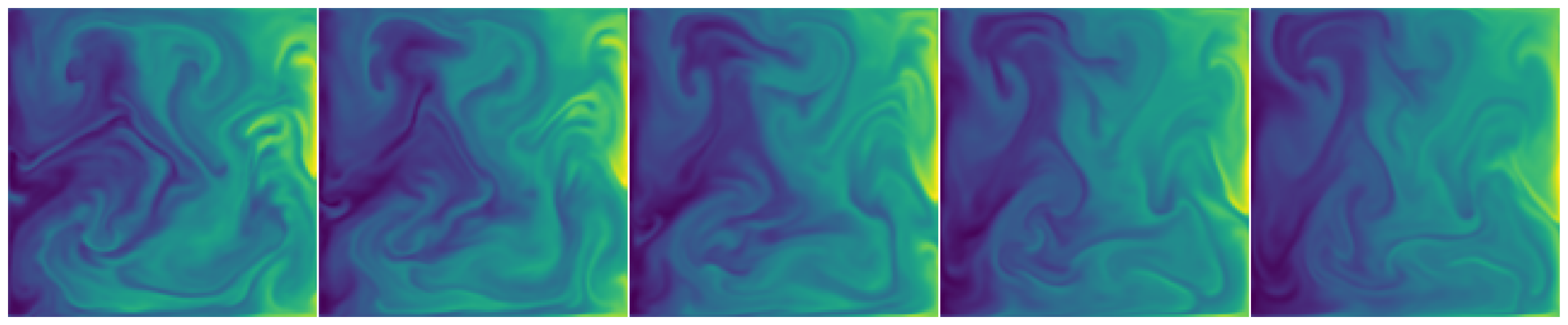}}
        \end{subfigure}\\
        \makebox[0pt]{\rotatebox[origin=c]{90}{
            \textbf{\small Target}
        }\hspace*{1em}}%
        \begin{subfigure}[c]{\textwidth}
            \centering
            {\includegraphics[width=\textwidth]{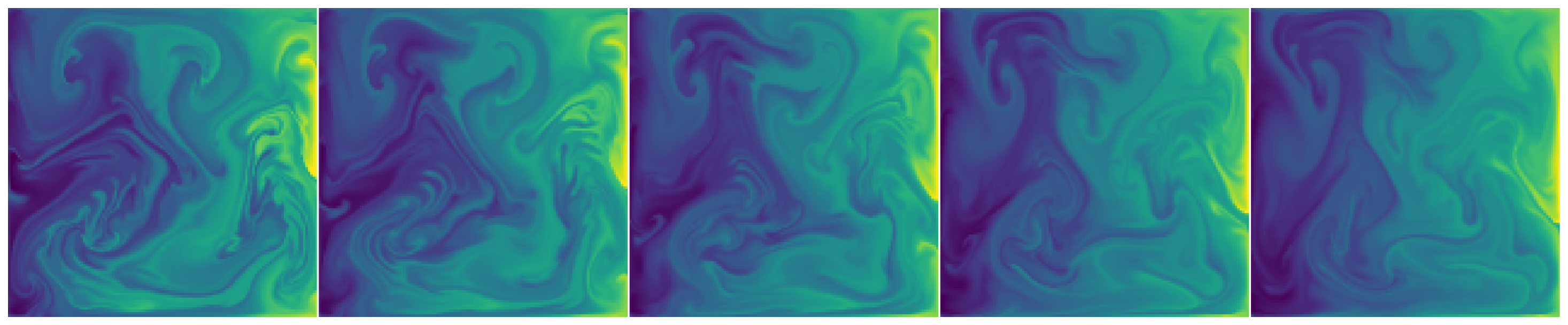}}   
        \end{subfigure}
        \makebox[0pt]{\rotatebox[origin=c]{90}{
            \textbf{\bigskip \small Prediction}
        }\hspace*{1em}}%
        \centering
        \begin{subfigure}[c]{\textwidth}
            \centering
            {\includegraphics[width=\textwidth]{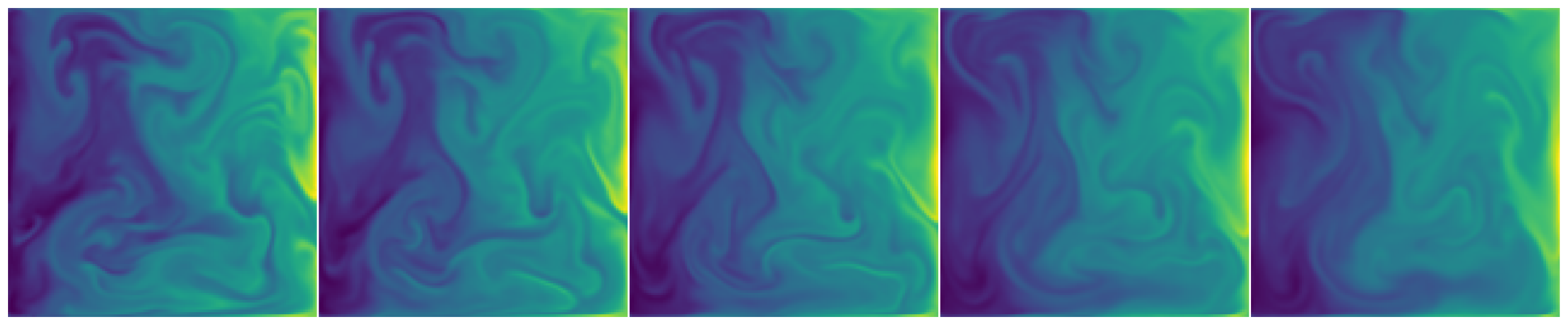}}
        \end{subfigure}\\
        \makebox[0pt]{\rotatebox[origin=c]{90}{
            \textbf{\small Target}
        }\hspace*{1em}}%
        \begin{subfigure}[c]{\textwidth}
            \centering
            {\includegraphics[width=\textwidth]{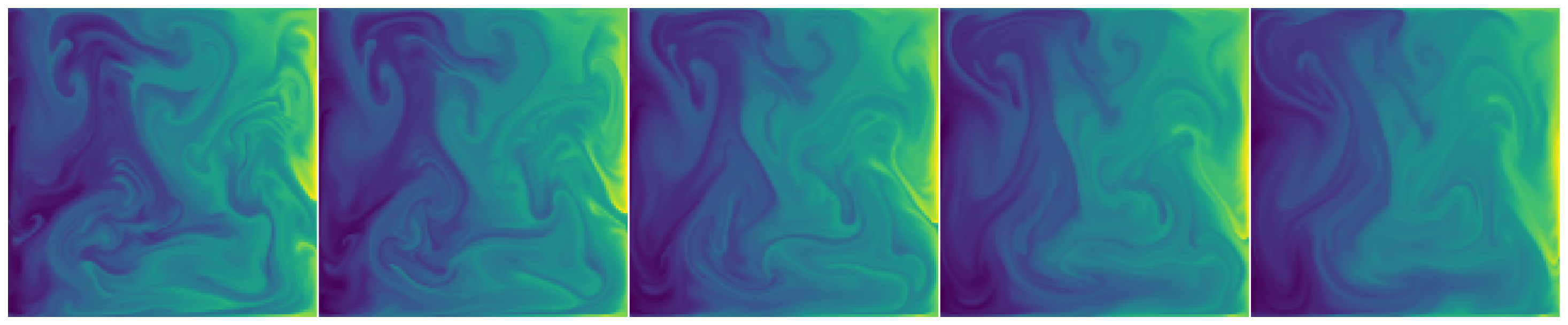}}   
        \end{subfigure}
        \caption{Scalar field}
    \end{subfigure}
    \begin{subfigure}[c]{0.8\textwidth}
        \makebox[0pt]{\rotatebox[origin=c]{90}{
            \textbf{\bigskip \small Prediction}
        }\hspace*{1em}}%
        \centering
        \begin{subfigure}[c]{\textwidth}
            \centering
            {\includegraphics[width=\textwidth]{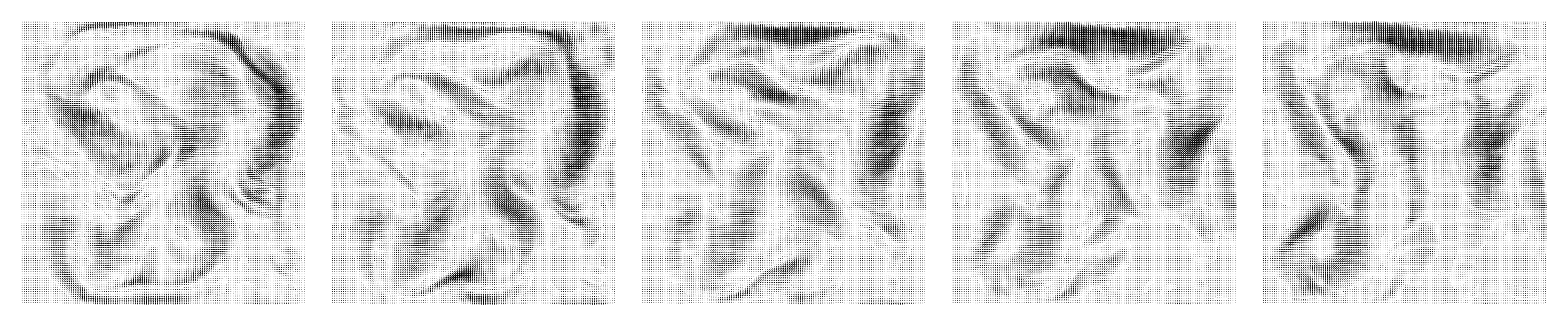}}
        \end{subfigure}\\
        \makebox[0pt]{\rotatebox[origin=c]{90}{
            \textbf{\small Target}
        }\hspace*{1em}}%
        \begin{subfigure}[c]{\textwidth}
            \centering
            {\includegraphics[width=\textwidth]{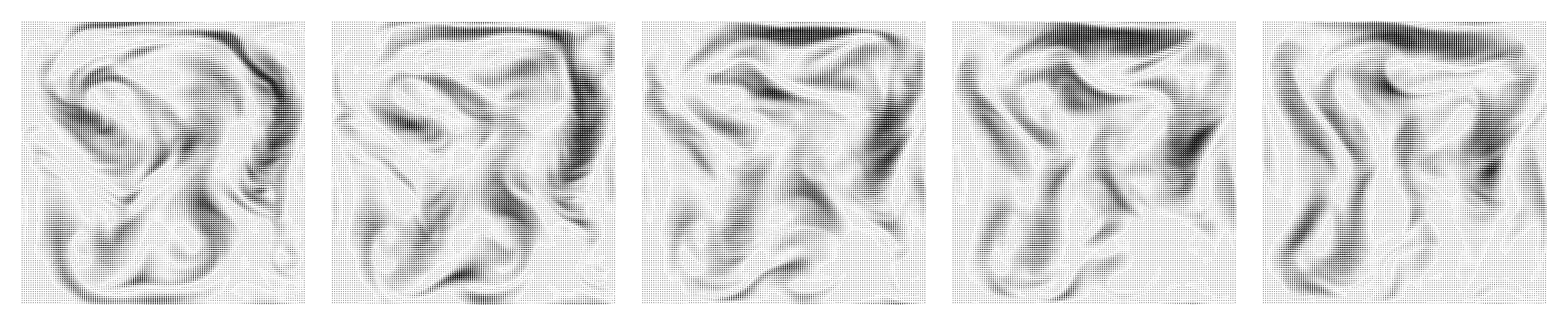}}   
        \end{subfigure}
        \makebox[0pt]{\rotatebox[origin=c]{90}{
            \textbf{\bigskip \small Prediction}
        }\hspace*{1em}}%
        \centering
        \begin{subfigure}[c]{\textwidth}
            \centering
            {\includegraphics[width=\textwidth]{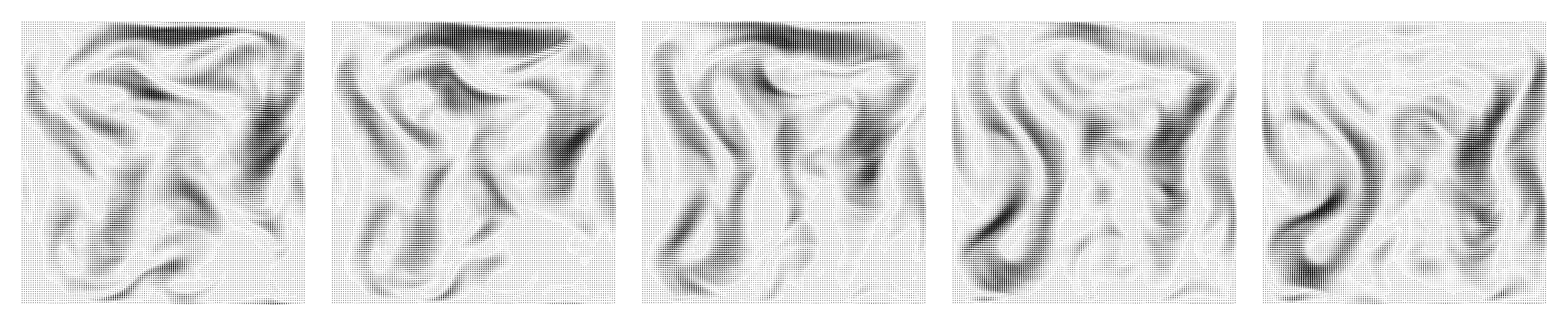}}
        \end{subfigure}\\
        \makebox[0pt]{\rotatebox[origin=c]{90}{
            \textbf{\small Target}
        }\hspace*{1em}}%
        \begin{subfigure}[c]{\textwidth}
            \centering
            {\includegraphics[width=\textwidth]{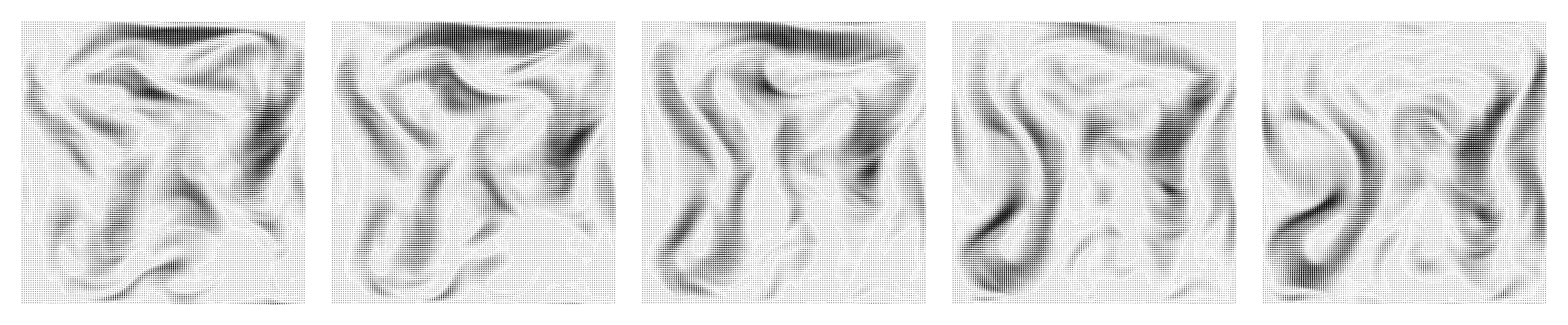}}   
        \end{subfigure}        
        \caption{Vector field}
    \end{subfigure}
    \caption{Example rollouts of the scalar and vector field of the Navier-Stokes experiments, obtained by a Clifford Fourier PDE surrogate and the ground truth.} %
    \label{fig:app_NS_rollout}
\end{figure}

\paragraph{Results.} Results are summarized in Figures~\ref{fig:app_fourier_NS_results},~\ref{fig:app_resnet_NS_results}, and detailed in Table~\ref{tab:app_NS_large}.
Figure~\ref{fig:app_NS_rollout} displays examples of Navier-Stokes rollouts of scalar and vector fields obtained by Clifford Fourier surrogates, and contrasts them with ground truth trajectories. 
For ResNet-like architectures, we observe that both CResNet and CResNet$_{\text{rot}}$ improve upon the ResNet baseline.
Additionally, we observe that rollout losses are also lower for the two Clifford based architectures, which we attribute to better and more stable models that do not overfit to one-step predictions so easily. 
Lastly, while in principle CResNet and CResNet$_{\text{rot}}$ based architectures are equally flexible, CResNet$_{\text{rot}}$ ones in general perform better than CResNet ones. 
For FNO and respective Clifford Fourier based (CFNO) architectures, the loss is in general much lower than for ResNet based architectures. CFNO architectures improve upon FNO architectures for all dataset sizes, and for one-step as well as rollout losses.

\begin{table*}[!htb]
    \centering
    \caption{Model comparison on four different metrics for neural PDE surrogates which are trained on Navier-Stokes training datasets of varying size. Error bars are obtained by running experiments with three different initial seeds.}
    \sisetup{separate-uncertainty=true}
    \resizebox{\textwidth}{!}{\begin{tabular}{rrS[table-format=1.5(6)]S[table-format=1.5(6)]S[table-format=1.5(6)]S[table-format=1.5(6)]}
        \toprule
        \multirow{2}{*}[-1em]{\textsc{Method}} & \multirow{2}{*}[-1em]{Trajs.} & \multicolumn{4}{c}{SMSE} \\%\addlinespace[2.5pt]
        \cmidrule(lr){3-6} \\
        {} & {} & {scalar} & {vector} & {onestep} & {rollout} \\
        \midrule
        ResNet & & 0.00300 \pm 0.00003 & 0.01255 \pm 0.00008 & 0.01553 \pm 0.00011 & 0.11362 \pm 0.00048 \\
        CResNet & 2080 & 0.00566 \pm 0.00062 & 0.02252 \pm 0.00284 & 0.02806 \pm 0.00346 & 0.15844 \pm 0.02677 \\
        CResNet$_{\text{rot}}$ &  & 0.00376 \pm 0.00028 & 0.01413 \pm 0.00116 & 0.01780 \pm 0.00143 & 0.10681 \pm 0.00476 \\
        \midrule
        ResNet &  & 0.00341 \pm 0.00028 & 0.01431 \pm 0.00102 & 0.01767 \pm 0.00138 & 0.13234 \pm 0.00020 \\
        CResNet & 5200 & 0.00265 \pm 0.00004 & 0.00988 \pm 0.00024 & 0.01250 \pm 0.00022 & 0.09975 \pm 0.00060 \\
        CResNet$_{\text{rot}}$ &  & 0.00234 \pm 0.00014 & 0.00857 \pm 0.00066 & 0.01087 \pm 0.00074 & 0.09427 \pm 0.00071 \\
        \midrule
        ResNet & & 0.00321 \pm 0.00004 & 0.01337 \pm 0.00044 & 0.01653 \pm 0.00048 & 0.13802 \pm 0.00223 \\
        CResNet & 10400 & 0.00315 \pm 0.00006 & 0.01162 \pm 0.00019 & 0.01473 \pm 0.00018 & 0.10671 \pm 0.00246 \\
        CResNet$_{\text{rot}}$ &  & 0.00201 \pm 0.00020 & 0.00719 \pm 0.00074 & 0.00917 \pm 0.00090 & 0.10005 \pm 0.00229 \\
        \midrule
        ResNet &  & 0.00342 \pm 0.00003 & 0.01379 \pm 0.00079 & 0.01716 \pm 0.00091 & 0.13030 \pm 0.00379 \\
        CResNet & 15600 & 0.00285 \pm 0.00019 & 0.01076 \pm 0.00051 & 0.01357 \pm 0.00063 & 0.10372 \pm 0.00269 \\
        CResNet$_{\text{rot}}$ &  & 0.00204 \pm 0.00014 & 0.00736 \pm 0.00069 & 0.00938 \pm 0.00087 & 0.09799 \pm 0.00139 \\
        \midrule
        FNO & 2080 & 0.00318 \pm 0.00021 & 0.00613 \pm 0.00044 & 0.00931 \pm 0.00064 & 0.04281 \pm 0.00300 \\
        CFNO &  & 0.00266 \pm 0.00002 & 0.00484 \pm 0.00006 & 0.00749 \pm 0.00008 & 0.03461 \pm 0.00031 \\
        \midrule
        FNO & 5200 & 0.00204 \pm 0.00004 & 0.00332 \pm 0.00011 & 0.00536 \pm 0.00015 & 0.02684 \pm 0.00067 \\
        CFNO &  & 0.00189 \pm 0.00001 & 0.00293 \pm 0.00002 & 0.00482 \pm 0.00003 & 0.02430 \pm 0.00012 \\
        \midrule
        FNO & 10400 & 0.00156 \pm 0.00003 & 0.00220 \pm 0.00007 & 0.00375 \pm 0.00010 & 0.02005 \pm 0.00042 \\
        CFNO &  & 0.00148 \pm 0.00001 & 0.00205 \pm 0.00001 & 0.00353 \pm 0.00002 & 0.01886 \pm 0.00006 \\
        \bottomrule
    \end{tabular}}
    \label{tab:app_NS_large}
\end{table*}

\strut\newpage

\clearpage

\subsection{Shallow water equations.}
The shallow water equations~\citep{vreugdenhil1994numerical} describe a thin layer of fluid of constant density in hydrostatic
balance, bounded from below by the bottom topography and from above by a free surface. For example, the deep water propagation of a tsunami can be described by the shallow water equations, and so can a simple weather model.
The shallow water equations read:
\begin{align}
    \frac{\partial v_x}{\partial t} + v_x \frac{\partial v_x}{\partial x} + v_y \frac{\partial v_x}{\partial y} + g\frac{\partial \eta}{\partial x} & = 0 \ , \nonumber \\
    \frac{\partial v_y}{\partial t} + v_x \frac{\partial v_y}{\partial x} + v_y \frac{\partial v_y}{\partial y} + g\frac{\partial \eta}{\partial y} & = 0 \ , \nonumber \\ 
    \frac{\partial \eta}{\partial t} + \frac{\partial}{\partial x} \bigg[ (\eta + h) v_x \bigg] + \frac{\partial}{\partial y} \bigg[ (\eta + h) v_y \bigg] & = 0 \ ,
\end{align}
where $v_x$ is the velocity in the $x$-direction, or zonal velocity,
$v_y$ is the velocity in the $y$-direction, or meridional velocity,
$g$ is the acceleration due to gravity,
$\eta(x,y)$ is the vertical displacement of free surface, which
subsequently is used to derive pressure fields; $h(x,y)$ is the topography of the earth's surface. 
We modify the implementation in \texttt{SpeedyWeather.jl}\footnote{\url{https://github.com/milankl/SpeedyWeather.jl}}\citep{milan_klower_2022_6788067} to further randomize initial conditions to generate our dataset.
\texttt{SpeedyWeather.jl} combines the shallow water equations with spherical harmonics for the linear terms and Gaussian grid for the non-linear terms with the appropriate spectral transforms. It internally uses a leapfrog time scheme with a Robert and William's filter to dampen the computational modes and achieve 3rd oder accuracy. \texttt{SpeedyWeather.jl} is based on the atmospheric general circulation model \texttt{SPEEDY} in Fortran~\citep{molteni2003atmospheric,Kucharski2013}.

\paragraph{Equation details.}
We obtain data for the 2D shallow water equations on a grid with spatial resolution of $192 \times 96$ ($\Delta x=1.875\degree$, $\Delta y=3.75\degree$), and temporal resolution of $\Delta t = \SI{6}\hour$. The equation is solved on a closed domain with periodic boundary conditions. We rollout the simulation for $20$ days and sample every $\SI{6}\hour$. Here $20$ days is of course not the actual simulation time but rather the simulated time. Trajectories contain scalar pressure and  wind vector fields at $84$ different time points. 

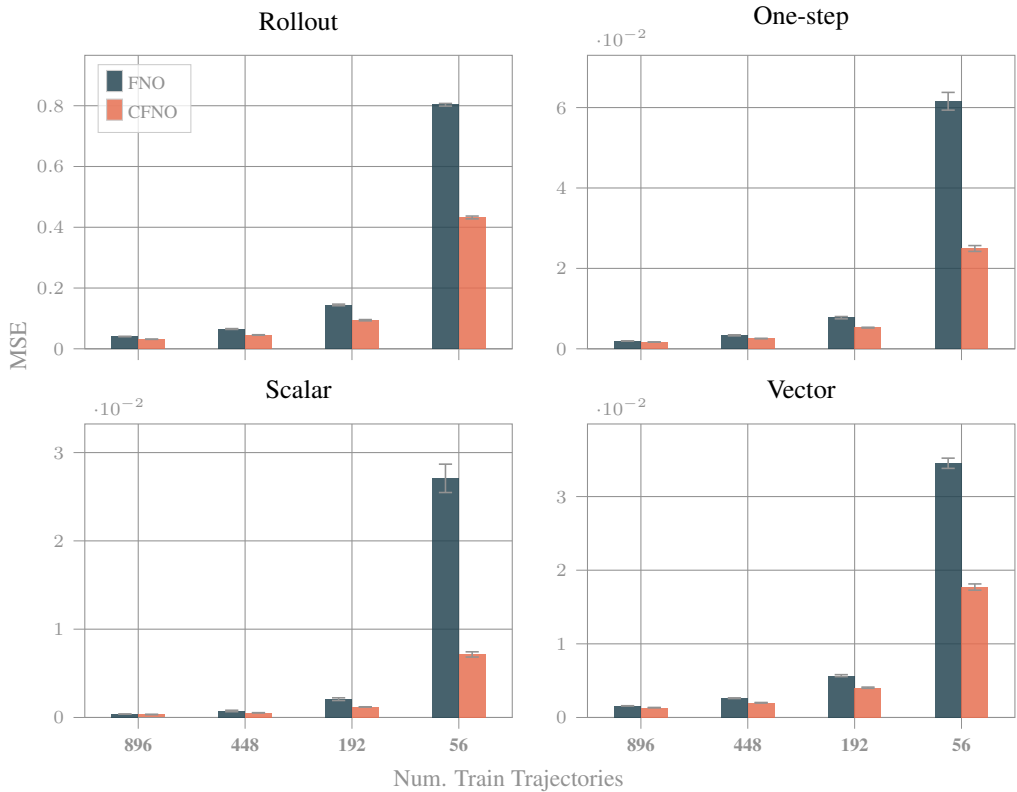
\begin{figure}[!htb]
    \centering
    {\input{tikz/weather/bar_fourier_2e-4_hist=2.tex}}
    \caption{Results on the shallow water equations obtained by Fourier based architectures using a \textbf{two timestep history input}. Unrolled loss, one-step loss, scalar loss and vector loss are reported for FNO and CFNO architectures. Models are trained on three training sets with increasing number of trajectories.}
    \label{fig:app_SW_FNO_results_hist2}
\end{figure}

\begin{figure}[tb]
    \centering
    \begin{subfigure}[c]{\textwidth}
        \makebox[0pt]{\rotatebox[origin=c]{90}{
            \textbf{\bigskip \footnotesize Prediction}
        }\hspace*{1em}}%
        \begin{subfigure}[c]{\textwidth}
            \centering
            {\includegraphics[width=\textwidth]{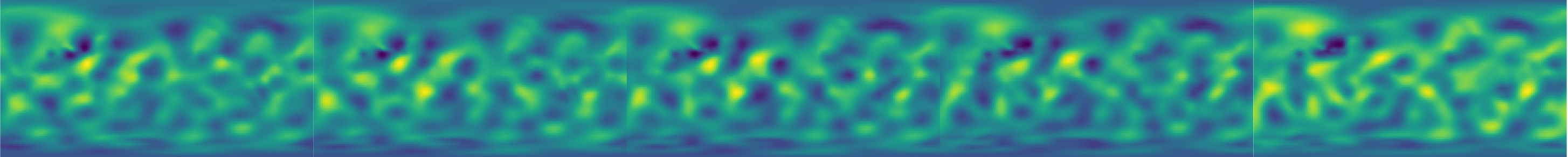}}
        \end{subfigure}
        \makebox[0pt]{\rotatebox[origin=c]{90}{
            \textbf{\bigskip \footnotesize Target}
        }\hspace*{1em}}%
        \begin{subfigure}[c]{\textwidth}
            \centering
            {\includegraphics[width=\textwidth]{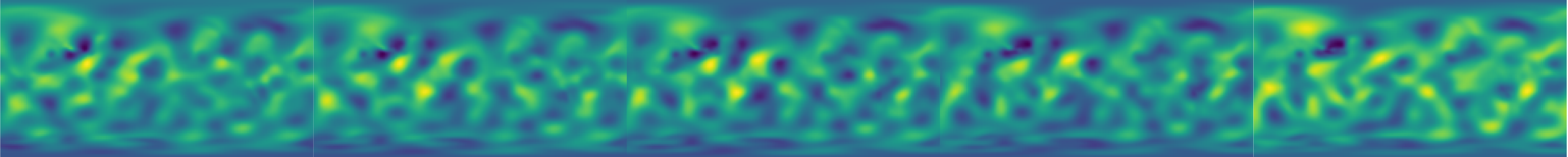}}    
        \end{subfigure}
        \makebox[0pt]{\rotatebox[origin=c]{90}{
            \textbf{\bigskip \footnotesize Prediction}
        }\hspace*{1em}}%
        \begin{subfigure}[c]{\textwidth}
            \centering
            {\includegraphics[width=\textwidth]{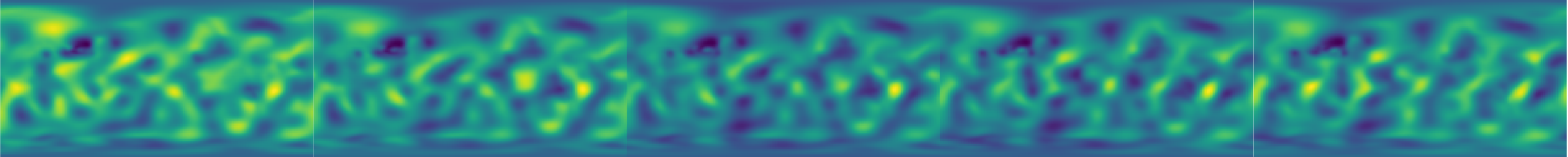}}
        \end{subfigure}
        \makebox[0pt]{\rotatebox[origin=c]{90}{
            \textbf{\bigskip \footnotesize Target}
        }\hspace*{1em}}%
        \begin{subfigure}[c]{\textwidth}
            \centering
            {\includegraphics[width=\textwidth]{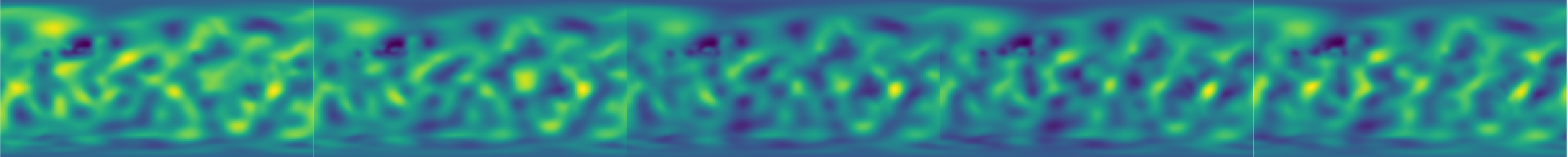}}    
        \end{subfigure}        
        \caption{Scalar field}
    \end{subfigure}
    \begin{subfigure}[c]{\textwidth}
        \makebox[0pt]{\rotatebox[origin=c]{90}{
            \textbf{\bigskip \footnotesize Prediction}
        }\hspace*{1em}}%
        \begin{subfigure}[c]{\textwidth}
            \centering
            {\includegraphics[width=\textwidth]{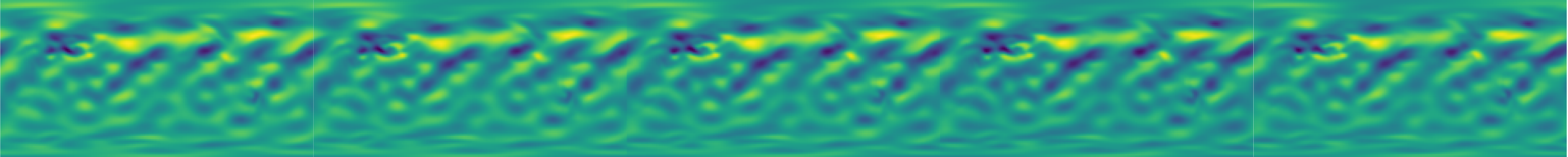}}
        \end{subfigure}
            \makebox[0pt]{\rotatebox[origin=c]{90}{
                \textbf{\bigskip \footnotesize Target}
            }\hspace*{1em}}%
        \begin{subfigure}[c]{\textwidth}
            \centering
            {\includegraphics[width=\textwidth]{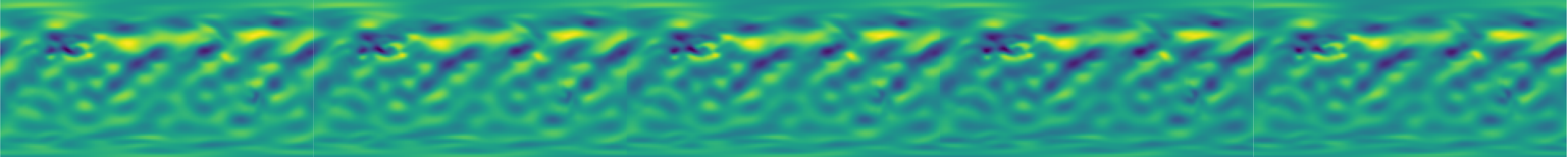}}    
        \end{subfigure}
        \makebox[0pt]{\rotatebox[origin=c]{90}{
            \textbf{\bigskip \footnotesize Prediction}
        }\hspace*{1em}}%
        \begin{subfigure}[c]{\textwidth}
            \centering
            {\includegraphics[width=\textwidth]{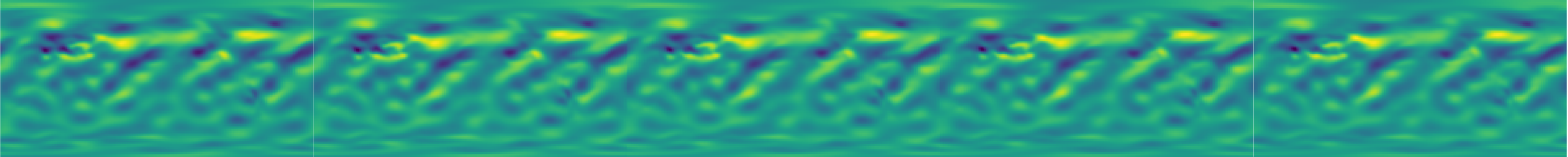}}
        \end{subfigure}
            \makebox[0pt]{\rotatebox[origin=c]{90}{
                \textbf{\bigskip \footnotesize Target}
            }\hspace*{1em}}%
        \begin{subfigure}[c]{\textwidth}
            \centering
            {\includegraphics[width=\textwidth]{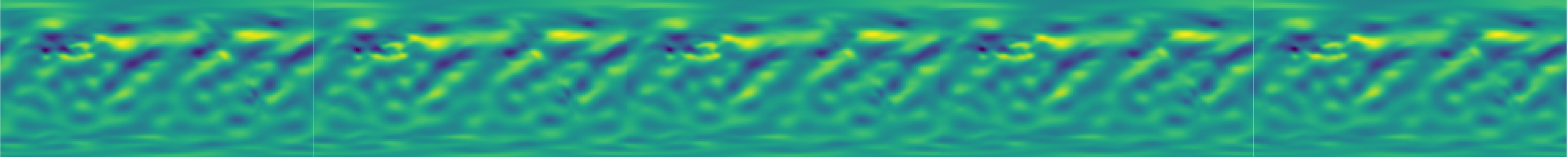}}    
        \end{subfigure}        
        \caption{Vector field ($x$-component)}
    \end{subfigure}
    \begin{subfigure}[c]{\textwidth}
        \makebox[0pt]{\rotatebox[origin=c]{90}{
            \textbf{\bigskip \footnotesize Prediction}
        }\hspace*{1em}}%
        \begin{subfigure}[c]{\textwidth}
            \centering
            {\includegraphics[width=\textwidth]{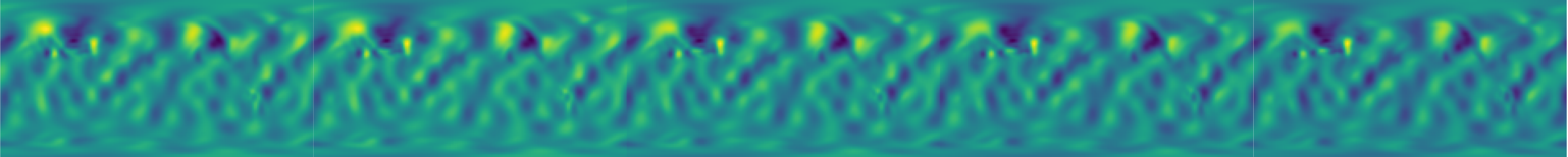}}
        \end{subfigure}
            \makebox[0pt]{\rotatebox[origin=c]{90}{
                \textbf{\bigskip \footnotesize Target}
            }\hspace*{1em}}%
        \begin{subfigure}[c]{\textwidth}
            \centering
            {\includegraphics[width=\textwidth]{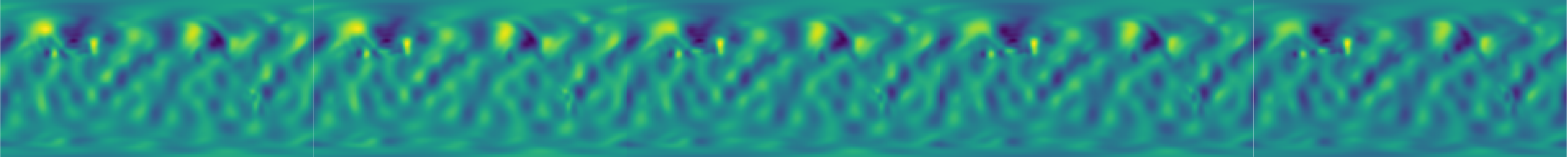}}    
        \end{subfigure}
        \makebox[0pt]{\rotatebox[origin=c]{90}{
            \textbf{\bigskip \footnotesize Prediction}
        }\hspace*{1em}}%
        \begin{subfigure}[c]{\textwidth}
            \centering
            {\includegraphics[width=\textwidth]{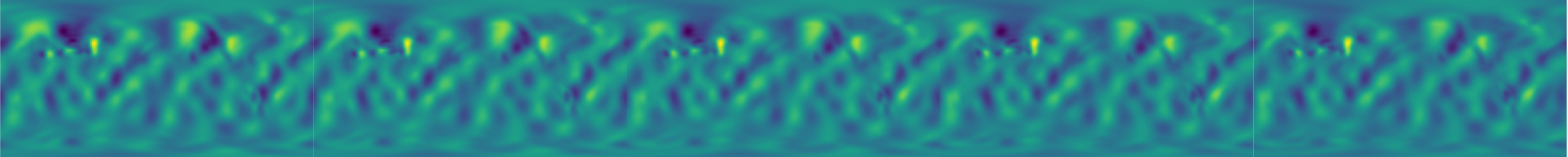}}
        \end{subfigure}
            \makebox[0pt]{\rotatebox[origin=c]{90}{
                \textbf{\bigskip \footnotesize Target}
            }\hspace*{1em}}%
        \begin{subfigure}[c]{\textwidth}
            \centering
            {\includegraphics[width=\textwidth]{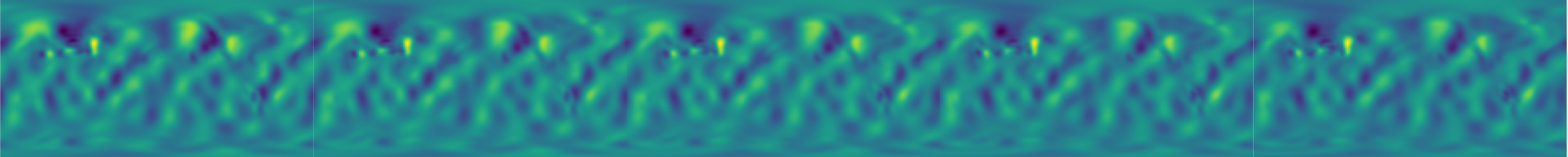}}    
        \end{subfigure}        
        \caption{Vector field ($y$-component)}
    \end{subfigure}
    \caption{Example rollouts of the scalar and vector field of the shallow water experiments, obtained by a Clifford Fourier PDE surrogate (top) and the ground truth (bottom).}
    \label{fig:app_SW_rollout}
\end{figure}

\paragraph{Results.} Results are summarized in Figures~\ref{fig:app_SW_FNO_results_hist2},~\ref{fig:app_SW_FNO_results_hist4},~\ref{fig:app_SW_ResNet_results_hist2}, and detailed in Tables~\ref{tab:app_SW_large_hist2},~\ref{tab:app_SW_large_hist4}.
Figure~\ref{fig:app_SW_rollout} displays examples of shallow water equations rollouts of scalar pressure and vector wind fields obtained by Clifford Fourier surrogate models, and contrasts them with ground truth trajectories. The predictions are fairly indistinguishable from ground truth trajectories.
We observe similar results than for the Navier-Stokes experiments. However,
performance differences between baseline and Clifford architectures are even more pronounced, which we attribute to the stronger coupling of the scalar and the vector fields. 
For ResNet-like architectures, CResNet and CResNet$_{\text{rot}}$ improve upon the ResNet baseline, rollout losses are much lower for the two Clifford based architectures, and CResNet$_{\text{rot}}$ based architectures in general perform better than CResNet based ones. 
For Fourier based architectures, the loss is in general much lower than for ResNet based architectures (a training set size of 56 trajectories yields similar (C)FNO test set performance than a training set size of 896 trajectories for ResNet based architectures). CFNO architectures improve upon FNO architectures for all dataset sizes, and for one-step as well as rollout losses, which is especially pronounced for low number of training trajectories.

\begin{figure}
    \centering
    {\input{tikz/weather/bar_fourier_2e-4_hist=4.tex}}
    \caption{Results on the shallow water equations obtained by Fourier based architectures using a \textbf{four timestep history input}. Rollout loss, one-step loss, scalar loss and vector loss are reported for FNO and CFNO architectures. Models are trained on three training sets with increasing number of trajectories.}
    \label{fig:app_SW_FNO_results_hist4}
\end{figure}
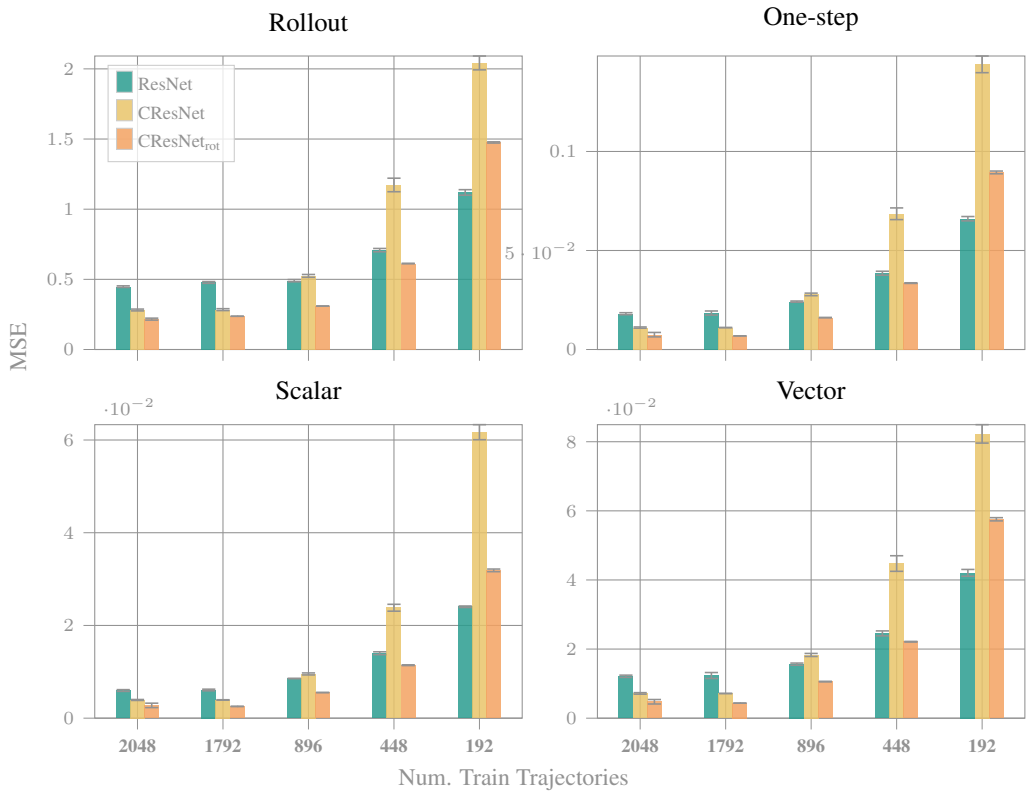
\begin{figure}
    \centering
    {\input{tikz/weather/bar_resnet_2e-4_hist=2}}
    \caption{Results on the shallow water equations obtained by ResNet based architectures using a \textbf{two timestep history input}. Rollout loss, one-step loss, scalar loss and vector loss are reported for ResNet, CResNet, and CResNet$_{\text{rot}}$ architectures. Models are trained on training sets with increasing number of trajectories. ResNet based architectures have a much higher loss than FNO based architectures in the low data regime, where possibly smearing and averaging operations are learned first.}
    \label{fig:app_SW_ResNet_results_hist2}
\end{figure}

\begin{table*}[htb]
    \centering
    \caption{Model comparison on four different metrics for neural PDE surrogates which are trained on the shallow water equations training datasets of varying size. Results are obtained by using a \textbf{two timestep history input}. Error bars are obtained by running experiments with three different initial seeds.}
    \sisetup{separate-uncertainty=true}
    {\begin{tabular}{rrS[table-format=1.4(5)]S[table-format=1.4(5)]S[table-format=1.4(5)]S[table-format=1.4(5)]}
        \toprule
        \multirow{2}{*}[-1em]{\textsc{Method}} & \multirow{2}{*}[-1em]{Trajs.} & \multicolumn{4}{c}{SMSE} \\ %
        \cmidrule(lr){3-6}
        {} & {} & {scalar} & {vector} & {onestep} & {rollout} \\
        \midrule
        ResNet &  & 0.0240 \pm 0.0002 & 0.0421 \pm 0.0010 & 0.0661 \pm 0.0011 & 1.1195 \pm 0.0197 \\
        CResNet & 192 & 0.0617 \pm 0.0016 & 0.0823 \pm 0.0027 & 0.1440 \pm 0.0042 & 2.0423 \pm 0.0494 \\
        CResNet$_{\text{rot}}$ &  & 0.0319\pm 0.0003 & 0.0576 \pm 0.0005 & 0.0894 \pm 0.0007 & 1.4756 \pm 0.0044 \\
        \midrule
        ResNet &  & 0.0140 \pm 0.0003 & 0.0245 \pm 0.0007 & 0.0385 \pm 0.0010 & 0.7083 \pm 0.0119 \\
        CResNet & 448 & 0.0238 \pm 0.0007 & 0.0448 \pm 0.0023 & 0.0685 \pm 0.0030 & 1.1727 \pm 0.0483 \\
        CResNet$_{\text{rot}}$ &  & 0.0114 \pm 0.0001 & 0.0221 \pm 0.0001 & 0.0335 \pm 0.0002 & 0.6127 \pm 0.0018 \\
        \midrule
        ResNet &  & 0.0086 \pm 0.0000 & 0.0156 \pm 0.0003 & 0.0242 \pm 0.0003 & 0.4904 \pm 0.0080\\
        CResNet & 896 & 0.0095 \pm 0.0002 & 0.0183\pm 0.0004 & 0.0278\pm 0.0006 & 0.5247 \pm 0.0101 \\
        CResNet$_{\text{rot}}$ &  & 0.0055\pm 0.0000 & 0.0106\pm 0.0001 & 0.0161\pm 0.0001 & 0.3096\pm 0.0010\\
        \midrule
        ResNet &  & 0.0061\pm 0.0002 & 0.0123\pm 0.0009 & 0.0184\pm 0.0010 & 0.4780\pm 0.0062 \\
        CResNet & 1792 & 0.0039\pm 0.0000 & 0.0071\pm 0.0000 & 0.0111\pm 0.0001 & 0.2842\pm 0.0067 \\
        CResNet$_{\text{rot}}$ &  & 0.0025\pm 0.0000 & 0.0044\pm 0.0000 & 0.0069\pm 0.0000 & 0.2370\pm 0.0000 \\
        \midrule
        ResNet &  & 0.0060\pm 0.0002 & 0.0121\pm 0.0003 & 0.0181\pm 0.0005 & 0.4480\pm 0.0058\\
        CResNet & 2048 & 0.0039\pm 0.0001& 0.0072\pm 0.0002 & 0.0111\pm 0.0003 & 0.2816\pm 0.0065\\
        CResNet$_{\text{rot}}$ &  & 0.0028\pm 0.0005  & 0.048\pm 0.0006 & 0.0075\pm 0.0011 & 0.2164\pm 0.0070\\
        \midrule
        FNO & 56 & 0.0271\pm 0.0016 & 0.0345\pm 0.0007 & 0.0616\pm 0.0022 & 0.8032\pm 0.0043 \\
        CFNO & {} & 0.0071\pm 0.0003 & 0.0177\pm 0.0004 & 0.0250\pm 0.0007 & 0.4323\pm 0.0046 \\
        \midrule
        FNO & 192 & 0.0021\pm 0.0002 & 0.0057\pm 0.0001 & 0.0077\pm 0.0003 & 0.1444\pm 0.0026 \\
        CFNO &  {} & 0.0012\pm 0.0000 & 0.0040\pm 0.0001 & 0.0053 \pm 0.0001 & 0.0941\pm 0.0021 \\
        \midrule
        FNO & 448 & 0.0007\pm 0.0001 & 0.0026\pm 0.0000 & 0.0034\pm 0.0001 & 0.0651\pm 0.0014 \\
        CFNO & {} & 0.0005\pm 0.0000 & 0.0020\pm 0.0000 & 0.0026\pm 0.0001 & 0.0455\pm 0.0009  \\
        \midrule
        FNO & 896 & 0.0004\pm 0.0000 & 0.0016 \pm 0.0000 & 0.0020\pm 0.0001 & 0.0404\pm 0.0005 \\
        CFNO &  {} & 0.0003\pm 0.0000 & 0.0013 \pm 0.0000 & 0.0017\pm 0.0001 & 0.0315\pm 0.0004 \\
        \bottomrule
    \end{tabular}}
    \label{tab:app_SW_large_hist2}
\end{table*}

\begin{table*}[htb]
    \centering
    \caption{Model comparison on four different metrics for neural PDE surrogates which are trained on the shallow water equations training datasets of varying size. Results are obtained by using a \textbf{four timestep history input}. Error bars are obtained by running experiments with three different initial seeds.}
    \sisetup{separate-uncertainty=true}
    {\begin{tabular}{rrS[table-format=1.4(5)]S[table-format=1.4(5)]S[table-format=1.4(5)]S[table-format=1.4(5)]}
        \toprule
        \multirow{2}{*}[-1em]{\textsc{Method}} & \multirow{2}{*}[-1em]{Trajs.} & \multicolumn{4}{c}{SMSE} \\%\addlinespace[5pt]
        \cmidrule(lr){3-6}
        {} & {} & {scalar} & {vector} & {onestep} & {rollout} \\
        \midrule
        FNO & 56 & 0.0276 \pm 0.0017 & 0.0388 \pm 0.0023 & 0.0663 \pm 0.0038 & 0.6821 \pm 0.0379 \\
        CFNO &  & 0.0093 \pm 0.0003 & 0.0252 \pm 0.0005 & 0.0345 \pm 0.0009 & 0.4357 \pm 0.0056\\
        \midrule
        FNO & 192 & 0.0033 \pm 0.0007 & 0.0069 \pm 0.0009 & 0.0102 \pm 0.0015 & 0.1612 \pm 0.0057 \\
        CFNO &  & 0.0015 \pm 0.0001 & 0.0050 \pm 0.0003 & 0.0065 \pm 0.0003 & 0.1023 \pm 0.0026 \\
        \midrule
        FNO & 448 & 0.0009 \pm 0.0001 & 0.0023 \pm 0.0002 & 0.0032 \pm 0.0003 & 0.0687 \pm 0.0023 \\
        CFNO &  & 0.0010 \pm 0.0006 & 0.0039 \pm 0.0027 & 0.0050 \pm 0.0033 & 0.1156 \pm 0.0913 \\
        \midrule
        FNO & 896 & 0.0004 \pm 0.0001 & 0.0012 \pm 0.0001 & 0.0016 \pm 0.0001 & 0.0436 \pm 0.0011 \\
        CFNO &  & 0.0003 \pm 0.0000 & 0.0012 \pm 0.0001 & 0.0015 \pm 0.0001 & 0.0353 \pm 0.0010 \\
        \bottomrule
    \end{tabular}}\label{tab:app_SW_large_hist4}
\end{table*}

\strut\newpage

\clearpage

\subsection{Maxwell's equations in matter in 3D.}
Electromagnetic simulations play a critical role in understanding light–matter interaction and designing optical elements. 
Neural networks have  been already successful applied in  inverse-designing  photonic  structures~\citep{ma2021deep, lim2022maxwellnet}.

Maxwell's equations in matter read:
\begin{align}
    \nabla \cdot D & = \rho \ \ && \text{Gauss's law} \\
    \nabla \cdot B & = 0 \ \ && \text{Gauss's law for magnetism} \\
    \nabla \times E & = - \frac{\partial B}{\partial t} \ \ && \text{Faraday's law of induction} \\
    \nabla \times H & = \frac{\partial D}{\partial t} + \vj \ \ && \text{Ampère's circuital law}
\end{align}
In isotropic media, the displacement field $D$ is related to the electrical field via $D = \epsilon_0 \epsilon_r E$, where $\epsilon_0$ is the permittivity of free space and $\epsilon_r$ is the permittivity of the media.
Similarly, the magnetization field $H$ in isotropic media is related to the magnetic field $B$ via $H = \mu_0 \mu_r B$, where $\mu_0$ is the permeability of free space and $\mu_r$ is the permeability of the media. Lastly, $\vj$ is the electric current density and $\rho$ the total electric charge density.

We propagate the solution of Maxwell's equation in matter using a finite-difference time-domain method\footnote{\url{https://github.com/flaport/fdtd}},
where the discretized Maxwell's equations are solved in a leapfrog manner.
First, the electric field vector components in a volume of space are solved at a given instant in time. Second, the magnetic field vector components in the same spatial volume are solved at the next instant in time.

\begin{figure}[!htb]
    \centering
    \begin{subfigure}[b]{0.325\textwidth}
        \resizebox{\columnwidth}{!}{\input{tikz/maxwell/trajfig/maxwell_1.tex}}
    \end{subfigure}
    \begin{subfigure}[b]{0.325\textwidth}
        \resizebox{\columnwidth}{!}{\input{tikz/maxwell/trajfig/maxwell_2.tex}}
    \end{subfigure}
    \begin{subfigure}[b]{0.325\textwidth}
        \resizebox{\columnwidth}{!}{\input{tikz/maxwell/trajfig/maxwell_3.tex}}
    \end{subfigure}
    \caption{An example propagation of the displacement field $D$ and the magnetization field $H$. Shown are the field components for an arbitrary slice of the $x\mathrm{-}y$ plane.}
    \label{fig:app_maxwell}
\end{figure}
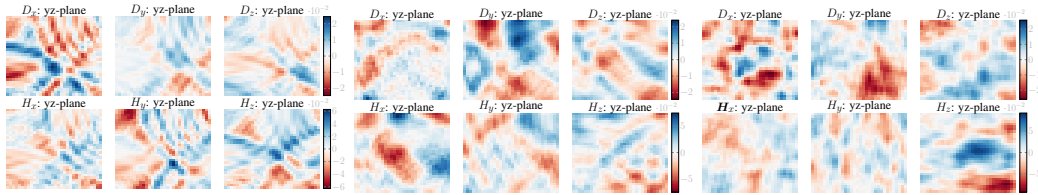

\paragraph{Equation details.}
We obtain data for the 3D Maxwell's equations on a grid with spatial resolution of $32 \times 32 \times 32$ ($\Delta x = \Delta y = \Delta z = 5\cdot 10^{-7}m$), and temporal resolution of $\Delta t = \SI{50}{\second}$. 
We randomly place 18 (6 in the $x\mathrm{-}y$ plane, 6 in the $x\mathrm{-}z$ plane, 6 in the $y\mathrm{-}z$ plane)  different light sources outside a cube which emit light with different amplitude and different phase shifts, causing the resulting $D$ and $H$ fields to interfere with each other. The wavelength of the emitted light is $10^{-5}m$. 
The equation is solved on a closed domain with periodic boundary conditions. We run the simulation for \SI{400}{\second} and sample data every \SI{50}{\second}. Trajectories contain displacement $D$ and the magnetization field $H$ components. Exemplary trajectories are shown in Figure~\ref{fig:app_maxwell}.

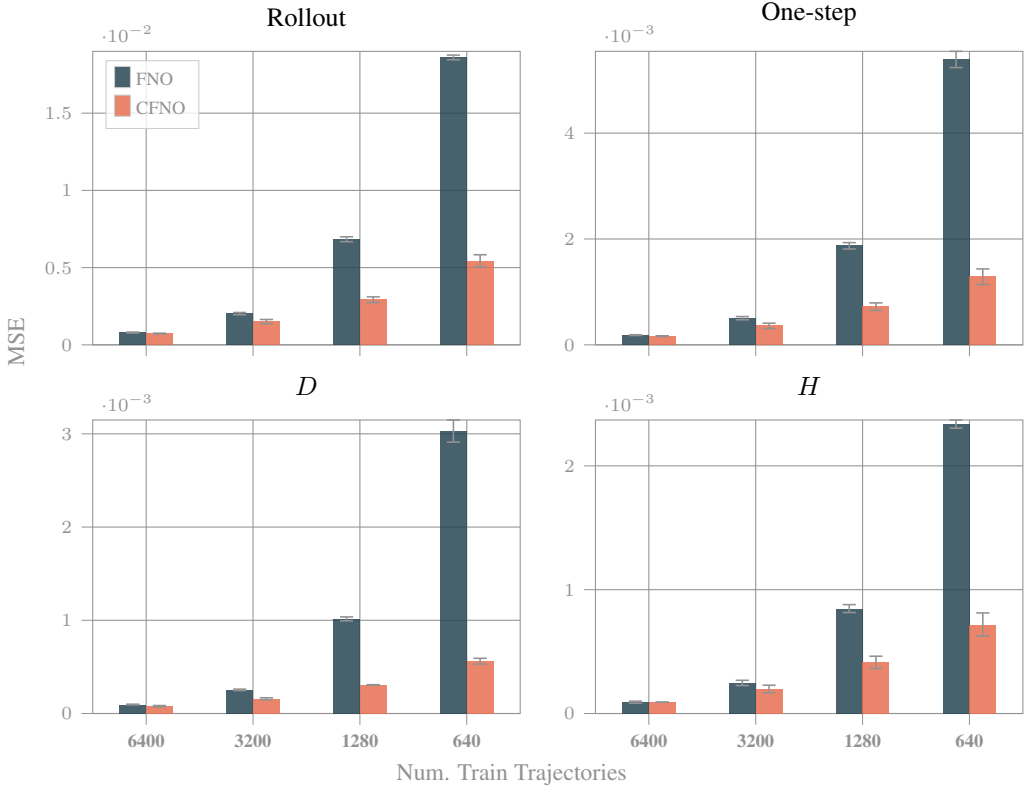
\begin{figure}[!htb]
    \centering
    {\input{tikz/maxwell/bar_fourier_1e-4_hist=2.tex}}
    \caption{Results on the Maxwell equations obtained by Fourier based architectures using a \textbf{two timestep history input}. Rollout loss, one-step loss, displacement field $D$ loss, and magnetization field $H$ loss are reported for FNO and CFNO architectures. Models are trained on four training sets with increasing number of trajectories.}
    \label{fig:app_Maxwell_FNO_results_hist2}
\end{figure}

\paragraph{Results.}
Results are summarized in Figure~\ref{fig:app_Maxwell_FNO_results_hist2} and detailed in Table~\ref{tab:app_Maxwell_large_hist2}.

\begin{table*}[!htb]
    \centering
    \caption{Model comparison on four different metrics for neural PDE surrogates which are trained on the Maxwell equations training datasets of varying size. Results are obtained by using a \textbf{two timestep history input}. Error bars are obtained by running experiments with three different initial seeds.}
    \sisetup{separate-uncertainty=true}
    {\begin{tabular}{rrS[table-format=1.4(5)]S[table-format=1.5(5)]S[table-format=1.4(5)]S[table-format=1.4(5)]}
        \toprule
        \multirow{2}{*}[-1em]{\textsc{Method}} & \multirow{2}{*}[-1em]{Trajs.} & \multicolumn{4}{c}{SMSE} \\%\addlinespace[5pt]
        \cmidrule(lr){3-6}
        {} & {} & {$D$} & {$H$} & {onestep} & {rollout} \\
        \midrule

        FNO & 640 & 0.0030\pm 0.0006 & 0.00233\pm 0.00050 & 0.0054\pm 0.0011& 0.0186\pm 0.0083 \\
        CFNO &  & 0.0006\pm 0.0001 & 0.00072\pm 0.00010 & 0.0013\pm 0.0002& 0.0054\pm 0.0023 \\
        \midrule
        FNO & 1280 & 0.0010\pm 0.0002 & 0.00085\pm 0.00020 & 0.0019\pm 0.0004& 0.0068\pm 0.0036 \\
        CFNO &  & 0.0003\pm 0.0001 & 0.00041\pm 0.00010 & 0.0007\pm 0.0002& 0.0029\pm 0.0016 \\
        \midrule
        FNO & 3200 & 0.0003\pm 0.0001 & 0.00025\pm 0.00010 & 0.0005\pm 0.0001& 0.0020\pm 0.0011 \\
        CFNO &  & 0.0002\pm 0.0000 & 0.00020\pm 0.00010 & 0.0004\pm 0.0001& 0.0015\pm 0.0009 \\
        \midrule
        FNO & 6400 & 0.0001\pm 0.0000 & 0.00009\pm 0.00000 & 0.0002\pm 0.0000& 0.0008\pm 0.0004 \\
        CFNO &  & 0.0001\pm 0.0000 & 0.00009\pm 0.00000 & 0.0002\pm 0.0000& 0.0007\pm 0.0004 \\
        \bottomrule
    \end{tabular}}\label{tab:app_Maxwell_large_hist2}
\end{table*}

%% file: tikz/smoke/bar_resnet_2e-4.tex
\begin{tikzpicture}

\definecolor{burlywood233196106}{RGB}{233,196,106}
\definecolor{darkgray176}{RGB}{146,146,146}
\definecolor{lightgray204}{RGB}{204,204,204}
\definecolor{lightseagreen42157143}{RGB}{42,157,143}
\definecolor{sandybrown24416297}{RGB}{244,162,97}

\begin{groupplot}[group style={group size=2 by 2}, height=0.24\textheight, width=0.52\textwidth, yticklabel style={color=darkgray176, font=\bfseries\scriptsize}, xticklabel style={color=darkgray176, font=\bfseries\scriptsize},]
\nextgroupplot[
axis line style={darkgray176},
legend cell align={left},
legend style={
  fill opacity=0.8,
  draw opacity=1,
  text=darkgray176,
  text opacity=1,
  at={(0.03,0.97)},
  anchor=north west,
  draw=lightgray204,
  font=\scriptsize
},
tick align=outside,
tick pos=left,
title={Rollout},
x grid style={darkgray176},
xmajorgrids,
xmin=-0.5, xmax=3.5,
xtick style={color=darkgray176},
xticklabels={},
y grid style={darkgray176},
ymajorgrids,
ymin=0, ymax=0.22,
ytick style={color=darkgray176}
]
\draw[draw=none,fill=lightseagreen42157143,fill opacity=0.85] (axis cs:-0.25,0) rectangle (axis cs:-0.0833333333333333,0.130299009382725);
\addlegendimage{ybar,ybar legend,draw=none,fill=lightseagreen42157143,fill opacity=0.85}
\addlegendentry{ResNet}

\draw[draw=none,fill=lightseagreen42157143,fill opacity=0.85] (axis cs:0.75,0) rectangle (axis cs:0.916666666666667,0.138020850718021);
\draw[draw=none,fill=lightseagreen42157143,fill opacity=0.85] (axis cs:1.75,0) rectangle (axis cs:1.91666666666667,0.132339775562286);
\draw[draw=none,fill=lightseagreen42157143,fill opacity=0.85] (axis cs:2.75,0) rectangle (axis cs:2.91666666666667,0.113620246450106);
\draw[draw=none,fill=burlywood233196106,fill opacity=0.85] (axis cs:-0.0833333333333333,0) rectangle (axis cs:0.0833333333333333,0.103724729269743);
\addlegendimage{ybar,ybar legend,draw=none,fill=burlywood233196106,fill opacity=0.85}
\addlegendentry{CResNet}

\draw[draw=none,fill=burlywood233196106,fill opacity=0.85] (axis cs:0.916666666666667,0) rectangle (axis cs:1.08333333333333,0.106710869818926);
\draw[draw=none,fill=burlywood233196106,fill opacity=0.85] (axis cs:1.91666666666667,0) rectangle (axis cs:2.08333333333333,0.0997475385665894);
\draw[draw=none,fill=burlywood233196106,fill opacity=0.85] (axis cs:2.91666666666667,0) rectangle (axis cs:3.08333333333333,0.158435072749853);
\draw[draw=none,fill=sandybrown24416297,fill opacity=0.85] (axis cs:0.0833333333333333,0) rectangle (axis cs:0.25,0.0979935452342033);
\addlegendimage{ybar,ybar legend,draw=none,fill=sandybrown24416297,fill opacity=0.85}
\addlegendentry{CResNet$_\text{rot}$}

\draw[draw=none,fill=sandybrown24416297,fill opacity=0.85] (axis cs:1.08333333333333,0) rectangle (axis cs:1.25,0.100053422152996);
\draw[draw=none,fill=sandybrown24416297,fill opacity=0.85] (axis cs:2.08333333333333,0) rectangle (axis cs:2.25,0.0942723527550697);
\draw[draw=none,fill=sandybrown24416297,fill opacity=0.85] (axis cs:3.08333333333333,0) rectangle (axis cs:3.25,0.106807375326753);
\path [draw=darkgray176, semithick]
(axis cs:-0.166666666666667,0.126509562134743)
--(axis cs:-0.166666666666667,0.134088456630707);

\path [draw=darkgray176, semithick]
(axis cs:0.833333333333333,0.135785982012749)
--(axis cs:0.833333333333333,0.140255719423294);

\path [draw=darkgray176, semithick]
(axis cs:1.83333333333333,0.132139414548874)
--(axis cs:1.83333333333333,0.132540136575699);

\path [draw=darkgray176, semithick]
(axis cs:2.83333333333333,0.113144599008285)
--(axis cs:2.83333333333333,0.114095893891927);

\addplot [semithick, darkgray176, mark=-, mark size=2.5, mark options={solid}, only marks]
table {%
-0.166666666666667 0.126509562134743
0.833333333333333 0.135785982012749
1.83333333333333 0.132139414548874
2.83333333333333 0.113144599008285
};

\addplot [semithick, darkgray176, mark=-, mark size=2.5, mark options={solid}, only marks]
table {%
-0.166666666666667 0.134088456630707
0.833333333333333 0.140255719423294
1.83333333333333 0.132540136575699
2.83333333333333 0.114095893891927
};

\path [draw=darkgray176, semithick]
(axis cs:0,0.101037546992302)
--(axis cs:0,0.106411911547184);

\path [draw=darkgray176, semithick]
(axis cs:1,0.104249946773052)
--(axis cs:1,0.109171792864799);

\path [draw=darkgray176, semithick]
(axis cs:2,0.099151685833931)
--(axis cs:2,0.100343391299248);

\path [draw=darkgray176, semithick]
(axis cs:3,0.131660307721713)
--(axis cs:3,0.185209837777993);

\addplot [semithick, darkgray176, mark=-, mark size=2.5, mark options={solid}, only marks]
table {%
0 0.101037546992302
1 0.104249946773052
2 0.099151685833931
3 0.131660307721713
};

\addplot [semithick, darkgray176, mark=-, mark size=2.5, mark options={solid}, only marks]
table {%
0 0.106411911547184
1 0.109171792864799
2 0.100343391299248
3 0.185209837777993
};

\path [draw=darkgray176, semithick]
(axis cs:0.166666666666667,0.0966045185923576)
--(axis cs:0.166666666666667,0.099382571876049);

\path [draw=darkgray176, semithick]
(axis cs:1.16666666666667,0.0977661535143852)
--(axis cs:1.16666666666667,0.102340690791607);

\path [draw=darkgray176, semithick]
(axis cs:2.16666666666667,0.0935615301132202)
--(axis cs:2.16666666666667,0.0949831753969193);

\path [draw=darkgray176, semithick]
(axis cs:3.16666666666667,0.102051166183622)
--(axis cs:3.16666666666667,0.111563584469883);

\addplot [semithick, darkgray176, mark=-, mark size=2.5, mark options={solid}, only marks]
table {%
0.166666666666667 0.0966045185923576
1.16666666666667 0.0977661535143852
2.16666666666667 0.0935615301132202
3.16666666666667 0.102051166183622
};

\addplot [semithick, darkgray176, mark=-, mark size=2.5, mark options={solid}, only marks]
table {%
0.166666666666667 0.099382571876049
1.16666666666667 0.102340690791607
2.16666666666667 0.0949831753969193
3.16666666666667 0.111563584469883
};

\nextgroupplot[
axis line style={darkgray176},
tick align=outside,
tick pos=left,
title={One-step},
x grid style={darkgray176},
xmajorgrids,
xmin=-0.5, xmax=3.5,
xtick style={color=darkgray176},
xticklabels={},
y grid style={darkgray176},
ymajorgrids,
ymin=0, ymax=0.0315228385071227,
ytick style={color=darkgray176}
]
\draw[draw=none,fill=lightseagreen42157143,fill opacity=0.85] (axis cs:-0.25,0) rectangle (axis cs:-0.0833333333333333,0.0171611784026027);
\draw[draw=none,fill=lightseagreen42157143,fill opacity=0.85] (axis cs:0.75,0) rectangle (axis cs:0.916666666666667,0.0165270520374179);
\draw[draw=none,fill=lightseagreen42157143,fill opacity=0.85] (axis cs:1.75,0) rectangle (axis cs:1.91666666666667,0.0176742104813457);
\draw[draw=none,fill=lightseagreen42157143,fill opacity=0.85] (axis cs:2.75,0) rectangle (axis cs:2.91666666666667,0.0155297961706916);
\draw[draw=none,fill=burlywood233196106,fill opacity=0.85] (axis cs:-0.0833333333333333,0) rectangle (axis cs:0.0833333333333333,0.0135723566636443);
\draw[draw=none,fill=burlywood233196106,fill opacity=0.85] (axis cs:0.916666666666667,0) rectangle (axis cs:1.08333333333333,0.0147318239323795);
\draw[draw=none,fill=burlywood233196106,fill opacity=0.85] (axis cs:1.91666666666667,0) rectangle (axis cs:2.08333333333333,0.0124978306703269);
\draw[draw=none,fill=burlywood233196106,fill opacity=0.85] (axis cs:2.91666666666667,0) rectangle (axis cs:3.08333333333333,0.0280586214115222);
\draw[draw=none,fill=sandybrown24416297,fill opacity=0.85] (axis cs:0.0833333333333333,0) rectangle (axis cs:0.25,0.00938191823661327);
\draw[draw=none,fill=sandybrown24416297,fill opacity=0.85] (axis cs:1.08333333333333,0) rectangle (axis cs:1.25,0.00917232036590576);
\draw[draw=none,fill=sandybrown24416297,fill opacity=0.85] (axis cs:2.08333333333333,0) rectangle (axis cs:2.25,0.0108740576542914);
\draw[draw=none,fill=sandybrown24416297,fill opacity=0.85] (axis cs:3.08333333333333,0) rectangle (axis cs:3.25,0.0177996177226305);
\path [draw=darkgray176, semithick]
(axis cs:-0.166666666666667,0.0162548813968897)
--(axis cs:-0.166666666666667,0.0180674754083157);

\path [draw=darkgray176, semithick]
(axis cs:0.833333333333333,0.0160455927252769)
--(axis cs:0.833333333333333,0.0170085113495588);

\path [draw=darkgray176, semithick]
(axis cs:1.83333333333333,0.0162966046482325)
--(axis cs:1.83333333333333,0.0190518163144588);

\path [draw=darkgray176, semithick]
(axis cs:2.83333333333333,0.0154201983547894)
--(axis cs:2.83333333333333,0.0156393939865938);

\addplot [semithick, darkgray176, mark=-, mark size=2.5, mark options={solid}, only marks]
table {%
-0.166666666666667 0.0162548813968897
0.833333333333333 0.0160455927252769
1.83333333333333 0.0162966046482325
2.83333333333333 0.0154201983547894
};
\addplot [semithick, darkgray176, mark=-, mark size=2.5, mark options={solid}, only marks]
table {%
-0.166666666666667 0.0180674754083157
0.833333333333333 0.0170085113495588
1.83333333333333 0.0190518163144588
2.83333333333333 0.0156393939865938
};
\path [draw=darkgray176, semithick]
(axis cs:0,0.012942329980433)
--(axis cs:0,0.0142023833468556);

\path [draw=darkgray176, semithick]
(axis cs:1,0.0145516116172075)
--(axis cs:1,0.0149120362475514);

\path [draw=darkgray176, semithick]
(axis cs:2,0.0122731691226363)
--(axis cs:2,0.0127224922180176);

\path [draw=darkgray176, semithick]
(axis cs:3,0.0245944043159218)
--(axis cs:3,0.0315228385071227);

\addplot [semithick, darkgray176, mark=-, mark size=2.5, mark options={solid}, only marks]
table {%
0 0.012942329980433
1 0.0145516116172075
2 0.0122731691226363
3 0.0245944043159218
};
\addplot [semithick, darkgray176, mark=-, mark size=2.5, mark options={solid}, only marks]
table {%
0 0.0142023833468556
1 0.0149120362475514
2 0.0127224922180176
3 0.0315228385071227
};
\path [draw=darkgray176, semithick]
(axis cs:0.166666666666667,0.00851487554609776)
--(axis cs:0.166666666666667,0.0102489609271288);

\path [draw=darkgray176, semithick]
(axis cs:1.16666666666667,0.00827652681618929)
--(axis cs:1.16666666666667,0.0100681139156222);

\path [draw=darkgray176, semithick]
(axis cs:2.16666666666667,0.0101350694894791)
--(axis cs:2.16666666666667,0.0116130458191037);

\path [draw=darkgray176, semithick]
(axis cs:3.16666666666667,0.0163710070609056)
--(axis cs:3.16666666666667,0.0192282283843554);

\addplot [semithick, darkgray176, mark=-, mark size=2.5, mark options={solid}, only marks]
table {%
0.166666666666667 0.00851487554609776
1.16666666666667 0.00827652681618929
2.16666666666667 0.0101350694894791
3.16666666666667 0.0163710070609056
};
\addplot [semithick, darkgray176, mark=-, mark size=2.5, mark options={solid}, only marks]
table {%
0.166666666666667 0.0102489609271288
1.16666666666667 0.0100681139156222
2.16666666666667 0.0116130458191037
3.16666666666667 0.0192282283843554
};

\nextgroupplot[
axis line style={darkgray176},
tick align=outside,
tick pos=left,
title={Scalar},
x grid style={darkgray176},
xmajorgrids,
xmin=-0.5, xmax=3.5,
xtick style={color=darkgray176},
xtick={0,1,2,3},
xticklabels={15600,10400,5200,2080},
y grid style={darkgray176},
ymajorgrids,
ymin=0, ymax=0.00628483705614781,
ytick style={color=darkgray176}
]
\draw[draw=none,fill=lightseagreen42157143,fill opacity=0.85] (axis cs:-0.25,0) rectangle (axis cs:-0.0833333333333333,0.00342281768098474);
\draw[draw=none,fill=lightseagreen42157143,fill opacity=0.85] (axis cs:0.75,0) rectangle (axis cs:0.916666666666667,0.00321320071816444);
\draw[draw=none,fill=lightseagreen42157143,fill opacity=0.85] (axis cs:1.75,0) rectangle (axis cs:1.91666666666667,0.00341478304471821);
\draw[draw=none,fill=lightseagreen42157143,fill opacity=0.85] (axis cs:2.75,0) rectangle (axis cs:2.91666666666667,0.00299611423785488);
\draw[draw=none,fill=burlywood233196106,fill opacity=0.85] (axis cs:-0.0833333333333333,0) rectangle (axis cs:0.0833333333333333,0.00285253173206002);
\draw[draw=none,fill=burlywood233196106,fill opacity=0.85] (axis cs:0.916666666666667,0) rectangle (axis cs:1.08333333333333,0.00315066485200077);
\draw[draw=none,fill=burlywood233196106,fill opacity=0.85] (axis cs:1.91666666666667,0) rectangle (axis cs:2.08333333333333,0.0026495506754145);
\draw[draw=none,fill=burlywood233196106,fill opacity=0.85] (axis cs:2.91666666666667,0) rectangle (axis cs:3.08333333333333,0.00566063479830821);
\draw[draw=none,fill=sandybrown24416297,fill opacity=0.85] (axis cs:0.0833333333333333,0) rectangle (axis cs:0.25,0.00204496778314933);
\draw[draw=none,fill=sandybrown24416297,fill opacity=0.85] (axis cs:1.08333333333333,0) rectangle (axis cs:1.25,0.00200622208649293);
\draw[draw=none,fill=sandybrown24416297,fill opacity=0.85] (axis cs:2.08333333333333,0) rectangle (axis cs:2.25,0.00233566120732576);
\draw[draw=none,fill=sandybrown24416297,fill opacity=0.85] (axis cs:3.08333333333333,0) rectangle (axis cs:3.25,0.00375594369446238);
\path [draw=darkgray176, semithick]
(axis cs:-0.166666666666667,0.00338851055130363)
--(axis cs:-0.166666666666667,0.00345712481066585);

\path [draw=darkgray176, semithick]
(axis cs:0.833333333333333,0.00317814922891557)
--(axis cs:0.833333333333333,0.00324825220741332);

\path [draw=darkgray176, semithick]
(axis cs:1.83333333333333,0.00313666672445834)
--(axis cs:1.83333333333333,0.00369289936497808);

\path [draw=darkgray176, semithick]
(axis cs:2.83333333333333,0.00296998394463667)
--(axis cs:2.83333333333333,0.00302224453107309);

\addplot [semithick, darkgray176, mark=-, mark size=2.5, mark options={solid}, only marks]
table {%
-0.166666666666667 0.00338851055130363
0.833333333333333 0.00317814922891557
1.83333333333333 0.00313666672445834
2.83333333333333 0.00296998394463667
};
\addplot [semithick, darkgray176, mark=-, mark size=2.5, mark options={solid}, only marks]
table {%
-0.166666666666667 0.00345712481066585
0.833333333333333 0.00324825220741332
1.83333333333333 0.00369289936497808
2.83333333333333 0.00302224453107309
};
\path [draw=darkgray176, semithick]
(axis cs:0,0.00266174576245248)
--(axis cs:0,0.00304331770166755);

\path [draw=darkgray176, semithick]
(axis cs:1,0.0030923169106245)
--(axis cs:1,0.00320901279337704);

\path [draw=darkgray176, semithick]
(axis cs:2,0.00260587618686259)
--(axis cs:2,0.00269322516396642);

\path [draw=darkgray176, semithick]
(axis cs:3,0.00503643254046861)
--(axis cs:3,0.00628483705614781);

\addplot [semithick, darkgray176, mark=-, mark size=2.5, mark options={solid}, only marks]
table {%
0 0.00266174576245248
1 0.0030923169106245
2 0.00260587618686259
3 0.00503643254046861
};
\addplot [semithick, darkgray176, mark=-, mark size=2.5, mark options={solid}, only marks]
table {%
0 0.00304331770166755
1 0.00320901279337704
2 0.00269322516396642
3 0.00628483705614781
};
\path [draw=darkgray176, semithick]
(axis cs:0.166666666666667,0.00190595758613199)
--(axis cs:0.166666666666667,0.00218397798016667);

\path [draw=darkgray176, semithick]
(axis cs:1.16666666666667,0.00180466764140874)
--(axis cs:1.16666666666667,0.00220777653157711);

\path [draw=darkgray176, semithick]
(axis cs:2.16666666666667,0.00219845585525036)
--(axis cs:2.16666666666667,0.00247286655940115);

\path [draw=darkgray176, semithick]
(axis cs:3.16666666666667,0.00347608549932591)
--(axis cs:3.16666666666667,0.00403580188959885);

\addplot [semithick, darkgray176, mark=-, mark size=2.5, mark options={solid}, only marks]
table {%
0.166666666666667 0.00190595758613199
1.16666666666667 0.00180466764140874
2.16666666666667 0.00219845585525036
3.16666666666667 0.00347608549932591
};
\addplot [semithick, darkgray176, mark=-, mark size=2.5, mark options={solid}, only marks]
table {%
0.166666666666667 0.00218397798016667
1.16666666666667 0.00220777653157711
2.16666666666667 0.00247286655940115
3.16666666666667 0.00403580188959885
};

\nextgroupplot[
axis line style={darkgray176},
tick align=outside,
tick pos=left,
title={Vector},
x grid style={darkgray176},
xmajorgrids,
xmin=-0.5, xmax=3.5,
xtick style={color=darkgray176},
xtick={0,1,2,3},
xticklabels={15600,10400,5200,2080},
y grid style={darkgray176},
ymajorgrids,
ymin=0, ymax=0.0253607963913037,
ytick style={color=darkgray176}
]
\draw[draw=none,fill=lightseagreen42157143,fill opacity=0.85] (axis cs:-0.25,0) rectangle (axis cs:-0.0833333333333333,0.0137891881167889);
\draw[draw=none,fill=lightseagreen42157143,fill opacity=0.85] (axis cs:0.75,0) rectangle (axis cs:0.916666666666667,0.0133660403080285);
\draw[draw=none,fill=lightseagreen42157143,fill opacity=0.85] (axis cs:1.75,0) rectangle (axis cs:1.91666666666667,0.0143121690489352);
\draw[draw=none,fill=lightseagreen42157143,fill opacity=0.85] (axis cs:2.75,0) rectangle (axis cs:2.91666666666667,0.0125500916813811);
\draw[draw=none,fill=burlywood233196106,fill opacity=0.85] (axis cs:-0.0833333333333333,0) rectangle (axis cs:0.0833333333333333,0.0107616372406483);
\draw[draw=none,fill=burlywood233196106,fill opacity=0.85] (axis cs:0.916666666666667,0) rectangle (axis cs:1.08333333333333,0.0116225210949779);
\draw[draw=none,fill=burlywood233196106,fill opacity=0.85] (axis cs:1.91666666666667,0) rectangle (axis cs:2.08333333333333,0.00988295860588551);
\draw[draw=none,fill=burlywood233196106,fill opacity=0.85] (axis cs:2.91666666666667,0) rectangle (axis cs:3.08333333333333,0.0225241749236981);
\draw[draw=none,fill=sandybrown24416297,fill opacity=0.85] (axis cs:0.0833333333333333,0) rectangle (axis cs:0.25,0.00735877081751823);
\draw[draw=none,fill=sandybrown24416297,fill opacity=0.85] (axis cs:1.08333333333333,0) rectangle (axis cs:1.25,0.00719424546696246);
\draw[draw=none,fill=sandybrown24416297,fill opacity=0.85] (axis cs:2.08333333333333,0) rectangle (axis cs:2.25,0.0085710296407342);
\draw[draw=none,fill=sandybrown24416297,fill opacity=0.85] (axis cs:3.08333333333333,0) rectangle (axis cs:3.25,0.0141349481418729);
\path [draw=darkgray176, semithick]
(axis cs:-0.166666666666667,0.0129963112995028)
--(axis cs:-0.166666666666667,0.0145820649340749);

\path [draw=darkgray176, semithick]
(axis cs:0.833333333333333,0.0129242315888405)
--(axis cs:0.833333333333333,0.0138078490272164);

\path [draw=darkgray176, semithick]
(axis cs:1.83333333333333,0.0132906939834356)
--(axis cs:1.83333333333333,0.0153336441144347);

\path [draw=darkgray176, semithick]
(axis cs:2.83333333333333,0.0124663129965943)
--(axis cs:2.83333333333333,0.0126338703661678);

\addplot [semithick, darkgray176, mark=-, mark size=2.5, mark options={solid}, only marks]
table {%
-0.166666666666667 0.0129963112995028
0.833333333333333 0.0129242315888405
1.83333333333333 0.0132906939834356
2.83333333333333 0.0124663129965943
};
\addplot [semithick, darkgray176, mark=-, mark size=2.5, mark options={solid}, only marks]
table {%
-0.166666666666667 0.0145820649340749
0.833333333333333 0.0138078490272164
1.83333333333333 0.0153336441144347
2.83333333333333 0.0126338703661678
};
\path [draw=darkgray176, semithick]
(axis cs:0,0.0102559085935354)
--(axis cs:0,0.0112673658877611);

\path [draw=darkgray176, semithick]
(axis cs:1,0.0114304386079311)
--(axis cs:1,0.0118146035820246);

\path [draw=darkgray176, semithick]
(axis cs:2,0.00964150577783585)
--(axis cs:2,0.0101244114339352);

\path [draw=darkgray176, semithick]
(axis cs:3,0.0196875534560925)
--(axis cs:3,0.0253607963913037);

\addplot [semithick, darkgray176, mark=-, mark size=2.5, mark options={solid}, only marks]
table {%
0 0.0102559085935354
1 0.0114304386079311
2 0.00964150577783585
3 0.0196875534560925
};
\addplot [semithick, darkgray176, mark=-, mark size=2.5, mark options={solid}, only marks]
table {%
0 0.0112673658877611
1 0.0118146035820246
2 0.0101244114339352
3 0.0253607963913037
};
\path [draw=darkgray176, semithick]
(axis cs:0.166666666666667,0.00667345151305199)
--(axis cs:0.166666666666667,0.00804409012198448);

\path [draw=darkgray176, semithick]
(axis cs:1.16666666666667,0.00645315228030086)
--(axis cs:1.16666666666667,0.00793533865362406);

\path [draw=darkgray176, semithick]
(axis cs:2.16666666666667,0.00791426002979279)
--(axis cs:2.16666666666667,0.00922779925167561);

\path [draw=darkgray176, semithick]
(axis cs:3.16666666666667,0.0129745258908798)
--(axis cs:3.16666666666667,0.0152953703928659);

\addplot [semithick, darkgray176, mark=-, mark size=2.5, mark options={solid}, only marks]
table {%
0.166666666666667 0.00667345151305199
1.16666666666667 0.00645315228030086
2.16666666666667 0.00791426002979279
3.16666666666667 0.0129745258908798
};
\addplot [semithick, darkgray176, mark=-, mark size=2.5, mark options={solid}, only marks]
table {%
0.166666666666667 0.00804409012198448
1.16666666666667 0.00793533865362406
2.16666666666667 0.00922779925167561
3.16666666666667 0.0152953703928659
};
\end{groupplot}

\draw ({$(current bounding box.south west)!0.02!(current bounding box.south east)$}|-{$(current bounding box.south west)!0.5!(current bounding box.north west)$}) node[
  scale=0.9,
  anchor=west,
  text=darkgray176,
  rotate=90.0
]{MSE};
\draw ({$(current bounding box.south west)!0.5!(current bounding box.south east)$}|-{$(current bounding box.south west)!-0.05!(current bounding box.north west)$}) node[
  scale=0.9,
  anchor=south,
  text=darkgray176,
  rotate=0.0
]{Num. Train Trajectories};
\end{tikzpicture}

%% file: tikz/smoke/bar_fourier_2e-4.tex
\begin{tikzpicture}

\definecolor{darkgray176}{RGB}{146,146,146}
\definecolor{darkslategray387083}{RGB}{38,70,83}
\definecolor{lightgray204}{RGB}{204,204,204}
\definecolor{tomato23111181}{RGB}{231,111,81}

\begin{groupplot}[group style={group size=2 by 2}, height=0.2\textheight, width=0.52\textwidth, yticklabel style={color=darkgray176, font=\bfseries\scriptsize}, xticklabel style={color=darkgray176, font=\bfseries\scriptsize},]
\nextgroupplot[
axis line style={darkgray176},
legend cell align={left},
legend style={
  fill opacity=0.8,
  draw opacity=1,
  text=darkgray176,
  text opacity=1,
  at={(0.03,0.97)},
  anchor=north west,
  draw=lightgray204,
  font=\scriptsize,
},
tick align=outside,
tick pos=left,
title={Rollout},
x grid style={darkgray176},
xmajorgrids,
xmin=-0.5, xmax=2.5,
xtick style={color=darkgray176},
xticklabels={},
y grid style={darkgray176},
ymajorgrids,
ymin=0, ymax=0.045806930862091,
ytick style={color=darkgray176}
]
\draw[draw=none,fill=darkslategray387083,fill opacity=0.85] (axis cs:-0.25,0) rectangle (axis cs:0,0.0200465209782124);
\addlegendimage{ybar,ybar legend,draw=none,fill=darkslategray387083,fill opacity=0.85}
\addlegendentry{FNO}

\draw[draw=none,fill=darkslategray387083,fill opacity=0.85] (axis cs:0.75,0) rectangle (axis cs:1,0.0268410226951043);
\draw[draw=none,fill=darkslategray387083,fill opacity=0.85] (axis cs:1.75,0) rectangle (axis cs:2,0.0428087487816811);
\draw[draw=none,fill=tomato23111181,fill opacity=0.85] (axis cs:0,0) rectangle (axis cs:0.25,0.018858910848697);
\addlegendimage{ybar,ybar legend,draw=none,fill=tomato23111181,fill opacity=0.85}
\addlegendentry{CFNO}

\draw[draw=none,fill=tomato23111181,fill opacity=0.85] (axis cs:1,0) rectangle (axis cs:1.25,0.0243015394856532);
\draw[draw=none,fill=tomato23111181,fill opacity=0.85] (axis cs:2,0) rectangle (axis cs:2.25,0.0346080536643664);
\path [draw=darkgray176, semithick]
(axis cs:-0.125,0.0196278563494451)
--(axis cs:-0.125,0.0204651856069796);

\path [draw=darkgray176, semithick]
(axis cs:0.875,0.0261708690555498)
--(axis cs:0.875,0.0275111763346587);

\path [draw=darkgray176, semithick]
(axis cs:1.875,0.0398105667012712)
--(axis cs:1.875,0.045806930862091);

\addplot [semithick, darkgray176, mark=-, mark size=2.5, mark options={solid}, only marks]
table {%
-0.125 0.0196278563494451
0.875 0.0261708690555498
1.875 0.0398105667012712
};

\addplot [semithick, darkgray176, mark=-, mark size=2.5, mark options={solid}, only marks]
table {%
-0.125 0.0204651856069796
0.875 0.0275111763346587
1.875 0.045806930862091
};

\path [draw=darkgray176, semithick]
(axis cs:0.125,0.0188005149575856)
--(axis cs:0.125,0.0189173067398085);

\path [draw=darkgray176, semithick]
(axis cs:1.125,0.0241859783887698)
--(axis cs:1.125,0.0244171005825367);

\path [draw=darkgray176, semithick]
(axis cs:2.125,0.0342985319480809)
--(axis cs:2.125,0.0349175753806519);

\addplot [semithick, darkgray176, mark=-, mark size=2.5, mark options={solid}, only marks]
table {%
0.125 0.0188005149575856
1.125 0.0241859783887698
2.125 0.0342985319480809
};

\addplot [semithick, darkgray176, mark=-, mark size=2.5, mark options={solid}, only marks]
table {%
0.125 0.0189173067398085
1.125 0.0244171005825367
2.125 0.0349175753806519
};

\nextgroupplot[
axis line style={darkgray176},
tick align=outside,
tick pos=left,
title={One-step},
x grid style={darkgray176},
xmajorgrids,
xmin=-0.5, xmax=2.5,
xtick style={color=darkgray176},
xticklabels={},
y grid style={darkgray176},
ymajorgrids,
ymin=0, ymax=0.00995087522806456,
ytick style={color=darkgray176}
]
\draw[draw=none,fill=darkslategray387083,fill opacity=0.85] (axis cs:-0.25,0) rectangle (axis cs:0,0.00375463627278805);
\draw[draw=none,fill=darkslategray387083,fill opacity=0.85] (axis cs:0.75,0) rectangle (axis cs:1,0.00536326132714748);
\draw[draw=none,fill=darkslategray387083,fill opacity=0.85] (axis cs:1.75,0) rectangle (axis cs:2,0.00930879730731249);
\draw[draw=none,fill=tomato23111181,fill opacity=0.85] (axis cs:0,0) rectangle (axis cs:0.25,0.00352552028683325);
\draw[draw=none,fill=tomato23111181,fill opacity=0.85] (axis cs:1,0) rectangle (axis cs:1.25,0.00481796295692523);
\draw[draw=none,fill=tomato23111181,fill opacity=0.85] (axis cs:2,0) rectangle (axis cs:2.25,0.00748796761035919);
\path [draw=darkgray176, semithick]
(axis cs:-0.125,0.00365053952712522)
--(axis cs:-0.125,0.00385873301845087);

\path [draw=darkgray176, semithick]
(axis cs:0.875,0.00521129983448025)
--(axis cs:0.875,0.00551522281981471);

\path [draw=darkgray176, semithick]
(axis cs:1.875,0.00866671938656042)
--(axis cs:1.875,0.00995087522806456);

\addplot [semithick, darkgray176, mark=-, mark size=2.5, mark options={solid}, only marks]
table {%
-0.125 0.00365053952712522
0.875 0.00521129983448025
1.875 0.00866671938656042
};
\addplot [semithick, darkgray176, mark=-, mark size=2.5, mark options={solid}, only marks]
table {%
-0.125 0.00385873301845087
0.875 0.00551522281981471
1.875 0.00995087522806456
};
\path [draw=darkgray176, semithick]
(axis cs:0.125,0.00350685575591489)
--(axis cs:0.125,0.0035441848177516);

\path [draw=darkgray176, semithick]
(axis cs:1.125,0.00478344830091218)
--(axis cs:1.125,0.00485247761293829);

\path [draw=darkgray176, semithick]
(axis cs:2.125,0.0074067950706876)
--(axis cs:2.125,0.00756914015003078);

\addplot [semithick, darkgray176, mark=-, mark size=2.5, mark options={solid}, only marks]
table {%
0.125 0.00350685575591489
1.125 0.00478344830091218
2.125 0.0074067950706876
};
\addplot [semithick, darkgray176, mark=-, mark size=2.5, mark options={solid}, only marks]
table {%
0.125 0.0035441848177516
1.125 0.00485247761293829
2.125 0.00756914015003078
};

\nextgroupplot[
axis line style={darkgray176},
tick align=outside,
tick pos=left,
title={Scalar},
x grid style={darkgray176},
xmajorgrids,
xmin=-0.5, xmax=2.5,
xtick style={color=darkgray176},
xtick={0,1,2},
xticklabels={10400,5200,2080},
y grid style={darkgray176},
ymajorgrids,
ymin=0, ymax=0.00338618714533136,
ytick style={color=darkgray176}
]
\draw[draw=none,fill=darkslategray387083,fill opacity=0.85] (axis cs:-0.25,0) rectangle (axis cs:0,0.00155954090102265);
\draw[draw=none,fill=darkslategray387083,fill opacity=0.85] (axis cs:0.75,0) rectangle (axis cs:1,0.0020413954431812);
\draw[draw=none,fill=darkslategray387083,fill opacity=0.85] (axis cs:1.75,0) rectangle (axis cs:2,0.00317945354618132);
\draw[draw=none,fill=tomato23111181,fill opacity=0.85] (axis cs:0,0) rectangle (axis cs:0.25,0.00148463617855062);
\draw[draw=none,fill=tomato23111181,fill opacity=0.85] (axis cs:1,0) rectangle (axis cs:1.25,0.00189025641884655);
\draw[draw=none,fill=tomato23111181,fill opacity=0.85] (axis cs:2,0) rectangle (axis cs:2.25,0.00265722582116723);
\path [draw=darkgray176, semithick]
(axis cs:-0.125,0.00152613954725554)
--(axis cs:-0.125,0.00159294225478977);

\path [draw=darkgray176, semithick]
(axis cs:0.875,0.00199893915789489)
--(axis cs:0.875,0.0020838517284675);

\path [draw=darkgray176, semithick]
(axis cs:1.875,0.00297271994703128)
--(axis cs:1.875,0.00338618714533136);

\addplot [semithick, darkgray176, mark=-, mark size=2.5, mark options={solid}, only marks]
table {%
-0.125 0.00152613954725554
0.875 0.00199893915789489
1.875 0.00297271994703128
};
\addplot [semithick, darkgray176, mark=-, mark size=2.5, mark options={solid}, only marks]
table {%
-0.125 0.00159294225478977
0.875 0.0020838517284675
1.875 0.00338618714533136
};
\path [draw=darkgray176, semithick]
(axis cs:0.125,0.00147615638481039)
--(axis cs:0.125,0.00149311597229085);

\path [draw=darkgray176, semithick]
(axis cs:1.125,0.0018777824436274)
--(axis cs:1.125,0.0019027303940657);

\path [draw=darkgray176, semithick]
(axis cs:2.125,0.00263332275096929)
--(axis cs:2.125,0.00268112889136517);

\addplot [semithick, darkgray176, mark=-, mark size=2.5, mark options={solid}, only marks]
table {%
0.125 0.00147615638481039
1.125 0.0018777824436274
2.125 0.00263332275096929
};
\addplot [semithick, darkgray176, mark=-, mark size=2.5, mark options={solid}, only marks]
table {%
0.125 0.00149311597229085
1.125 0.0019027303940657
2.125 0.00268112889136517
};

\nextgroupplot[
axis line style={darkgray176},
tick align=outside,
tick pos=left,
title={Vector},
x grid style={darkgray176},
xmajorgrids,
xmin=-0.5, xmax=2.5,
xtick style={color=darkgray176},
xtick={0,1,2},
xticklabels={10400,5200,2080},
y grid style={darkgray176},
ymajorgrids,
ymin=0, ymax=0.00656785585242912,
ytick style={color=darkgray176}
]
\draw[draw=none,fill=darkslategray387083,fill opacity=0.85] (axis cs:-0.25,0) rectangle (axis cs:0,0.00219666243841251);
\draw[draw=none,fill=darkslategray387083,fill opacity=0.85] (axis cs:0.75,0) rectangle (axis cs:1,0.0033237913933893);
\draw[draw=none,fill=darkslategray387083,fill opacity=0.85] (axis cs:1.75,0) rectangle (axis cs:2,0.0061319552672406);
\draw[draw=none,fill=tomato23111181,fill opacity=0.85] (axis cs:0,0) rectangle (axis cs:0.25,0.0020461850023518);
\draw[draw=none,fill=tomato23111181,fill opacity=0.85] (axis cs:1,0) rectangle (axis cs:1.25,0.00293456367217004);
\draw[draw=none,fill=tomato23111181,fill opacity=0.85] (axis cs:2,0) rectangle (axis cs:2.25,0.00484006572514772);
\path [draw=darkgray176, semithick]
(axis cs:-0.125,0.00212595746625731)
--(axis cs:-0.125,0.0022673674105677);

\path [draw=darkgray176, semithick]
(axis cs:0.875,0.00321436273481674)
--(axis cs:0.875,0.00343322005196187);

\path [draw=darkgray176, semithick]
(axis cs:1.875,0.00569605468205209)
--(axis cs:1.875,0.00656785585242912);

\addplot [semithick, darkgray176, mark=-, mark size=2.5, mark options={solid}, only marks]
table {%
-0.125 0.00212595746625731
0.875 0.00321436273481674
1.875 0.00569605468205209
};
\addplot [semithick, darkgray176, mark=-, mark size=2.5, mark options={solid}, only marks]
table {%
-0.125 0.0022673674105677
0.875 0.00343322005196187
1.875 0.00656785585242912
};
\path [draw=darkgray176, semithick]
(axis cs:0.125,0.0020353653148594)
--(axis cs:0.125,0.0020570046898442);

\path [draw=darkgray176, semithick]
(axis cs:1.125,0.00291187404124132)
--(axis cs:1.125,0.00295725330309877);

\path [draw=darkgray176, semithick]
(axis cs:2.125,0.00478187535702887)
--(axis cs:2.125,0.00489825609326658);

\addplot [semithick, darkgray176, mark=-, mark size=2.5, mark options={solid}, only marks]
table {%
0.125 0.0020353653148594
1.125 0.00291187404124132
2.125 0.00478187535702887
};
\addplot [semithick, darkgray176, mark=-, mark size=2.5, mark options={solid}, only marks]
table {%
0.125 0.0020570046898442
1.125 0.00295725330309877
2.125 0.00489825609326658
};
\end{groupplot}

\draw ({$(current bounding box.south west)!-0.01!(current bounding box.south east)$}|-{$(current bounding box.south west)!0.5!(current bounding box.north west)$}) node[
  scale=0.9,
  anchor=west,
  text=darkgray176,
  rotate=90.0
]{MSE};
\draw ({$(current bounding box.south west)!0.5!(current bounding box.south east)$}|-{$(current bounding box.south west)!-0.052!(current bounding box.north west)$}) node[
  scale=0.9,
  anchor=south,
  text=darkgray176,
  rotate=0.0
]{Num. Train Trajectories};
\end{tikzpicture}

%% file: tikz/weather/bar_fourier_2e-4_hist=2.tex
\begin{tikzpicture}

\definecolor{darkgray176}{RGB}{146,146,146}
\definecolor{darkslategray387083}{RGB}{38,70,83}
\definecolor{lightgray204}{RGB}{204,204,204}
\definecolor{tomato23111181}{RGB}{231,111,81}

\begin{groupplot}[group style={group size=2 by 2},height=0.24\textheight, width=0.52\textwidth, yticklabel style={color=darkgray176, font=\bfseries\scriptsize}, xticklabel style={color=darkgray176, font=\bfseries\scriptsize},]
\nextgroupplot[
axis line style={darkgray176},
legend cell align={left},
legend style={
  fill opacity=0.8,
  draw opacity=1,
  text=darkgray176,  
  text opacity=1,
  at={(0.03,0.97)},
  anchor=north west,
  draw=lightgray204,
  font=\scriptsize,
},
tick align=outside,
tick pos=left,
title={Rollout},
x grid style={darkgray176},
xmajorgrids,
xmin=-0.5, xmax=3.5,
xtick style={color=darkgray176},
xticklabels={},
y grid style={darkgray176},
ymajorgrids,
ymin=0, ymax=0.965912721905391,
ytick style={color=darkgray176}
]
\draw[draw=none,fill=darkslategray387083,fill opacity=0.85] (axis cs:-0.25,0) rectangle (axis cs:0,0.0404472816735506);
\addlegendimage{ybar,ybar legend,draw=none,fill=darkslategray387083,fill opacity=0.85}
\addlegendentry{FNO}

\draw[draw=none,fill=darkslategray387083,fill opacity=0.85] (axis cs:0.75,0) rectangle (axis cs:1,0.0651295781135559);
\draw[draw=none,fill=darkslategray387083,fill opacity=0.85] (axis cs:1.75,0) rectangle (axis cs:2,0.144377276301384);
\draw[draw=none,fill=darkslategray387083,fill opacity=0.85] (axis cs:2.75,0) rectangle (axis cs:3,0.80318542321523);
\draw[draw=none,fill=tomato23111181,fill opacity=0.85] (axis cs:0,0) rectangle (axis cs:0.25,0.0314969004442294);
\addlegendimage{ybar,ybar legend,draw=none,fill=tomato23111181,fill opacity=0.85}
\addlegendentry{CFNO}

\draw[draw=none,fill=tomato23111181,fill opacity=0.85] (axis cs:1,0) rectangle (axis cs:1.25,0.0454886717100938);
\draw[draw=none,fill=tomato23111181,fill opacity=0.85] (axis cs:2,0) rectangle (axis cs:2.25,0.0941112736860911);
\draw[draw=none,fill=tomato23111181,fill opacity=0.85] (axis cs:3,0) rectangle (axis cs:3.25,0.432298913598061);
\path [draw=darkgray176, semithick]
(axis cs:-0.125,0.0399861969053745)
--(axis cs:-0.125,0.0409083664417267);

\path [draw=darkgray176, semithick]
(axis cs:0.875,0.0637602210044861)
--(axis cs:0.875,0.0664989352226257);

\path [draw=darkgray176, semithick]
(axis cs:1.875,0.141744792461395)
--(axis cs:1.875,0.147009760141373);

\path [draw=darkgray176, semithick]
(axis cs:2.875,0.79883819854832)
--(axis cs:2.875,0.807532647882141);

\addplot [semithick, darkgray176, mark=-, mark size=2.5, mark options={solid}, only marks]
table {%
-0.125 0.0399861969053745
0.875 0.0637602210044861
1.875 0.141744792461395
2.875 0.79883819854832
};

\addplot [semithick, darkgray176, mark=-, mark size=2.5, mark options={solid}, only marks]
table {%
-0.125 0.0409083664417267
0.875 0.0664989352226257
1.875 0.147009760141373
2.875 0.807532647882141
};

\path [draw=darkgray176, semithick]
(axis cs:0.125,0.0310836300189506)
--(axis cs:0.125,0.0319101708695083);

\path [draw=darkgray176, semithick]
(axis cs:1.125,0.0446319874578019)
--(axis cs:1.125,0.0463453559623857);

\path [draw=darkgray176, semithick]
(axis cs:2.125,0.0919993273599173)
--(axis cs:2.125,0.096223220012265);

\path [draw=darkgray176, semithick]
(axis cs:3.125,0.42769256234169)
--(axis cs:3.125,0.436905264854431);

\addplot [semithick, darkgray176, mark=-, mark size=2.5, mark options={solid}, only marks]
table {%
0.125 0.0310836300189506
1.125 0.0446319874578019
2.125 0.0919993273599173
3.125 0.42769256234169
};

\addplot [semithick, darkgray176, mark=-, mark size=2.5, mark options={solid}, only marks]
table {%
0.125 0.0319101708695083
1.125 0.0463453559623857
2.125 0.096223220012265
3.125 0.436905264854431
};

\nextgroupplot[
axis line style={darkgray176},
tick align=outside,
tick pos=left,
title={One-step},
x grid style={darkgray176},
xmajorgrids,
xmin=-0.5, xmax=3.5,
xtick style={color=darkgray176},
xticklabels={},
y grid style={darkgray176},
ymajorgrids,
ymin=0, ymax=0.0729974239669811,
ytick style={color=darkgray176}
]

\draw[draw=none,fill=darkslategray387083,fill opacity=0.85] (axis cs:-0.25,0) rectangle (axis cs:0,0.0019612645264715);
\draw[draw=none,fill=darkslategray387083,fill opacity=0.85] (axis cs:0.75,0) rectangle (axis cs:1,0.00335354392882437);
\draw[draw=none,fill=darkslategray387083,fill opacity=0.85] (axis cs:1.75,0) rectangle (axis cs:2,0.00773894088342786);
\draw[draw=none,fill=darkslategray387083,fill opacity=0.85] (axis cs:2.75,0) rectangle (axis cs:3,0.0615671922763189);
\draw[draw=none,fill=tomato23111181,fill opacity=0.85] (axis cs:0,0) rectangle (axis cs:0.25,0.00169398433839281);
\draw[draw=none,fill=tomato23111181,fill opacity=0.85] (axis cs:1,0) rectangle (axis cs:1.25,0.00256044402097662);
\draw[draw=none,fill=tomato23111181,fill opacity=0.85] (axis cs:2,0) rectangle (axis cs:2.25,0.00526664933810631);
\draw[draw=none,fill=tomato23111181,fill opacity=0.85] (axis cs:3,0) rectangle (axis cs:3.25,0.0249517727643251);
\path [draw=darkgray176, semithick]
(axis cs:-0.125,0.00193743291310966)
--(axis cs:-0.125,0.00198509613983333);

\path [draw=darkgray176, semithick]
(axis cs:0.875,0.00322146224789321)
--(axis cs:0.875,0.00348562560975552);

\path [draw=darkgray176, semithick]
(axis cs:1.875,0.00744888000190258)
--(axis cs:1.875,0.00802900176495314);

\path [draw=darkgray176, semithick]
(axis cs:2.875,0.0593614324910978)
--(axis cs:2.875,0.0637729520615399);

\addplot [semithick, darkgray176, mark=-, mark size=2.5, mark options={solid}, only marks]
table {%
-0.125 0.00193743291310966
0.875 0.00322146224789321
1.875 0.00744888000190258
2.875 0.0593614324910978
};
\addplot [semithick, darkgray176, mark=-, mark size=2.5, mark options={solid}, only marks]
table {%
-0.125 0.00198509613983333
0.875 0.00348562560975552
1.875 0.00802900176495314
2.875 0.0637729520615399
};
\path [draw=darkgray176, semithick]
(axis cs:0.125,0.0016719076421672)
--(axis cs:0.125,0.00171606103461843);

\path [draw=darkgray176, semithick]
(axis cs:1.125,0.00252204231508529)
--(axis cs:1.125,0.00259884572686795);

\path [draw=darkgray176, semithick]
(axis cs:2.125,0.00517335413449837)
--(axis cs:2.125,0.00535994454171426);

\path [draw=darkgray176, semithick]
(axis cs:3.125,0.0242259856313467)
--(axis cs:3.125,0.0256775598973036);

\addplot [semithick, darkgray176, mark=-, mark size=2.5, mark options={solid}, only marks]
table {%
0.125 0.0016719076421672
1.125 0.00252204231508529
2.125 0.00517335413449837
3.125 0.0242259856313467
};
\addplot [semithick, darkgray176, mark=-, mark size=2.5, mark options={solid}, only marks]
table {%
0.125 0.00171606103461843
1.125 0.00259884572686795
2.125 0.00535994454171426
3.125 0.0256775598973036
};

\nextgroupplot[
axis line style={darkgray176},
tick align=outside,
tick pos=left,
title={Scalar},
x grid style={darkgray176},
xmajorgrids,
xmin=-0.5, xmax=3.5,
xtick style={color=darkgray176},
xtick={0,1,2,3},
xticklabels={896,448,192,56},
y grid style={darkgray176},
ymajorgrids,
ymin=0, ymax=0.033241735005843,
ytick style={color=darkgray176}
]

\draw[draw=none,fill=darkslategray387083,fill opacity=0.85] (axis cs:-0.25,0) rectangle (axis cs:0,0.000380174606107175);
\draw[draw=none,fill=darkslategray387083,fill opacity=0.85] (axis cs:0.75,0) rectangle (axis cs:1,0.000735001929569989);
\draw[draw=none,fill=darkslategray387083,fill opacity=0.85] (axis cs:1.75,0) rectangle (axis cs:2,0.00207265466451645);
\draw[draw=none,fill=darkslategray387083,fill opacity=0.85] (axis cs:2.75,0) rectangle (axis cs:3,0.0270830808828274);
\draw[draw=none,fill=tomato23111181,fill opacity=0.85] (axis cs:0,0) rectangle (axis cs:0.25,0.000339371268637478);
\draw[draw=none,fill=tomato23111181,fill opacity=0.85] (axis cs:1,0) rectangle (axis cs:1.25,0.000537565967533737);
\draw[draw=none,fill=tomato23111181,fill opacity=0.85] (axis cs:2,0) rectangle (axis cs:2.25,0.00119873388515164);
\draw[draw=none,fill=tomato23111181,fill opacity=0.85] (axis cs:3,0) rectangle (axis cs:3.25,0.00713966740295291);
\path [draw=darkgray176, semithick]
(axis cs:-0.125,0.000373343005776405)
--(axis cs:-0.125,0.000387006206437945);

\path [draw=darkgray176, semithick]
(axis cs:0.875,0.000651557231321931)
--(axis cs:0.875,0.000818446627818048);

\path [draw=darkgray176, semithick]
(axis cs:1.875,0.00191920017823577)
--(axis cs:1.875,0.00222610915079713);

\path [draw=darkgray176, semithick]
(axis cs:2.875,0.0254767664339195)
--(axis cs:2.875,0.0286893953317354);

\addplot [semithick, darkgray176, mark=-, mark size=2.5, mark options={solid}, only marks]
table {%
-0.125 0.000373343005776405
0.875 0.000651557231321931
1.875 0.00191920017823577
2.875 0.0254767664339195
};
\addplot [semithick, darkgray176, mark=-, mark size=2.5, mark options={solid}, only marks]
table {%
-0.125 0.000387006206437945
0.875 0.000818446627818048
1.875 0.00222610915079713
2.875 0.0286893953317354
};
\path [draw=darkgray176, semithick]
(axis cs:0.125,0.000337295133042398)
--(axis cs:0.125,0.000341447404232559);

\path [draw=darkgray176, semithick]
(axis cs:1.125,0.000535040853532917)
--(axis cs:1.125,0.000540091081534558);

\path [draw=darkgray176, semithick]
(axis cs:2.125,0.00118679065597426)
--(axis cs:2.125,0.00121067711432903);

\path [draw=darkgray176, semithick]
(axis cs:3.125,0.00684307143092155)
--(axis cs:3.125,0.00743626337498426);

\addplot [semithick, darkgray176, mark=-, mark size=2.5, mark options={solid}, only marks]
table {%
0.125 0.000337295133042398
1.125 0.000535040853532917
2.125 0.00118679065597426
3.125 0.00684307143092155
};
\addplot [semithick, darkgray176, mark=-, mark size=2.5, mark options={solid}, only marks]
table {%
0.125 0.000341447404232559
1.125 0.000540091081534558
2.125 0.00121067711432903
3.125 0.00743626337498426
};

\nextgroupplot[
axis line style={darkgray176},
tick align=outside,
tick pos=left,
title={Vector},
x grid style={darkgray176},
xmajorgrids,
xmin=-0.5, xmax=3.5,
xtick style={color=darkgray176},
xtick={0,1,2,3},
xticklabels={896,448,192,56},
y grid style={darkgray176},
ymajorgrids,
ymin=0, ymax=0.0398588895525499,
ytick style={color=darkgray176}
]
\draw[draw=none,fill=darkslategray387083,fill opacity=0.85] (axis cs:-0.25,0) rectangle (axis cs:0,0.0015783368726261);
\draw[draw=none,fill=darkslategray387083,fill opacity=0.85] (axis cs:0.75,0) rectangle (axis cs:1,0.00261437462177128);
\draw[draw=none,fill=darkslategray387083,fill opacity=0.85] (axis cs:1.75,0) rectangle (axis cs:2,0.00567777291871607);
\draw[draw=none,fill=darkslategray387083,fill opacity=0.85] (axis cs:2.75,0) rectangle (axis cs:3,0.0345232759912809);
\draw[draw=none,fill=tomato23111181,fill opacity=0.85] (axis cs:0,0) rectangle (axis cs:0.25,0.00133951132496198);
\draw[draw=none,fill=tomato23111181,fill opacity=0.85] (axis cs:1,0) rectangle (axis cs:1.25,0.0020014406957974);
\draw[draw=none,fill=tomato23111181,fill opacity=0.85] (axis cs:2,0) rectangle (axis cs:2.25,0.00404283155997594);
\draw[draw=none,fill=tomato23111181,fill opacity=0.85] (axis cs:3,0) rectangle (axis cs:3.25,0.0177130829542875);
\path [draw=darkgray176, semithick]
(axis cs:-0.125,0.00156129465904087)
--(axis cs:-0.125,0.00159537908621132);

\path [draw=darkgray176, semithick]
(axis cs:0.875,0.00256556365638971)
--(axis cs:0.875,0.00266318558715284);

\path [draw=darkgray176, semithick]
(axis cs:1.875,0.00554017908871174)
--(axis cs:1.875,0.00581536674872041);

\path [draw=darkgray176, semithick]
(axis cs:2.875,0.0338292774561872)
--(axis cs:2.875,0.0352172745263745);

\addplot [semithick, darkgray176, mark=-, mark size=2.5, mark options={solid}, only marks]
table {%
-0.125 0.00156129465904087
0.875 0.00256556365638971
1.875 0.00554017908871174
2.875 0.0338292774561872
};
\addplot [semithick, darkgray176, mark=-, mark size=2.5, mark options={solid}, only marks]
table {%
-0.125 0.00159537908621132
0.875 0.00266318558715284
1.875 0.00581536674872041
2.875 0.0352172745263745
};
\path [draw=darkgray176, semithick]
(axis cs:0.125,0.00131961396784156)
--(axis cs:0.125,0.0013594086820824);

\path [draw=darkgray176, semithick]
(axis cs:1.125,0.00196566665873776)
--(axis cs:1.125,0.00203721473285704);

\path [draw=darkgray176, semithick]
(axis cs:2.125,0.00396073990623359)
--(axis cs:2.125,0.00412492321371829);

\path [draw=darkgray176, semithick]
(axis cs:3.125,0.0172892417758703)
--(axis cs:3.125,0.0181369241327047);

\addplot [semithick, darkgray176, mark=-, mark size=2.5, mark options={solid}, only marks]
table {%
0.125 0.00131961396784156
1.125 0.00196566665873776
2.125 0.00396073990623359
3.125 0.0172892417758703
};
\addplot [semithick, darkgray176, mark=-, mark size=2.5, mark options={solid}, only marks]
table {%
0.125 0.0013594086820824
1.125 0.00203721473285704
2.125 0.00412492321371829
3.125 0.0181369241327047
};
\end{groupplot}

\draw ({$(current bounding box.south west)!-0.01!(current bounding box.south east)$}|-{$(current bounding box.south west)!0.5!(current bounding box.north west)$}) node[
  scale=0.9,
  anchor=west,
  text=darkgray176,
  rotate=90.0
]{MSE};
\draw ({$(current bounding box.south west)!0.5!(current bounding box.south east)$}|-{$(current bounding box.south west)!-0.052!(current bounding box.north west)$}) node[
  scale=0.9,
  anchor=south,
  text=darkgray176,
  rotate=0.0
]{Num. Train Trajectories};
\end{tikzpicture}

%% file: tikz/weather/bar_fourier_2e-4_hist=4.tex
\begin{tikzpicture}

\definecolor{darkgray176}{RGB}{146,146,146}
\definecolor{darkslategray387083}{RGB}{38,70,83}
\definecolor{lightgray204}{RGB}{204,204,204}
\definecolor{tomato23111181}{RGB}{231,111,81}

\begin{groupplot}[group style={group size=2 by 2},height=0.24\textheight, width=0.52\textwidth, yticklabel style={color=darkgray176, font=\bfseries\scriptsize}, xticklabel style={color=darkgray176, font=\bfseries\scriptsize},]
\nextgroupplot[
axis line style={darkgray176},
legend cell align={left},
legend style={
  fill opacity=0.8,
  draw opacity=1,
  text=darkgray176,  
  text opacity=1,
  at={(0.03,0.97)},
  anchor=north west,
  draw=lightgray204,
  font=\scriptsize
},
tick align=outside,
tick pos=left,
title={Rollout},
x grid style={darkgray176},
xmajorgrids,
xmin=-0.5, xmax=3.5,
xtick style={color=darkgray176},
xticklabels={},
y grid style={darkgray176},
ymajorgrids,
ymin=0, ymax=0.720084678410775,
ytick style={color=darkgray176}
]
\draw[draw=none,fill=darkslategray387083,fill opacity=0.85] (axis cs:-0.25,0) rectangle (axis cs:0,0.0435867011547089);
\addlegendimage{ybar,ybar legend,draw=none,fill=darkslategray387083,fill opacity=0.85}
\addlegendentry{FNO}

\draw[draw=none,fill=darkslategray387083,fill opacity=0.85] (axis cs:0.75,0) rectangle (axis cs:1,0.0687271803617477);
\draw[draw=none,fill=darkslategray387083,fill opacity=0.85] (axis cs:1.75,0) rectangle (axis cs:2,0.161195285618305);
\draw[draw=none,fill=darkslategray387083,fill opacity=0.85] (axis cs:2.75,0) rectangle (axis cs:3,0.682143847147624);
\draw[draw=none,fill=tomato23111181,fill opacity=0.85] (axis cs:0,0) rectangle (axis cs:0.25,0.035269087801377);
\addlegendimage{ybar,ybar legend,draw=none,fill=tomato23111181,fill opacity=0.85}
\addlegendentry{CFNO}

\draw[draw=none,fill=tomato23111181,fill opacity=0.85] (axis cs:1,0) rectangle (axis cs:1.25,0.115606271972259);
\draw[draw=none,fill=tomato23111181,fill opacity=0.85] (axis cs:2,0) rectangle (axis cs:2.25,0.102290324866772);
\draw[draw=none,fill=tomato23111181,fill opacity=0.85] (axis cs:3,0) rectangle (axis cs:3.25,0.435696184635162);
\path [draw=darkgray176, semithick]
(axis cs:-0.125,0.0424671471118927)
--(axis cs:-0.125,0.044706255197525);

\path [draw=darkgray176, semithick]
(axis cs:0.875,0.0663914382457733)
--(axis cs:0.875,0.0710629224777222);

\path [draw=darkgray176, semithick]
(axis cs:1.875,0.155511781573296)
--(axis cs:1.875,0.166878789663315);

\path [draw=darkgray176, semithick]
(axis cs:2.875,0.644203015884472)
--(axis cs:2.875,0.720084678410775);

\addplot [semithick, darkgray176, mark=-, mark size=2.5, mark options={solid}, only marks]
table {%
-0.125 0.0424671471118927
0.875 0.0663914382457733
1.875 0.155511781573296
2.875 0.644203015884472
};

\addplot [semithick, darkgray176, mark=-, mark size=2.5, mark options={solid}, only marks]
table {%
-0.125 0.044706255197525
0.875 0.0710629224777222
1.875 0.166878789663315
2.875 0.720084678410775
};

\path [draw=darkgray176, semithick]
(axis cs:0.125,0.0342283195089997)
--(axis cs:0.125,0.0363098560937543);

\path [draw=darkgray176, semithick]
(axis cs:1.125,0.0243398149346986)
--(axis cs:1.125,0.206872729009819);

\path [draw=darkgray176, semithick]
(axis cs:2.125,0.0997205815830133)
--(axis cs:2.125,0.10486006815053);

\path [draw=darkgray176, semithick]
(axis cs:3.125,0.430129303082868)
--(axis cs:3.125,0.441263066187457);

\addplot [semithick, darkgray176, mark=-, mark size=2.5, mark options={solid}, only marks]
table {%
0.125 0.0342283195089997
1.125 0.0243398149346986
2.125 0.0997205815830133
3.125 0.430129303082868
};

\addplot [semithick, darkgray176, mark=-, mark size=2.5, mark options={solid}, only marks]
table {%
0.125 0.0363098560937543
1.125 0.206872729009819
2.125 0.10486006815053
3.125 0.441263066187457
};

\nextgroupplot[
axis line style={darkgray176},
tick align=outside,
tick pos=left,
title={One-step},
x grid style={darkgray176},
xmajorgrids,
xmin=-0.5, xmax=3.5,
xtick style={color=darkgray176},
xticklabels={},
y grid style={darkgray176},
ymajorgrids,
ymin=0, ymax=0.0700700133749393,
ytick style={color=darkgray176}
]
\draw[draw=none,fill=darkslategray387083,fill opacity=0.85] (axis cs:-0.25,0) rectangle (axis cs:0,0.00160222325939685);
\draw[draw=none,fill=darkslategray387083,fill opacity=0.85] (axis cs:0.75,0) rectangle (axis cs:1,0.00317146989982575);
\draw[draw=none,fill=darkslategray387083,fill opacity=0.85] (axis cs:1.75,0) rectangle (axis cs:2,0.0101928263902664);
\draw[draw=none,fill=darkslategray387083,fill opacity=0.85] (axis cs:2.75,0) rectangle (axis cs:3,0.0663080140948296);
\draw[draw=none,fill=tomato23111181,fill opacity=0.85] (axis cs:0,0) rectangle (axis cs:0.25,0.0015353843724976);
\draw[draw=none,fill=tomato23111181,fill opacity=0.85] (axis cs:1,0) rectangle (axis cs:1.25,0.00495798823734124);
\draw[draw=none,fill=tomato23111181,fill opacity=0.85] (axis cs:2,0) rectangle (axis cs:2.25,0.00654531518618266);
\draw[draw=none,fill=tomato23111181,fill opacity=0.85] (axis cs:3,0) rectangle (axis cs:3.25,0.0344935059547424);
\path [draw=darkgray176, semithick]
(axis cs:-0.125,0.00146273232530802)
--(axis cs:-0.125,0.00174171419348568);

\path [draw=darkgray176, semithick]
(axis cs:0.875,0.00282761082053185)
--(axis cs:0.875,0.00351532897911966);

\path [draw=darkgray176, semithick]
(axis cs:1.875,0.00867895688861609)
--(axis cs:1.875,0.0117066958919168);

\path [draw=darkgray176, semithick]
(axis cs:2.875,0.0625460148147199)
--(axis cs:2.875,0.0700700133749393);

\addplot [semithick, darkgray176, mark=-, mark size=2.5, mark options={solid}, only marks]
table {%
-0.125 0.00146273232530802
0.875 0.00282761082053185
1.875 0.00867895688861609
2.875 0.0625460148147199
};
\addplot [semithick, darkgray176, mark=-, mark size=2.5, mark options={solid}, only marks]
table {%
-0.125 0.00174171419348568
0.875 0.00351532897911966
1.875 0.0117066958919168
2.875 0.0700700133749393
};
\path [draw=darkgray176, semithick]
(axis cs:0.125,0.00145913546190758)
--(axis cs:0.125,0.00161163328308762);

\path [draw=darkgray176, semithick]
(axis cs:1.125,0.00162538502169844)
--(axis cs:1.125,0.00829059145298405);

\path [draw=darkgray176, semithick]
(axis cs:2.125,0.00620032109778849)
--(axis cs:2.125,0.00689030927457683);

\path [draw=darkgray176, semithick]
(axis cs:3.125,0.0336275204750069)
--(axis cs:3.125,0.0353594914344779);

\addplot [semithick, darkgray176, mark=-, mark size=2.5, mark options={solid}, only marks]
table {%
0.125 0.00145913546190758
1.125 0.00162538502169844
2.125 0.00620032109778849
3.125 0.0336275204750069
};
\addplot [semithick, darkgray176, mark=-, mark size=2.5, mark options={solid}, only marks]
table {%
0.125 0.00161163328308762
1.125 0.00829059145298405
2.125 0.00689030927457683
3.125 0.0353594914344779
};

\nextgroupplot[
axis line style={darkgray176},
tick align=outside,
tick pos=left,
title={Scalar},
x grid style={darkgray176},
xmajorgrids,
xmin=-0.5, xmax=3.5,
xtick style={color=darkgray176},
xtick={0,1,2,3},
xticklabels={896,448,192,56},
y grid style={darkgray176},
ymajorgrids,
ymin=0, ymax=0.0293359217637108,
ytick style={color=darkgray176}
]
\draw[draw=none,fill=darkslategray387083,fill opacity=0.85] (axis cs:-0.25,0) rectangle (axis cs:0,0.000387216598028317);
\draw[draw=none,fill=darkslategray387083,fill opacity=0.85] (axis cs:0.75,0) rectangle (axis cs:1,0.000864618254126981);
\draw[draw=none,fill=darkslategray387083,fill opacity=0.85] (axis cs:1.75,0) rectangle (axis cs:2,0.00331727811135352);
\draw[draw=none,fill=darkslategray387083,fill opacity=0.85] (axis cs:2.75,0) rectangle (axis cs:3,0.0276119622091452);
\draw[draw=none,fill=tomato23111181,fill opacity=0.85] (axis cs:0,0) rectangle (axis cs:0.25,0.00033950200304389);
\draw[draw=none,fill=tomato23111181,fill opacity=0.85] (axis cs:1,0) rectangle (axis cs:1.25,0.00104329721458877);
\draw[draw=none,fill=tomato23111181,fill opacity=0.85] (axis cs:2,0) rectangle (axis cs:2.25,0.00152281365202119);
\draw[draw=none,fill=tomato23111181,fill opacity=0.85] (axis cs:3,0) rectangle (axis cs:3.25,0.0092858737334609);
\path [draw=darkgray176, semithick]
(axis cs:-0.125,0.000316955323796719)
--(axis cs:-0.125,0.000457477872259915);

\path [draw=darkgray176, semithick]
(axis cs:0.875,0.000734147091861814)
--(axis cs:0.875,0.000995089416392148);

\path [draw=darkgray176, semithick]
(axis cs:1.875,0.00266182841733098)
--(axis cs:1.875,0.00397272780537605);

\path [draw=darkgray176, semithick]
(axis cs:2.875,0.0258880026545796)
--(axis cs:2.875,0.0293359217637108);

\addplot [semithick, darkgray176, mark=-, mark size=2.5, mark options={solid}, only marks]
table {%
-0.125 0.000316955323796719
0.875 0.000734147091861814
1.875 0.00266182841733098
2.875 0.0258880026545796
};
\addplot [semithick, darkgray176, mark=-, mark size=2.5, mark options={solid}, only marks]
table {%
-0.125 0.000457477872259915
0.875 0.000995089416392148
1.875 0.00397272780537605
2.875 0.0293359217637108
};
\path [draw=darkgray176, semithick]
(axis cs:0.125,0.000324376823843428)
--(axis cs:0.125,0.000354627182244352);

\path [draw=darkgray176, semithick]
(axis cs:1.125,0.000393386412034826)
--(axis cs:1.125,0.00169320801714272);

\path [draw=darkgray176, semithick]
(axis cs:2.125,0.00144645230060805)
--(axis cs:2.125,0.00159917500343433);

\path [draw=darkgray176, semithick]
(axis cs:3.125,0.00896296962995554)
--(axis cs:3.125,0.00960877783696626);

\addplot [semithick, darkgray176, mark=-, mark size=2.5, mark options={solid}, only marks]
table {%
0.125 0.000324376823843428
1.125 0.000393386412034826
2.125 0.00144645230060805
3.125 0.00896296962995554
};
\addplot [semithick, darkgray176, mark=-, mark size=2.5, mark options={solid}, only marks]
table {%
0.125 0.000354627182244352
1.125 0.00169320801714272
2.125 0.00159917500343433
3.125 0.00960877783696626
};

\nextgroupplot[
axis line style={darkgray176},
tick align=outside,
tick pos=left,
title={Vector},
x grid style={darkgray176},
xmajorgrids,
xmin=-0.5, xmax=3.5,
xtick style={color=darkgray176},
xtick={0,1,2,3},
xticklabels={896,448,192,56},
y grid style={darkgray176},
ymajorgrids,
ymin=0, ymax=0.0411172385581451,
ytick style={color=darkgray176}
]
\draw[draw=none,fill=darkslategray387083,fill opacity=0.85] (axis cs:-0.25,0) rectangle (axis cs:0,0.00121412967564538);
\draw[draw=none,fill=darkslategray387083,fill opacity=0.85] (axis cs:0.75,0) rectangle (axis cs:1,0.00230442336760461);
\draw[draw=none,fill=darkslategray387083,fill opacity=0.85] (axis cs:1.75,0) rectangle (axis cs:2,0.00688759656623006);
\draw[draw=none,fill=darkslategray387083,fill opacity=0.85] (axis cs:2.75,0) rectangle (axis cs:3,0.0387884577115377);
\draw[draw=none,fill=tomato23111181,fill opacity=0.85] (axis cs:0,0) rectangle (axis cs:0.25,0.00118881196249276);
\draw[draw=none,fill=tomato23111181,fill opacity=0.85] (axis cs:1,0) rectangle (axis cs:1.25,0.00389967346563935);
\draw[draw=none,fill=tomato23111181,fill opacity=0.85] (axis cs:2,0) rectangle (axis cs:2.25,0.00502337189391255);
\draw[draw=none,fill=tomato23111181,fill opacity=0.85] (axis cs:3,0) rectangle (axis cs:3.25,0.025177472581466);
\path [draw=darkgray176, semithick]
(axis cs:-0.125,0.0011449916055426)
--(axis cs:-0.125,0.00128326774574816);

\path [draw=darkgray176, semithick]
(axis cs:0.875,0.00209183944389224)
--(axis cs:0.875,0.00251700729131699);

\path [draw=darkgray176, semithick]
(axis cs:1.875,0.00602894322946668)
--(axis cs:1.875,0.00774624990299344);

\path [draw=darkgray176, semithick]
(axis cs:2.875,0.0364596768649302)
--(axis cs:2.875,0.0411172385581451);

\addplot [semithick, darkgray176, mark=-, mark size=2.5, mark options={solid}, only marks]
table {%
-0.125 0.0011449916055426
0.875 0.00209183944389224
1.875 0.00602894322946668
2.875 0.0364596768649302
};
\addplot [semithick, darkgray176, mark=-, mark size=2.5, mark options={solid}, only marks]
table {%
-0.125 0.00128326774574816
0.875 0.00251700729131699
1.875 0.00774624990299344
2.875 0.0411172385581451
};
\path [draw=darkgray176, semithick]
(axis cs:0.125,0.00112756473820786)
--(axis cs:0.125,0.00125005918677767);

\path [draw=darkgray176, semithick]
(axis cs:1.125,0.00122310911112378)
--(axis cs:1.125,0.00657623782015493);

\path [draw=darkgray176, semithick]
(axis cs:2.125,0.00475292370432748)
--(axis cs:2.125,0.00529382008349763);

\path [draw=darkgray176, semithick]
(axis cs:3.125,0.0246283865440642)
--(axis cs:3.125,0.0257265586188679);

\addplot [semithick, darkgray176, mark=-, mark size=2.5, mark options={solid}, only marks]
table {%
0.125 0.00112756473820786
1.125 0.00122310911112378
2.125 0.00475292370432748
3.125 0.0246283865440642
};
\addplot [semithick, darkgray176, mark=-, mark size=2.5, mark options={solid}, only marks]
table {%
0.125 0.00125005918677767
1.125 0.00657623782015493
2.125 0.00529382008349763
3.125 0.0257265586188679
};
\end{groupplot}

\draw ({$(current bounding box.south west)!-0.01!(current bounding box.south east)$}|-{$(current bounding box.south west)!0.5!(current bounding box.north west)$}) node[
  scale=0.9,
  anchor=west,
  text=darkgray176,
  rotate=90.0
]{MSE};
\draw ({$(current bounding box.south west)!0.5!(current bounding box.south east)$}|-{$(current bounding box.south west)!-0.05!(current bounding box.north west)$}) node[
  scale=0.9,
  anchor=south,
  text=darkgray176,
  rotate=0.0
]{Num. Train Trajectories};
\end{tikzpicture}

%% file: tikz/weather/bar_resnet_2e-4_hist=2.tex
\begin{tikzpicture}

\definecolor{burlywood233196106}{RGB}{233,196,106}
\definecolor{darkgray176}{RGB}{146,146,146}
\definecolor{lightgray204}{RGB}{204,204,204}
\definecolor{lightseagreen42157143}{RGB}{42,157,143}
\definecolor{sandybrown24416297}{RGB}{244,162,97}

\begin{groupplot}[group style={group size=2 by 2},,height=0.24\textheight, width=0.52\textwidth, yticklabel style={color=darkgray176, font=\bfseries\scriptsize}, xticklabel style={color=darkgray176, font=\bfseries\scriptsize},]
\nextgroupplot[
axis line style={darkgray176},
legend cell align={left},
legend style={
  fill opacity=0.8,
  draw opacity=1,
  text=darkgray176,  
  text opacity=1,
  at={(0.03,0.97)},
  anchor=north west,
  draw=lightgray204,
  font=\scriptsize,    
},
tick align=outside,
tick pos=left,
title={Rollout},
x grid style={darkgray176},
xmajorgrids,
xmin=-0.5, xmax=4.5,
xtick style={color=darkgray176},
xticklabels={},
y grid style={darkgray176},
ymajorgrids,
ymin=0, ymax=2.0916576385498,
ytick style={color=darkgray176}
]
\draw[draw=none,fill=lightseagreen42157143,fill opacity=0.85] (axis cs:-0.25,0) rectangle (axis cs:-0.0833333333333333,0.44796983897686);
\addlegendimage{ybar,ybar legend,draw=none,fill=lightseagreen42157143,fill opacity=0.85}
\addlegendentry{ResNet}

\draw[draw=none,fill=lightseagreen42157143,fill opacity=0.85] (axis cs:0.75,0) rectangle (axis cs:0.916666666666667,0.478021115064621);
\draw[draw=none,fill=lightseagreen42157143,fill opacity=0.85] (axis cs:1.75,0) rectangle (axis cs:1.91666666666667,0.490442663431168);
\draw[draw=none,fill=lightseagreen42157143,fill opacity=0.85] (axis cs:2.75,0) rectangle (axis cs:2.91666666666667,0.708285570144653);
\draw[draw=none,fill=lightseagreen42157143,fill opacity=0.85] (axis cs:3.75,0) rectangle (axis cs:3.91666666666667,1.11948978900909);
\draw[draw=none,fill=burlywood233196106,fill opacity=0.85] (axis cs:-0.0833333333333333,0) rectangle (axis cs:0.0833333333333333,0.281607732176781);
\addlegendimage{ybar,ybar legend,draw=none,fill=burlywood233196106,fill opacity=0.85}
\addlegendentry{CResNet}

\draw[draw=none,fill=burlywood233196106,fill opacity=0.85] (axis cs:0.916666666666667,0) rectangle (axis cs:1.08333333333333,0.284194797277451);
\draw[draw=none,fill=burlywood233196106,fill opacity=0.85] (axis cs:1.91666666666667,0) rectangle (axis cs:2.08333333333333,0.524738281965256);
\draw[draw=none,fill=burlywood233196106,fill opacity=0.85] (axis cs:2.91666666666667,0) rectangle (axis cs:3.08333333333333,1.17265355587006);
\draw[draw=none,fill=burlywood233196106,fill opacity=0.85] (axis cs:3.91666666666667,0) rectangle (axis cs:4.08333333333333,2.04229319095612);
\draw[draw=none,fill=sandybrown24416297,fill opacity=0.85] (axis cs:0.0833333333333333,0) rectangle (axis cs:0.25,0.216447204351425);
\addlegendimage{ybar,ybar legend,draw=none,fill=sandybrown24416297,fill opacity=0.85}
\addlegendentry{CResNet$_\text{rot}$}

\draw[draw=none,fill=sandybrown24416297,fill opacity=0.85] (axis cs:1.08333333333333,0) rectangle (axis cs:1.25,0.236994981765747);
\draw[draw=none,fill=sandybrown24416297,fill opacity=0.85] (axis cs:2.08333333333333,0) rectangle (axis cs:2.25,0.309622228145599);
\draw[draw=none,fill=sandybrown24416297,fill opacity=0.85] (axis cs:3.08333333333333,0) rectangle (axis cs:3.25,0.612715363502502);
\draw[draw=none,fill=sandybrown24416297,fill opacity=0.85] (axis cs:4.08333333333333,0) rectangle (axis cs:4.25,1.47560441493988);
\path [draw=darkgray176, semithick]
(axis cs:-0.166666666666667,0.442171990871429)
--(axis cs:-0.166666666666667,0.453767687082291);

\path [draw=darkgray176, semithick]
(axis cs:0.833333333333333,0.471808791160583)
--(axis cs:0.833333333333333,0.484233438968658);

\path [draw=darkgray176, semithick]
(axis cs:1.83333333333333,0.482428193092346)
--(axis cs:1.83333333333333,0.498457133769989);

\path [draw=darkgray176, semithick]
(axis cs:2.83333333333333,0.696405410766602)
--(axis cs:2.83333333333333,0.720165729522705);

\path [draw=darkgray176, semithick]
(axis cs:3.83333333333333,1.09977757930756)
--(axis cs:3.83333333333333,1.13920199871063);

\addplot [semithick, darkgray176, mark=-, mark size=2.5, mark options={solid}, only marks]
table {%
-0.166666666666667 0.442171990871429
0.833333333333333 0.471808791160583
1.83333333333333 0.482428193092346
2.83333333333333 0.696405410766602
3.83333333333333 1.09977757930756
};

\addplot [semithick, darkgray176, mark=-, mark size=2.5, mark options={solid}, only marks]
table {%
-0.166666666666667 0.453767687082291
0.833333333333333 0.484233438968658
1.83333333333333 0.498457133769989
2.83333333333333 0.720165729522705
3.83333333333333 1.13920199871063
};
\path [draw=darkgray176, semithick]
(axis cs:0,0.275118321180344)
--(axis cs:0,0.288097143173218);

\path [draw=darkgray176, semithick]
(axis cs:1,0.277455687522888)
--(axis cs:1,0.290933907032013);

\path [draw=darkgray176, semithick]
(axis cs:2,0.514678657054901)
--(axis cs:2,0.53479790687561);

\path [draw=darkgray176, semithick]
(axis cs:3,1.12437319755554)
--(axis cs:3,1.22093391418457);

\path [draw=darkgray176, semithick]
(axis cs:4,1.99292874336243)
--(axis cs:4,2.0916576385498);

\addplot [semithick, darkgray176, mark=-, mark size=2.5, mark options={solid}, only marks]
table {%
0 0.275118321180344
1 0.277455687522888
2 0.514678657054901
3 1.12437319755554
4 1.99292874336243
};

\addplot [semithick, darkgray176, mark=-, mark size=2.5, mark options={solid}, only marks]
table {%
0 0.288097143173218
1 0.290933907032013
2 0.53479790687561
3 1.22093391418457
4 2.0916576385498
};

\path [draw=darkgray176, semithick]
(axis cs:0.166666666666667,0.209446549415588)
--(axis cs:0.166666666666667,0.223447859287262);

\path [draw=darkgray176, semithick]
(axis cs:1.16666666666667,0.236994981765747)
--(axis cs:1.16666666666667,0.236994981765747);

\path [draw=darkgray176, semithick]
(axis cs:2.16666666666667,0.308647722005844)
--(axis cs:2.16666666666667,0.310596734285355);

\path [draw=darkgray176, semithick]
(axis cs:3.16666666666667,0.610884606838226)
--(axis cs:3.16666666666667,0.614546120166779);

\path [draw=darkgray176, semithick]
(axis cs:4.16666666666667,1.47121810913086)
--(axis cs:4.16666666666667,1.4799907207489);

\addplot [semithick, darkgray176, mark=-, mark size=2.5, mark options={solid}, only marks]
table {%
0.166666666666667 0.209446549415588
1.16666666666667 0.236994981765747
2.16666666666667 0.308647722005844
3.16666666666667 0.610884606838226
4.16666666666667 1.47121810913086
};

\addplot [semithick, darkgray176, mark=-, mark size=2.5, mark options={solid}, only marks]
table {%
0.166666666666667 0.223447859287262
1.16666666666667 0.236994981765747
2.16666666666667 0.310596734285355
3.16666666666667 0.614546120166779
4.16666666666667 1.4799907207489
};

\nextgroupplot[
axis line style={darkgray176},
tick align=outside,
tick pos=left,
title={One-step},
x grid style={darkgray176},
xmajorgrids,
xmin=-0.5, xmax=4.5,
xtick style={color=darkgray176},
xticklabels={},
y grid style={darkgray176},
ymajorgrids,
ymin=0, ymax=0.148205906152725,
ytick style={color=darkgray176}
]
\draw[draw=none,fill=lightseagreen42157143,fill opacity=0.85] (axis cs:-0.25,0) rectangle (axis cs:-0.0833333333333333,0.0180717324838042);
\draw[draw=none,fill=lightseagreen42157143,fill opacity=0.85] (axis cs:0.75,0) rectangle (axis cs:0.916666666666667,0.0184197016060352);
\draw[draw=none,fill=lightseagreen42157143,fill opacity=0.85] (axis cs:1.75,0) rectangle (axis cs:1.91666666666667,0.0241788029670715);
\draw[draw=none,fill=lightseagreen42157143,fill opacity=0.85] (axis cs:2.75,0) rectangle (axis cs:2.91666666666667,0.038529496639967);
\draw[draw=none,fill=lightseagreen42157143,fill opacity=0.85] (axis cs:3.75,0) rectangle (axis cs:3.91666666666667,0.0660587251186371);
\draw[draw=none,fill=burlywood233196106,fill opacity=0.85] (axis cs:-0.0833333333333333,0) rectangle (axis cs:0.0833333333333333,0.011072451248765);
\draw[draw=none,fill=burlywood233196106,fill opacity=0.85] (axis cs:0.916666666666667,0) rectangle (axis cs:1.08333333333333,0.01107277860865);
\draw[draw=none,fill=burlywood233196106,fill opacity=0.85] (axis cs:1.91666666666667,0) rectangle (axis cs:2.08333333333333,0.0278224851936102);
\draw[draw=none,fill=burlywood233196106,fill opacity=0.85] (axis cs:2.91666666666667,0) rectangle (axis cs:3.08333333333333,0.0685449503362179);
\draw[draw=none,fill=burlywood233196106,fill opacity=0.85] (axis cs:3.91666666666667,0) rectangle (axis cs:4.08333333333333,0.143995463848114);
\draw[draw=none,fill=sandybrown24416297,fill opacity=0.85] (axis cs:0.0833333333333333,0) rectangle (axis cs:0.25,0.00749733252450824);
\draw[draw=none,fill=sandybrown24416297,fill opacity=0.85] (axis cs:1.08333333333333,0) rectangle (axis cs:1.25,0.00688420748338103);
\draw[draw=none,fill=sandybrown24416297,fill opacity=0.85] (axis cs:2.08333333333333,0) rectangle (axis cs:2.25,0.0161148635670543);
\draw[draw=none,fill=sandybrown24416297,fill opacity=0.85] (axis cs:3.08333333333333,0) rectangle (axis cs:3.25,0.0335362777113914);
\draw[draw=none,fill=sandybrown24416297,fill opacity=0.85] (axis cs:4.08333333333333,0) rectangle (axis cs:4.25,0.0894178673624992);
\path [draw=darkgray176, semithick]
(axis cs:-0.166666666666667,0.017560001462698)
--(axis cs:-0.166666666666667,0.0185834635049105);

\path [draw=darkgray176, semithick]
(axis cs:0.833333333333333,0.0173870921134949)
--(axis cs:0.833333333333333,0.0194523110985756);

\path [draw=darkgray176, semithick]
(axis cs:1.83333333333333,0.0238740108907223)
--(axis cs:1.83333333333333,0.0244835950434208);

\path [draw=darkgray176, semithick]
(axis cs:2.83333333333333,0.0375280454754829)
--(axis cs:2.83333333333333,0.039530947804451);

\path [draw=darkgray176, semithick]
(axis cs:3.83333333333333,0.0649633407592773)
--(axis cs:3.83333333333333,0.0671541094779968);

\addplot [semithick, darkgray176, mark=-, mark size=2.5, mark options={solid}, only marks]
table {%
-0.166666666666667 0.017560001462698
0.833333333333333 0.0173870921134949
1.83333333333333 0.0238740108907223
2.83333333333333 0.0375280454754829
3.83333333333333 0.0649633407592773
};
\addplot [semithick, darkgray176, mark=-, mark size=2.5, mark options={solid}, only marks]
table {%
-0.166666666666667 0.0185834635049105
0.833333333333333 0.0194523110985756
1.83333333333333 0.0244835950434208
2.83333333333333 0.039530947804451
3.83333333333333 0.0671541094779968
};
\path [draw=darkgray176, semithick]
(axis cs:0,0.0107529880478978)
--(axis cs:0,0.0113919144496322);

\path [draw=darkgray176, semithick]
(axis cs:1,0.0110111711546779)
--(axis cs:1,0.0111343860626221);

\path [draw=darkgray176, semithick]
(axis cs:2,0.0271749682724476)
--(axis cs:2,0.0284700021147728);

\path [draw=darkgray176, semithick]
(axis cs:3,0.0655909776687622)
--(axis cs:3,0.0714989230036736);

\path [draw=darkgray176, semithick]
(axis cs:4,0.139785021543503)
--(axis cs:4,0.148205906152725);

\addplot [semithick, darkgray176, mark=-, mark size=2.5, mark options={solid}, only marks]
table {%
0 0.0107529880478978
1 0.0110111711546779
2 0.0271749682724476
3 0.0655909776687622
4 0.139785021543503
};
\addplot [semithick, darkgray176, mark=-, mark size=2.5, mark options={solid}, only marks]
table {%
0 0.0113919144496322
1 0.0111343860626221
2 0.0284700021147728
3 0.0714989230036736
4 0.148205906152725
};
\path [draw=darkgray176, semithick]
(axis cs:0.166666666666667,0.00637686531990767)
--(axis cs:0.166666666666667,0.00861779972910881);

\path [draw=darkgray176, semithick]
(axis cs:1.16666666666667,0.00688420748338103)
--(axis cs:1.16666666666667,0.00688420748338103);

\path [draw=darkgray176, semithick]
(axis cs:2.16666666666667,0.0160245858132839)
--(axis cs:2.16666666666667,0.0162051413208246);

\path [draw=darkgray176, semithick]
(axis cs:3.16666666666667,0.0333758853375912)
--(axis cs:3.16666666666667,0.0336966700851917);

\path [draw=darkgray176, semithick]
(axis cs:4.16666666666667,0.0887224078178406)
--(axis cs:4.16666666666667,0.0901133269071579);

\addplot [semithick, darkgray176, mark=-, mark size=2.5, mark options={solid}, only marks]
table {%
0.166666666666667 0.00637686531990767
1.16666666666667 0.00688420748338103
2.16666666666667 0.0160245858132839
3.16666666666667 0.0333758853375912
4.16666666666667 0.0887224078178406
};
\addplot [semithick, darkgray176, mark=-, mark size=2.5, mark options={solid}, only marks]
table {%
0.166666666666667 0.00861779972910881
1.16666666666667 0.00688420748338103
2.16666666666667 0.0162051413208246
3.16666666666667 0.0336966700851917
4.16666666666667 0.0901133269071579
};

\nextgroupplot[
axis line style={darkgray176},
tick align=outside,
tick pos=left,
title={Scalar},
x grid style={darkgray176},
xmajorgrids,
xmin=-0.5, xmax=4.5,
xtick style={color=darkgray176},
xtick={0,1,2,3,4},
xticklabels={2048,1792,896,448,192},
y grid style={darkgray176},
ymajorgrids,
ymin=0, ymax=0.063310518860817,
ytick style={color=darkgray176}
]
\draw[draw=none,fill=lightseagreen42157143,fill opacity=0.85] (axis cs:-0.25,0) rectangle (axis cs:-0.0833333333333333,0.00596607336774468);
\draw[draw=none,fill=lightseagreen42157143,fill opacity=0.85] (axis cs:0.75,0) rectangle (axis cs:0.916666666666667,0.00607453542761505);
\draw[draw=none,fill=lightseagreen42157143,fill opacity=0.85] (axis cs:1.75,0) rectangle (axis cs:1.91666666666667,0.00855469610542059);
\draw[draw=none,fill=lightseagreen42157143,fill opacity=0.85] (axis cs:2.75,0) rectangle (axis cs:2.91666666666667,0.0140063273720443);
\draw[draw=none,fill=lightseagreen42157143,fill opacity=0.85] (axis cs:3.75,0) rectangle (axis cs:3.91666666666667,0.0240334495902061);
\draw[draw=none,fill=burlywood233196106,fill opacity=0.85] (axis cs:-0.0833333333333333,0) rectangle (axis cs:0.0833333333333333,0.00392377167008817);
\draw[draw=none,fill=burlywood233196106,fill opacity=0.85] (axis cs:0.916666666666667,0) rectangle (axis cs:1.08333333333333,0.0039281650679186);
\draw[draw=none,fill=burlywood233196106,fill opacity=0.85] (axis cs:1.91666666666667,0) rectangle (axis cs:2.08333333333333,0.00953843770548701);
\draw[draw=none,fill=burlywood233196106,fill opacity=0.85] (axis cs:2.91666666666667,0) rectangle (axis cs:3.08333333333333,0.0238178269937634);
\draw[draw=none,fill=burlywood233196106,fill opacity=0.85] (axis cs:3.91666666666667,0) rectangle (axis cs:4.08333333333333,0.061688020825386);
\draw[draw=none,fill=sandybrown24416297,fill opacity=0.85] (axis cs:0.0833333333333333,0) rectangle (axis cs:0.25,0.00275734916795045);
\draw[draw=none,fill=sandybrown24416297,fill opacity=0.85] (axis cs:1.08333333333333,0) rectangle (axis cs:1.25,0.00253422977402806);
\draw[draw=none,fill=sandybrown24416297,fill opacity=0.85] (axis cs:2.08333333333333,0) rectangle (axis cs:2.25,0.00552859017625451);
\draw[draw=none,fill=sandybrown24416297,fill opacity=0.85] (axis cs:3.08333333333333,0) rectangle (axis cs:3.25,0.0114229419268668);
\draw[draw=none,fill=sandybrown24416297,fill opacity=0.85] (axis cs:4.08333333333333,0) rectangle (axis cs:4.25,0.0318812001496553);
\path [draw=darkgray176, semithick]
(axis cs:-0.166666666666667,0.00577771803364158)
--(axis cs:-0.166666666666667,0.00615442870184779);

\path [draw=darkgray176, semithick]
(axis cs:0.833333333333333,0.00589599180966616)
--(axis cs:0.833333333333333,0.00625307904556394);

\path [draw=darkgray176, semithick]
(axis cs:1.83333333333333,0.00855275522917509)
--(axis cs:1.83333333333333,0.00855663698166609);

\path [draw=darkgray176, semithick]
(axis cs:2.83333333333333,0.0136962765827775)
--(axis cs:2.83333333333333,0.0143163781613111);

\path [draw=darkgray176, semithick]
(axis cs:3.83333333333333,0.0238702539354563)
--(axis cs:3.83333333333333,0.024196645244956);

\addplot [semithick, darkgray176, mark=-, mark size=2.5, mark options={solid}, only marks]
table {%
-0.166666666666667 0.00577771803364158
0.833333333333333 0.00589599180966616
1.83333333333333 0.00855275522917509
2.83333333333333 0.0136962765827775
3.83333333333333 0.0238702539354563
};
\addplot [semithick, darkgray176, mark=-, mark size=2.5, mark options={solid}, only marks]
table {%
-0.166666666666667 0.00615442870184779
0.833333333333333 0.00625307904556394
1.83333333333333 0.00855663698166609
2.83333333333333 0.0143163781613111
3.83333333333333 0.024196645244956
};
\path [draw=darkgray176, semithick]
(axis cs:0,0.00382177997380495)
--(axis cs:0,0.00402576336637139);

\path [draw=darkgray176, semithick]
(axis cs:1,0.00388074549846351)
--(axis cs:1,0.00397558463737369);

\path [draw=darkgray176, semithick]
(axis cs:2,0.00932062137871981)
--(axis cs:2,0.00975625403225422);

\path [draw=darkgray176, semithick]
(axis cs:3,0.023075083270669)
--(axis cs:3,0.0245605707168579);

\path [draw=darkgray176, semithick]
(axis cs:4,0.0600655227899551)
--(axis cs:4,0.063310518860817);

\addplot [semithick, darkgray176, mark=-, mark size=2.5, mark options={solid}, only marks]
table {%
0 0.00382177997380495
1 0.00388074549846351
2 0.00932062137871981
3 0.023075083270669
4 0.0600655227899551
};
\addplot [semithick, darkgray176, mark=-, mark size=2.5, mark options={solid}, only marks]
table {%
0 0.00402576336637139
1 0.00397558463737369
2 0.00975625403225422
3 0.0245605707168579
4 0.063310518860817
};
\path [draw=darkgray176, semithick]
(axis cs:0.166666666666667,0.0022805945482105)
--(axis cs:0.166666666666667,0.0032341037876904);

\path [draw=darkgray176, semithick]
(axis cs:1.16666666666667,0.00253422977402806)
--(axis cs:1.16666666666667,0.00253422977402806);

\path [draw=darkgray176, semithick]
(axis cs:2.16666666666667,0.00551193580031395)
--(axis cs:2.16666666666667,0.00554524455219507);

\path [draw=darkgray176, semithick]
(axis cs:3.16666666666667,0.0113413389772177)
--(axis cs:3.16666666666667,0.0115045448765159);

\path [draw=darkgray176, semithick]
(axis cs:4.16666666666667,0.0315973982214928)
--(axis cs:4.16666666666667,0.0321650020778179);

\addplot [semithick, darkgray176, mark=-, mark size=2.5, mark options={solid}, only marks]
table {%
0.166666666666667 0.0022805945482105
1.16666666666667 0.00253422977402806
2.16666666666667 0.00551193580031395
3.16666666666667 0.0113413389772177
4.16666666666667 0.0315973982214928
};
\addplot [semithick, darkgray176, mark=-, mark size=2.5, mark options={solid}, only marks]
table {%
0.166666666666667 0.0032341037876904
1.16666666666667 0.00253422977402806
2.16666666666667 0.00554524455219507
3.16666666666667 0.0115045448765159
4.16666666666667 0.0321650020778179
};

\nextgroupplot[
axis line style={darkgray176},
tick align=outside,
tick pos=left,
title={Vector},
x grid style={darkgray176},
xmajorgrids,
xmin=-0.5, xmax=4.5,
xtick style={color=darkgray176},
xtick={0,1,2,3,4},
xticklabels={2048,1792,896,448,192},
y grid style={darkgray176},
ymajorgrids,
ymin=0, ymax=0.0849384739995003,
ytick style={color=darkgray176}
]
\draw[draw=none,fill=lightseagreen42157143,fill opacity=0.85] (axis cs:-0.25,0) rectangle (axis cs:-0.0833333333333333,0.0121299182064831);
\draw[draw=none,fill=lightseagreen42157143,fill opacity=0.85] (axis cs:0.75,0) rectangle (axis cs:0.916666666666667,0.0123499832116067);
\draw[draw=none,fill=lightseagreen42157143,fill opacity=0.85] (axis cs:1.75,0) rectangle (axis cs:1.91666666666667,0.0156368631869555);
\draw[draw=none,fill=lightseagreen42157143,fill opacity=0.85] (axis cs:2.75,0) rectangle (axis cs:2.91666666666667,0.0245414953678846);
\draw[draw=none,fill=lightseagreen42157143,fill opacity=0.85] (axis cs:3.75,0) rectangle (axis cs:3.91666666666667,0.0420597866177559);
\draw[draw=none,fill=burlywood233196106,fill opacity=0.85] (axis cs:-0.0833333333333333,0) rectangle (axis cs:0.0833333333333333,0.00716561218723655);
\draw[draw=none,fill=burlywood233196106,fill opacity=0.85] (axis cs:0.916666666666667,0) rectangle (axis cs:1.08333333333333,0.00714993057772517);
\draw[draw=none,fill=burlywood233196106,fill opacity=0.85] (axis cs:1.91666666666667,0) rectangle (axis cs:2.08333333333333,0.0182941788807511);
\draw[draw=none,fill=burlywood233196106,fill opacity=0.85] (axis cs:2.91666666666667,0) rectangle (axis cs:3.08333333333333,0.044754346832633);
\draw[draw=none,fill=burlywood233196106,fill opacity=0.85] (axis cs:3.91666666666667,0) rectangle (axis cs:4.08333333333333,0.0822731778025627);
\draw[draw=none,fill=sandybrown24416297,fill opacity=0.85] (axis cs:0.0833333333333333,0) rectangle (axis cs:0.25,0.00475162081420422);
\draw[draw=none,fill=sandybrown24416297,fill opacity=0.85] (axis cs:1.08333333333333,0) rectangle (axis cs:1.25,0.00435626972466707);
\draw[draw=none,fill=sandybrown24416297,fill opacity=0.85] (axis cs:2.08333333333333,0) rectangle (axis cs:2.25,0.0105918091721833);
\draw[draw=none,fill=sandybrown24416297,fill opacity=0.85] (axis cs:3.08333333333333,0) rectangle (axis cs:3.25,0.0221252283081412);
\draw[draw=none,fill=sandybrown24416297,fill opacity=0.85] (axis cs:4.08333333333333,0) rectangle (axis cs:4.25,0.0575585011392832);
\path [draw=darkgray176, semithick]
(axis cs:-0.166666666666667,0.0118048600852489)
--(axis cs:-0.166666666666667,0.0124549763277173);

\path [draw=darkgray176, semithick]
(axis cs:0.833333333333333,0.0114900982007384)
--(axis cs:0.833333333333333,0.0132098682224751);

\path [draw=darkgray176, semithick]
(axis cs:1.83333333333333,0.0153260733932257)
--(axis cs:1.83333333333333,0.0159476529806852);

\path [draw=darkgray176, semithick]
(axis cs:2.83333333333333,0.0238264929503202)
--(axis cs:2.83333333333333,0.025256497785449);

\path [draw=darkgray176, semithick]
(axis cs:3.83333333333333,0.0410754755139351)
--(axis cs:3.83333333333333,0.0430440977215767);

\addplot [semithick, darkgray176, mark=-, mark size=2.5, mark options={solid}, only marks]
table {%
-0.166666666666667 0.0118048600852489
0.833333333333333 0.0114900982007384
1.83333333333333 0.0153260733932257
2.83333333333333 0.0238264929503202
3.83333333333333 0.0410754755139351
};
\addplot [semithick, darkgray176, mark=-, mark size=2.5, mark options={solid}, only marks]
table {%
-0.166666666666667 0.0124549763277173
0.833333333333333 0.0132098682224751
1.83333333333333 0.0159476529806852
2.83333333333333 0.025256497785449
3.83333333333333 0.0430440977215767
};
\path [draw=darkgray176, semithick]
(axis cs:0,0.00694545032456517)
--(axis cs:0,0.00738577404990792);

\path [draw=darkgray176, semithick]
(axis cs:1,0.00714098988100886)
--(axis cs:1,0.00715887127444148);

\path [draw=darkgray176, semithick]
(axis cs:2,0.0178525373339653)
--(axis cs:2,0.018735820427537);

\path [draw=darkgray176, semithick]
(axis cs:3,0.0424975752830505)
--(axis cs:3,0.0470111183822155);

\path [draw=darkgray176, semithick]
(axis cs:4,0.0796078816056252)
--(axis cs:4,0.0849384739995003);

\addplot [semithick, darkgray176, mark=-, mark size=2.5, mark options={solid}, only marks]
table {%
0 0.00694545032456517
1 0.00714098988100886
2 0.0178525373339653
3 0.0424975752830505
4 0.0796078816056252
};
\addplot [semithick, darkgray176, mark=-, mark size=2.5, mark options={solid}, only marks]
table {%
0 0.00738577404990792
1 0.00715887127444148
2 0.018735820427537
3 0.0470111183822155
4 0.0849384739995003
};
\path [draw=darkgray176, semithick]
(axis cs:0.166666666666667,0.00410480704158545)
--(axis cs:0.166666666666667,0.00539843458682299);

\path [draw=darkgray176, semithick]
(axis cs:1.16666666666667,0.00435626972466707)
--(axis cs:1.16666666666667,0.00435626972466707);

\path [draw=darkgray176, semithick]
(axis cs:2.16666666666667,0.010513917542994)
--(axis cs:2.16666666666667,0.0106697008013725);

\path [draw=darkgray176, semithick]
(axis cs:3.16666666666667,0.0220324099063873)
--(axis cs:3.16666666666667,0.0222180467098951);

\path [draw=darkgray176, semithick]
(axis cs:4.16666666666667,0.057091947644949)
--(axis cs:4.16666666666667,0.0580250546336174);

\addplot [semithick, darkgray176, mark=-, mark size=2.5, mark options={solid}, only marks]
table {%
0.166666666666667 0.00410480704158545
1.16666666666667 0.00435626972466707
2.16666666666667 0.010513917542994
3.16666666666667 0.0220324099063873
4.16666666666667 0.057091947644949
};
\addplot [semithick, darkgray176, mark=-, mark size=2.5, mark options={solid}, only marks]
table {%
0.166666666666667 0.00539843458682299
1.16666666666667 0.00435626972466707
2.16666666666667 0.0106697008013725
3.16666666666667 0.0222180467098951
4.16666666666667 0.0580250546336174
};
\end{groupplot}

\draw ({$(current bounding box.south west)!-0.02!(current bounding box.south east)$}|-{$(current bounding box.south west)!0.5!(current bounding box.north west)$}) node[
  scale=0.9,
  anchor=west,
  text=darkgray176,
  rotate=90.0
]{MSE};
\draw ({$(current bounding box.south west)!0.5!(current bounding box.south east)$}|-{$(current bounding box.south west)!-0.05!(current bounding box.north west)$}) node[
  scale=0.9,
  anchor=south,
  text=darkgray176,
  rotate=0.0
]{Num. Train Trajectories};
\end{tikzpicture}

%% file: tikz/maxwell/trajfig/maxwell_1.tex
\begin{tikzpicture}

\definecolor{darkgray176}{RGB}{176,176,176}

\begin{groupplot}[group style={group size=3 by 2}, yticklabel style={color=darkgray176, font=\bfseries\huge}, xticklabel style={color=darkgray176, font=\bfseries}, title style={font=\Huge,yshift=-1em},]
\nextgroupplot[
axis line style={darkgray176},
hide x axis,
hide y axis,
tick align=outside,
tick pos=left,
title={$D_x$: yz-plane},
x grid style={darkgray176},
xmin=-0.5, xmax=31.5,
xtick style={color=darkgray176},
y dir=reverse,
y grid style={darkgray176},
ymajorgrids,
ymin=-0.5, ymax=31.5,
ytick style={color=darkgray176}
]
\addplot graphics [includegraphics cmd=\pgfimage,xmin=-0.5, xmax=31.5, ymin=31.5, ymax=-0.5] {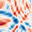};

\nextgroupplot[
axis line style={darkgray176},
hide x axis,
hide y axis,
tick align=outside,
tick pos=left,
title={$D_y$: yz-plane},
x grid style={darkgray176},
xmin=-0.5, xmax=31.5,
xtick style={color=darkgray176},
y dir=reverse,
y grid style={darkgray176},
ymajorgrids,
ymin=-0.5, ymax=31.5,
ytick style={color=darkgray176}
]
\addplot graphics [includegraphics cmd=\pgfimage,xmin=-0.5, xmax=31.5, ymin=31.5, ymax=-0.5] {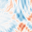};

\nextgroupplot[
axis line style={darkgray176},
colorbar,
colorbar style={ylabel={}},
colormap={mymap}{[1pt]
  rgb(0pt)=(0.403921568627451,0,0.12156862745098);
  rgb(1pt)=(0.698039215686274,0.0941176470588235,0.168627450980392);
  rgb(2pt)=(0.83921568627451,0.376470588235294,0.301960784313725);
  rgb(3pt)=(0.956862745098039,0.647058823529412,0.509803921568627);
  rgb(4pt)=(0.992156862745098,0.858823529411765,0.780392156862745);
  rgb(5pt)=(0.968627450980392,0.968627450980392,0.968627450980392);
  rgb(6pt)=(0.819607843137255,0.898039215686275,0.941176470588235);
  rgb(7pt)=(0.572549019607843,0.772549019607843,0.870588235294118);
  rgb(8pt)=(0.262745098039216,0.576470588235294,0.764705882352941);
  rgb(9pt)=(0.129411764705882,0.4,0.674509803921569);
  rgb(10pt)=(0.0196078431372549,0.188235294117647,0.380392156862745)
},
hide x axis,
hide y axis,
point meta max=0.0244226295115271,
point meta min=-0.0244226295115271,
tick align=outside,
tick pos=left,
title={$D_z$: yz-plane},
x grid style={darkgray176},
xmin=-0.5, xmax=31.5,
xtick style={color=darkgray176},
y dir=reverse,
y grid style={darkgray176},
ymajorgrids,
ymin=-0.5, ymax=31.5,
ytick style={color=darkgray176}
]
\addplot graphics [includegraphics cmd=\pgfimage,xmin=-0.5, xmax=31.5, ymin=31.5, ymax=-0.5] {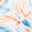};

\nextgroupplot[
axis line style={darkgray176},
hide x axis,
hide y axis,
tick align=outside,
tick pos=left,
title={$H_x$: yz-plane},
x grid style={darkgray176},
xmin=-0.5, xmax=31.5,
xtick style={color=darkgray176},
y dir=reverse,
y grid style={darkgray176},
ymajorgrids,
ymin=-0.5, ymax=31.5,
ytick style={color=darkgray176}
]
\addplot graphics [includegraphics cmd=\pgfimage,xmin=-0.5, xmax=31.5, ymin=31.5, ymax=-0.5] {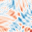};

\nextgroupplot[
axis line style={darkgray176},
hide x axis,
hide y axis,
tick align=outside,
tick pos=left,
title={$H_y$: yz-plane},
x grid style={darkgray176},
xmin=-0.5, xmax=31.5,
xtick style={color=darkgray176},
y dir=reverse,
y grid style={darkgray176},
ymajorgrids,
ymin=-0.5, ymax=31.5,
ytick style={color=darkgray176}
]
\addplot graphics [includegraphics cmd=\pgfimage,xmin=-0.5, xmax=31.5, ymin=31.5, ymax=-0.5] {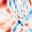};

\nextgroupplot[
axis line style={darkgray176},
colorbar,
colorbar style={ylabel={}},
colormap={mymap}{[1pt]
  rgb(0pt)=(0.403921568627451,0,0.12156862745098);
  rgb(1pt)=(0.698039215686274,0.0941176470588235,0.168627450980392);
  rgb(2pt)=(0.83921568627451,0.376470588235294,0.301960784313725);
  rgb(3pt)=(0.956862745098039,0.647058823529412,0.509803921568627);
  rgb(4pt)=(0.992156862745098,0.858823529411765,0.780392156862745);
  rgb(5pt)=(0.968627450980392,0.968627450980392,0.968627450980392);
  rgb(6pt)=(0.819607843137255,0.898039215686275,0.941176470588235);
  rgb(7pt)=(0.572549019607843,0.772549019607843,0.870588235294118);
  rgb(8pt)=(0.262745098039216,0.576470588235294,0.764705882352941);
  rgb(9pt)=(0.129411764705882,0.4,0.674509803921569);
  rgb(10pt)=(0.0196078431372549,0.188235294117647,0.380392156862745)
},
hide x axis,
hide y axis,
point meta max=0.0625548835776461,
point meta min=-0.0625548835776461,
tick align=outside,
tick pos=left,
title={$H_z$: yz-plane},
x grid style={darkgray176},
xmin=-0.5, xmax=31.5,
xtick style={color=darkgray176},
y dir=reverse,
y grid style={darkgray176},
ymajorgrids,
ymin=-0.5, ymax=31.5,
ytick style={color=darkgray176}
]
\addplot graphics [includegraphics cmd=\pgfimage,xmin=-0.5, xmax=31.5, ymin=31.5, ymax=-0.5] {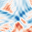};
\end{groupplot}

\end{tikzpicture}

%% file: tikz/maxwell/trajfig/maxwell_2.tex
\begin{tikzpicture}

\definecolor{darkgray176}{RGB}{176,176,176}

\begin{groupplot}[group style={group size=3 by 2}, yticklabel style={color=darkgray176, font=\bfseries\huge}, xticklabel style={color=darkgray176, font=\bfseries}, title style={font=\Huge, yshift=-1em},]
\nextgroupplot[
axis line style={darkgray176},
hide x axis,
hide y axis,
tick align=outside,
tick pos=left,
title={$D_x$: yz-plane},
x grid style={darkgray176},
xmin=-0.5, xmax=31.5,
xtick style={color=darkgray176},
y dir=reverse,
y grid style={darkgray176},
ymajorgrids,
ymin=-0.5, ymax=31.5,
ytick style={color=darkgray176}
]
\addplot graphics [includegraphics cmd=\pgfimage,xmin=-0.5, xmax=31.5, ymin=31.5, ymax=-0.5] {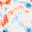};

\nextgroupplot[
axis line style={darkgray176},
hide x axis,
hide y axis,
tick align=outside,
tick pos=left,
title={$D_y$: yz-plane},
x grid style={darkgray176},
xmin=-0.5, xmax=31.5,
xtick style={color=darkgray176},
y dir=reverse,
y grid style={darkgray176},
ymajorgrids,
ymin=-0.5, ymax=31.5,
ytick style={color=darkgray176}
]
\addplot graphics [includegraphics cmd=\pgfimage,xmin=-0.5, xmax=31.5, ymin=31.5, ymax=-0.5] {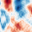};

\nextgroupplot[
axis line style={darkgray176},
colorbar,
colorbar style={ylabel={}},
colormap={mymap}{[1pt]
  rgb(0pt)=(0.403921568627451,0,0.12156862745098);
  rgb(1pt)=(0.698039215686274,0.0941176470588235,0.168627450980392);
  rgb(2pt)=(0.83921568627451,0.376470588235294,0.301960784313725);
  rgb(3pt)=(0.956862745098039,0.647058823529412,0.509803921568627);
  rgb(4pt)=(0.992156862745098,0.858823529411765,0.780392156862745);
  rgb(5pt)=(0.968627450980392,0.968627450980392,0.968627450980392);
  rgb(6pt)=(0.819607843137255,0.898039215686275,0.941176470588235);
  rgb(7pt)=(0.572549019607843,0.772549019607843,0.870588235294118);
  rgb(8pt)=(0.262745098039216,0.576470588235294,0.764705882352941);
  rgb(9pt)=(0.129411764705882,0.4,0.674509803921569);
  rgb(10pt)=(0.0196078431372549,0.188235294117647,0.380392156862745)
},
hide x axis,
hide y axis,
point meta max=0.0239905632012178,
point meta min=-0.0239905632012178,
tick align=outside,
tick pos=left,
title={$D_z$: yz-plane},
x grid style={darkgray176},
xmin=-0.5, xmax=31.5,
xtick style={color=darkgray176},
y dir=reverse,
y grid style={darkgray176},
ymajorgrids,
ymin=-0.5, ymax=31.5,
ytick style={color=darkgray176}
]
\addplot graphics [includegraphics cmd=\pgfimage,xmin=-0.5, xmax=31.5, ymin=31.5, ymax=-0.5] {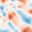};

\nextgroupplot[
axis line style={darkgray176},
hide x axis,
hide y axis,
tick align=outside,
tick pos=left,
title={$H_x$: yz-plane},
x grid style={darkgray176},
xmin=-0.5, xmax=31.5,
xtick style={color=darkgray176},
y dir=reverse,
y grid style={darkgray176},
ymajorgrids,
ymin=-0.5, ymax=31.5,
ytick style={color=darkgray176}
]
\addplot graphics [includegraphics cmd=\pgfimage,xmin=-0.5, xmax=31.5, ymin=31.5, ymax=-0.5] {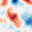};

\nextgroupplot[
axis line style={darkgray176},
hide x axis,
hide y axis,
tick align=outside,
tick pos=left,
title={$H_y$: yz-plane},
x grid style={darkgray176},
xmin=-0.5, xmax=31.5,
xtick style={color=darkgray176},
y dir=reverse,
y grid style={darkgray176},
ymajorgrids,
ymin=-0.5, ymax=31.5,
ytick style={color=darkgray176}
]
\addplot graphics [includegraphics cmd=\pgfimage,xmin=-0.5, xmax=31.5, ymin=31.5, ymax=-0.5] {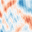};

\nextgroupplot[
axis line style={darkgray176},
colorbar,
colorbar style={ylabel={}},
colormap={mymap}{[1pt]
  rgb(0pt)=(0.403921568627451,0,0.12156862745098);
  rgb(1pt)=(0.698039215686274,0.0941176470588235,0.168627450980392);
  rgb(2pt)=(0.83921568627451,0.376470588235294,0.301960784313725);
  rgb(3pt)=(0.956862745098039,0.647058823529412,0.509803921568627);
  rgb(4pt)=(0.992156862745098,0.858823529411765,0.780392156862745);
  rgb(5pt)=(0.968627450980392,0.968627450980392,0.968627450980392);
  rgb(6pt)=(0.819607843137255,0.898039215686275,0.941176470588235);
  rgb(7pt)=(0.572549019607843,0.772549019607843,0.870588235294118);
  rgb(8pt)=(0.262745098039216,0.576470588235294,0.764705882352941);
  rgb(9pt)=(0.129411764705882,0.4,0.674509803921569);
  rgb(10pt)=(0.0196078431372549,0.188235294117647,0.380392156862745)
},
hide x axis,
hide y axis,
point meta max=0.0739032650629722,
point meta min=-0.0739032650629722,
tick align=outside,
tick pos=left,
title={$H_z$: yz-plane},
x grid style={darkgray176},
xmin=-0.5, xmax=31.5,
xtick style={color=darkgray176},
y dir=reverse,
y grid style={darkgray176},
ymajorgrids,
ymin=-0.5, ymax=31.5,
ytick style={color=darkgray176}
]
\addplot graphics [includegraphics cmd=\pgfimage,xmin=-0.5, xmax=31.5, ymin=31.5, ymax=-0.5] {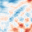};
\end{groupplot}

\end{tikzpicture}

%% file: tikz/maxwell/trajfig/maxwell_3.tex
\begin{tikzpicture}

\definecolor{darkgray176}{RGB}{176,176,176}

\begin{groupplot}[group style={group size=3 by 2},yticklabel style={color=darkgray176, font=\bfseries\huge}, xticklabel style={color=darkgray176, font=\bfseries}, title style={font=\Huge,yshift=-1em},]
\nextgroupplot[
axis line style={darkgray176},
hide x axis,
hide y axis,
tick align=outside,
tick pos=left,
title={$D_x$: yz-plane},
x grid style={darkgray176},
xmin=-0.5, xmax=31.5,
xtick style={color=darkgray176},
y dir=reverse,
y grid style={darkgray176},
ymajorgrids,
ymin=-0.5, ymax=31.5,
ytick style={color=darkgray176}
]
\addplot graphics [includegraphics cmd=\pgfimage,xmin=-0.5, xmax=31.5, ymin=31.5, ymax=-0.5] {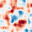};

\nextgroupplot[
axis line style={darkgray176},
hide x axis,
hide y axis,
tick align=outside,
tick pos=left,
title={$D_y$: yz-plane},
x grid style={darkgray176},
xmin=-0.5, xmax=31.5,
xtick style={color=darkgray176},
y dir=reverse,
y grid style={darkgray176},
ymajorgrids,
ymin=-0.5, ymax=31.5,
ytick style={color=darkgray176}
]
\addplot graphics [includegraphics cmd=\pgfimage,xmin=-0.5, xmax=31.5, ymin=31.5, ymax=-0.5] {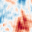};

\nextgroupplot[
axis line style={darkgray176},
colorbar,
colorbar style={ylabel={}},
colormap={mymap}{[1pt]
  rgb(0pt)=(0.403921568627451,0,0.12156862745098);
  rgb(1pt)=(0.698039215686274,0.0941176470588235,0.168627450980392);
  rgb(2pt)=(0.83921568627451,0.376470588235294,0.301960784313725);
  rgb(3pt)=(0.956862745098039,0.647058823529412,0.509803921568627);
  rgb(4pt)=(0.992156862745098,0.858823529411765,0.780392156862745);
  rgb(5pt)=(0.968627450980392,0.968627450980392,0.968627450980392);
  rgb(6pt)=(0.819607843137255,0.898039215686275,0.941176470588235);
  rgb(7pt)=(0.572549019607843,0.772549019607843,0.870588235294118);
  rgb(8pt)=(0.262745098039216,0.576470588235294,0.764705882352941);
  rgb(9pt)=(0.129411764705882,0.4,0.674509803921569);
  rgb(10pt)=(0.0196078431372549,0.188235294117647,0.380392156862745)
},
hide x axis,
hide y axis,
point meta max=0.025903497208879,
point meta min=-0.025903497208879,
tick align=outside,
tick pos=left,
title={$D_z$: yz-plane},
x grid style={darkgray176},
xmin=-0.5, xmax=31.5,
xtick style={color=darkgray176},
y dir=reverse,
y grid style={darkgray176},
ymajorgrids,
ymin=-0.5, ymax=31.5,
ytick style={color=darkgray176}
]
\addplot graphics [includegraphics cmd=\pgfimage,xmin=-0.5, xmax=31.5, ymin=31.5, ymax=-0.5] {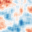};

\nextgroupplot[
axis line style={darkgray176},
hide x axis,
hide y axis,
tick align=outside,
tick pos=left,
title={$\mH_x$: yz-plane},
x grid style={darkgray176},
xmin=-0.5, xmax=31.5,
xtick style={color=darkgray176},
y dir=reverse,
y grid style={darkgray176},
ymajorgrids,
ymin=-0.5, ymax=31.5,
ytick style={color=darkgray176}
]
\addplot graphics [includegraphics cmd=\pgfimage,xmin=-0.5, xmax=31.5, ymin=31.5, ymax=-0.5] {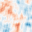};

\nextgroupplot[
axis line style={darkgray176},
hide x axis,
hide y axis,
tick align=outside,
tick pos=left,
title={$H_y$: yz-plane},
x grid style={darkgray176},
xmin=-0.5, xmax=31.5,
xtick style={color=darkgray176},
y dir=reverse,
y grid style={darkgray176},
ymajorgrids,
ymin=-0.5, ymax=31.5,
ytick style={color=darkgray176}
]
\addplot graphics [includegraphics cmd=\pgfimage,xmin=-0.5, xmax=31.5, ymin=31.5, ymax=-0.5] {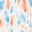};

\nextgroupplot[
axis line style={darkgray176},
colorbar,
colorbar style={ylabel={}},
colormap={mymap}{[1pt]
  rgb(0pt)=(0.403921568627451,0,0.12156862745098);
  rgb(1pt)=(0.698039215686274,0.0941176470588235,0.168627450980392);
  rgb(2pt)=(0.83921568627451,0.376470588235294,0.301960784313725);
  rgb(3pt)=(0.956862745098039,0.647058823529412,0.509803921568627);
  rgb(4pt)=(0.992156862745098,0.858823529411765,0.780392156862745);
  rgb(5pt)=(0.968627450980392,0.968627450980392,0.968627450980392);
  rgb(6pt)=(0.819607843137255,0.898039215686275,0.941176470588235);
  rgb(7pt)=(0.572549019607843,0.772549019607843,0.870588235294118);
  rgb(8pt)=(0.262745098039216,0.576470588235294,0.764705882352941);
  rgb(9pt)=(0.129411764705882,0.4,0.674509803921569);
  rgb(10pt)=(0.0196078431372549,0.188235294117647,0.380392156862745)
},
hide x axis,
hide y axis,
point meta max=0.0919556767587935,
point meta min=-0.0919556767587935,
tick align=outside,
tick pos=left,
title={$H_z$: yz-plane},
x grid style={darkgray176},
xmin=-0.5, xmax=31.5,
xtick style={color=darkgray176},
y dir=reverse,
y grid style={darkgray176},
ymajorgrids,
ymin=-0.5, ymax=31.5,
ytick style={color=darkgray176}
]
\addplot graphics [includegraphics cmd=\pgfimage,xmin=-0.5, xmax=31.5, ymin=31.5, ymax=-0.5] {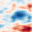};
\end{groupplot}

\end{tikzpicture}

%% file: tikz/maxwell/bar_fourier_1e-4_hist=2.tex
\begin{tikzpicture}

\definecolor{darkgray176}{RGB}{146,146,146}
\definecolor{darkslategray387083}{RGB}{38,70,83}
\definecolor{lightgray204}{RGB}{204,204,204}
\definecolor{tomato23111181}{RGB}{231,111,81}

\begin{groupplot}[group style={group size=2 by 2},height=0.24\textheight, width=0.52\textwidth, yticklabel style={color=darkgray176, font=\bfseries\scriptsize}, xticklabel style={color=darkgray176, font=\bfseries\scriptsize},]
\nextgroupplot[
axis line style={darkgray176},
legend cell align={left},
legend style={
  fill opacity=0.8,
  draw opacity=1,
  text=darkgray176,    
  text opacity=1,
  at={(0.03,0.97)},
  anchor=north west,
  draw=lightgray204,
  font=\scriptsize,  
},
tick align=outside,
tick pos=left,
title={Rollout},
x grid style={darkgray176},
xmajorgrids,
xmin=-0.5, xmax=3.5,
xtick style={color=darkgray176},
xticklabels={},
y grid style={darkgray176},
ymajorgrids,
ymin=0, ymax=0.019,
ytick style={color=darkgray176}
]
\draw[draw=none,fill=darkslategray387083,fill opacity=0.85] (axis cs:-0.25,0) rectangle (axis cs:0,0.000798066088464111);
\addlegendimage{ybar,ybar legend,draw=none,fill=darkslategray387083,fill opacity=0.85}
\addlegendentry{FNO}

\draw[draw=none,fill=darkslategray387083,fill opacity=0.85] (axis cs:0.75,0) rectangle (axis cs:1,0.00202110704655449);
\draw[draw=none,fill=darkslategray387083,fill opacity=0.85] (axis cs:1.75,0) rectangle (axis cs:2,0.00683706564207872);
\draw[draw=none,fill=darkslategray387083,fill opacity=0.85] (axis cs:2.75,0) rectangle (axis cs:3,0.0186062703529994);
\draw[draw=none,fill=tomato23111181,fill opacity=0.85] (axis cs:0,0) rectangle (axis cs:0.25,0.000729633262380958);
\addlegendimage{ybar,ybar legend,draw=none,fill=tomato23111181,fill opacity=0.85}
\addlegendentry{CFNO}

\draw[draw=none,fill=tomato23111181,fill opacity=0.85] (axis cs:1,0) rectangle (axis cs:1.25,0.00150580428695927);
\draw[draw=none,fill=tomato23111181,fill opacity=0.85] (axis cs:2,0) rectangle (axis cs:2.25,0.00292138056829572);
\draw[draw=none,fill=tomato23111181,fill opacity=0.85] (axis cs:3,0) rectangle (axis cs:3.25,0.00543286480630438);
\path [draw=darkgray176, semithick]
(axis cs:-0.125,0.000768150435760617)
--(axis cs:-0.125,0.000827981741167605);

\path [draw=darkgray176, semithick]
(axis cs:0.875,0.00194724577447932)
--(axis cs:0.875,0.00209496831862966);

\path [draw=darkgray176, semithick]
(axis cs:1.875,0.00667927902110079)
--(axis cs:1.875,0.00699485226305664);

\path [draw=darkgray176, semithick]
(axis cs:2.875,0.0184555451955739)
--(axis cs:2.875,0.0187569955104248);

\addplot [semithick, darkgray176, mark=-, mark size=2.5, mark options={solid}, only marks]
table {%
-0.125 0.000768150435760617
0.875 0.00194724577447932
1.875 0.00667927902110079
2.875 0.0184555451955739
};

\addplot [semithick, darkgray176, mark=-, mark size=2.5, mark options={solid}, only marks]
table {%
-0.125 0.000827981741167605
0.875 0.00209496831862966
1.875 0.00699485226305664
2.875 0.0187569955104248
};

\path [draw=darkgray176, semithick]
(axis cs:0.125,0.000703814090229571)
--(axis cs:0.125,0.000755452434532344);

\path [draw=darkgray176, semithick]
(axis cs:1.125,0.00136982230374258)
--(axis cs:1.125,0.00164178627017596);

\path [draw=darkgray176, semithick]
(axis cs:2.125,0.00273767498107032)
--(axis cs:2.125,0.00310508615552112);

\path [draw=darkgray176, semithick]
(axis cs:3.125,0.00503351547625453)
--(axis cs:3.125,0.00583221413635422);

\addplot [semithick, darkgray176, mark=-, mark size=2.5, mark options={solid}, only marks]
table {%
0.125 0.000703814090229571
1.125 0.00136982230374258
2.125 0.00273767498107032
3.125 0.00503351547625453
};

\addplot [semithick, darkgray176, mark=-, mark size=2.5, mark options={solid}, only marks]
table {%
0.125 0.000755452434532344
1.125 0.00164178627017596
2.125 0.00310508615552112
3.125 0.00583221413635422
};

\nextgroupplot[
axis line style={darkgray176},
tick align=outside,
tick pos=left,
title={One-step},
x grid style={darkgray176},
xmajorgrids,
xmin=-0.5, xmax=3.5,
xtick style={color=darkgray176},
xticklabels={},
y grid style={darkgray176},
ymajorgrids,
ymin=0, ymax=0.00554542343454997,
ytick style={color=darkgray176}
]
\draw[draw=none,fill=darkslategray387083,fill opacity=0.85] (axis cs:-0.25,0) rectangle (axis cs:0,0.00018444857414579);
\draw[draw=none,fill=darkslategray387083,fill opacity=0.85] (axis cs:0.75,0) rectangle (axis cs:1,0.000504025966317082);
\draw[draw=none,fill=darkslategray387083,fill opacity=0.85] (axis cs:1.75,0) rectangle (axis cs:2,0.00187148499147346);
\draw[draw=none,fill=darkslategray387083,fill opacity=0.85] (axis cs:2.75,0) rectangle (axis cs:3,0.0053920663582782);
\draw[draw=none,fill=tomato23111181,fill opacity=0.85] (axis cs:0,0) rectangle (axis cs:0.25,0.00016511094145244);
\draw[draw=none,fill=tomato23111181,fill opacity=0.85] (axis cs:1,0) rectangle (axis cs:1.25,0.000359014433342963);
\draw[draw=none,fill=tomato23111181,fill opacity=0.85] (axis cs:2,0) rectangle (axis cs:2.25,0.000721845911660542);
\draw[draw=none,fill=tomato23111181,fill opacity=0.85] (axis cs:3,0) rectangle (axis cs:3.25,0.00128668829953919);
\path [draw=darkgray176, semithick]
(axis cs:-0.125,0.000177403169800527)
--(axis cs:-0.125,0.000191493978491053);

\path [draw=darkgray176, semithick]
(axis cs:0.875,0.00047348665450292)
--(axis cs:0.875,0.000534565278131245);

\path [draw=darkgray176, semithick]
(axis cs:1.875,0.00181122968849896)
--(axis cs:1.875,0.00193174029444795);

\path [draw=darkgray176, semithick]
(axis cs:2.875,0.00523870928200642)
--(axis cs:2.875,0.00554542343454997);

\addplot [semithick, darkgray176, mark=-, mark size=2.5, mark options={solid}, only marks]
table {%
-0.125 0.000177403169800527
0.875 0.00047348665450292
1.875 0.00181122968849896
2.875 0.00523870928200642
};
\addplot [semithick, darkgray176, mark=-, mark size=2.5, mark options={solid}, only marks]
table {%
-0.125 0.000191493978491053
0.875 0.000534565278131245
1.875 0.00193174029444795
2.875 0.00554542343454997
};
\path [draw=darkgray176, semithick]
(axis cs:0.125,0.000159301547682844)
--(axis cs:0.125,0.000170920335222036);

\path [draw=darkgray176, semithick]
(axis cs:1.125,0.000309258094822485)
--(axis cs:1.125,0.000408770771863442);

\path [draw=darkgray176, semithick]
(axis cs:2.125,0.00065093732831237)
--(axis cs:2.125,0.000792754495008715);

\path [draw=darkgray176, semithick]
(axis cs:3.125,0.00113897136082729)
--(axis cs:3.125,0.00143440523825109);

\addplot [semithick, darkgray176, mark=-, mark size=2.5, mark options={solid}, only marks]
table {%
0.125 0.000159301547682844
1.125 0.000309258094822485
2.125 0.00065093732831237
3.125 0.00113897136082729
};
\addplot [semithick, darkgray176, mark=-, mark size=2.5, mark options={solid}, only marks]
table {%
0.125 0.000170920335222036
1.125 0.000408770771863442
2.125 0.000792754495008715
3.125 0.00143440523825109
};

\nextgroupplot[
axis line style={darkgray176},
tick align=outside,
tick pos=left,
title={$D$},
x grid style={darkgray176},
xmajorgrids,
xmin=-0.5, xmax=3.5,
xtick style={color=darkgray176},
xtick={0,1,2,3},
xticklabels={6400,3200,1280,640},
y grid style={darkgray176},
ymajorgrids,
ymin=0, ymax=0.00314842976348749,
ytick style={color=darkgray176}
]
\draw[draw=none,fill=darkslategray387083,fill opacity=0.85] (axis cs:-0.25,0) rectangle (axis cs:0,9.62285557761788e-05);
\draw[draw=none,fill=darkslategray387083,fill opacity=0.85] (axis cs:0.75,0) rectangle (axis cs:1,0.000254062974515061);
\draw[draw=none,fill=darkslategray387083,fill opacity=0.85] (axis cs:1.75,0) rectangle (axis cs:2,0.00101505417842418);
\draw[draw=none,fill=darkslategray387083,fill opacity=0.85] (axis cs:2.75,0) rectangle (axis cs:3,0.00302958597118656);
\draw[draw=none,fill=tomato23111181,fill opacity=0.85] (axis cs:0,0) rectangle (axis cs:0.25,7.78766952862497e-05);
\draw[draw=none,fill=tomato23111181,fill opacity=0.85] (axis cs:1,0) rectangle (axis cs:1.25,0.000158705661306158);
\draw[draw=none,fill=tomato23111181,fill opacity=0.85] (axis cs:2,0) rectangle (axis cs:2.25,0.000306312266426782);
\draw[draw=none,fill=tomato23111181,fill opacity=0.85] (axis cs:3,0) rectangle (axis cs:3.25,0.000561348395422101);
\path [draw=darkgray176, semithick]
(axis cs:-0.125,9.26291977521032e-05)
--(axis cs:-0.125,9.98279138002545e-05);

\path [draw=darkgray176, semithick]
(axis cs:0.875,0.000246864331170557)
--(axis cs:0.875,0.000261261617859565);

\path [draw=darkgray176, semithick]
(axis cs:1.875,0.000994028559029631)
--(axis cs:1.875,0.00103607979781873);

\path [draw=darkgray176, semithick]
(axis cs:2.875,0.00291074217888563)
--(axis cs:2.875,0.00314842976348749);

\addplot [semithick, darkgray176, mark=-, mark size=2.5, mark options={solid}, only marks]
table {%
-0.125 9.26291977521032e-05
0.875 0.000246864331170557
1.875 0.000994028559029631
2.875 0.00291074217888563
};
\addplot [semithick, darkgray176, mark=-, mark size=2.5, mark options={solid}, only marks]
table {%
-0.125 9.98279138002545e-05
0.875 0.000261261617859565
1.875 0.00103607979781873
2.875 0.00314842976348749
};
\path [draw=darkgray176, semithick]
(axis cs:0.125,7.02799952705391e-05)
--(axis cs:0.125,8.54733953019604e-05);

\path [draw=darkgray176, semithick]
(axis cs:1.125,0.000148629646322856)
--(axis cs:1.125,0.00016878167628946);

\path [draw=darkgray176, semithick]
(axis cs:2.125,0.000303448705720059)
--(axis cs:2.125,0.000309175827133505);

\path [draw=darkgray176, semithick]
(axis cs:3.125,0.000530372339273714)
--(axis cs:3.125,0.000592324451570488);

\addplot [semithick, darkgray176, mark=-, mark size=2.5, mark options={solid}, only marks]
table {%
0.125 7.02799952705391e-05
1.125 0.000148629646322856
2.125 0.000303448705720059
3.125 0.000530372339273714
};
\addplot [semithick, darkgray176, mark=-, mark size=2.5, mark options={solid}, only marks]
table {%
0.125 8.54733953019604e-05
1.125 0.00016878167628946
2.125 0.000309175827133505
3.125 0.000592324451570488
};

\nextgroupplot[
axis line style={darkgray176},
tick align=outside,
tick pos=left,
title={$H$ },
x grid style={darkgray176},
xmajorgrids,
xmin=-0.5, xmax=3.5,
xtick style={color=darkgray176},
xtick={0,1,2,3},
xticklabels={6400,3200,1280,640},
y grid style={darkgray176},
ymajorgrids,
ymin=0, ymax=0.00237038152002518,
ytick style={color=darkgray176}
]
\draw[draw=none,fill=darkslategray387083,fill opacity=0.85] (axis cs:-0.25,0) rectangle (axis cs:0,9.15731034183409e-05);
\draw[draw=none,fill=darkslategray387083,fill opacity=0.85] (axis cs:0.75,0) rectangle (axis cs:1,0.000247196158549438);
\draw[draw=none,fill=darkslategray387083,fill opacity=0.85] (axis cs:1.75,0) rectangle (axis cs:2,0.000847450263487796);
\draw[draw=none,fill=darkslategray387083,fill opacity=0.85] (axis cs:2.75,0) rectangle (axis cs:3,0.00233744333187739);
\draw[draw=none,fill=tomato23111181,fill opacity=0.85] (axis cs:0,0) rectangle (axis cs:0.25,9.11770039238036e-05);
\draw[draw=none,fill=tomato23111181,fill opacity=0.85] (axis cs:1,0) rectangle (axis cs:1.25,0.000197908676151807);
\draw[draw=none,fill=tomato23111181,fill opacity=0.85] (axis cs:2,0) rectangle (axis cs:2.25,0.000413191960736488);
\draw[draw=none,fill=tomato23111181,fill opacity=0.85] (axis cs:3,0) rectangle (axis cs:3.25,0.000718661292921752);
\path [draw=darkgray176, semithick]
(axis cs:-0.125,8.329791307915e-05)
--(axis cs:-0.125,9.98482937575318e-05);

\path [draw=darkgray176, semithick]
(axis cs:0.875,0.000226104151179159)
--(axis cs:0.875,0.000268288165919717);

\path [draw=darkgray176, semithick]
(axis cs:1.875,0.000815372121325113)
--(axis cs:1.875,0.000879528405650479);

\path [draw=darkgray176, semithick]
(axis cs:2.875,0.0023045051437296)
--(axis cs:2.875,0.00237038152002518);

\addplot [semithick, darkgray176, mark=-, mark size=2.5, mark options={solid}, only marks]
table {%
-0.125 8.329791307915e-05
0.875 0.000226104151179159
1.875 0.000815372121325113
2.875 0.0023045051437296
};
\addplot [semithick, darkgray176, mark=-, mark size=2.5, mark options={solid}, only marks]
table {%
-0.125 9.98482937575318e-05
0.875 0.000268288165919717
1.875 0.000879528405650479
2.875 0.00237038152002518
};
\path [draw=darkgray176, semithick]
(axis cs:0.125,9.07674548216164e-05)
--(axis cs:0.125,9.15865530259907e-05);

\path [draw=darkgray176, semithick]
(axis cs:1.125,0.000167447471434703)
--(axis cs:1.125,0.000228369880868911);

\path [draw=darkgray176, semithick]
(axis cs:2.125,0.000363675421156484)
--(axis cs:2.125,0.000462708500316493);

\path [draw=darkgray176, semithick]
(axis cs:3.125,0.000625028568316928)
--(axis cs:3.125,0.000812294017526576);

\addplot [semithick, darkgray176, mark=-, mark size=2.5, mark options={solid}, only marks]
table {%
0.125 9.07674548216164e-05
1.125 0.000167447471434703
2.125 0.000363675421156484
3.125 0.000625028568316928
};
\addplot [semithick, darkgray176, mark=-, mark size=2.5, mark options={solid}, only marks]
table {%
0.125 9.15865530259907e-05
1.125 0.000228369880868911
2.125 0.000462708500316493
3.125 0.000812294017526576
};
\end{groupplot}

\draw ({$(current bounding box.south west)!-0.02!(current bounding box.south east)$}|-{$(current bounding box.south west)!0.5!(current bounding box.north west)$}) node[
  scale=0.9,
  anchor=west,
  text=darkgray176,
  rotate=90.0
]{MSE};
\draw ({$(current bounding box.south west)!0.5!(current bounding box.south east)$}|-{$(current bounding box.south west)!-0.05!(current bounding box.north west)$}) node[
  scale=0.9,
  anchor=south,
  text=darkgray176,
  rotate=0.0
]{Num. Train Trajectories};
\end{tikzpicture}

%% file: appendix/relatedwork.tex
\section{Related work}\label{app:relatedwork}
This appendix supports detailed discussions of how our work relates to complex and quaternion neural networks, to work on Clifford algebras and Clifford Fourier transforms in computer vision, to Fourier Neural Operators, equivariant neural networks and geometric deep learning approaches, to neural operator learning and neural PDE surrogates.

\paragraph{Clifford (geometric) algebra and Clifford Fourier transform.}
(Real) Clifford algebras (also known as geometric algebras), as an extension of elementary algebra to work with geometrical objects such as vectors are extensively discussed in \citet{suter2003geometric, hestenes2003oersted, dorst2010geometric, hestenes2012new, renaud2020clifford}. Compared to other formalisms for manipulating geometric objects, Clifford algebras are tailored towards vector manipulation of objects of different dimensions. Hypercomplex and quaternion Fourier transforms are extensively discussed in~\citet{ell1992hypercomplex, ell1993quaternion, ell2006hypercomplex, ell2014quaternion}. This work heavily builds on the concepts introduced in~\citet{ebling2003clifford, ebling2005, ebling2006visualization}. Comprehensive summaries of Clifford and quaternion Fourier transforms can be found in~\citet{hitzer2013quaternion, brackx2013history, hitzer2021quaternion}. Clifford algebras and Clifford Fourier transforms are already deployed to solve PDEs numerically in~\citet{alfarraj2022geometric}. More precisely, the Clifford-Fourier transform is used to solve the mode decomposition process in PDE transforms.

\paragraph{Clifford neural networks.}
Neural networks in the Clifford domain were proposed already in 1994 by~\citet{pearson1994neural},  and later by~\citet{pearson2003clifford}. These works put the emphasis on the geometric perceptron~\citep{melnyk2021embed}, i.e. how to recast vanilla multilayer perceptrons (MLPs) as Clifford MLPs. Similarly, \citep{hoffmann2020algebranets} generalized from complex numbers and quaternions to a set of alternative algebras. Besides Clifford MLPs, Clifford algebras have been used in recurrent neural networks (RNNs)~\citep{kuroe2011models}, and have been used to formulate quantum neural networks~\citep{trindade2022clifford}. Their applicability to neural computing has been studied in general~\citep{buchholz2001introduction, buchholz2005theory}, exploring global exponential stabilities of Clifford MLPs with time-varying delays and impulsive effects.
Probably the most related wors are: (i) ~\citet{zang2022multi} who build geometric algebra convolution networks to process spatial and temporal data of 3D traffic data.
Multidimensional traffic parameters are encoded as multivectors which allows to model correlation between traffic data in both spatial and temporal domains. (ii)~\citet{spellings2021geometric} who build rotation- and permutation-equivariant graph network architectures based on geometric algebra products of node features. Higher order information is built from available node inputs.

In contrast to previous works, we are the first to introduce the multivector viewpoint of field components which allows us to effectively connect Clifford neural layers with the geometric structure of the input data.
We further connect neural Clifford convolutions on multivectors with various works on complex numbers and quaternions. We are further the first to introduce neural Clifford Fourier transforms.

\paragraph{Complex and quaternion neural networks.}
\citet{Trabelsi2018DeepCN} introduced the key components for complex-valued deep neural networks. More precisely, they introduced convolutional~\citep{lecun1998gradient} feed-forward and convolutional LSTM~\citep{shi2015convolutional, hochreiter1997long} networks, together with
complex batch-normalization, and complex weight initialization strategies.
Quaternions are a natural extension of complex neural networks. Already in classical computer vision, quaternions as hypercomplex convolution~\citep{sangwine2000colour} and hypercomplex correlation~\citep{moxey2003hypercomplex} techniques were introduced for color image processing.
Quaternion based deep learning architectures are a natural extension of complex neural networks.
In quaternion neural networks~\citep{Zhu_2018_ECCV, parcollet2018quaternion, gaudet2018deep, parcollet2018quaternionconv, parcollet2018quaternionconv2, parcollet2020survey, nguyen2021quaternion, moya2021trainable}, concepts such as complex convolution, complex batchnorm, and complex initialization are transfered from the complex numbers $\CC$, which are algebra-isomorph to $Cl(0,1)(\R)$
to $Cl(0,2)(\R)$, which is algebra-isomorph to the quaternions $\HH$.
Although \citet{hoffmann2020algebranets} generalized these from complex numbers and quaternions to a set of alternative algebras, their tasks did not really leverage any multivector structure in data.

\paragraph{Fourier Neural Operators.}
Fourier Neural Operators (FNOs)~\citep{li2020fourier}
have had tremendous impact towards improving neural PDE solver surrogates. Efficient implementations of FNO layers come as physics-informed neural networks (PINO)~\citep{li2021physics}, as U-shaped network architectures (UNO)~\citep{rahman2022u}, as spectral surrogate for vision transformer architectures~\citep{rao2021global, guibas2021adaptive}, as Markov neural operators (MNO) for chaotic systems (MNO)~\citep{li2021markov}, and as generative adversarial neural operators (GANOs)~\citep{rahman2022generative}. Applications range from weather forecasting~\citep{pathak2022fourcastnet}, $\text{CO}_2$-water multiphase problems~\citep{wen2022u}, multiscale method for crystal plasticity~\citep{liu2022learning}, seismic wave propagation~\citep{yang2021seismic}, photoaccustic wave propagation~\citep{guan2021fourier}, PDE-constrained control problems~\citep{hwang2022solving}, and for thermochemical curing of composites~\citep{chen2021residual}. Recently, FNOs have been successfully applied to PDEs on general geometries~\citep{li2022fourier}. Furthermore, universal approximation and error bounds have been studied for FNOs~\citep{kovachki2021universal}.

\paragraph{Neural PDE solvers/surrogates.}
The intersection of PDE solving, deep learning, fluid dynamics, and weather forecasting has developed into a very active hub of research lately~\citep{thuerey2021physics}.
We roughly group recent approaches to learn neural PDE surrogates and neural PDE solvers into three categories:
\begin{enumerate}[label=(\roman*)]
    \item \emph{hybrid approaches}, where neural networks augment numerical solvers or replace parts of numerical solvers;  
    \item \emph{direct approaches},
    \begin{enumerate}[label=(\alph*)]
    \item where the mapping from an initial state to a solution is learned, i.e. the solution function of the underlying PDE is approximated;
    \item where the mapping from an initial state to a final state of an underlying PDE, i.e. the solution operator is learned, ideally as mapping between function spaces to be able to generalize across e.g. parameters.
    \end{enumerate}
\end{enumerate}

Ad (i): Neural networks augment numerical solvers by learning data-driven discretizations for PDEs~\citep{bar2019learning} or by controlling learned approximations inside the calculation of standard numerical solver used for computational fluid dynamics~\citep{kochkov2021machine}.
In \citet{GreenfeldGBYK19}, a prolongation is learned which maps from discretized PDE solutions to multigrid solutions, \citet{HsiehZEME19} learn to modify the updates of an existing solver, \citet{praditia2021finite} adopt the numerical structure of Finite Volume Methods (FVMs), and~\citet{um2020solver} learn a correction function of conventional PDE solvers to improve accuracy. All these approaches are hybrid approaches~\citep{GarciaAW19}, where the computational graph of the solver is preserved and heuristically-chosen parameters are predicted with a neural network. 
A different flavor of hybrid approaches can be assigned to the works of~\citet{sanchez2021simulate, pfaff2020learning, mayr2021boundary} who predict accelerations of particles/meshes to numerical update the respective positions. Finally, PauliNet~\citep{hermann2020deep} and FermiNet~\citep{pfau2020ab} approximate wave-functions of many-electron systems, and thus replace the hand-crafted ansatz which is conventionally used in variational quantum Monte Carlo methods.

Ad (ii.a): \citet{sirignano2018dgm, han2018solving} approximate the solution of high-dimensional Black-Scholes and Hamilton-Jacobi-Bellman equations, respectively. Physics-informed neural networks (PINNs)~\citep{raissi2019physics} embed the underlying physics in the training process, and can be used to solve both forward~\citep{jin2021nsfnets} as well as backward~\citep{raissi2020hidden} dynamics. \citet{zubov2021neuralpde} allow automating many of these aspects under a single coherent interface.

Ad (ii.b): \citet{guo2016convolutional} learned a surrogate CNN-based model to approximate steady-state flow field predictions, similarly~\citet{bhatnagar2019prediction} trained a surrogate CNN-based model to predict solutions for unseen flow conditions and geometries, and~\citet{zhu2018bayesian} used Baysian CNNs for surrogate PDE modeling and uncertainty quantification. Fourier Neural Operators (FNOs)~\citep{li2020fourier}
proposed the mapping from parameter space to solution spaces, and had tremendous impact towards improving neural PDE solver surrogates. In parallel, \citet{lu2021learning} introduced DeepONet, which learns mappings between function spaces, and was successfully applied to many parametric ODEs and PDEs. Both, FNOs and DeepONets have been combined with PINNs and trained in a physics-informed style~\citep{li2022fourier, wang2021learning}. A comprehensive comparison of these two neural operator approaches is done by~\citet{lu2022comprehensive}. 
Other directions include the modeling of PDE solution operators via latent space models, transformers, and graph neural networks (GNNs). 
\citet{wu2022learning} present the modeling of the systems dynamics in a latent space with fixed dimension where the latent modeling is done via MLPs, and the encoding and decoding via CNNs, which can also be replaced by graph neural networks (GNNs). \citet{cao2021choose} propose the Galerkin transformer, a simple attention based operator learning method without softmax normalization, LOCA (Learning Operators with Coupled Attention)~\citep{kissas2022learning} maps the input functions to a finite set of features and attends to them by output query locations, and \citet{Li2022TransformerFP} propose a transformer which provides a flexible way to implicitly exploit the patterns within inputs. \citet{brandstetter2022message} formulated a message passing neural network approach that representationally contains several conventional numerical PDE solving schemes.
Further GNN based approaches are~\citet{lotzsch2022learning} who learn the operator for boundary value problems on finite element method (FEM)~\citep{brenner2008mathematical} ground truth data, and~\citet{lienen2022learning} who derive their GNN models from 
FEM in a principled way.

A practical use case for neural PDE surrogates is replacing expensive classical PDE solvers. 
There is however a major chicken-and-egg problem here~\citep{brandstetter2022lie, shi2022lordnet}: obtaining high quality ground truth training data for neural PDE surrogates often requires using these expensive solvers.
Minimizing this data requirement is beginning to be approached in recent works.
\citet{geneva2020modeling, wandel2020learning, wandel2022spline} achieve ``data-free'' training in various settings.
``Data-free'' refers to the self-supervised training steps, which are done without ground truth data.
The current state-of-art generic approach is introduced in~\citet{shi2022lordnet} as the mean squared residual (MSR) loss constructed by the discretized PDE itself.
However, for e.g. generating realistic initial conditions numerical solvers are still needed. 
\citet{pestourie2021physics} identify how incorporating limited physical knowledge in the form of a low-fidelity ``coarse'' solver can allow training PDE surrogate models with an order of magnitude less data.
Another direction to improve data efficiency is by exploiting the Lie point symmetries of the underlying PDEs, either via data augmentation~\citep{brandstetter2022lie} or by building equivariant PDE surrogates~\citep{wang2020incorporating}.
Our current work in a way also improves data efficiency by capturing the inductive bias appropriate for multivector fields.
Overall we believe hybrids of such approaches are going to be necessary for making neural PDE surrogates of practical use in many domains.

Neural PDE surrogates for fluid flow and weather forecasting applications are gaining momentum. In weather forecasting, \citet{pathak2022fourcastnet} introduced FourCastNet
as high-resolution weather modeling built on Adaptive Fourier Neural Operators \citep{guibas2021adaptive}, \citet{keisler2022forecasting} successfully applied a graph neural network based approach to weather forecasting, \citet{rasp2021data} achieved data-driven medium-range weather prediction with a ResNet which was pretrained on climate simulations, \citet{weyn2020improving}
use CNNs on a cubed sphere for global weather prediction, \citet{weyn2021sub} forecast weather sub-seasonally with a large ensemble of deep-learning weather prediction models, \citet{arcomano2020machine} build a reservoir computing-based, low-resolution, global prediction model, and MetNet~\citep{sonderby2020metnet} takes as input radar and satellite data to forecast probabilistic precipitation maps. 
Finally, data assimilation is improved by deep learning techniques in~\citet{frerix2021variational} and~\citet{maulik2022efficient}.
Similarly, in fluid dynamics, \citet{ma2021physics} applied U-Nets~\citep{ronneberger2015u} to achieve physics-driven learning of steady Navier-Stokes equations, \citet{stachenfeld2021learned} learned coarse models for turbulence simulations, TF-Net~\citep{wang2020towards} introduced domain-specific variations of U-Nets along with trainable spectral filters in a coupled model of Reynolds-averaged
Navier-Stokes and Large Eddy Simulation.

This exhaustive list of neural PDE solver surrogates shows that many of the architectures are based on convolutional or Fourier layers. For these two, Clifford layers are applicable as a drop-in replacement in almost all cases.
For graph neural network and attention based architectures, we leave the implementation of respective Clifford counterparts to future work.

\paragraph{Geometric deep learning.}
The core idea of geometric deep learning~\citep{bronstein2017geometric, bronstein2021geometric} is to exploit underlying low-dimensionality
and structure of the physical world, in order to design deep learning models which can better learn in high dimensional spaces. 
Incorporating underlying symmetries would be one way to achieve this.
If done correctly, it can drastically shrink the search space, which has proven to be quite successful in multiple scenarios.
The most obvious examples are CNNs~\citep{fukushima1982neocognitron, lecun1998gradient}, where the convolution operation commutes with the shift operator, and thus provides a way to equip layers and subsequently networks with translation equivariant operations.
Group convolution networks~\citep{cohen2016group, kondor2018generalization, cohen2019general} generalize equivariant layers beyond translations, i.e. provide a concept of how to build general layers that are equivariant to a broader range of groups, such as rotation groups. An appealing way of how to build such group equivariant layers is via so-called steerable basis functions~\citep{hel1998canonical}, which allow to write transformation by specific groups as a linear combination of a fixed, finite set of basis functions. This concept leads to steerable group convolution approaches~\citep{cohen2016steerable, worrall2017harmonic}. Two concrete examples are: (i) circular harmonics, which are respective basis functions for building layers that are equivariant to the group SO($2$), the rotation group in $2$ dimensions~\citep{worrall2017harmonic, weiler2019general}; (ii) spherical harmonics, which are respective basis functions for building layers that are equivariant to the group SO($3$), the rotation group in 3 dimensions~\citep{weiler20183d, geiger2022e3nn, brandstetter2021geometric}. The similarity to multivector fields becomes more obvious if we have a closer look at spherical harmonics, which are defined as homogeneous polynomials of degree $l$, where the $l=0$ case corresponds to scalars, the $l=1$ case to vectors, and $l\geq2$ to higher order objects. Finally, \citet{jenner2021steerable} built steerable PDE operators such as curl or divergence as equivariant neural network components.

\paragraph{Grouped convolution.}
In their seminal work, \citet{krizhevsky2012imagenet} introduced filter grouping, which allowed them to reduce the parameters in CNNs. The respective grouped convolutions (not to be confused with \textit{group} convolutions) divide the filter maps at channel dimension, as the channel dimension most of the time increases strongly for deeper layers, and thus dominates the parameter count. Subsequent work showed that it is beneficial to additionally shuffle the channels for each filter group~\citep{zhang2018shufflenet}, and to adaptively recalibrate channel-wise feature responses~\citep{hu2018squeeze}. All these approaches can be seen in the wider spectrum of effective model scaling~\citep{tan2019efficientnet, sandler2018mobilenetv2}.

Clifford convolutions in contrast do not have groupings in the channel dimensions, but instead group together elements as multivectors. In Clifford convolution, the Clifford kernel is therefore a constrained object where weight blocks appear multiple times (due to the nature of the geometric product). Thus, Clifford convolutions are more parameter efficient than standard convolutions, and all tricks of effective model scaling could in principle be applied on top of Clifford convolutions. Findings from \citet{hoffmann2020algebranets} with respect to higher compute density of alternative algebras are applicable to our work as well.

%% file: appendix/glossary.tex
\section{Glossary}
This short appendix summarizes notations used throughout the paper (Table~\ref{tab:app_notation_summary}), and contrasts the most fundamental concepts which arise when using Clifford algebras. 

\begin{table}[!htb]
    \centering
    \caption{Notations used throughout the paper.}
    \label{tab:app_notation_summary}
    \begin{tabular}{ll}
    \toprule
        \textbf{Notation} &  \textbf{Meaning} \\
    \midrule
         $e_1$, $e_2$, $e_3$  & Basis vectors of the \textit{generating} vector space of the Clifford algebra. \\
         $e_{i} \wedge e_{j}$ & Wedge (outer) product of basis vectors $e_{i}$ and $e_{j}$. \\
         $e_{i} \cdot e_{j} = \langle e_1, e_j \rangle$ & Inner product of basis vectors $e_{i}$ and $e_{j}$. \\
         $e_1e_2$, $e_3e_1$, $e_2e_3$  & Basis bivectors of the vector space of the Clifford algebra. \\
         $e_1e_2e_3$  & Basis trivector of the vector space of the Clifford algebra. \\
         $i_2=e_1e_2$ & Pseudoscalar for Clifford algebras of grade 2. \\
         $i_3=e_1e_2e_3$ & Pseudoscalar for Clifford algebras of grade 3. \\
         $x$ & Euclidean vector $\in \R^n$. \\
         $x \wedge y$ & wedge (outer) product of Euclidean vectors $x$ and $y$. \\
         $x \cdot y = \langle x,y \rangle$ & Inner product of vectors $x$ and $y$. \\
         $\va$ & Multivector. \\
         $\va\vb$ & Geometric product of multivectors $\va$ and $\vb$. \\
         $ \hat{\imath}, \hat{\jmath}, \hat{k}$ & Base elements of quaternions. \\
    \bottomrule
    \end{tabular}
\end{table}

\paragraph{Geometric, Exterior, and Clifford algebras.}
A geometric algebra is a Clifford algebra of the real numbers. Since we are only using $Cl_{2,0}(\R)$, $Cl_{0,2}(\R)$, and $Cl_{3,0}(\R)$, we are effectively working with geometric algebras.
The exterior or Grassmann algebra is built up from the same concepts of scalars, vectors, bivectors, \ldots, $k$-vectors, but only exterior (wedge) products exist. Therefore, the exterior algebra has a zero quadratic form (all base vectors square to zero). Clifford algebras are a generalization thereof with nonzero quadratic forms.

\paragraph{Complex numbers, quaternions, hypercomplex numbers.}
Hypercomplex numbers are elements of finite-dimensional algebras over the real numbers that are unital, i.e. contain a multiplicative identity element, but not necessarily associative or commutative. Elements are generated for a basis $\{\hat{\imath}, 
\hat{\jmath}, \ldots\}$ such that $\hat{\imath}^2, \hat{\jmath}^2, \ldots \in \{-1,0,1\}$. Complex numbers, quaternions, octonions are all hypercomplex numbers which can be characterized by different Clifford algebras. The bivector, trivector (and higher objects) of the Clifford algebras directly translate into basis elements of the respective algebras. For example, quaternions (which are of the form $a + b\hat{\imath} + c\hat{\jmath} + d\hat{k}$, where $\hat{\imath}^2=\hat{\jmath}^2=\hat{k}^2=-1$) are isomorphic to the Clifford algebra $Cl_{0,2}(\R)$ where the basis element $e_1$, $e_2$, and $e_1e_2$ directly translate to $\hat{\imath}$, $\hat{\jmath}$, $\hat{k}$. 

\paragraph{Spinor.}
Spinors arise naturally in discussions of the Lorentz group,
the group to describe transformations in special relativity.
One could say that a spinor is the most basic sort of
mathematical object that can be Lorentz-transformed.
In its essence, a spinor is a complex two-component vector-like quantity in which rotations and Lorentz boosts (relativistic translations) are built into the overall formalism. 
More generally, spinors are elements of complex vector spaces
that can be associated with Euclidean vector spaces. However, unlike vectors, spinors transform to their negative when the space is rotated by $360\degree$.
In this work, the subalgebra $Cl^0(2,0)(\R)$, spanned by even-graded basis elements of $Cl_{2,0}(\R)$, i.e. $1$ and $e_1e_2$, determines the space of spinors via linear combinations of $1$ and $e_1e_2$. It is thus isomorphic to the field of complex numbers $\CC$. Most notably, spinors of $Cl_{2,0}(\R)$ commute with the Fourier kernel, whereas vectors do not. For a detailed introduction to spinors we recommend~\citet{steane2013introduction}, and the comprehensive physics book of~\citet{schwichtenberg2015physics}.

\paragraph{Pseudoscalar.}
A pseudoscalar -- unlike a scalar -- changes sign when you invert the coordinate axis.
The easiest example of a pseudoscalar is the scalar triplet product of three arbitrary vectors of an Euclidean vector space $x,y,z \in \R^n$ with inner product $\langle .,. \rangle$. The scalar triplet product becomes negative for any parity inversion, i.e. swapping any two of the three operands: $x \cdot (y \times z) = -x \cdot (z \times y) = - y \cdot (x \times z) = - z \cdot (y \times x)$.

\paragraph{Scalar field, vector field.}
A \emph{field} is any (physical) quantity which takes on different values at different points in space (space-time). A scalar field is map $\mathbb{D} \rightarrow \R$, where $\mathbb{D} \subseteq \R^n$.
A vector field is map $\mathbb{D} \rightarrow \R^n$, where $\mathbb{D} \subseteq \R^n$. For example, $n=2$ results in a vector field in plane, and $n=3$ in a vector field in space.
For an interesting history of the evolution of the concept of fields in physics we recommend~\citet{mirowski1991more,mcmullin2002origins}. In Table~\ref{tab:app_various_fields}, we list various important vector and scalar fields for comparison.

\begin{table}[!htb]
    \centering
    \caption{Examples of various vector and scalar fields. Vector fields ascribe a vector to each point in space, e.g. force, electric current (stream of charged particles), or velocity. Scalar fields on the other hand collate each field point with a scalar value such as temperature. }
    \label{tab:app_various_fields}
    \begin{threeparttable}[t]
    \centering
    \begin{tabular}{llll}
        \toprule
        Example & Field quantity & Type & Coordinates  \\
        \midrule
         Gravitational field (strength) & Force per unit mass ($N/kg$) & Vector & $\R^3 \rightarrow \R^3$ \\
         Electric field (strength) & Force per unit electric charge ($N/C$) & Vector & $\R^3 \rightarrow \R^3$ \\
         Magnetic field (strength) & Electric current per meter ($A/m$) & Vector &  $\R^3 \rightarrow \R^3$ \\
         Pressure field & Force per unit square ($N/m^2$) & & \\
         & = energy per unit volume ($J/m^3$) & Scalar & $\mathbb{R}^3 \rightarrow \R$ \\
         Mean sea level pressure & Pressure field at mean sea level & Scalar & $\R^2 \rightarrow \R$ \\
         Flow velocity field & Change of point along its streamline\tnote{1}~~($v$) & Vector & $\R^2\rightarrow \R^2$, $\R^3 \rightarrow \R^3$ \\
         Flow speed field & Length of flow velocity vector ($\lvert v \rvert$) & Scalar & $\R^2\rightarrow \R$, $\R^3 \rightarrow \R$ \\
         Wind velocity field & Air flow velocity field ($v$) & Vector & $\R^2\rightarrow \R^2$, $\R^3 \rightarrow \R^3$\\
         Temperature field & Temperature at space point ($K$) & Scalar & $\R^3 \rightarrow \R$\\
         Signed distance field (SDF) & Signed distance & Scalar & $\R^3 \rightarrow \R$ \\
         Occupancy field & Occupancy & Scalar & $\R^3 \rightarrow \R$\\
        \bottomrule
    \end{tabular}
    \begin{tablenotes}
     \footnotesize
     \item[1] Streamlines are a family of curves whose tangent vectors constitute the velocity vector field of the flow. Streamlines differ over time when the flow of a fluid changes. The flow velocity vector field itself shows the direction in which a massless fluid element will travel at any spatial coordinate in time, and therefore describes and characterizes a fluid. 
   \end{tablenotes}
\end{threeparttable}
\end{table}